\RequirePackage{fix-cm}
\documentclass[twocolumn]{svjour3} 

\usepackage{xspace}
\usepackage{color}

\definecolor{gray}{rgb}{0.35,0.35,0.35}
\definecolor{MyBlue}{rgb}{0,0.2,0.8}
\definecolor{MyRed}{rgb}{0.8,0.2,0}
\definecolor{MyGreen}{rgb}{0.0,0.5,0.1}
\definecolor{MyGray}{rgb}{0.4,0.4,0.4}

\long\def\ignorethis#1{}

\newlength\figcca%
\newlength\figccb%
\newlength\figccc%
\newlength\figccd

\ignorethis{
\newcommand*\patchAmsMathEnvironmentForLineno[1]{%
  \expandafter\let\csname old#1\expandafter\endcsname\csname #1\endcsname
  \expandafter\let\csname oldend#1\expandafter\endcsname\csname end#1\endcsname
  \renewenvironment{#1}%
     {\linenomath\csname old#1\endcsname}%
     {\csname oldend#1\endcsname\endlinenomath}}%
\newcommand*\patchBothAmsMathEnvironmentsForLineno[1]{%
  \patchAmsMathEnvironmentForLineno{#1}%
  \patchAmsMathEnvironmentForLineno{#1*}}%
\AtBeginDocument{%
\patchBothAmsMathEnvironmentsForLineno{equation}%
\patchBothAmsMathEnvironmentsForLineno{align}%
\patchBothAmsMathEnvironmentsForLineno{flalign}%
\patchBothAmsMathEnvironmentsForLineno{alignat}%
\patchBothAmsMathEnvironmentsForLineno{gather}%
\patchBothAmsMathEnvironmentsForLineno{multline}%
} 
}
\usepackage{times}
\usepackage{epsfig}
\usepackage{graphicx}
\usepackage{amsmath}
\usepackage{amssymb}
\usepackage{mathrsfs}
\usepackage{array}
\usepackage{booktabs}
\usepackage[lined,ruled,linesnumbered]{algorithm2e}
\usepackage{multirow}
\usepackage{threeparttable}
\usepackage{wasysym}
\usepackage{xcolor}

\usepackage[normalem]{ulem}
\useunder{\uline}{\ul}{}

\smartqed 
\journalname{International Journal of Computer Vision}

\usepackage[breaklinks=true,bookmarks=false]{hyperref}

\DeclareMathOperator*{\argmax}{argmax}

\newcommand{\bx}{\mathbf{x}}
\newcommand{\bbX}{\mathbb{X}}
\newcommand{\bX}{\mathcal{X}}
\newcommand{\by}{\mathbf{y}}
\newcommand{\bY}{\mathcal{Y}}

\newcommand{\bK}{\mathcal{K}}

\newcommand{\bh}{\mathbf{h}}
\newcommand{\bbf}{\mathbf{f}}
\newcommand{\bF}{\mathbf{F}}
\newcommand{\fF}{\mathscr{F}}
\newcommand{\bw}{\mathbf{w}}
\newcommand{\bW}{\mathcal{W}}
\newcommand{\bz}{\mathbf{z}}
\newcommand{\bv}{\mathbf{v}}

\newcommand{\bb}{\mathbf{b}}

\begin{document}
	
	\title{Adaptive Correlation Filters with Long-Term and Short-Term Memory for Object Tracking
	}
	
	
	\author{Chao Ma \and Jia-Bin Huang \and Xiaokang Yang \and Ming-Hsuan Yang
	}
	
	
	\institute{C. Ma \at  Shanghai Jiao Tong University, Shanghai, 200240, P. R. China. \email{chaoma@sjtu.edu.cn}
		\at Australian Centre for Robotic Vision, The University of Adelaide, Adelaide, 5000, Australia. 
		\email{c.ma@adelaide.edu.au}
		\and J.-B. Huang \at Virginia Tech, VA, 24060. 
		\email{jbhuang@vt.edu}
		\and X. Yang \at Shanghai Jiao Tong University, Shanghai, 200240, P. R. China.
		\email{xkyang@sjtu.edu.cn}
		\and M.-H. Yang \at University of California at Merced, CA, 95344.
		\email{mhyang@ucmerced.edu}
	}
	
	\date{Received: date / Accepted: date}
	
	\maketitle
	
	\begin{abstract}
		Object tracking is challenging as target objects often undergo drastic appearance changes over time. 
		Recently, adaptive correlation filters have been successfully applied to object tracking. 
		However, tracking algorithms relying on highly adaptive correlation filters are prone to drift due to noisy updates. 
		Moreover, as these algorithms do not maintain long-term memory of target appearance, they cannot recover from tracking failures caused by heavy occlusion or target disappearance in the camera view. 
		In this paper, we propose to learn multiple adaptive correlation filters with both long-term and short-term memory of target appearance for robust object tracking. 
		First, we learn a kernelized correlation filter with an aggressive learning rate for locating target objects precisely. 
		We take into account the appropriate size of surrounding context and the feature representations. 
		Second, we learn a correlation filter over a feature pyramid centered at the estimated target position for predicting scale changes. 
		Third, we learn a complementary correlation filter with a conservative learning rate to maintain  long-term memory of target appearance. 
		We use the output responses of this long-term filter to determine if tracking failure occurs. 
		In the case of tracking failures, we apply an incrementally learned detector to recover the target position in a sliding window fashion. 
		Extensive experimental results on large-scale benchmark datasets demonstrate that the proposed algorithm performs favorably against the state-of-the-art methods in terms of efficiency, accuracy, and robustness.
	\end{abstract}

	\begin{figure*}
		\centering
		\setlength{\tabcolsep}{1pt}
		\begin{tabular}{cccc}
			\includegraphics[trim = 0mm 10mm 15mm 0mm, clip, width=.245\textwidth]{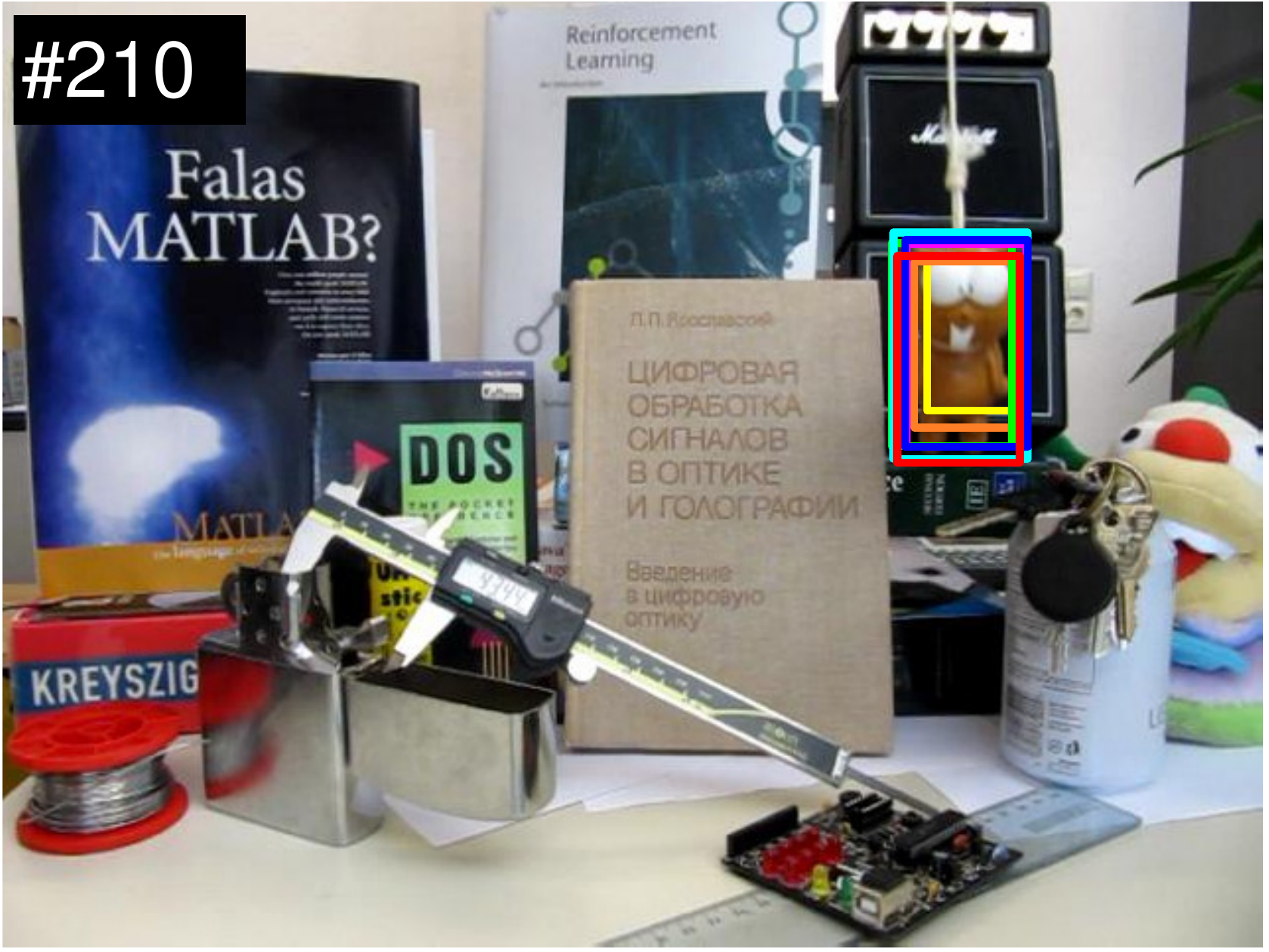} &
			\includegraphics[trim = 0mm 10mm 15mm 0mm, clip, width=.245\textwidth]{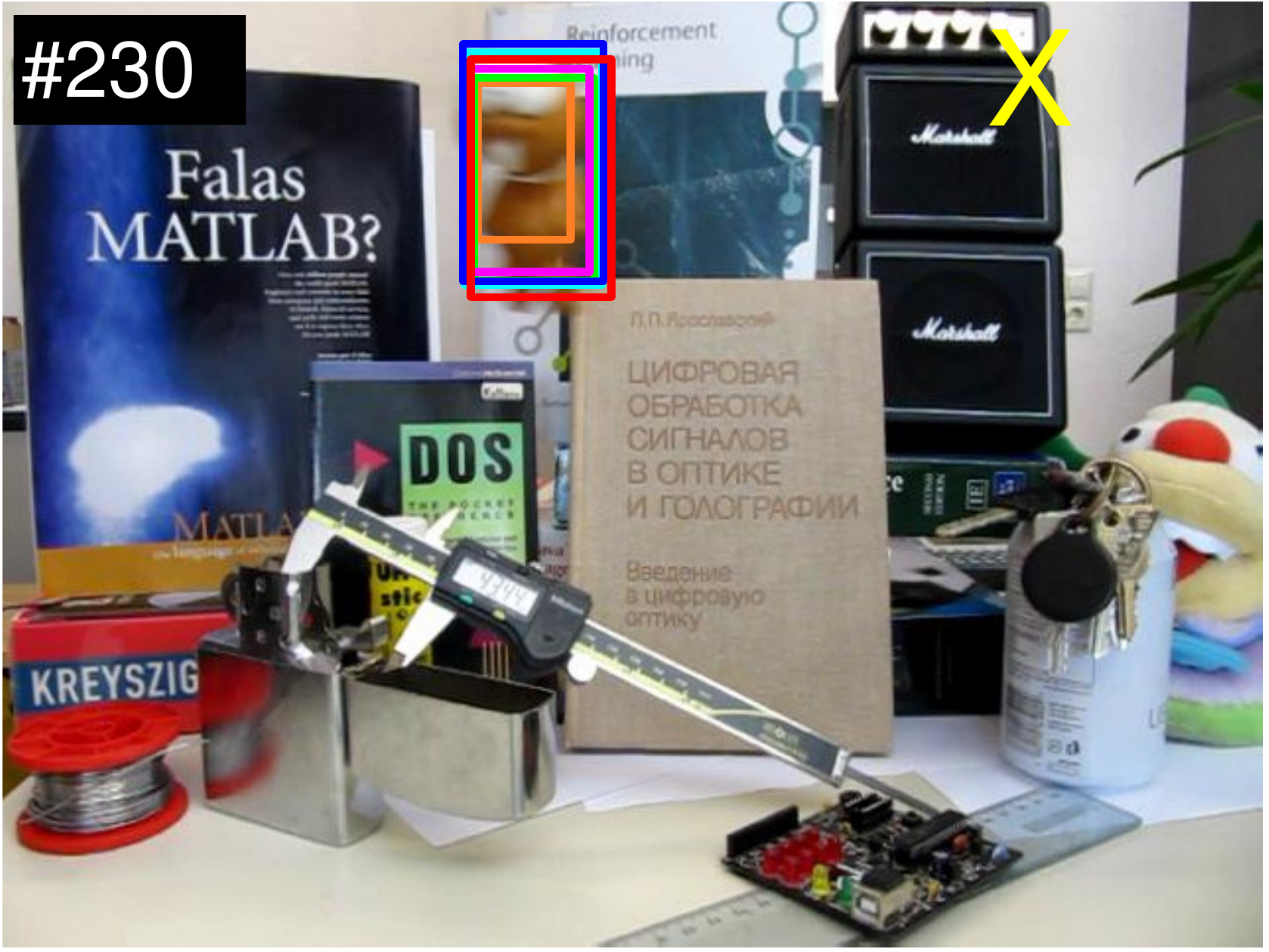} &
			\includegraphics[trim = 0mm 10mm 15mm 0mm, clip, width=.245\textwidth]{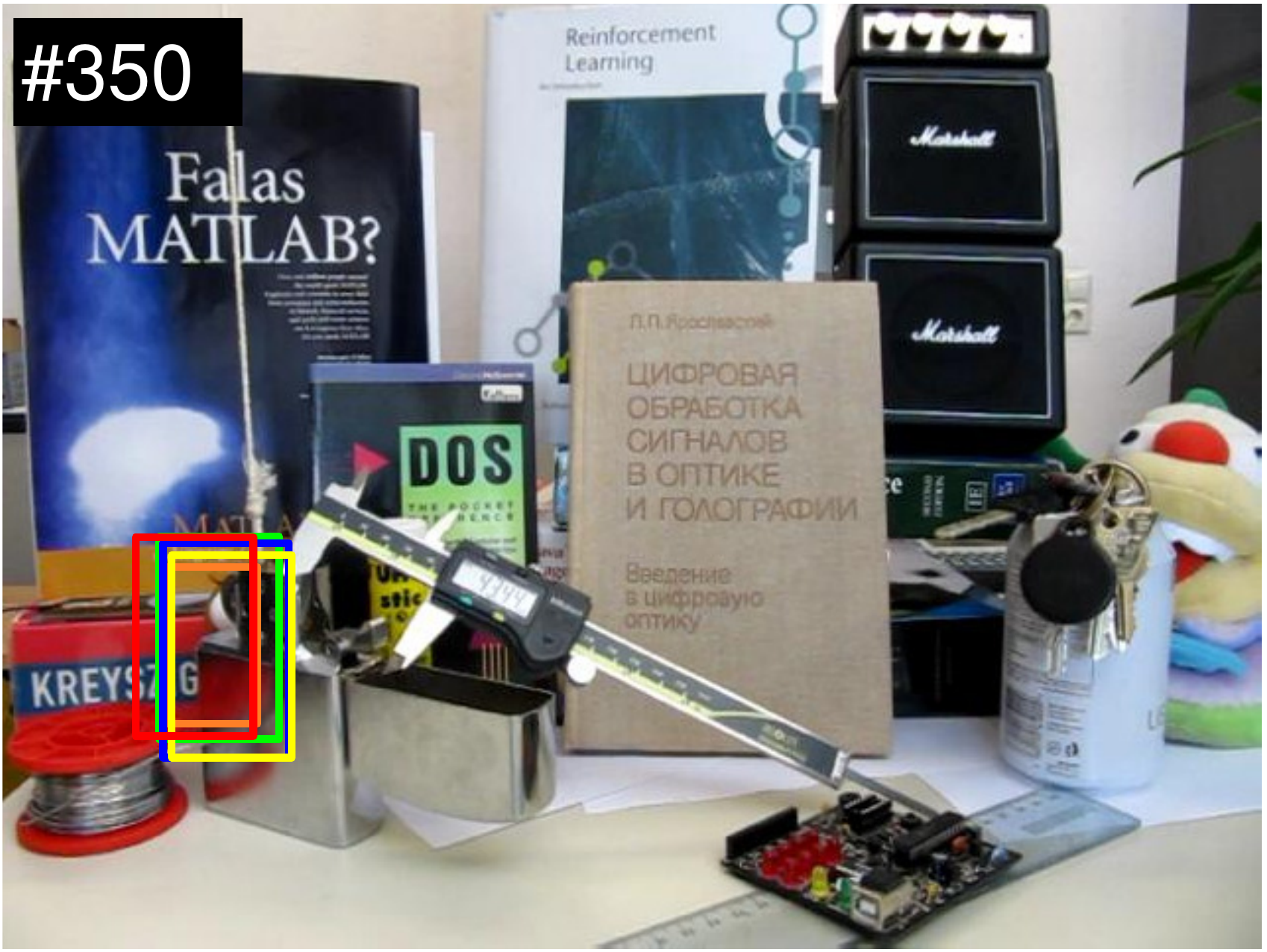} &
			\includegraphics[trim = 0mm 10mm 15mm 0mm, clip, width=.245\textwidth]{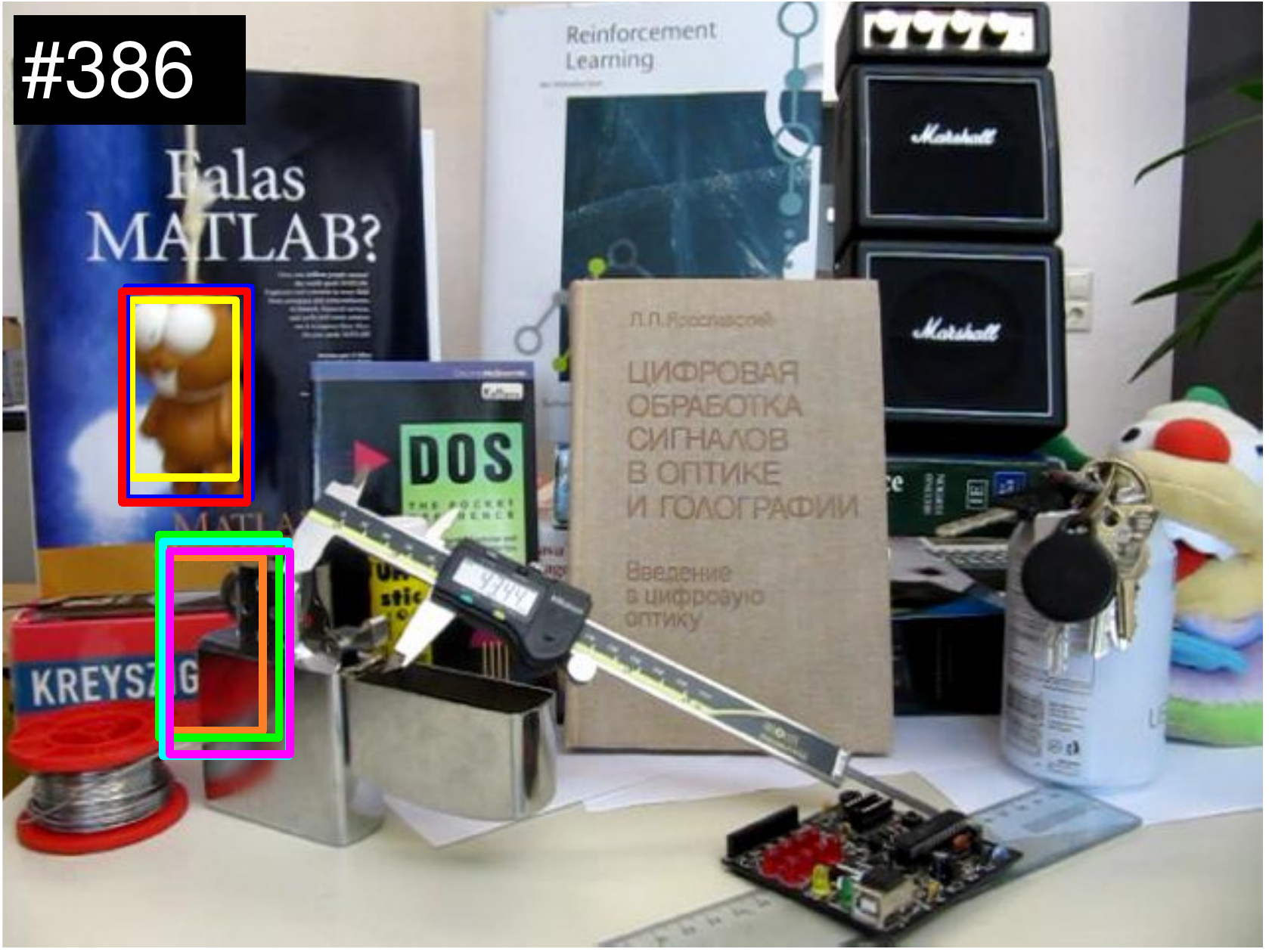}\\ 
		\end{tabular}
		\includegraphics[width=.92\textwidth]{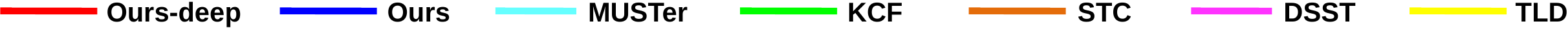}
		\caption{\textbf{Effectiveness of long-term memory of target appearance for tracking.}
			Sample tracking results on the \textit{lemming} sequence \cite{DBLP:conf/cvpr/WuLY13} by our approach, MUSTer \cite{Hong_2015_CVPR}, KCF \cite{DBLP:journals/pami/HenriquesC0B15}, STC \cite{DBLP:conf/eccv/ZhangZLZY14}, DSST \cite{DBLP:conf/bmvc/DanelljanKFW14} and TLD \cite{DBLP:journals/pami/KalalMM12} ($\times$: no tracking output from TLD \cite{DBLP:journals/pami/KalalMM12}). 
			Our tracker learns adaptive correlation filters with short-term memory for translation and scale estimation. 
			Compared to the TLD \cite{DBLP:journals/pami/KalalMM12} tracker, the proposed tracking algorithm is more robust to abrupt motion and significant deformation in the 230th frame. 
			Our trackers explicitly capture long-term memory of target appearance.
			As a result, our methods can recover the lost target after persistent occlusion in the 386th frame. 
			The other state-of-the-art correlation trackers (MUSTer \cite{Hong_2015_CVPR}, KCF \cite{DBLP:journals/pami/HenriquesC0B15}, DSST \cite{DBLP:conf/bmvc/DanelljanKFW14} and STC \cite{DBLP:conf/eccv/ZhangZLZY14}) fail to handle such tracking failures. 
		}
		\label{fig:demo}
	\end{figure*}
	
	\section{Introduction}
	\label{sec: introduction}
	Object tracking is one of the fundamental problems in computer vision with numerous applications including surveillance, human-computer interaction, and autonomous vehicle navigation \cite{DBLP:journals/csur/YilmazJS06,DBLP:journals/tist/LiHSZDH13,DBLP:journals/pami/SmeuldersCCCDS14}. 
	Given a generic target object specified by a bounding box in the first frame, the goal of object tracking is to estimate the unknown target states, e.g., position and scale, in the subsequent frames. 
	Despite significant progress in the last decade, object tracking remains challenging due to the large appearance variation caused by deformation, sudden motion, illumination change, heavy occlusion, and target disappearance in the camera view, to name a few. 
	To cope with this appearance variation over time, adaptive correlation filters have been applied for object tracking. 
	However, existing tracking algorithms relying on such highly adaptive models do not maintain  long-term memory of target appearance and thus are prone to drift in the case of noisy updates.
	In this paper, we propose to employ \emph{multiple} adaptive correlation filters with both long-term and short-term memory for robust object tracking.
	
	Correlation filters have attracted considerable attention in the object tracking community \cite{DBLP:conf/cvpr/BolmeBDL10,DBLP:conf/eccv/HenriquesCMB12,DBLP:conf/bmvc/DanelljanKFW14,DBLP:conf/eccv/ZhangZLZY14,DBLP:conf/eccv/LiZ14,Hong_2015_CVPR,Liu_2015_CVPR,Ma_2015_CVPR,DBLP:journals/pami/HenriquesC0B15,ma_2015_icip,iccv15/DanelljanHKF15,cvpr16/BertinettoVGMT16} in recent years. 
	We attribute the effectiveness of correlation filters for object tracking to the following three important characteristics.
	First, correlation filter based tracking algorithms can achieve high tracking speed by computing the spatial correlation efficiently in the Fourier domain. 
	The use of kernel tricks \cite{DBLP:journals/pami/HenriquesC0B15,DBLP:conf/cvpr/DanelljanKFW14,DBLP:conf/eccv/LiZ14} further improves the tracking accuracy without significantly increasing the computational complexity. 
	Second, correlation filters naturally take the surrounding visual context into account and provide more discriminative information than the appearance models \cite{DBLP:journals/pami/KalalMM12,DBLP:conf/cvpr/SupancicR13} constructed based on target objects only. 
	For example, even if a target object undergoes heavy occlusion, contextual cues can still help infer the target position \cite{DBLP:conf/eccv/ZhangZLZY14}.
	Third, learning correlation filters is equivalent to a regression problem \cite{DBLP:journals/pami/HenriquesC0B15,DBLP:conf/eccv/HenriquesCMB12}, where the circularly shifted versions of input image patches are regressed to \emph{soft} labels, 
	e.g., generated by a Gaussian function with a narrow bandwidth ranging from zero to one. 
	This differs from existing tracking-by-detection approaches \cite{DBLP:journals/pami/Avidan07,DBLP:journals/pami/BabenkoYB11,DBLP:conf/iccv/HareST11} where \emph{binary} (hard-thresholded) sample patches are densely or randomly drawn around the estimated target positions to incrementally train discriminative classifiers. 
	Hence, correlation filter based trackers can alleviate the inevitable ambiguity of assigning positive and negative labels to those highly spatially correlated samples.
	
	However, existing correlation filter based trackers \cite{DBLP:conf/cvpr/BolmeBDL10,DBLP:conf/eccv/HenriquesCMB12,DBLP:conf/bmvc/DanelljanKFW14,DBLP:conf/eccv/ZhangZLZY14,DBLP:conf/eccv/LiZ14} have several limitations. 
	These methods adopt moving average schemes with high learning rates to update the learned filters for handling appearance variations over time.
	Since such highly adaptive update schemes can only maintain a short-term memory of target appearance, these methods are thus prone to drift due to the noisy updates \cite{DBLP:journals/pami/MatthewsIB04}, and cannot recover from tracking failures as long-term memory of target appearance is not maintained. 
	Figure \ref{fig:demo} shows an example highlighting these issues.
	The state-of-the-art correlation filter trackers (KCF \cite{DBLP:journals/pami/HenriquesC0B15}, STC \cite{DBLP:conf/eccv/ZhangZLZY14}, and DSST \cite{DBLP:conf/bmvc/DanelljanKFW14}) tend to drift caused by noisy updates in the 350th frame and fail to recover in the 386th frame after a long-duration occlusion.
	
	In this paper, we address the stability-adaptivity dilemma \cite{DBLP:journals/cogsci/Grossberg87,DBLP:conf/cvpr/SantnerLSPB10} by strategically leveraging both the short-term and long-term memory of target appearance.
	Specifically, we exploit three types of correlation filters: (1) translation filter, (2) scale filter, and (3) long-term filter.
	First, we learn a correlation filter for estimating the target translation.
	To improve location accuracy, we introduce histogram of local intensities (HOI) as complementary features to the commonly used histogram of oriented gradients (HOG). 
	We show that the combined features increase the discriminative strength between the target and its surrounding background.
	Second, we learn a scale correlation filter by regressing the feature pyramid of the target object to a one-dimensional scale space for estimating the scale variation.
	Third, we learn and update a long-term filter using confidently tracked sample patches. 
	For each tracked result, we compute the confidence score using the long-term filter to determine whether tracking failures occur.
	When the confidence score is below a certain threshold, we activate an online trained detector to recover the target object.
	
	The main contribution of this work is an effective approach that best exploits three types of correlation filters for robust object tracking. 
	Specifically, we make the following three contributions:
	\begin{itemize}
		\item We show that adaptive correlation filters are competent in estimating translation and scale changes, as well as determining whether tracking failures occur. 
		Compared to our previous work in \cite{Ma_2015_CVPR}, we employ a different detection module with incremental updates using an efficient passive-aggressive scheme \cite{DBLP:journals/jmlr/CrammerDKSS06}.
		\item We systematically analyze the effect of different feature types and the size of surrounding context area for designing effective correlation filters. We also provide thorough ablation study to investigate the contribution of the design choices.
		\item We discuss and compare the proposed algorithm with the 
		concurrent work \cite{Hong_2015_CVPR} in details.
		We evaluate the proposed algorithm and present extensive comparisons with the state-of-the-art trackers on both the OTB2013 \cite{DBLP:conf/cvpr/WuLY13} and OTB2015 \cite{DBLP:journals/pami/WuLY15} datasets as well as on additional 10 challenging sequences from \cite{DBLP:conf/eccv/ZhangMS14}.
	\end{itemize}

	\section{Related Work}
	Object tracking has been an active research topic in computer vision. 
	In this section, we discuss the most closely related tracking-by-detection approaches.
	Comprehensive reviews on object tracking can be found in \cite{DBLP:journals/csur/YilmazJS06,DBLP:journals/tist/LiHSZDH13,DBLP:journals/pami/SmeuldersCCCDS14}.
	
	\subsection{Tracking-by-Detection} 
	Tracking-by-detection methods treat object tracking in each frame as a detection problem within a local search window, and often by incrementally learning classifiers to separate the target from its surrounding background. 
	To adapt to the appearance variations of the target, existing approaches typically draw positive and negative training sample patches around the estimated target location for updating the classifiers. 
	Two issues ensue with such approaches. 
	The first issue is the sampling ambiguity, i.e., a slight inaccuracy in the labeled samples may be accumulated over time and cause trackers to drift. 
	Numerous methods have been proposed to alleviate the sampling ambiguity.
	The main objective is to robustly update a discriminative classifier with noisy training samples.
	Examples include ensemble learning \cite{DBLP:journals/pami/Avidan07,DBLP:conf/iccv/BaiWSBM13}, semi-supervised learning \cite{DBLP:conf/eccv/GrabnerLB08}, multiple instance learning (MIL) \cite{DBLP:journals/pami/BabenkoYB11}, structure learning \cite{DBLP:conf/iccv/HareST11}, and transfer learning \cite{DBLP:conf/eccv/GaoLHX14}. 
	The second issue is the dilemma between stability and adaptivity when updating appearance models. 
	To strike a balance between the model stability and adaptivity, Kalal et al. \cite{DBLP:journals/pami/KalalMM12} decompose the tracking task into tracking, learning and detection (TLD) modules 
	where the tracking and detection modules facilitate each other, i.e., the results from the aggressively updated tracker provide additional training samples to conservatively update the detector. The online learned detector can be used to reinitialize the tracker when tracking failure occurs. Similar mechanisms have also been exploited in \cite{DBLP:conf/eccv/Pernici12,DBLP:conf/cvpr/SupancicR13,DBLP:conf/eccv/HuaAS14} to recover target objects from tracking failures. 
	Zhang et al. \cite{DBLP:conf/eccv/ZhangMS14} use multiple classifiers with different learning rates and design an entropy measure to fuse multiple tracking outputs. We also use an online trained detector for reinitializing the tracker as in \cite{DBLP:journals/pami/KalalMM12,DBLP:conf/eccv/ZhangMS14}. 
	However, we only activate the detector if the response from the long-term filter is lower than a certain threshold. 
	This approach helps improve efficiency because we do not active the detector in every frame as in \cite{DBLP:journals/pami/KalalMM12,DBLP:conf/eccv/ZhangMS14}.
	
	\begin{figure*}[t]
		\centering
		\includegraphics[width=.92\textwidth]{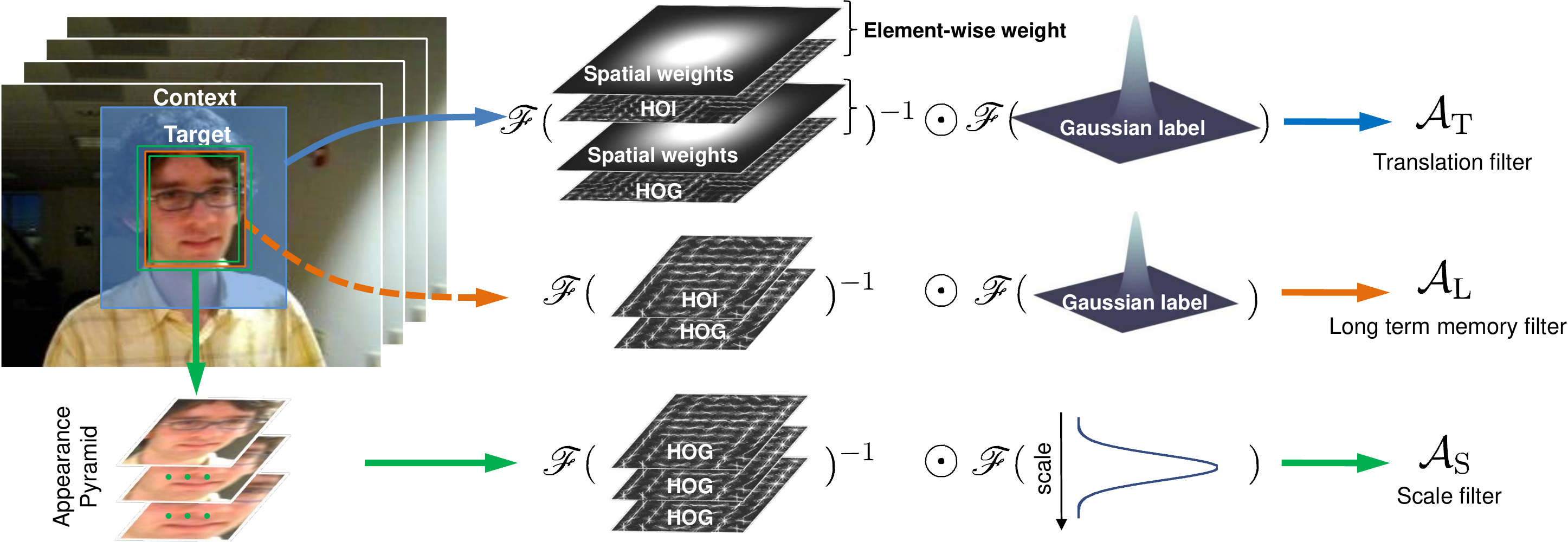} \\
		\caption{
			\textbf{Overview of three types of correlation filters: translation filter, scale filter, and long-term memory filter.}
			The translation filter $\mathcal{A}_\mathrm{T}$ with short-term memory adapts to appearance changes of the target and its surrounding context.
			The scale filter $\mathcal{A}_\mathrm{S}$ predicts scale variation of the target. 
			The long-term filter $\mathcal{A}_\mathrm{L}$ conservatively learns and maintains long-term memory of target appearance. 
			Also, we use the long-term filter $\mathcal{A}_\mathrm{L}$ to detect tracking failures by checking if the filter response is below a certain threshold.
			Note that the filter responses for the scale filter $\mathcal{A}_\mathrm{S}$ are one-dimensional while the filter responses of the other two filters are two-dimensional. 
			We train the translation filter $\mathcal{A}_\mathrm{T}$ using the histogram of local intensities as features in addition to the commonly used HOG features.
			Furthermore, we apply a layer of spatial weights to the feature space to cope with the discontinuity introduced by circular shifts. 
			The operator $\fF$ indicates the Fourier transformation, and $\odot$ is the Hadamard product. The dashed arrow in orange denotes a conservative update scheme for filter learning.
		}
		\label{fig:tworeg}
	\end{figure*}

	\subsection{Correlation Filter based Tracking}
	Correlation filters have been widely applied to various computer vision problems such as object detection and recognition \cite{MC/book/CPR}. 
	Recently, correlation filters have drawn significant attention in the object tracking community, owing to the computational efficiency and the effectiveness in alleviating the sampling ambiguity, i.e., learning correlation filters does not require hard-thresholded binary samples.
	In \cite{DBLP:conf/cvpr/BolmeBDL10} Bolme et al. learn a minimum output sum of squared error (MOSSE)~filter on the luminance channel for fast tracking. 
	Considerable efforts have since been made to improve the tracking performance using correlation filters.
	Extensions include kernelized correlation filters \cite{DBLP:conf/eccv/HenriquesCMB12}, multi-channel filters \cite{DBLP:journals/pami/HenriquesC0B15,DBLP:conf/cvpr/DanelljanKFW14,ma_2015_icip,DBLP:conf/iccv/GaloogahiSL13}, context learning \cite{DBLP:conf/eccv/ZhangZLZY14}, scale estimation \cite{DBLP:conf/bmvc/DanelljanKFW14,DBLP:conf/eccv/LiZ14}, and spatial regularization \cite{iccv15/DanelljanHKF15}.
	However, most of these approaches emphasize the adaptivity of the model and do not maintain a 
	long-term memory of target appearance. As a result, these models are prone to drift in the presence of occlusion and target disappearance in the camera view and are unable to recover targets from tracking failures either. 
	Our work builds upon correlation filter based trackers. 
	Unlike existing work that relies on only one correlation filter for translation estimation, we learn three complementary correlation filters for estimating target translation, predicting scale change, and determining whether tracking failure occurs.
	The most closely related work is that by Hong et al. \cite{Hong_2015_CVPR} (MUSTer) which also exploits the long-term and short-term memory for correlation filter based tracking. 
	The main difference lies in the model used for capturing the long-term memory of target appearance.
	The MUSTer tracker represents the target appearance using a pool of local features. 
	In contrast, our long-term correlation filter represents a target object with a holistic template. 
	We observe that it is often challenging to
	match two sets of local points of interest between two frames due to outliers. Figure \ref{fig:demo} shows one example where the MUSTer tracker fails to recover the target in the 386th frame, as few interest points are correctly matched.

	\section{Overview}
	We aim to exploit multiple correlation filters to handle the following three major challenges in object tracking: 
	(1) significant appearance changes over time, 
	(2) scale variation, and 
	(3) target recovery from tracking failures.
	First, existing trackers using one single correlation filter are unable to achieve these goals as it is difficult to strike a balance between the stability and adaptivity with only one module. 
	Second, while considerable efforts have been made to address the problem of scale prediction \cite{DBLP:conf/bmvc/DanelljanKFW14,DBLP:conf/eccv/LiZ14,DBLP:conf/eccv/ZhangZLZY14}, it remains an unsolved problem as slight inaccuracy in scale estimation causes significant performance loss of an appearance model. 
	Third, determining when tracking failures occur and re-detecting the target object from failures remain challenging. 
	In this work, we exploit three types of correlation filters with different levels of adaptiveness to address these issues. 
	Figure \ref{fig:tworeg} illustrates the construction of three correlation filters for object tracking. 
	We use two correlation filters with short-term memory, i.e., the translation filter $\mathcal{A}_\mathrm{T}$ and the scale filter $\mathcal{A}_\mathrm{S}$, for translation and scale estimation. 
	We learn a long-term filter $\mathcal{A}_\mathrm{L}$ to maintain long-term memory of target appearance for estimating the confidence of each tracked result.
	
\begin{figure*}
		\centering
		\includegraphics[width=\textwidth]{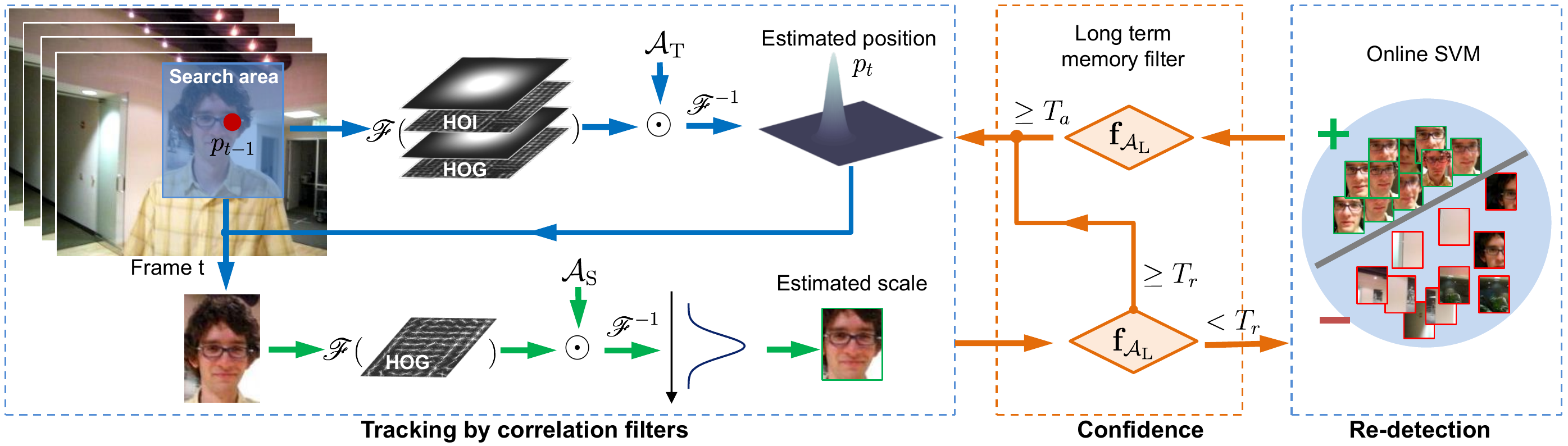} \\
		\caption{
			\textbf{Overview of the proposed algorithm.} 
			We decompose the tracking task into translation and scale estimation. We first infer the target position $p_t$ from correlation response map of the translation filter $\mathcal{A}_\mathrm{T}$, and then predict the scale change using the scale filter $\mathcal{A}_\mathrm{S}$. The correlation tracking module collaborates with the re-detection module built on SVM classifier via the long-term filter $\mathcal{A}_\mathrm{L}$. We activate the re-detection module when the correlation response of $\mathcal{A}_\mathrm{L}$ is below a given $T_r$. 
			Note that we conservatively adopt the detection result \emph{only when it is highly confident}, i.e., the correlation response of $\mathcal{A}_\mathrm{L}$ is above a given threshold $T_a$.
		}
		\label{fig:flowchart}
	\end{figure*}

	Figure \ref{fig:flowchart} illustrates the main steps of the proposed algorithm with the three correlation filters for object tracking. 
	Given an input frame, we first apply the translation filter $\mathcal{A}_\mathrm{T}$ for locating the target object in a search window centered at the position in the previous frame. 
	Once we obtain the estimated target position, we apply the scale filter $\mathcal{A}_\mathrm{S}$ to predict the scale changes. 
	For each tracked result, we use the long-term filter $\mathcal{A}_\mathrm{L}$ to determine whether tracking failure occurs (i.e., whether the confidence score is below a certain threshold). 
	In the cases where the tracker loses the target, we activate an online detector for recovering the lost or drifted target, and reinitialize our tracker. 
	While the proposed long-term filter itself can also be used as a detector, the computational load is high due to the use of high-dimensional features. 
	For computational efficiency, we build on an additional detection module using an online support vector machine (SVM). 
	We update both the detection module and the long-term filter with a conservative learning rate to capture the target appearance over a long temporal span.
	
	\section{Tracking Components}
	In this section, we describe the main components of the proposed tracking algorithm. 
	We first describe the kernelized correlation filters \cite{DBLP:conf/eccv/HenriquesCMB12}, and then present the schemes of learning these filters over multi-channel features \cite{DBLP:journals/pami/HenriquesC0B15} for translation estimation and scale prediction. 
	We then discuss how to learn correlation filters to capture the short-term and long-term memory of target appearance using different learning rates, as well as how these filters collaborate with each other for object tracking.
	Finally, we present an online detection module to recover the target when tracking failure occurs.
	
	\subsection{Kernelized Correlation Filters}
	\label{sec:tracking}
	Correlation filter based trackers \cite{DBLP:conf/cvpr/DanelljanKFW14,DBLP:conf/bmvc/DanelljanKFW14} achieve the state-of-the-art performance in the recent benchmark evaluations \cite{DBLP:conf/cvpr/WuLY13,Hadfield14d}. 
	The core idea of these methods is to regress the circularly shifted versions of the input image patch 
	to soft target scores (e.g., generated by a Gaussian function and decaying from 1 to 0 when the input images gradually shift away from the target center).
	The underlying assumption is that the circularly shifted versions of input images approximate the dense samples of target appearance \cite{DBLP:journals/pami/HenriquesC0B15}. 
	As learning correlation filters do not require binary (hard-threshold) samples, correlation filter based trackers effectively alleviate the sampling ambiguity that adversely affects most tracking-by-detection approaches. 
	By exploiting the redundancies in the set of shifted samples, correlation filters can be trained with a substantially large number of training samples efficiently using fast Fourier transform (FFT).
	This data augmentation helps discriminate the target from its surrounding background.
	
	Given one-dimensional data $\bx=\{x_1,x_2,\ldots,x_n\}^\top$, we denote its circularly shifted version with one entry by $\bx_1=\{x_n, x_1,x_2,\ldots,x_{n-1}\}^\top$. 
	All the circularly shifted versions of $\bx$ are concatenated to form the circulant matrix $\bbX=[\bx_1, \bx_2, \ldots, \bx_n]$. 
	Using the Discrete Fourier Transform (DFT), we can diagonalize the circulant matrix $\bbX$ as:
	\begin{equation}
	\bbX=\bF^\mathrm{H}\text{diag}\bigl(\fF(\bx)\bigr)\bF,
	\label{equ:cmatrix}
	\end{equation}
	where $\bF$ denotes the constant DFT matrix that transforms the data from the spatial domain into the Fourier domain. 
	We denote $\fF(\bx)$ as the Fourier transform of $\bx$, i.e., $\fF(\bx)=\bF\bx$ and 
	$\bF^\mathrm{H}$ as the Hermitian transpose of $\bF$.
	
	A linear correlation filter $f$ trained on an image patch $\bx$ of size $M\times N$ is equivalent to a ridge regression model, which considers all the circularly shifted versions of $\bx$ (in both horizontal and vertical directions) as training data. 
	Each example $\bx_i$, $i\in\{1,2,\ldots,M\}\times\{1,2,\ldots,N\}$ corresponds to
	a target score $y_i=\exp\left(-\frac{(m-M/2)^2+(n-N/2)^2}{2\sigma_0^2}\right)$, where $(m,n)$ indicates the shifted positions along horizontal and vertical directions. 
	The target center has a maximum score $y_i = 1$. 
	The score $y_i$ decays rapidly from 1 to 0 when the position $(m, n)$ is away from the desired target center.
	The kernel width $\sigma_0$ is a predefined parameter controlling the sensitivity of the score function. 
	The objective function of the linear ridge regression for learning the correlation filter is of the form:
	\begin{equation}
	\label{equ:filter}
	\min_\bw\sum_{i=1}^{M\times N}\bigl(f(\bx_{i})-y_i\bigr)^2+\lambda\|\bw\|^2,
	\end{equation}
	where $\lambda > 0$ is a regularization parameter. The solution to \eqref{equ:filter} is a linear estimator: $f(\bx)=\mathbf{w}^\top\bx$. 
	As a result, the ridge regression problem has the close-form solution with respect to the circulant matrix $\bbX$ as follows: 
	\begin{equation}
	\bw=(\bbX^\top\bbX+\lambda\mathbf{I})^{-1}\bbX^\top\by,
	\end{equation}
	where $\mathbf{I}$ is the identity matrix with the size of $MN\times MN$. 
	Substituting the DFT form of $\bx$ for the circulant matrix $\bbX$ using \eqref{equ:cmatrix}, 
	we have the solution in the Fourier domain as:
	\begin{equation}
	\bW=\frac{\bX\odot\overline{\bY}}{\bX\odot\overline{\bX}+\lambda},
	\end{equation} 
	where $\bW$, $\bX$, $\bY$ are the corresponding signals in the Fourier domain, and $\overline{\bX}$ is the complex conjugate of $\bX$.

	To strengthen the discrimination of the learned filters, Henriques et al. \cite{DBLP:conf/eccv/HenriquesCMB12} use a kernel $k$, $k(\bx,\bx')=\langle\phi(\bx),\phi(\bx')\rangle$ to learn correlation filters in a kernel space while maintaining the computational complexity as its linear counterpart. 
	Here the notation $\langle\cdot,\cdot\rangle$ denotes inner product. 
	The kernelized filter can be derived as $f(\bx)=\bw^\top\phi(\bx)=\sum_i a_i k(\bx_i,\bx')$, where $\mathbf{a}=\{a_i\}$ are the dual variables of $\bw$. For shift-invariant kernels, e.g., an RBF kernel, the dual coefficients $\mathbf{a}$ can be obtained using the circulant matrix in the Fourier domain \cite{DBLP:conf/cvpr/DanelljanKFW14,DBLP:journals/pami/HenriquesC0B15} as: 
	\begin{equation} 
	{\mathcal{A}}=\frac{\overline{\bY}}{\overline{\bK}^{\bx\bx'}+\lambda},
	\label{equ:A} 
	\end{equation} 
	where the matrix $\bK$ denotes the Fourier transform of the kernel correlation matrix $\mathbf{k}$, which is defined as: 
	\begin{equation}
	\mathbf{k}^{\bx\bx'}=\exp\left(-\frac{\|\bx\|^2+\|\bx'\|^2}{\sigma^2}-2\fF^{-1}\left(\bX\odot\overline{\bX}'\right)\right).
	\label{equ:skernel}
	\end{equation}
	As the computation involves only element-wise product, the overall complexity remains in linearithmic time, $\mathcal{O}(n\log n)$, where the input size $n = MN$.
	Given a new input frame, we apply a similar scheme in \eqref{equ:skernel} to compute the correlation response map. 
	Specifically, we first crop an image patch $\mathbf{z}$ centered at the location in the previous frame.
	We then compute the response map $\bbf$ using the learned target template $\tilde{\bx}$ in the Fourier domain as:
	\begin{equation}
	\bbf(\mathbf{z})=\fF^{-1}(\bK^{\tilde{\bx}\mathbf{z}}\odot\mathcal{A}).
	\label{equ:getres}
	\end{equation}
	We can then locate the target object by searching for the position with the maximum value in the response map $\bbf$. 
	Note that the maximum correlation response naturally reflects the similarity between a candidate patch and the learned filter.
	As we learn the filer in the scale space, the confidence score can be used to estimate scale changes (Section \ref{sec:scale}).
	Similarly, when the filter maintains long-term memory of target appearance, the confidence score can be used for determining tracking failures (Section \ref{sec:lstm}). Note that here we use two-dimensional long-term memory filter. In the case where an input image patch is not well aligned with the learned filter, the maximum value of the 2D response map can still effectively indicate the similarity between the image patch and learned filter.
	
	
	\subsection{Multi-Channel Correlation Filters}
	Appearance features play a critical role for object tracking. 
	In general, multi-channel features are more effective in separating the foreground and background than single channel features, e.g., intensity.
	With the use of the kernel trick, we can efficiently learn correlation filters over multi-channel features.
	Let $\bx=[\bx_1,\bx_2,\ldots,\bx_c]$ denote the multi-channel features for the target object, where $c$ is the total number of channels.
	The kernel correlation matrix $\mathbf{k}^{\bx\bx'}$ can be computed by a summation of the element-wise products over each feature channel in the Fourier domain. 
	Thus, we rewrite \eqref{equ:skernel} as follows:
	\begin{equation}
	\mathbf{k}^{\bx\bx'}=\exp\left(-\frac{\|\bx\|^2+\|\bx'\|^2}{\sigma^2}-2\fF^{-1}\left(\sum_{c}\bX_c\odot\overline{\bX}'_c\right)\right).
	\label{equ:mkernel}
	\end{equation}
	This formulation renders an efficient solution to incorporate different types of multi-channel features. 
	In this work, we use two complementary types of local statistical features for learning correlation filters: (1) histogram of oriented gradients (HOG) and (2) histogram of local intensities (HOI). 
	
	{\flushleft \bf Histogram of Oriented Gradients (HOG).} 
	As one of the most widely used feature descriptors for object detection \cite{DBLP:conf/cvpr/DalalT05}, the main idea of the HOG descriptor is to encode the local object appearance and shape by the distribution of oriented gradients.
	Each image patch is divided into small regions (cells), in which a histogram of oriented gradients is computed to form the descriptor. 
	These gradient-based descriptors are robust to illumination variation and local shape deformation and have been used to help learn discriminative correlation filters for object tracking \cite{DBLP:journals/pami/HenriquesC0B15,DBLP:conf/bmvc/DanelljanKFW14}.
	
	{\flushleft \bf Histogram of Local Intensities (HOI).}
	We observe that correlation filters learned from only HOG features 
	are not effective in handling cases where the images are heavily blurred, e.g., caused by abrupt motion, blurring, or illumination changes. 
	We attribute the performance degradation to the fact that intensity gradients are no longer discriminative when representing target objects undergoing abrupt motion. 
	In such cases, we observe that the intensity values between the target and its background remain distinctive. 
	However, learning correlation filters directly over intensities does not perform well \cite{DBLP:conf/eccv/HenriquesCMB12,DBLP:conf/eccv/ZhangZLZY14} (see Section \ref*{sec:overallperformance}) as intensity values are not robust to appearance variation, e.g., caused by deformation.
	We thus propose to compute the histogram of local intensities as a complementary of HOG features. 
	The proposed HOI features bear some resemblance to the distribution field 
	scheme \cite{DBLP:conf/cvpr/Sevilla-LaraL12,6755887} where the statistical properties 
	of pixel intensities are exploited as features.
	In contrast to computing the statistical properties over the whole image \cite{DBLP:conf/cvpr/Sevilla-LaraL12,6755887}, we use the histograms of local patches. 
	These local statistical features are more robust to appearance changes than pixel intensities. 
	In this work, we compute the histogram in a $6\times6$ local cell and quantize the intensity values into 8 bins. 
	To enhance the robustness to drastic illumination variation, we compute the HOI feature descriptors not only on the intensity channel but also on a transformed channel by applying a non-parametric local rank transformation \cite{Zabih:1994:NLT:200241.200258} to the intensity channel. 
	The intensity-based features (HOI) and the gradient-based features (HOG) are complementary to each other as they capture different aspects of target appearance. 
	We compute these two types of features using local histograms with the same spatial resolution (but with different channels). 
	Thus, the HOG and HOI features can be easily integrated into the kernelized correlation filters using \eqref{equ:mkernel}.

	%
	
	{\flushleft \bf Deep Convolutional Features.} 
	Deep features learned from  
	convolutional neural networks (CNNs) have recently been applied to 
	numerous vision problems~\cite{DBLP:conf/nips/KrizhevskySH12,DBLP:conf/cvpr/GirshickDDM14,DBLP:conf/cvpr/LongSD15}. 
	In this work, we exploit deep CNN features to complement 
	the proposed multi-channel HOG and HOI features for learning correlation filters.
	Specifically, we use the features from the last convolutional layer (\textit{conv5-4}) of the VGGNet-19 \cite{DBLP:journals/corr/SimonyanZ14a} as in \cite{DBLP:conf/iccv/MaHYY15}.
	Compared to the handcrafted HOI and HOG features that capture fine-grained spatial details of a target object, deep CNN features encode hierarchical and high-level semantic information 
	that is robust to significant appearance changes over time. 
	Our goal is to fully exploit the merits of both handcrafted and deep features. 
	Instead of learning a single correlation filter over 
	the concatenated features, we learn one correlation filter for each type of features.
	We infer target object location by combining the response maps from the two correlation filters.
	
	Given the correlation response maps $\bbf^h$ and $\bbf^d$ using handcrafted and deep features in \eqref{equ:getres}, we denote the probability of position $(i,j)\in\{1,2
	\ldots, M\}\times\{1,2, \ldots, N\}$ to be the center of the target by a distribution $f_{ij}^l$, where $\sum_{ij} f_{ij}^l=1$ and $l\in\{h,d\}$. 
	We determine the optimal distribution $\mathbf{q}$ by minimizing the Kullback-Leibler (KL) divergence of each response map $\bbf^l$ and $\mathbf{q}$ as follows:
	\begin{align}
	\label{equ:search}
	\argmax_{\mathbf{q}} ~ & ~ \sum_{l\in\{h,d\}} D_{KL}(\bbf^l\|\mathbf{q}), \\ 
	\text{s.t.} ~~ & ~ \sum_{i,j} q_{ij}=1, \nonumber 
	\end{align}
	where $f_{ij}$ and $q_{ij}$ are the elements of $\bbf$ and $\mathbf{q}$ at position $(i,j)$. 
	The KL divergence is defined as:
	\begin{equation}
	D_{KL}(\bbf^l\|\mathbf{q})=\sum f_{ij}^l\log\frac{f_{ij}^l}{q_{ij}}.
	\end{equation}
	Using the Lagrange multiplier method, the solution of $\mathbf{q}$ in \eqref{equ:search} is:
	\begin{equation}
	\mathbf{q}=\frac{\bbf^h\oplus \bbf^d}{2},
	\end{equation} 
	where $\oplus$ means element-wise addition. 
	An intuitive explanation is that the final probability distribution of object location $\mathbf{q}$ is the average of the response maps $\bbf^h$ and $\bbf^d$. 
	We locate the target object based on the maximum value of $\mathbf{q}$. 
	
\begin{figure}
		\centering
		\setlength{\tabcolsep}{1pt}
		\begin{tabular}{cc}
			\includegraphics[width=.23\textwidth]{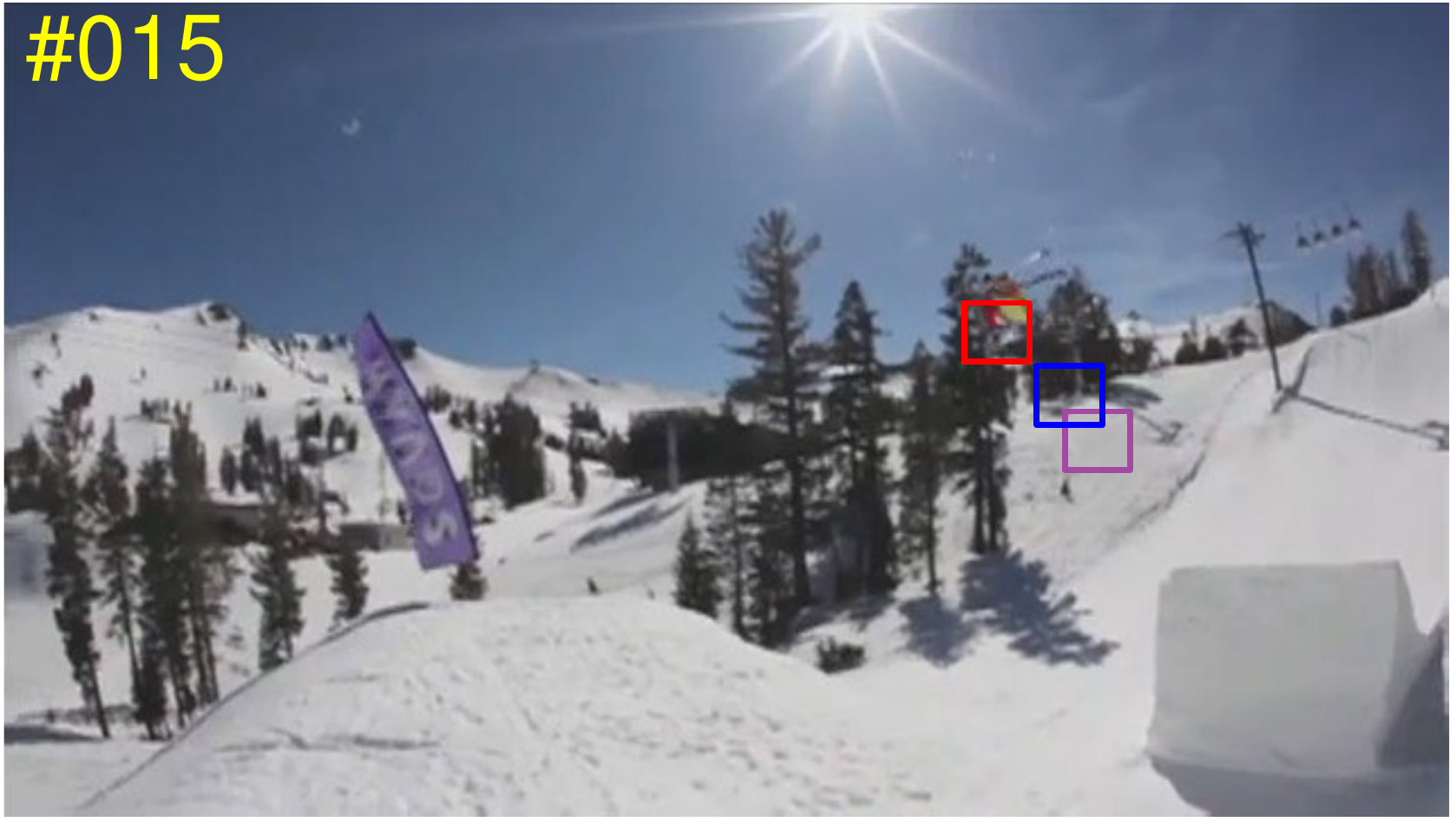} &
			\includegraphics[width=.23\textwidth]{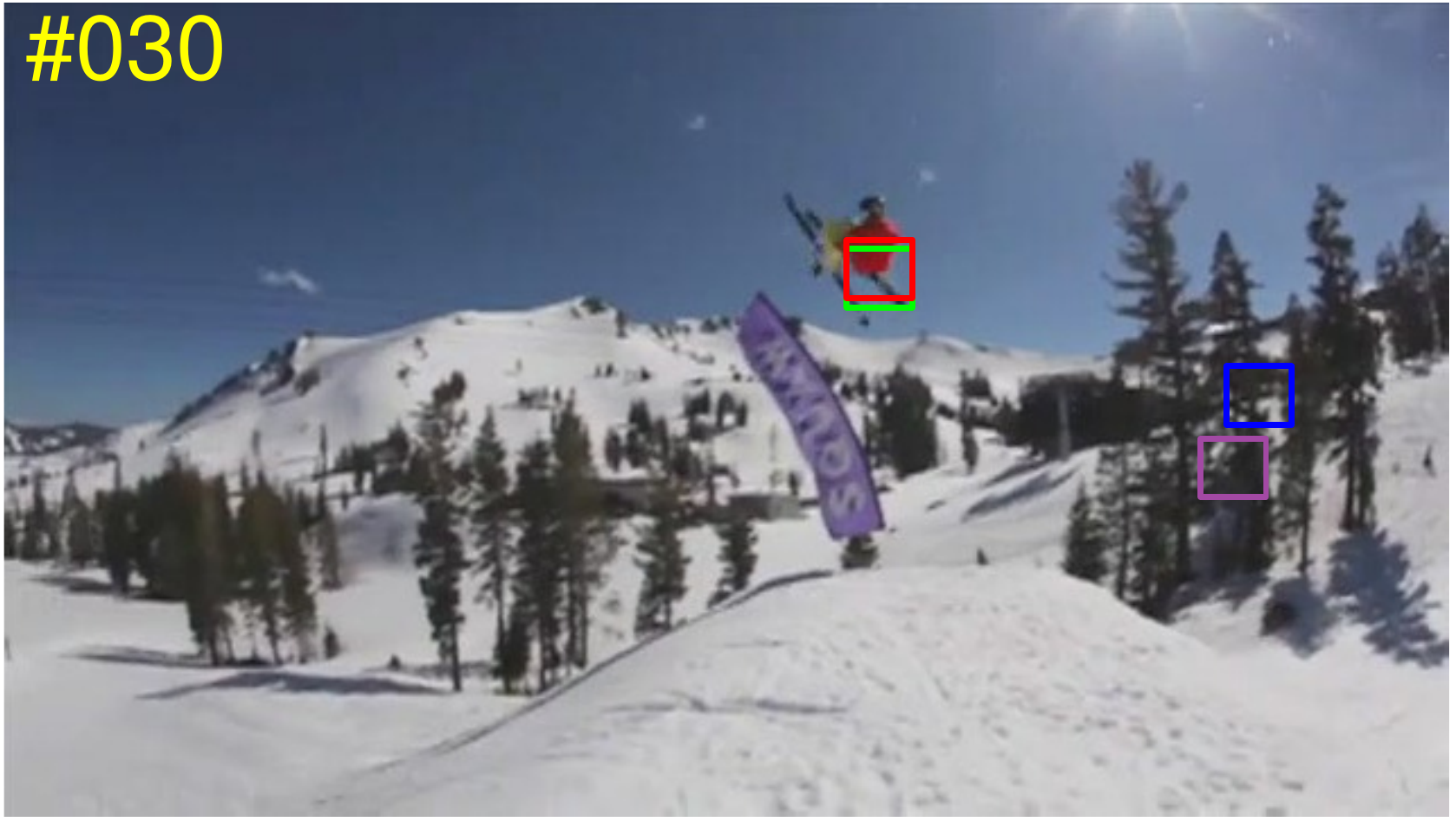} \\
			\includegraphics[width=.23\textwidth]{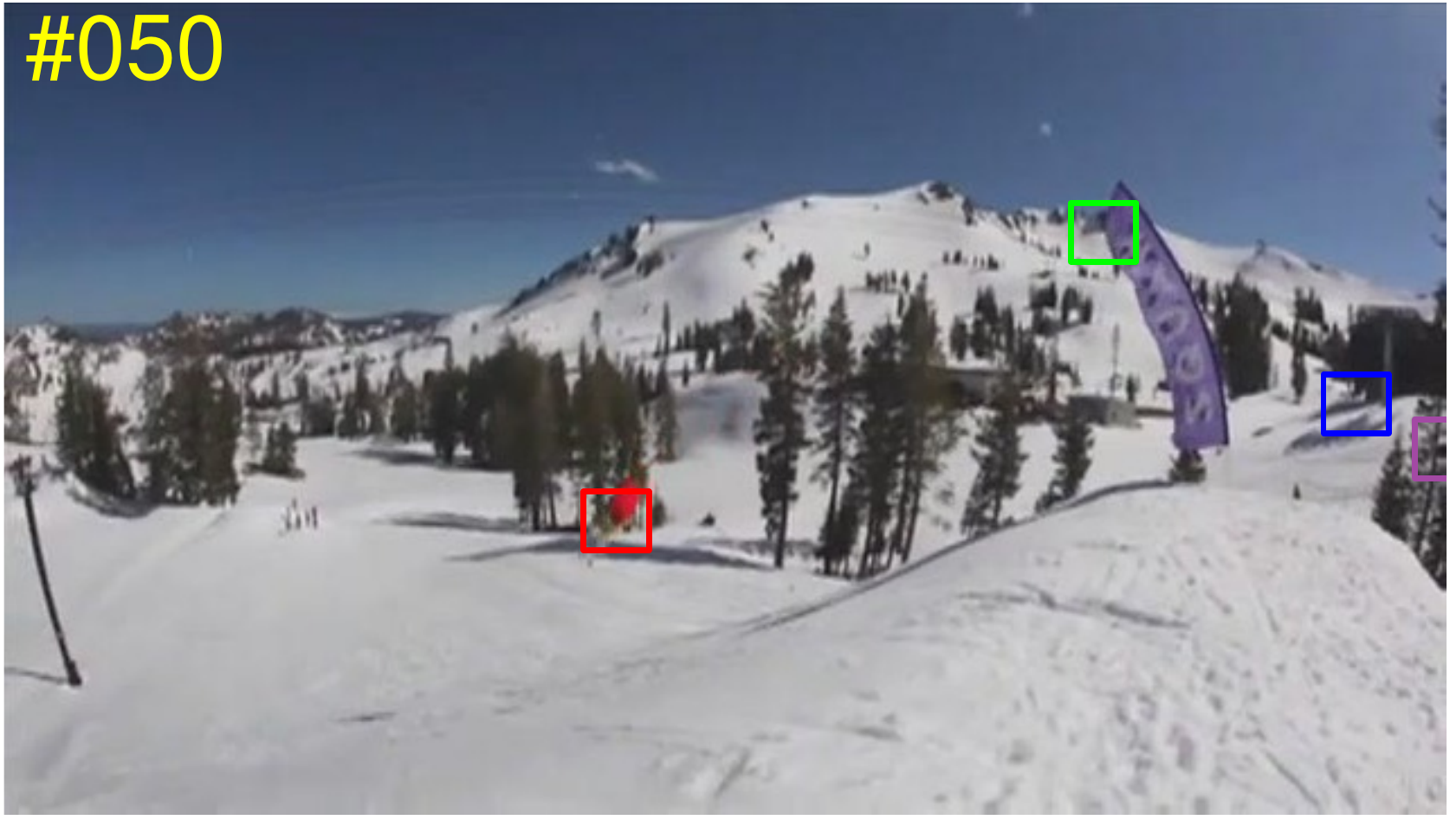} &
			\includegraphics[width=.23\textwidth]{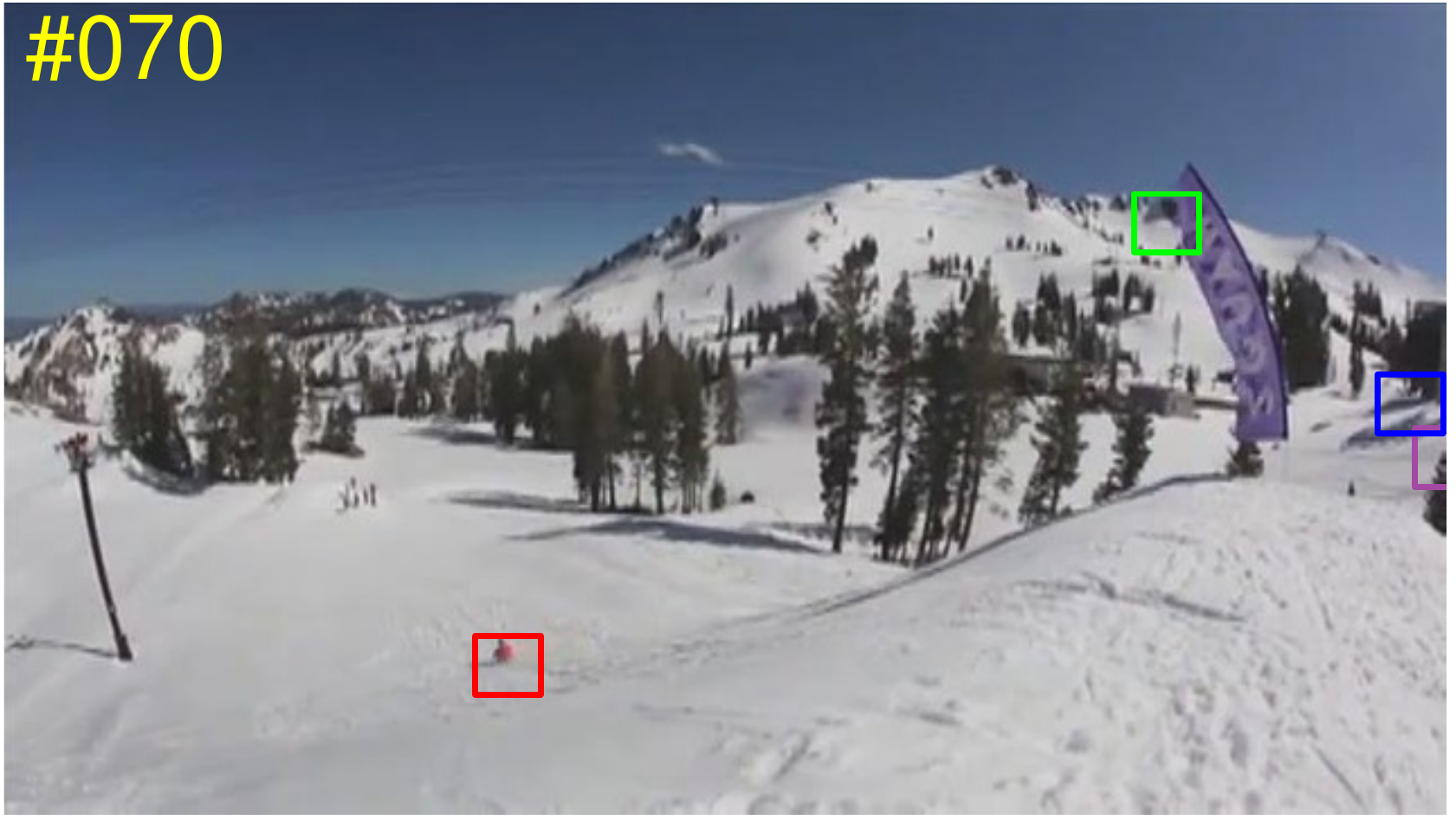}\\ 
		\end{tabular}
		\includegraphics[width=.4\textwidth]{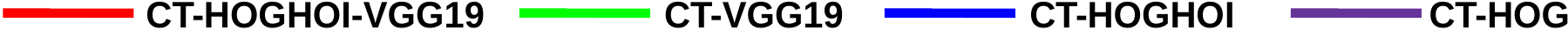}
		\caption{ 
			\textbf{Effectiveness of using both deep and handcrafted features.}
			Qualitative results on the \textit{skiing} sequence \cite{DBLP:conf/cvpr/WuLY13} using different types of features for learning the translation filter. 
			The CT-HOGHOI-VGG19 method effectively exploits the advantages of both deep and handcrafted features, and tracks the target object over the entire sequence.} 
		\label{fig:skiing}
	\end{figure}

\begin{figure}
		\centering
		\includegraphics[width=.44\textwidth]{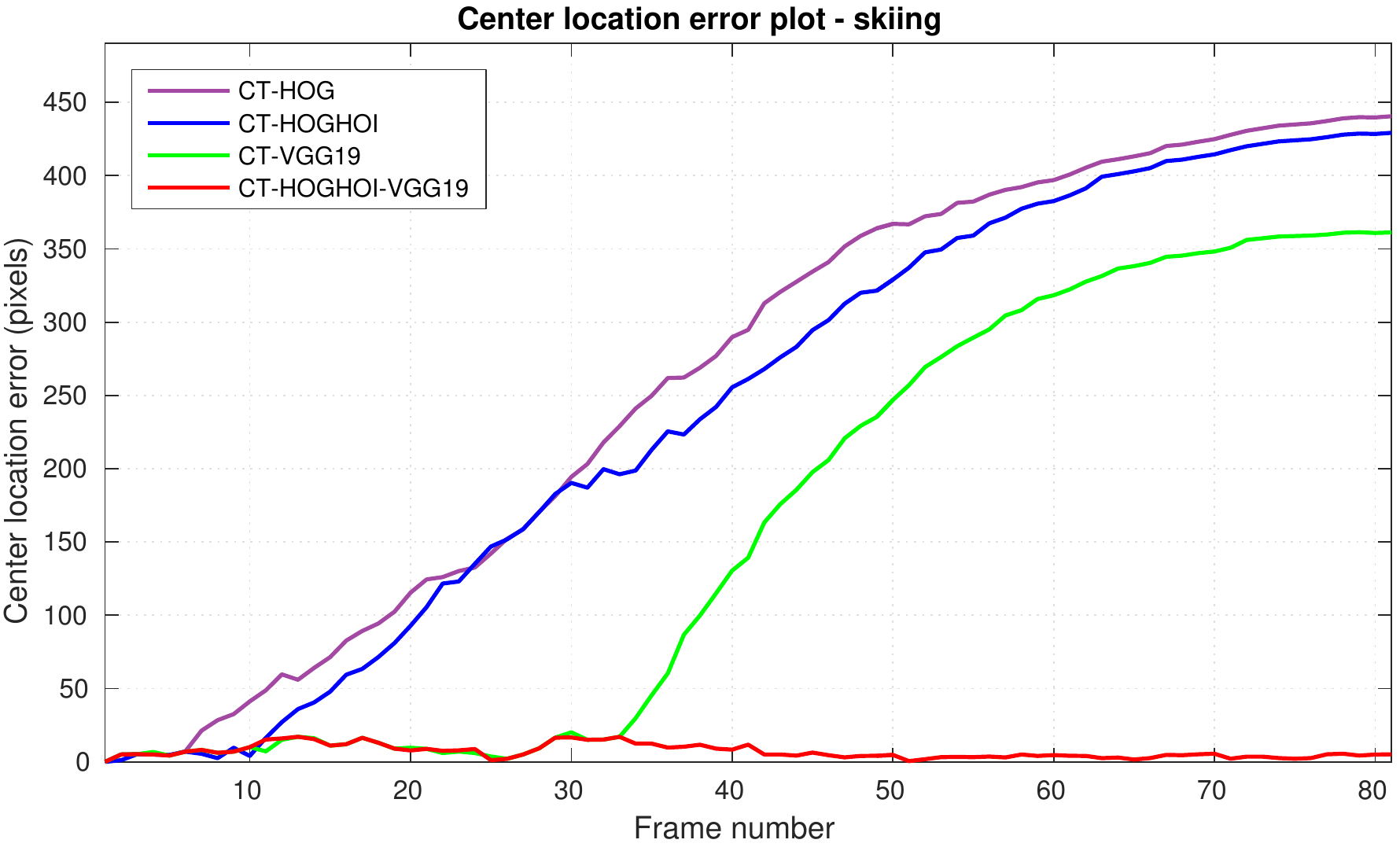}
		\caption{ 
			\textbf{Comparison of different features for precise localization.}
			We plot center location error on the \textit{skiing} sequence \cite{DBLP:conf/cvpr/WuLY13}. 
			Compared with other variants, the CT-HOGHOI-VGG19 method exploits both deep and handcrafted features, and achieves smallest center location error.
		}
		\label{fig:skiingcle}
	\end{figure}

	To demonstrate the effectiveness of exploiting both the deep (conv5-4 in VGGNet-19) and handcrafted (HOG-HOI) features, we show quantitative comparisons of the tracking results using four different types of features on the \emph{skiing} sequence in Figure \ref{fig:skiing}.
	Figure \ref{fig:skiingcle} shows the center location error plot over the entire sequence. 
	Note that the CT-HOGHOI-VGG19 approach exploits the merits of both deep and handcrafted features effectively, and thus successfully track the target skier over the entire sequence.
	On the other hand, other alternative approaches (using either deep or handcrafted features) do not achieve satisfactory results as shown by the large center location error in Figure \ref{fig:skiingcle}.

	\begin{figure*}
		\centering
		\small
		\setlength{\tabcolsep}{0mm}
		\begin{tabular}{cccc}
			\includegraphics[width=.24\textwidth]{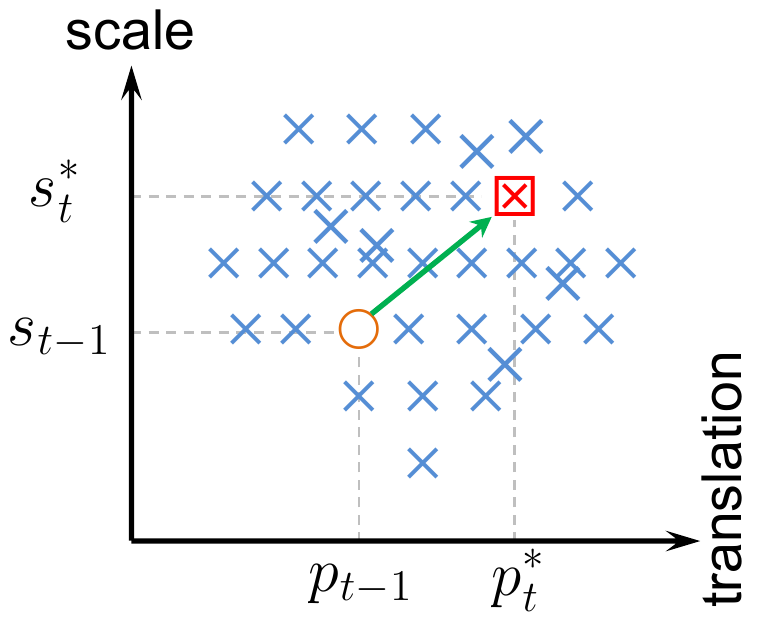}&
			\includegraphics[width=.24\textwidth]{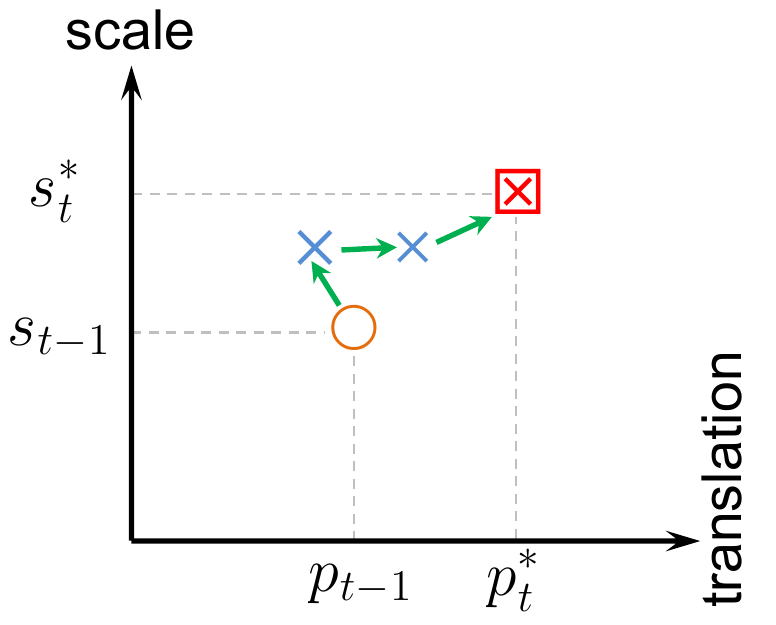}&
			\includegraphics[width=.24\textwidth]{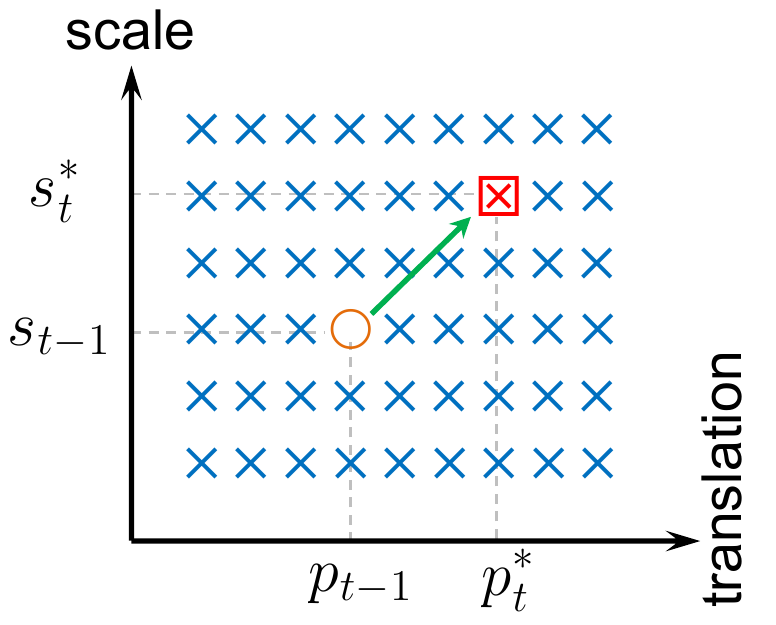}&
			\includegraphics[width=.24\textwidth]{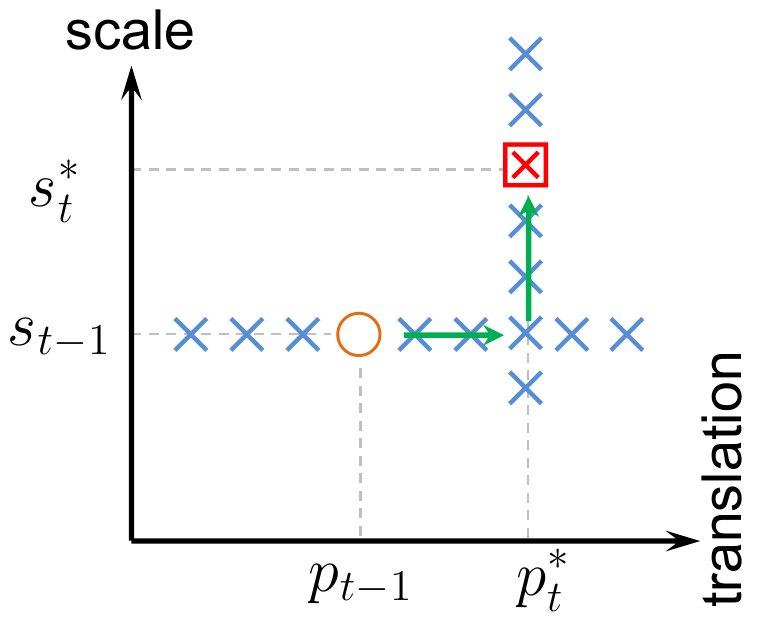} \\
			(a) Particle filter & (b) Gradient descent & (c) Exhaustive search & (d) Ours \\
		\end{tabular}
		\caption{
			\textbf{Four main schemes for state estimation in object tracking.} 
			The symbols $\color{orange}\ocircle$, $\color{blue}\times$ and $\color{red}\XBox$ denote the current, sampled and optimal states, respectively. 
			(a) The particle filter scheme such as \cite{DBLP:journals/tsp/ArulampalamMGC02} draws random samples (particles) to approximate the joint distribution of target states. 
			(b) The gradient descent method such as Lucas-Kanade \cite{DBLP:conf/ijcai/LucasK81} iteratively infers the local optimal translation and scale changes jointly. 
			(c) The exhaustive search scheme evaluates all possible states in a brutal force manner.
			(d) Our method first estimates the translation change first and then predicts scale change in a similar spirit to coordinate descent optimization. 
		}
		\label{fig:frame}
	\end{figure*}

	\subsection{Scale Regression Model via Correlation Filters}
	\label{sec:scale}
	We construct the feature pyramid of target appearance centered at the estimated location to train a scale regression model using correlation filters.
	Danelljan et al. \cite{DBLP:conf/bmvc/DanelljanKFW14} also learn a discriminative correlation filter for scale estimation. 
	Our method differs from \cite{DBLP:conf/bmvc/DanelljanKFW14} in that we \emph{do not} use the predicted scale changes to update the translation filter. 
	Let $W\times H$ be the target size and $N$ indicate the number of scales $S=\{\alpha^n|n=
	\lfloor- \frac{N-1}{2}\rfloor,\lfloor- \frac{N-3}{2}\rfloor, \ldots, \lfloor\frac{N-1}{2}\rfloor\}$, where $\alpha$ is a scaling factor, e.g., $\alpha=1.03$. 
	For each scale $s\in S$, we crop an image patch of size $sW\times sH$ centered at the estimated location. 
	We resize all these cropped patches to an uniform size of $W\times H$, and then extract HOG features from each sampled patch to form a feature pyramid containing the multi-scale representation of the target. 

	Let $\bx_s$ denote the target features in the $s$-th scale. 
	We assign $\bx_s$ with a regression target score $y_s=\exp(-\frac{(s-N/2)^2}{2\sigma_0^2})$. 
	As $\{y_s\}$ is one dimensional, we vectorize $\bx_s$ before applying \eqref{equ:A} to learn correlation filters. 
	With the use of vectorization, the output response score of \eqref{equ:getres} is a scalar indicating the similarity between $\bx_s$ and the learned filter. 
	To mitigate the ambiguity, we denote this output scalar from \eqref{equ:getres} as $g(\bx_s)$. 
	The optimal scale $s^*$ can then be inferred by:
	\begin{equation}
	s^*=\argmax_s\{g(\bx_s)\ | \ s\in S\}.
	\label{equ:scale}
	\end{equation}
	
	In Figure \ref{fig:frame}, we compare four different schemes for estimating the target states 
	in terms of translation and scale.
	As shown in Figure \ref{fig:frame}(d), our approach first estimates the translation, and then predicts the scale change in a spirit similar to the coordinate descent optimization.
	Our approach differs from several existing tracking schemes that jointly infer the translation and scale changes.
	For example, particle filter based tracking algorithms such as \cite{DBLP:journals/tsp/ArulampalamMGC02} draw random samples to approximate the distribution of target states containing translation and scale changes (Figure \ref{fig:frame}(a)). 
	The gradient descent method (e.g., Lucas-Kanade \cite{DBLP:conf/ijcai/LucasK81}) 
	iteratively infers the local optimal translation and scale changes (Figure \ref{fig:frame}(b)). 
	Although it is suboptimal to decompose the tracking task into two independent sub-tasks (i.e., translation and scale estimation) as shown in Figure \ref{fig:frame}(d), 
	our tracker not only alleviates the burden of densely evaluating the target states, but also avoids the noisy updates on the translation filter in case of inaccurate scale estimation. 
	We note that the DSST \cite{DBLP:conf/bmvc/DanelljanKFW14} also decomposes the tracking task into the translation and scale estimation sub-tasks. 
	Our approach differs the DSST in that we update the translation filter using the ground-truth scale in the first frame rather than the estimated scale in each frame. 
	We further alleviate the degradation of translation filter caused by inaccurate scale estimation.
	Experimental results (see Figure \ref{fig:component-scale} and Section \ref{sec:experiment}) show that our tracker significantly outperforms an alternative implementation (CT-HOGHOI-joint-scale-HOG), where the estimated scale change in each frame is used to update the translation filter.

	\subsection{Long-Term and Short-Term Memory}
	\label{sec:lstm}
	To adapt to appearance changes, we incrementally update the learned correlation filters over time.
	Since it is computationally expensive to update filters by directly minimizing the output errors over all tracked results \cite{DBLP:conf/cvpr/BoddetiKK13,DBLP:conf/iccv/GaloogahiSL13}, we use a moving average scheme for updating a single filter as follows:
	\begin{subequations}
		\label{equ:update}
		\begin{align}
		\tilde{\bx}^t&=(1-\eta)\tilde{\bx}^{t-1}+\eta\bx^t, \\
		\tilde{\mathbf{a}}^t&=(1-\eta)\tilde{\mathbf{a}}^{t-1}+\eta\mathbf{a}^t,
		\end{align}
	\end{subequations}
	where $t$ is the frame index and $\eta\in(0,1)$ is a learning rate. 
	This approach updates the filter at each frame and emphasizes the importance of model adaptivity with the short-term memory of target appearance. 
	Due to the effectiveness of this scheme in handling appearance changes, tracking algorithms \cite{DBLP:journals/pami/HenriquesC0B15,DBLP:conf/bmvc/DanelljanKFW14} achieve favorable performance in recent benchmark studies \cite{DBLP:conf/cvpr/WuLY13,Hadfield14d}. 
	However, these trackers are prone to drift when the training samples are noisy and unable to recover from tracking failures due to the lack of the long-term memory of target appearance. 
	In other words, the update scheme in \eqref{equ:update} assumes the tracked result in each frame is accurate. 
	In this work, we propose to learn a long-term filter $\mathcal{A}_\mathrm{L}$ to address this problem.
	As the output response map of the long-term filter $\mathcal{A}_\mathrm{L}$ in \eqref{equ:getres} is two-dimensional, we take the maximum value of the response map as the confidence score.
	To capture the long-term memory of target appearance for determining if tracking failures occur, we set a stability threshold $T_s$ to conservatively update the long-term filter $\mathcal{A}_\mathrm{L}$ for maintaining the model stability. 
	We update the filter using \eqref{equ:update} only if the confidence score, $\max\bigl(\bbf(\bz)\bigr)$, of the tracked object $\bz$ exceeds the stability threshold $T_s$. 
	Compared to the methods \cite{DBLP:conf/cvpr/SantnerLSPB10,DBLP:journals/tip/ZhongLY14} that use only the first frame as the long-term memory of target appearance, our long-term filter can adapt to appearance variation over a long time span.

	An alternative approach to maintain long-term memory of target appearance is to directly learn a long short-term memory (LSTM) network \cite{hochreiter1997long} from training sequences off-line. 
	%
	%
	We follow the project \cite{rolo} ({\small\url{https://github.com/Guanghan/ROLO}}) to implement a baseline tracker using the standard LSTM cell.
	We interpret the hidden state of the LSTM as the counterpart of the long-term correlation filter. 
	We use all the sequences on the VOT datasets \cite{Hadfield14d,DBLP:conf/iccvw/KristanMLFCFVHN15} as training data (excluding the overlapped sequences on the OTB2015 dataset \cite{DBLP:journals/pami/WuLY15}). 
	For each input feature vector $\bx_t$, the output cell state $\mathbf{c}_t$ and hidden state $\bh_t$ are: 
	\begin{eqnarray}
	\hat{\mathbf{g}}_f,\hat{\mathbf{g}}_i,\hat{\mathbf{g}}_o,\hat{\mathbf{u}}&=&\mathbf{W}^{xh}\bx_t+\mathbf{W}^{hh}\bh_{t-1}+\bb^h \\
	\mathbf{g}_f&=&\sigma(\hat{\mathbf{g}}_f) \\
	\mathbf{g}_i&=&\sigma(\hat{\mathbf{g}}_i) \\
	\mathbf{g}_o&=&\sigma(\hat{\mathbf{g}}_o) \\ \mathbf{u}&=&\tanh(\hat{\mathbf{u}}) \\
	\mathbf{c}_t&=&\mathbf{g}_f\odot\mathbf{c}_{t-1}+\mathbf{g}_i\odot\mathbf{u} \\
	\bh_t&=&\mathbf{g}_o\odot\tanh(\mathbf{c}_t)
	\end{eqnarray}
	where $\hat{\mathbf{g}}_f$, $\hat{\mathbf{g}}_i$, $\hat{\mathbf{g}}_o$ are the forget gates, input gates and output gates, respectively. 
	We denote $\bb$ as the hidden state biases, $\sigma(x)$ as the sigmoid function of $x$, $\frac{1}{1+e^{-x}}$, and $\odot$ as the element-wise multiplication. 
	Here, $\mathbf{W}^{xh}$ are the weights from the input $\bx_t$ to the hidden state, and $\mathbf{W}^{hh}$ are the weights between hidden states and are connected through time. 
	For simplicity, we set the size of the hidden state $\bh_t$ to be equal to the size of the input $\bx_t$. 
	We use the hidden state $\bh_t$ as the long-term filter to compute the correlation response map.

\begin{figure}
		\centering
		\includegraphics[width=.42\textwidth]{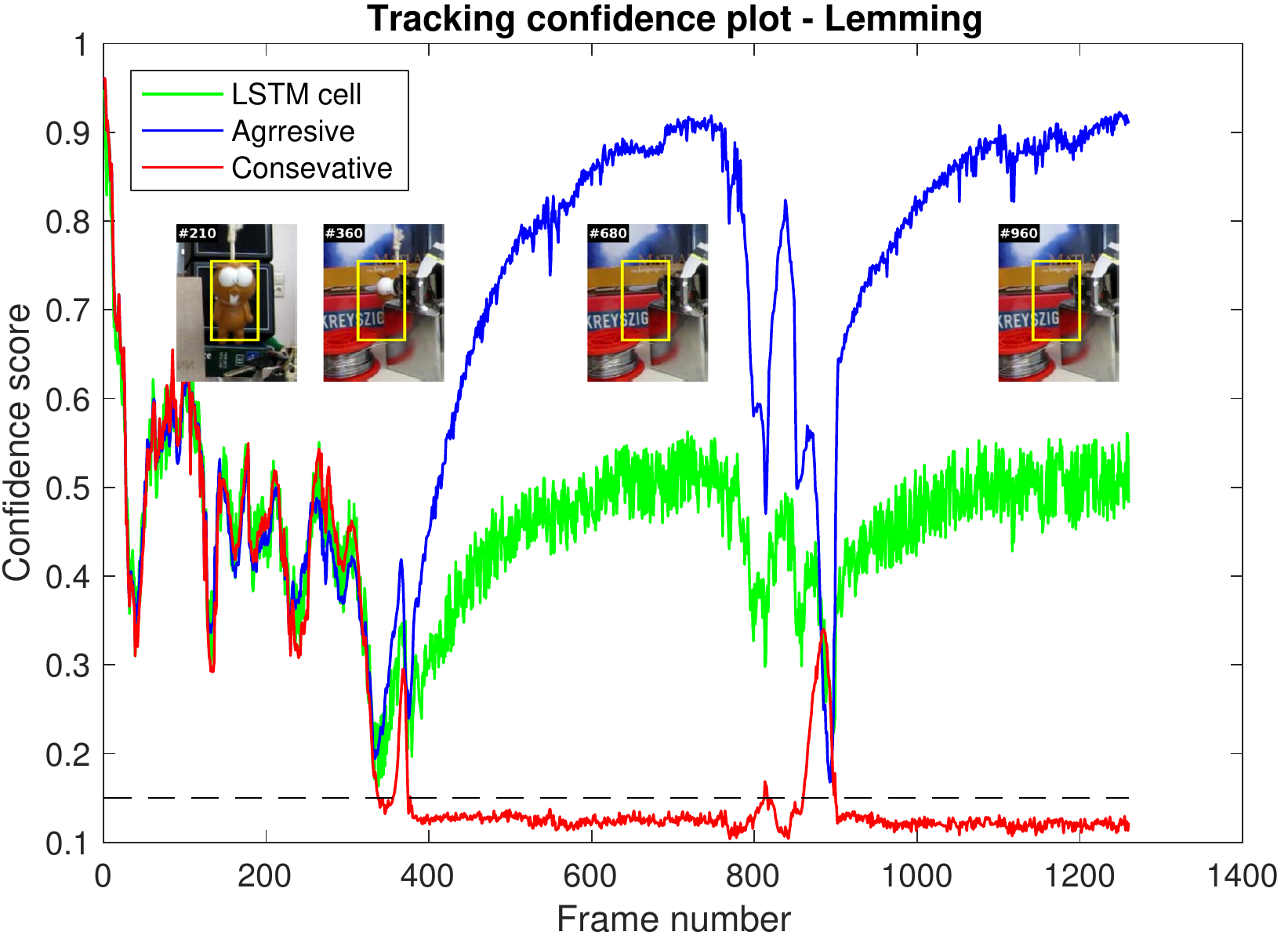}
		\caption{
			\textbf{Effect of long-term filters.}
			On the \textit{lemming} sequence,
			we use three different long-term filters to compute the confidence scores of the tracking results (yellow boxes) using the baseline CT-HOGHOI method \emph{without} incorporating a re-detection module.
			%
			%
			%
			%
			The conservatively updated correlation filter performs well in determining tracking failures as the confidence scores are generally below the threshold 0.15 after the 360-th frame.
		}
		\label{fig:confidence-nodetection}
	\end{figure}

	Using \textit{lemming} \cite{DBLP:conf/cvpr/WuLY13} sequence as an example, 
	we compare three types of long-term filters to compute the tracking confidence scores of the baseline CT-HOGHOI method in Figure \ref{fig:confidence-nodetection}.
	The baseline algorithm does not incorporate a re-detection module and thus fails to track the target after the 360-th frame. 
	The aggressively updated correlation filter and the LSTM hidden states gradually degrade due to noisy updates and cannot predict the tracking failures.
	In contrast, the conservatively learned correlation filter accurately predicts tracking failures (the confidence scores are generally below 0.15).
	
	\subsection{Three Types of Correlation Filters}
	For translation estimation, we enlarge the input bounding boxes of target objects to incorporate surrounding context to provide substantially more shift positions. 
	Compared to the tracking methods based on online classifiers \cite{DBLP:conf/cvpr/DinhVM11,DBLP:conf/eccv/ZhangGLA12,DBLP:journals/ijcv/Zhang0AYG15} that learn from \emph{sparse} samples (randomly drawn around the estimated target position), our approach based on correlation filters learns from \emph{dense} samples, i.e., all the circularly shifted versions of input features (see Section \ref{sec:tracking}). 
	The increase of training data facilitates discriminating the target from its background.
	For learning the scale filter and long-term filter , we do not incorporate contextual cues as the surrounding context often changes drastically over time and may adversely affect both the scale filter and the long-term filter. 
	Figure \ref{fig:tworeg} shows the three different correlation filters 
	with the update schemes, context size, and feature type. 
	We refer the readers to the ablation studies in Section \ref{sec:ablation} for justification of the design choices of the filters.
	Here, we summarize the three different correlation filters as follows:
	\begin{itemize} 
		\item The translation filter $\mathcal{A}_\mathrm{T}$ captures a short-term memory of target appearance. We exploit surrounding context information for learning the filter $\mathcal{A}_\mathrm{T}$. 
		To fully exploit the semantics within deep features and the fine-grained spatial details within hand-crafted features, we learn translation filters over deep and handcrafted features, respectively. 
		To alleviate the boundary discontinuities caused by the circular shifts, we use a two-dimensional cosine window to weight each channel of the input features.
		\item We learn the scale filter $\mathcal{A}_\mathrm{S}$ using HOG features only. We empirically find that adding HOI features does not improve the accuracy of scale estimation (See Figure \ref{fig:component-scale}). 
		Unlike the translation filter $\mathcal{A}_\mathrm{T}$, we extract the features directly from the target region without incorporating the surrounding contexts as they do not provide information about the scale changes of the target.
		\item We learn the long-term filter $\mathcal{A}_\mathrm{L}$ using a conservative learning rate to maintain the long-term memory of target appearance for determining the tracking failures. We learn the filter $\mathcal{A}_\mathrm{L}$ using both HOG and HOI features.
	\end{itemize}

	\subsection{Online Detector}
	\label{sec:svm}
	A robust tracking algorithm requires a detection module to recover the target from potential tracking failures caused by heavy occlusion or out-of-view movement. 
	For each tracked target $\bz$, we compute its confidence score as $C_L=\max\bigl(\bbf_{\mathcal{A}_\mathrm{L}}(\bz)\bigr)$ using the long-term filter $\mathcal{A}_\mathrm{L}$. 
	Unlike previous trackers \cite{DBLP:conf/eccv/Pernici12,DBLP:conf/cvpr/SupancicR13,DBLP:conf/eccv/HuaAS14} carrying out the detection at every frame, we activate the detector only if the confidence score $C_L$ is below a predefined re-detection threshold $T_r$.
	The main goal of using $T_r$ is to reduce the computational load by avoiding the sliding-window detection in each frame.
	For efficiency, we use an online SVM classifier as the detector rather than using the long-term filter $\mathcal{A}_\mathrm{L}$. 
	We incrementally train the SVM classifier by drawing dense training samples around the estimated position and scale change and assigning these samples with binary labels according to their overlap ratios similar to \cite{DBLP:conf/eccv/ZhangMS14}.
	In this work, we only take the translated samples for training to further reduce the computational burden.
	We use quantized color histogram as our feature representation where each channel in the CIE LAB space is quantized into four bins as in \cite{DBLP:conf/eccv/ZhangMS14}. 
	To improve robustness to dramatic illumination variation, we apply the non-parametric local rank transformation \cite{Zabih:1994:NLT:200241.200258} to the L channel.
	Given a training set $\{(\bv_i,c_i)|i=1,2,\ldots,N\}$ with $N$ samples in a frame, where $\bv_i$ is the feature vector generated by the $i$-th sample and $c_i\in\{+1,-1\}$ is the class label. 
	The objective function of solving the hyperplane $\bh$ of the SVM detector is
	\begin{eqnarray}
	\min_{\bh}\frac{\lambda}{2}\|\bh\|^2+\frac{1}{N}\sum_i\ell\bigl(\bh; (\bv_i,c_i)\bigr) \\
	\text{where} \quad \ell\bigl(\bh;(\bv,c)\bigr)=\max\{0,1-c\langle\bh,\bv\rangle\}. \nonumber
	\end{eqnarray} 
	The notation $\langle\bh,\bv\rangle$ indicates the inner product between $\bh$ and $\bv$.
	Unlike existing methods \cite{DBLP:journals/pami/Avidan04,DBLP:conf/eccv/ZhangMS14} that maintain a large pool of supported vectors for incremental update, we apply the passive-aggressive algorithm \cite{DBLP:journals/jmlr/CrammerDKSS06} to update the hyperplane parameters efficiently.
	\begin{equation}
	\bh\leftarrow\bh-\frac{\ell\bigl(\bh;(\bv,c)\bigr)}{\|\nabla_\bh\ell\bigl(\bh;(\bv,c)\bigr)\|^2+\frac{1}{2\tau}}\nabla_\bh\ell\bigl(\bh;(\bv,c)\bigr),
	\label{equ:svmupdate}
	\end{equation}
	where $\nabla_\bh\ell\bigl(\bh;(\bv,c)\bigr)$ is the gradient of the loss function in terms of $\bh$ and $\tau\in(0,+\infty)$ is a hyper-parameter that controls the update rate of $\bh$. 
	Similar to the long-term filter, we update the classifier parameters using \eqref{equ:svmupdate} only when the confidence score $C_L$ is above the threshold $T_s$.
	
	\begin{algorithm}
		\SetKwData{Left}{left}\SetKwData{This}{this}\SetKwData{Up}{up}
		\SetKwInOut{Input}{Input}\SetKwInOut{Output}{Output}
		\Input{Initial bounding box $\mathbf{b}_{t-1}=(x_{t-1}, y_{t-1},s_{t-1})$, $\mathcal{A}_\mathrm{T}$, $\mathcal{A}_\mathrm{S}$, $\mathcal{A}_\mathrm{L}$, $\bh$}
		\Output{Estimated bounding box $\mathbf{b}_t = (x_t, y_t,s_t)$}
		\BlankLine
		\Repeat{end of image sequence}{
			Crop out the image patch $\bz$ centered at ($x_{t-1},y_{t-1}$) and extract HOG and HOI features\; 
			\tcp{Translation estimation}
			Compute $\bbf_{\mathcal{A}_\mathrm{T}}(\bz)$ and estimate target position ($x_t, y_t$)\;
			\tcp{Scale estimation}
			Construct scale pyramid $\bz'$ around ($x_t, y_t$) and compute $\bbf_{\mathcal{A}_\mathrm{S}}(\bz')$ to infer $s_t$\;
			Crop out patch $\bz$ centered at $(x_t,y_t)$\;
			\tcp{Re-detection}
			\If{$\max(\bbf_{\mathcal{A}_\mathrm{L}}(\bz))<T_r$}{\label{ut}
				Activate detection module $\bh$ and find the candidate bounding boxes $\mathbf{B}$ with positive labels\;
				\lForEach{state $\mathbf{b}'$ in $\mathbf{B}$}{computing $\bbf_{\mathcal{A}_\mathrm{L}}(\mathbf{b'})$}
				\lIf{$\max{(\bbf_{\mathcal{A}_\mathrm{L}}(\mathbf{b'}))}>T_a$}{$\mathbf{b}_t=\mathbf{b}'$}
			}
			\tcp{Model update}
			Update $\mathcal{A}_\mathrm{T}$ and $\mathcal{A}_\mathrm{S}$\;
			\If{$\max(\bbf_{\mathcal{A}_\mathrm{L}}(\bz))>T_s$}{Update $\mathcal{A}_\mathrm{L}$ and $\bh$;}
		}
		\caption{Outline of the proposed tracking algorithm.}
		\label{algorithm}
	\end{algorithm}
	
	\section{Implementation Details}
	Figure \ref{fig:flowchart} presents the main steps of the proposed tracking algorithm. 
	We learn three types of correlation filters ($\mathcal{A}_\mathrm{T}$, $\mathcal{A}_\mathrm{S}$, $\mathcal{A}_\mathrm{L}$) for translation estimation, scale estimation, and capturing the long-term memory of target appearance, respectively. 
	We also build a re-detection module using SVM for recovering targets from tracking failures. 
	
	Algorithm \ref{algorithm} summarizes the proposed tracker. 
	Our translation filter $\mathcal{A}_\mathrm{T}$ separates the target object from the background by incorporating the contextual cues. 
	Existing methods \cite{DBLP:journals/pami/HenriquesC0B15,DBLP:conf/bmvc/DanelljanKFW14} use an enlarged target bounding box by a fixed ratio $r=2.5$ to incorporate the surrounding context.
	Our experimental analysis (see Section \ref{sec:pa}) shows that slightly increasing the context area leads to improved results. 
	We set the value of $r$ to a larger ratio of 2.8. 
	We also take the aspect ratio of target bounding box into consideration. 
	We observe that when the target (e.g., pedestrian) with a small aspect ratio, a smaller value for $r$ can decrease the unnecessary context area in the vertical direction. 
	To this end, we reduce the ratio $r$ by one-half in the vertical direction when the target with aspect ratio smaller than 0.5 (see Section \ref{sec:pa} for more detailed analysis).
	
	For training the SVM detector, we densely draw samples from a window centered at the estimated location. 
	We assign these samples with positive labels when their overlap ratios with the target bounding box are above 0.5, and assign them with negative labels when their overlap ratios are below 0.1. 
	We set the re-detection threshold $T_r = 0.15$ for activating the detection module and the acceptance threshold $T_a = 0.38$ for adopting the detection results. 
	These settings suggest that we conservatively adopt each detection result, as it would relocate the targets and reinitialize the tracking process. 
	We set the stability threshold $T_s$ to 0.38 to conservatively update the filter $\mathcal{A}_\mathrm{L}$ for maintaining the long-term memory of target appearance. 
	Note that all these threshold values compare with the confidence scores computed by the long-term filter $\mathcal{A}_\mathrm{L}$. 
	The regularization parameter of \eqref{equ:filter} is set to $\lambda=10^{-4}$. 
	The Gaussian kernel width $\sigma$ in \eqref{equ:mkernel} is set to 0.1, and the other kernel width $\sigma_0$ for generating the soft labels is proportional to the target size, i.e., $\sigma_0=0.1\times\sqrt{WH}$.
	We set the learning rate $\eta$ in \eqref{equ:update} to 0.01.
	For scale estimation, we use $N=21$ levels of the feature pyramid and set the scale factor $\alpha$ to 1.03.
	The hyper-parameter $\tau$ in \eqref{equ:svmupdate} is set to 1.
	We empirically determine all these parameters and fix them throughout all the experiments.
	The source code is available at {\small \url{https://sites.google.com/site/chaoma99/cf-lstm}}.
	
	\begin{table*}
		\centering
		\caption{Overall performance on the OTB2013 (I) \cite{DBLP:conf/cvpr/WuLY13} 
			and OTB2015 (II) \cite{DBLP:journals/pami/WuLY15} 
			datasets with the representative distance precision (DP) rate at a threshold of 20 pixels, overlap success (OS) rate at a threshold of 0.5 IoU, average center location error (CLE), and tracking speed in frame per second (FPS). 
			Best: bold; second best: underline.}
		\label{tb:comparison}
		\setlength{\tabcolsep}{.3em}
		\begin{tabular}{r*{16}{c}}
			\toprule
			& & Ours-deep & \ Ours \ & MUSTer & MEEM & TGPR & \ KCF \ & DSST & \ STC \ & \ \ CSK \ & Struck & \ SCM \ & \ MIL \ & \ TLD \ & LSHT & \ \ CT \ \ \\ 
			& & & & \cite{Hong_2015_CVPR} &
			\cite{DBLP:conf/eccv/ZhangMS14} &
			\cite{DBLP:conf/eccv/GaoLHX14} &
			\cite{DBLP:journals/pami/HenriquesC0B15} & 
			\cite{DBLP:conf/bmvc/DanelljanKFW14} &
			\cite{DBLP:conf/eccv/ZhangZLZY14} & 
			\cite{DBLP:conf/eccv/HenriquesCMB12} & 
			\cite{DBLP:conf/iccv/HareST11} & \cite{DBLP:journals/tip/ZhongLY14} & 
			\cite{DBLP:journals/pami/BabenkoYB11} &
			\cite{DBLP:journals/pami/KalalMM12} & \cite{DBLP:conf/cvpr/HeYLWY13} & 
			\cite{DBLP:conf/eccv/Zhang0Y12} \\\midrule
			\multirow{2}{*}{DP rate (\%)} & I & \textbf{87.8} & 84.8 & {\ul 86.5} & 83.0 & 70.5 & 74.1 & 73.9 & 54.7 & 54.5 & 65.6 & 64.9 & 47.5 & 60.8 & 56.1 & 40.6 \\
			& II & \textbf{82.5} & 76.2 & {\ul 77.4} & 78.1 & 64.3 & 69.2 & 69.5 & 50.7 & 51.6 & 63.5 & 57.2 & 43.9 & 59.2 & 49.7 & 35.9 \\\midrule
			\multirow{2}{*}{OS rate (\%)} & I & {\ul 79.9} & \textbf{81.3} & 78.4 & 69.6 & 62.8 & 62.2 & 59.3 & 36.5 & 44.3 & 55.9 & 61.6 & 37.3 & 52.1 & 45.7 & 34.1 \\
			& II & \textbf{73.9} & {\ul 70.1} & 68.3 & 62.2 & 53.5 & 54.8 & 53.7 & 31.4 & 41.3 & 51.6 & 51.2 & 33.1 & 49.7 & 38.8 & 27.8 \\\midrule
			\multirow{2}{*}{CLE (pixel)} & I & \textbf{15.4} & 26.9 & {\ul 17.3} & 20.9 & 51.3 & 35.5 & 40.7 & 80.5 & 88.8 & 50.6 & 54.1 & 62.3 & 48.1 & 55.7 & 78.9 \\
			& II & \textbf{24.3} & 37.1 & 31.5 & {\ul 27.7} & 55.5 & 45.0 & 47.7 & 86.2 & 305.9 & 47.1 & 61.6 & 72.1 & 60.0 & 68.2 & 80.1 \\\midrule
			\multirow{2}{*}{Speed (FPS)} & I & 14.4 & 21.6 & 13.6 & 20.8 & 0.66 & 245 & 43.0 & \textbf{687} & {\ul 269} & 10.0 & 0.37 & 28.1 & 21.7 & 39.6 & 38.8 \\
			& II & 13.8 & 20.7 & 14.2 & 20.8 & 0.64 & 243 & 40.9 & \textbf{653} & {\ul 248} & 9.84 & 0.36 & 28.0 & 23.3 & 39.9 & 44.4 \\\bottomrule
		\end{tabular}
	\end{table*}
	\begin{figure*}
		\centering
		\setlength{\tabcolsep}{.2em}
		\begin{tabular}{ccc}
			\includegraphics[width=.32\textwidth]{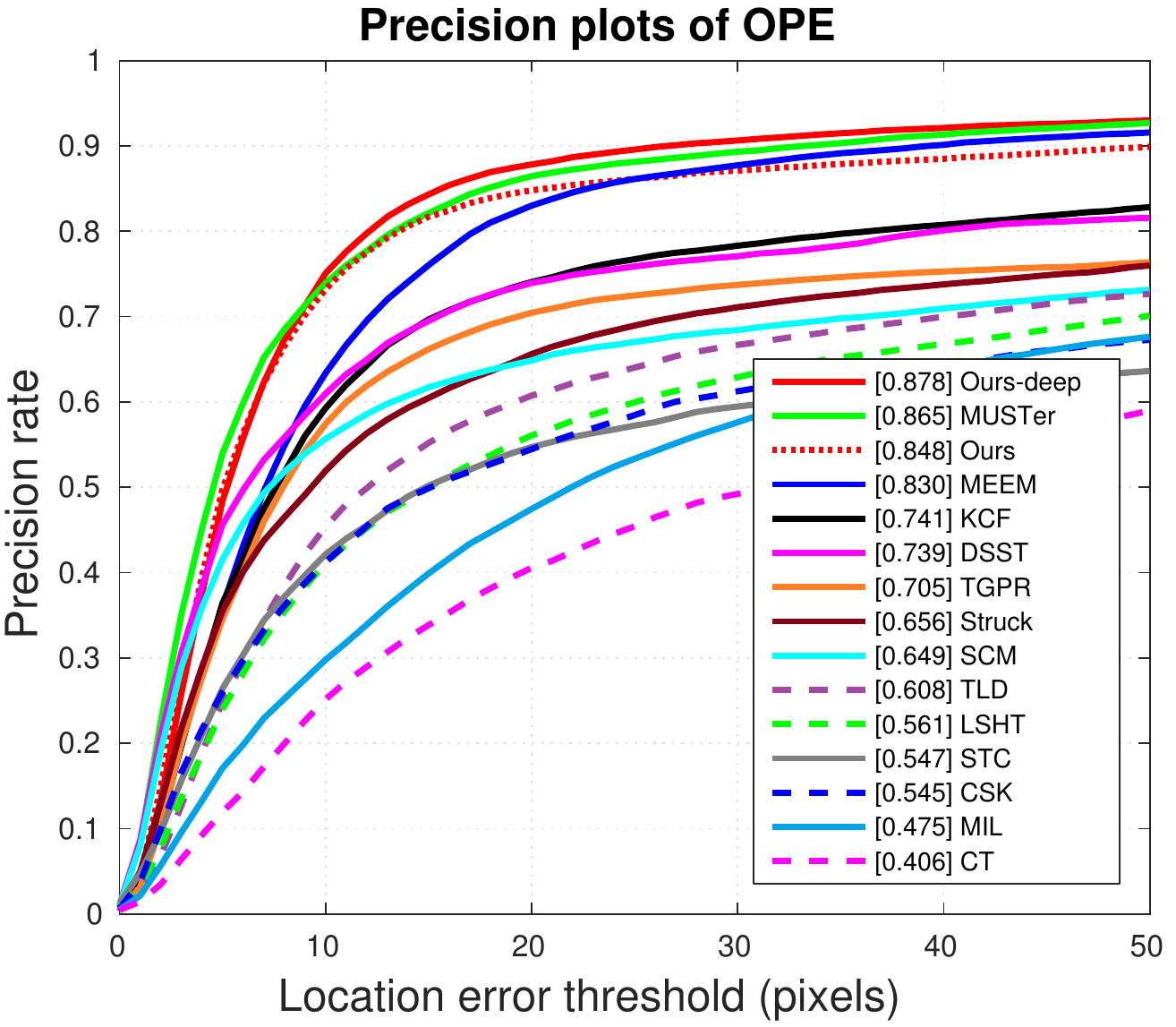}&
			\includegraphics[width=.32\textwidth]{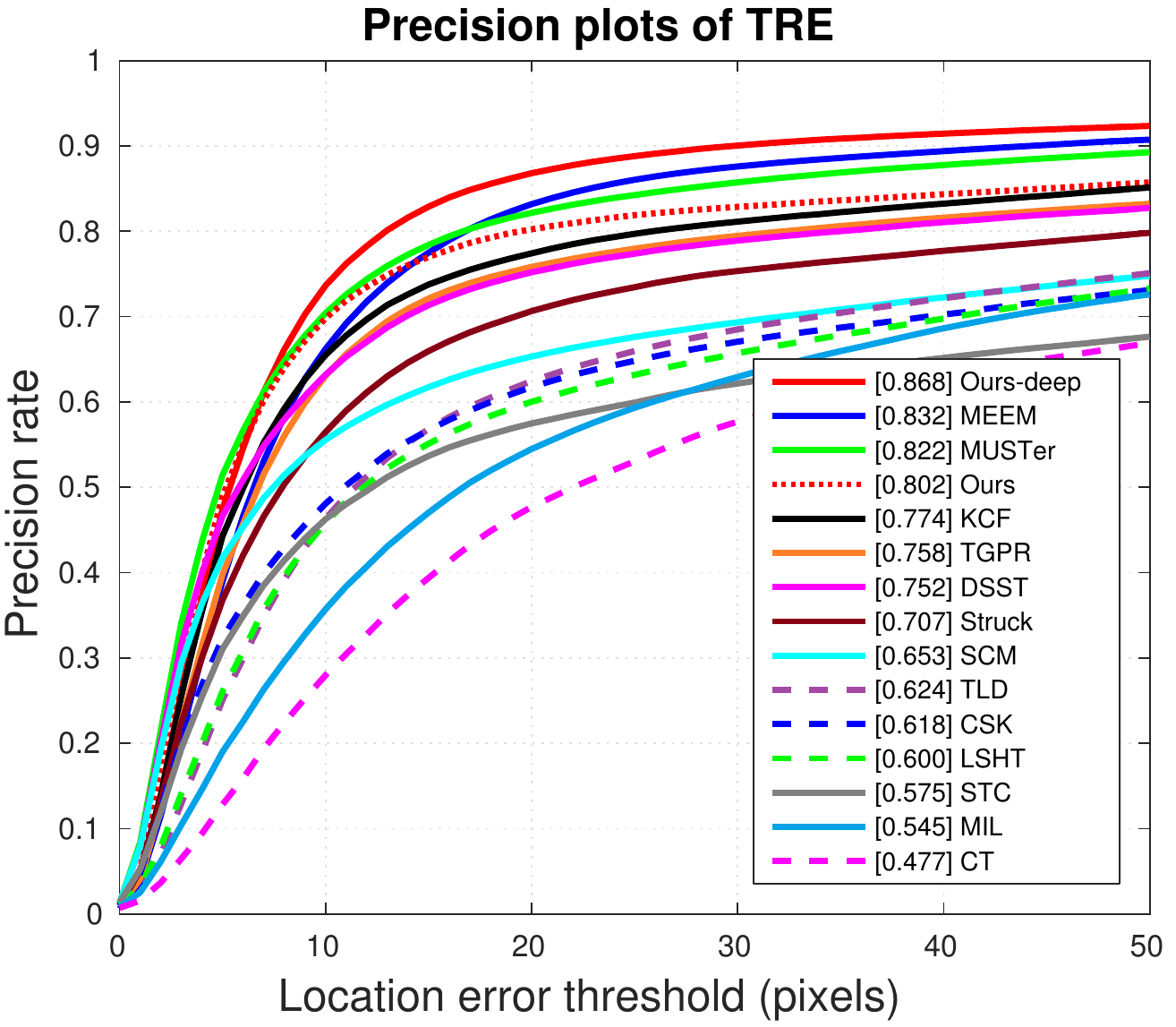}&
			\includegraphics[width=.32\textwidth]{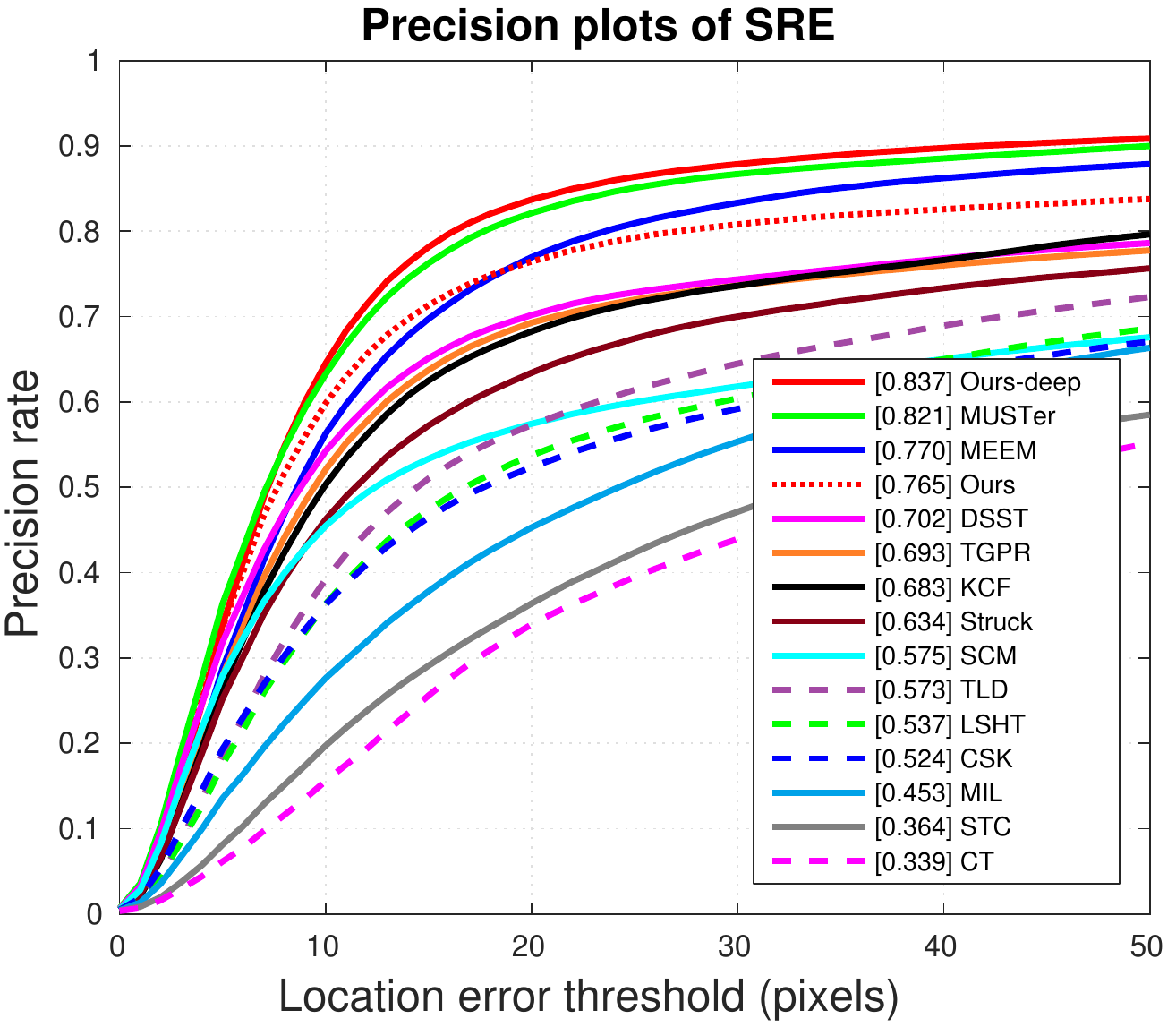}\\ 
			\includegraphics[width=.32\textwidth]{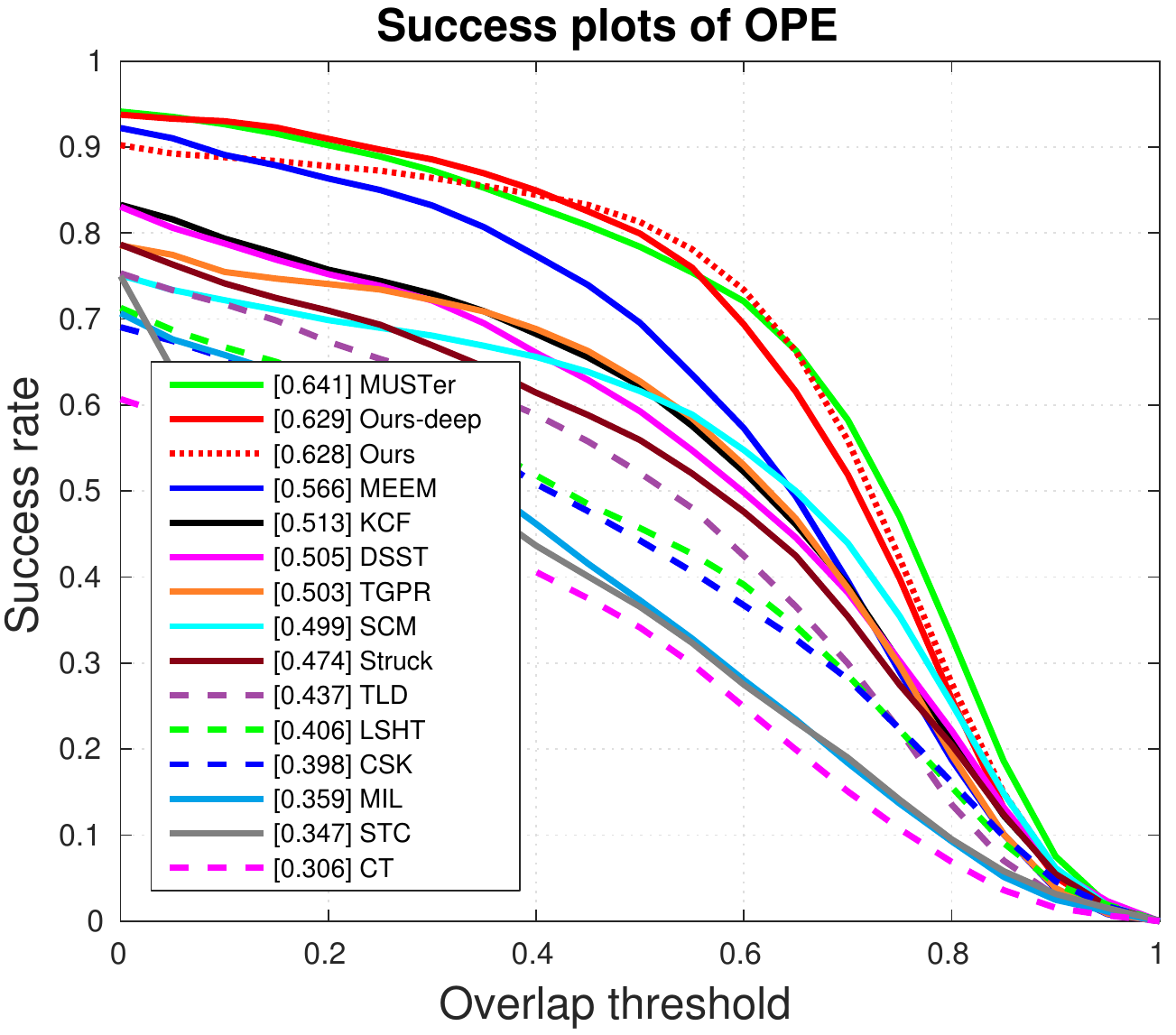}&
			\includegraphics[width=.32\textwidth]{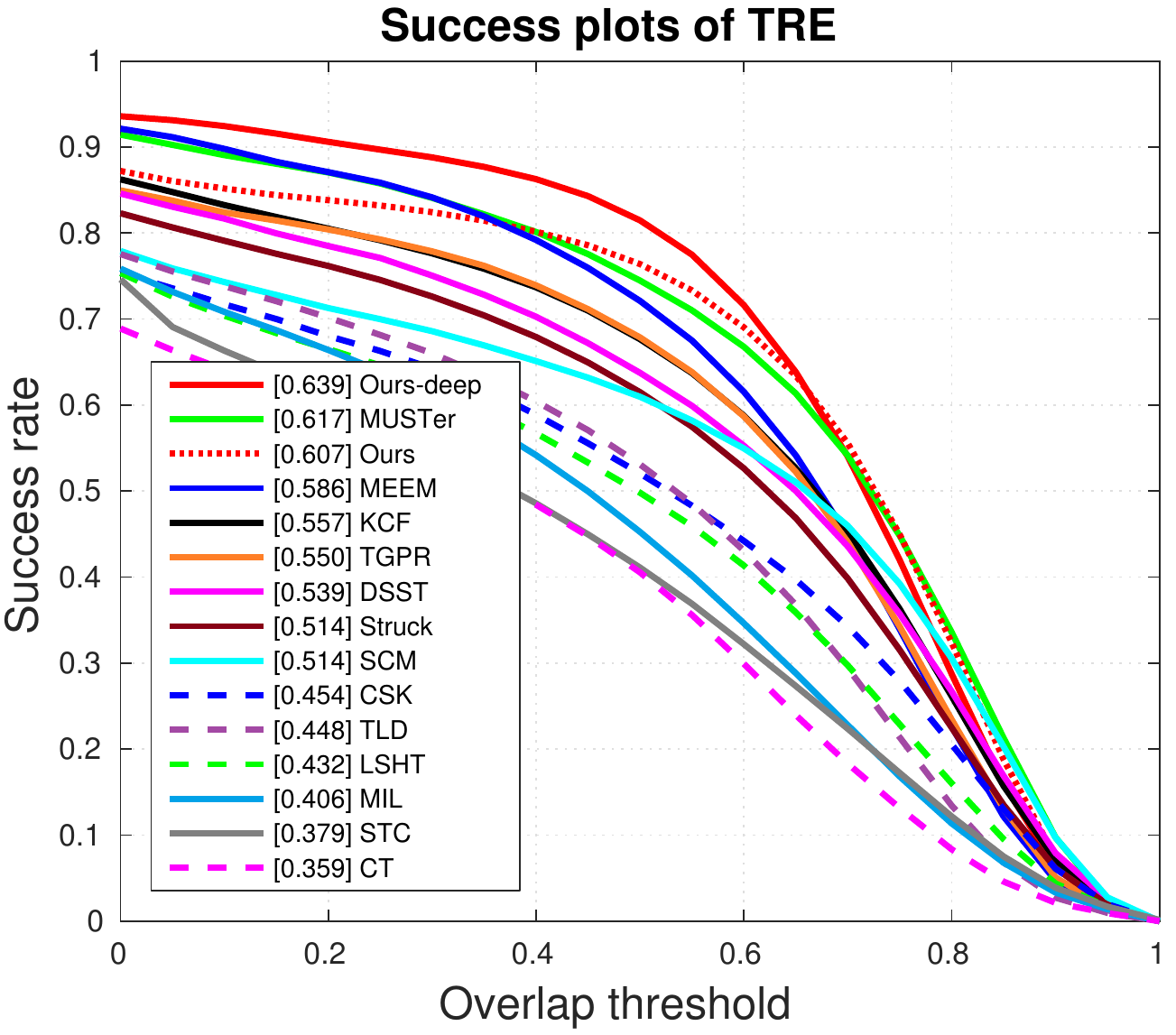}&
			\includegraphics[width=.32\textwidth]{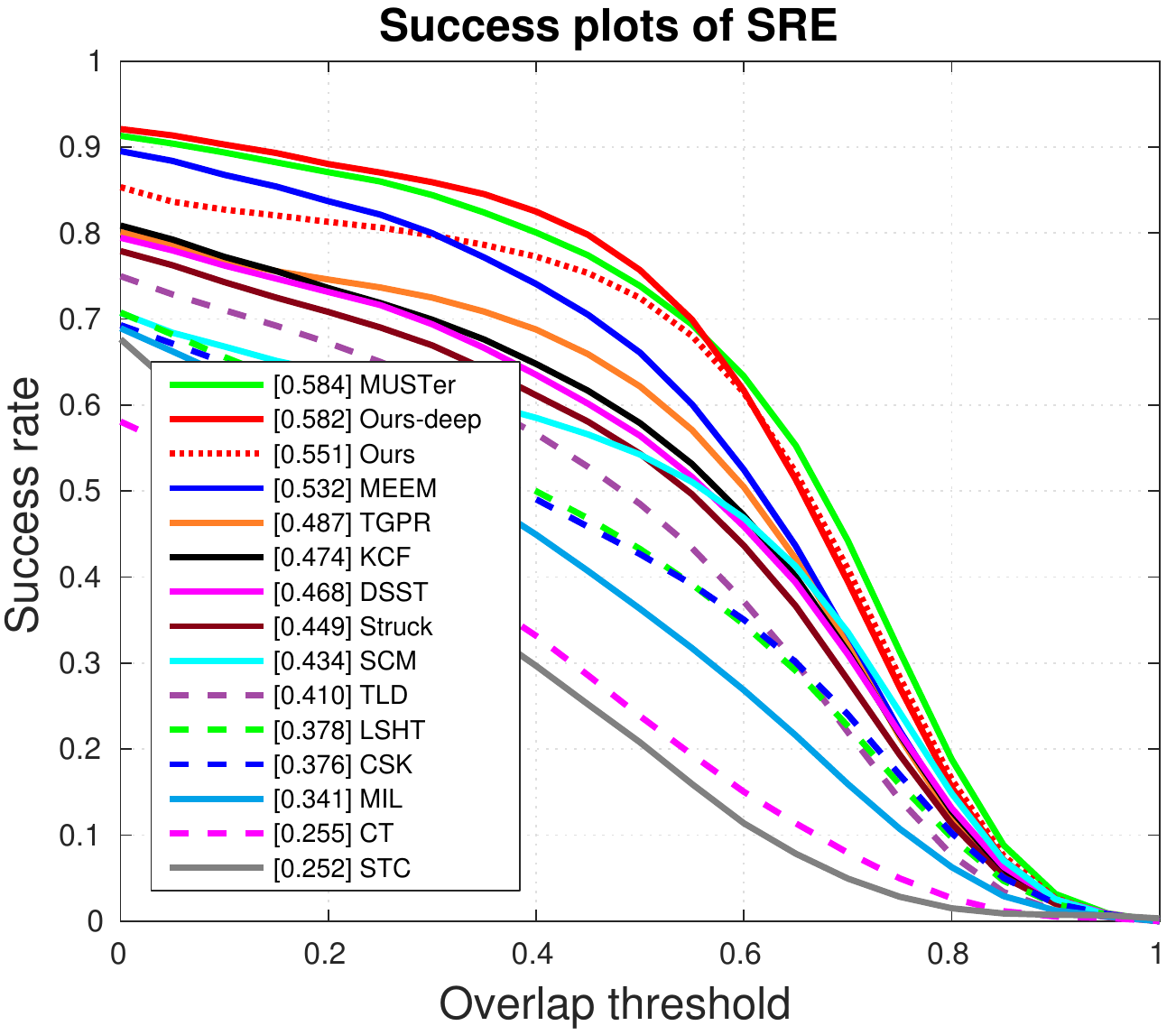}\\
		\end{tabular}
		\caption{
			\textbf{Quantitative evaluation on the OTB2013 dataset.}
			Overlap success and distance precision plots using one-pass evaluation (OPE), temporal robustness evaluation (TRE) and spatial robustness evaluation (SRE). The legend of precision plots shows the distance precision scores at 20 pixels, and the legend of success plots contains the overlap success scores with the area under the curve (AUC).}
		\label{fig:benchmark50}
	\end{figure*}
	\begin{figure*}[t]
		\centering
		\setlength{\tabcolsep}{.2em}
		\begin{tabular}{ccc}
			\includegraphics[width=.32\textwidth]{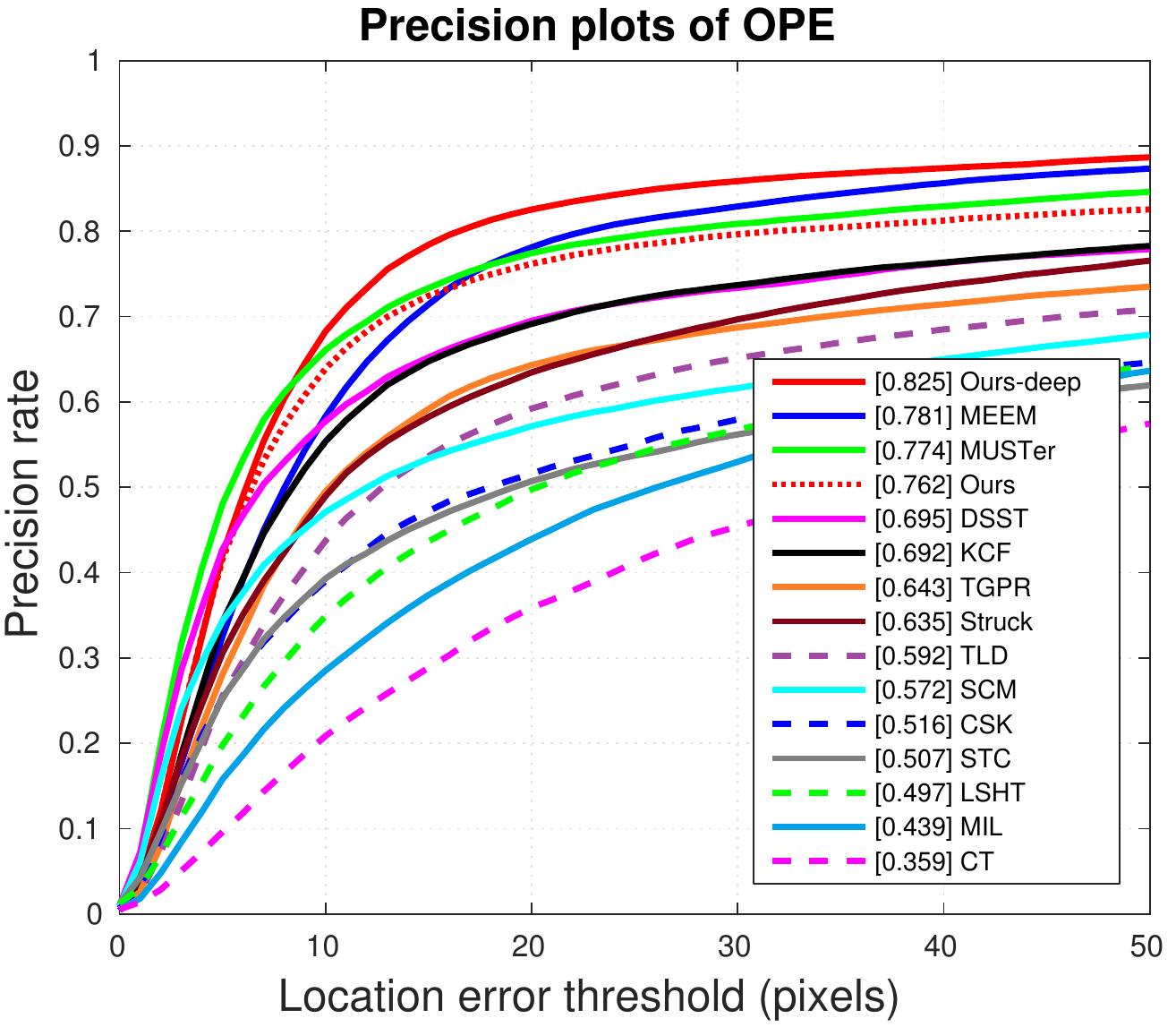} &
			\includegraphics[width=.32\textwidth]{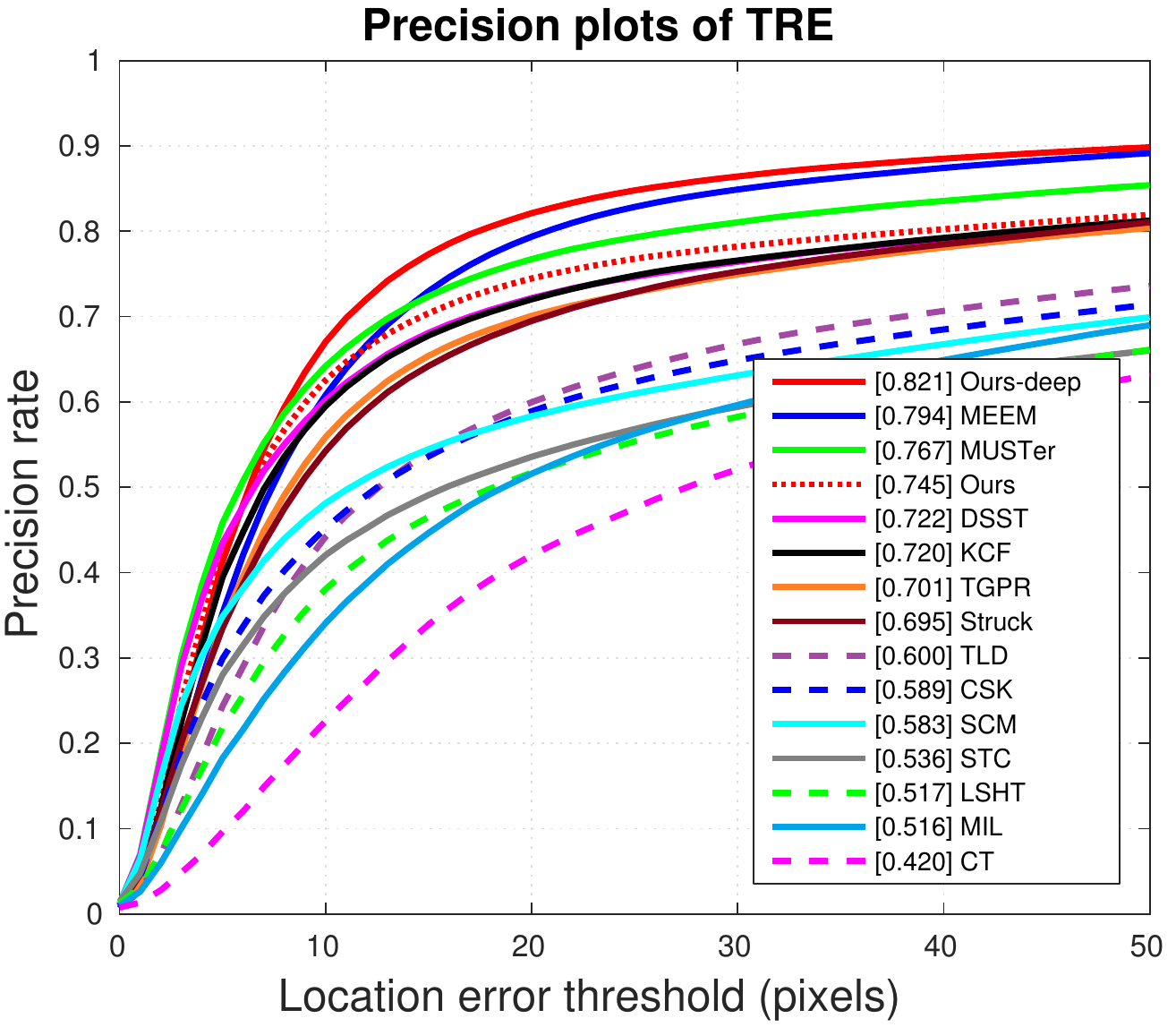} &
			\includegraphics[width=.32\textwidth]{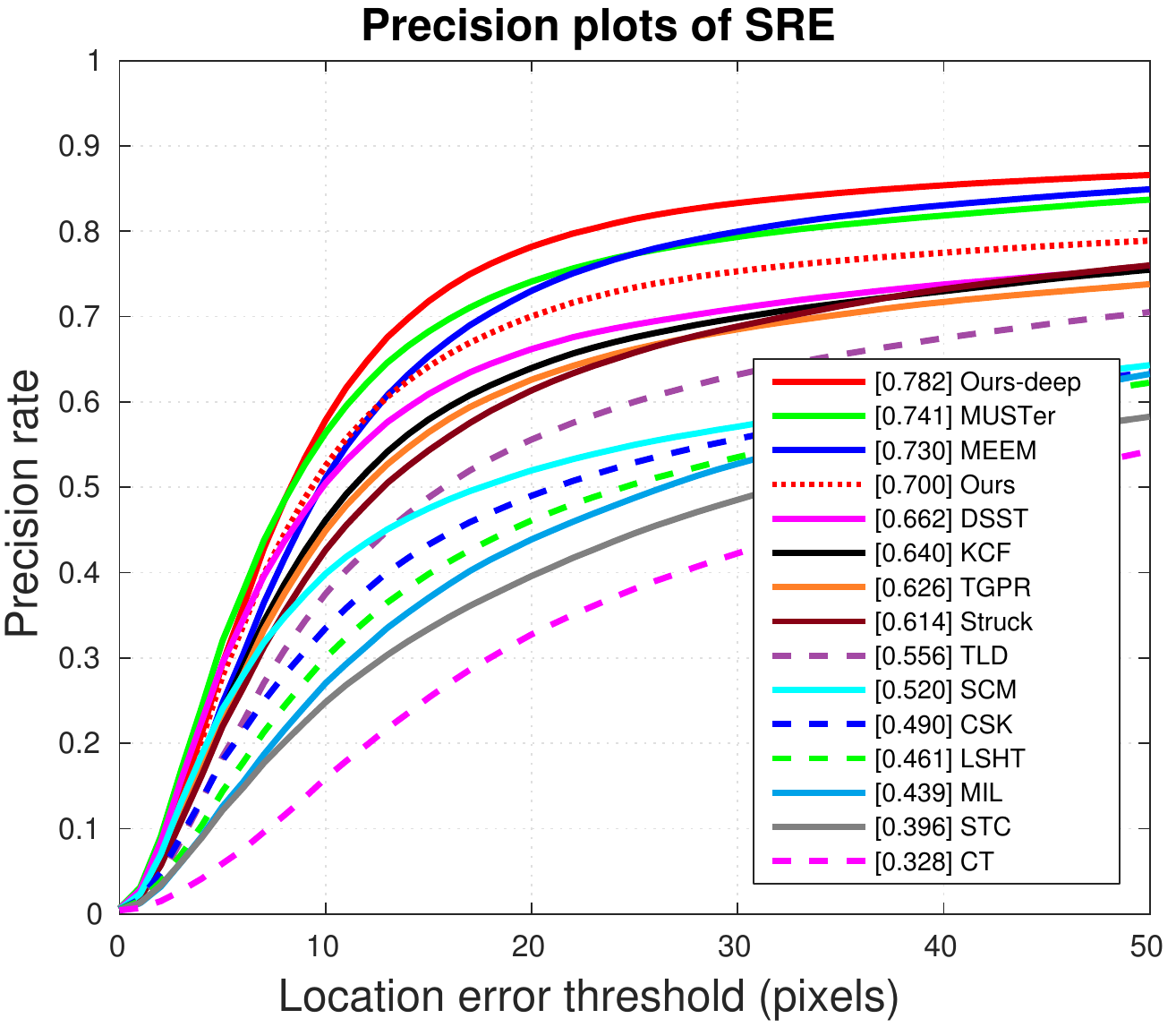} \\ 
			\includegraphics[width=.32\textwidth]{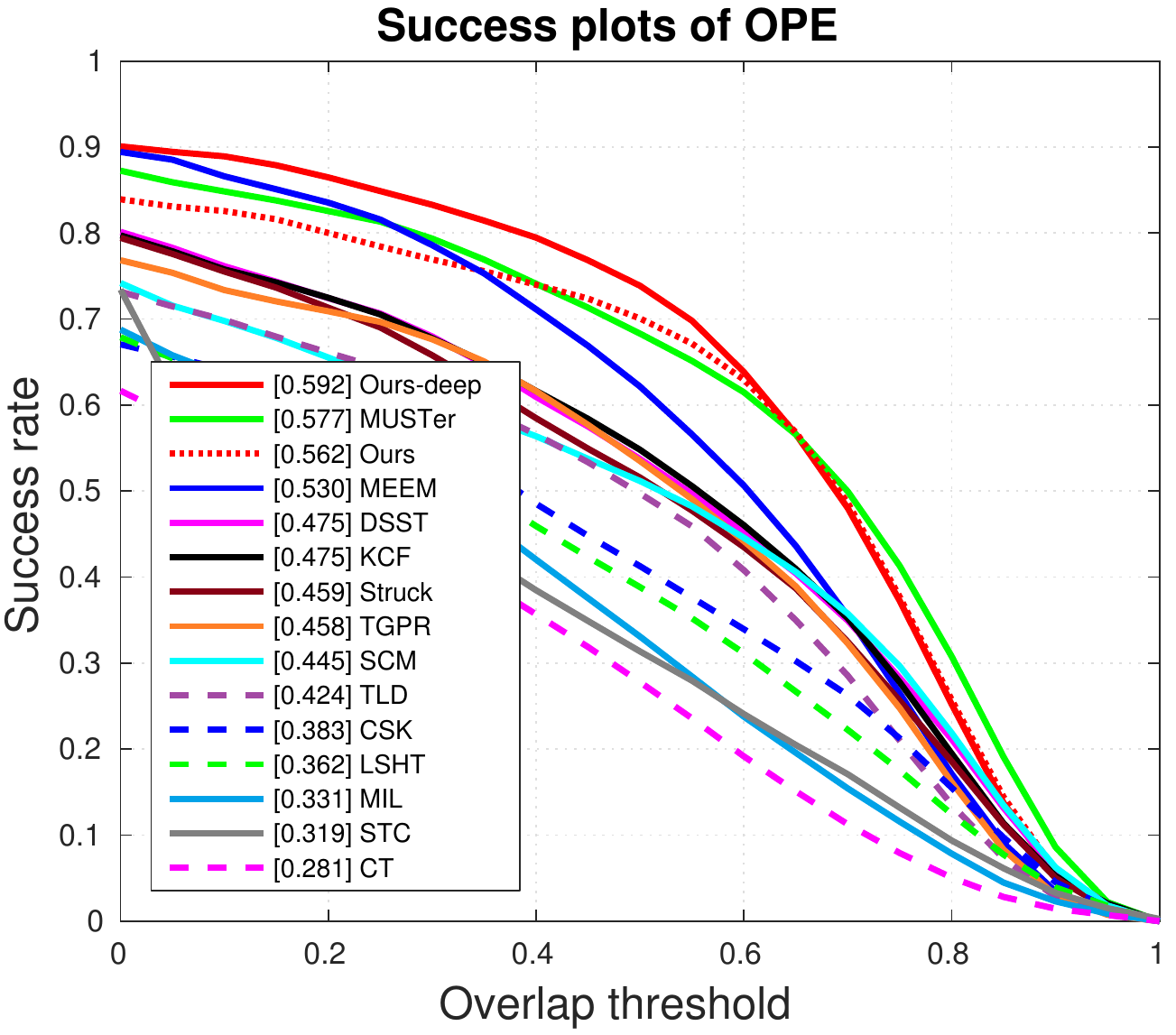} & 
			\includegraphics[width=.32\textwidth]{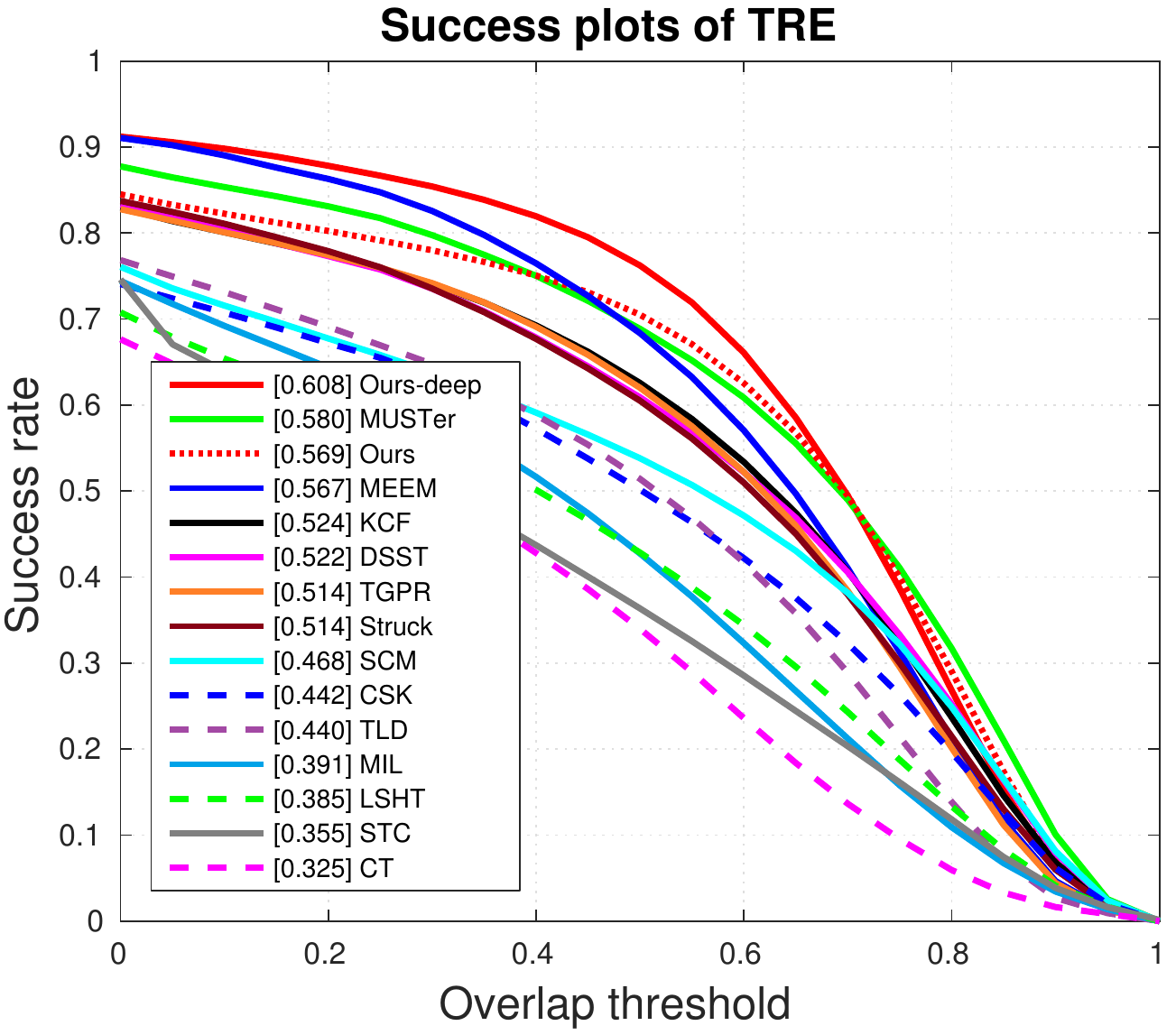} &
			\includegraphics[width=.32\textwidth]{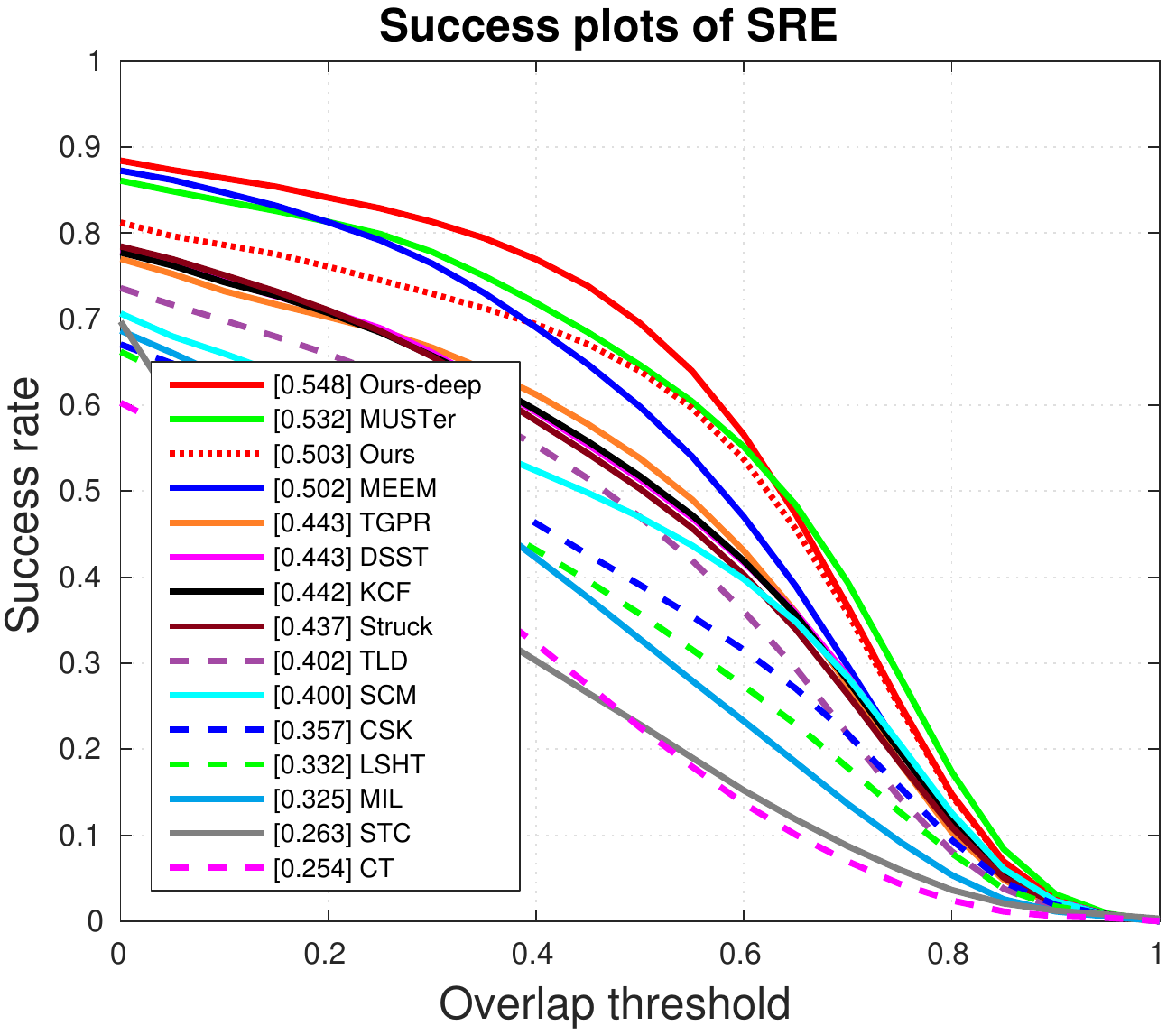} \\
		\end{tabular}
		\caption{
			\textbf{Quantitative evaluation on the OTB2015 dataset.}
			Overlap success and distance precision plots using the one-pass evaluation (OPE), temporal robustness evaluation (TRE) and spatial robustness evaluation (SRE). The legend of precision plots shows the distance precision scores at 20 pixels, and the legend of success plots contains the overlap success scores with the area under the curve (AUC).}
		\label{fig:benchmark100}
	\end{figure*}

	\section{Experimental Results}
	\label{sec:experiment}
	\subsection{Experimental Settings}
	{\flushleft \bf Datasets.} We evaluate the proposed algorithm on a large benchmark dataset \cite{DBLP:journals/pami/WuLY15} that contains 100 videos.
	To validate the effectiveness of the proposed re-detection module, we use additional ten sequences provided by \cite{DBLP:conf/eccv/ZhangMS14}.
	
	{\flushleft \bf Evaluation Metrics.} We evaluate the performance using two widely used metrics:
	\begin{itemize}
		\item Overlap success rate: the percentage of frames where the overlap ratio between predicted bounding box ($\mathbf{b}_1$) and ground truth bounding box ($\mathbf{b}_0$) is larger than a given threshold, i.e., $\frac{\mathbf{b}_1\cap\mathbf{b}_0}{\mathbf{b}_1\cup\mathbf{b}_0}>0.5$.
		\item Distance precision rate: the percentage of frames where the estimated center location error is smaller than a given distance threshold, e.g., 20 pixels.
	\end{itemize} 
	
	{\flushleft \bf Baseline Trackers.}
	We compare the proposed algorithm (1) using only handcrafted features (Ours) and (2) using both handcrafted and deep features (Ours-deep) with 13 state-of-the-art trackers. The compared trackers can be roughly grouped into three categories:
	\begin{itemize}
		\item Trackers using correlation filters including the MUSTer \cite{Hong_2015_CVPR}, 
		KCF \cite{DBLP:journals/pami/HenriquesC0B15}, DSST \cite{DBLP:conf/bmvc/DanelljanKFW14}, STC \cite{DBLP:conf/eccv/ZhangZLZY14}, and CSK \cite{DBLP:conf/eccv/HenriquesCMB12} methods.
		\item Trackers using single online classifier including the Struck \cite{DBLP:conf/iccv/HareST11}, 
		LSHT \cite{DBLP:conf/cvpr/HeYLWY13}, MIL \cite{DBLP:journals/pami/BabenkoYB11}, and CT \cite{DBLP:conf/eccv/Zhang0Y12} methods.
		\item Trackers using multiple online classifiers including the 
		MEEM \cite{DBLP:conf/eccv/ZhangMS14}, TGPR \cite{DBLP:conf/eccv/GaoLHX14}, SCM \cite{DBLP:journals/tip/ZhongLY14}, and TLD \cite{DBLP:journals/pami/KalalMM12} methods.
	\end{itemize}
	Since the baseline methods are not deep learning trackers, we report the results of our method using only handcrafted features throughout all quantitative comparisons.
	We use the evaluation protocol from the benchmark study \cite{DBLP:conf/cvpr/WuLY13}.
	We implemented the algorithm in MATLAB. 
	We conduct all the experimental results on a machine with an Intel I7-4770 3.40 GHz CPU and 32 GB RAM.
	More quantitative and qualitative evaluation results are available at {\small \url{https://sites.google.com/site/chaoma99/cf-lstm}}.
	
	\begin{table*}
		\centering
		\caption{Distance precision scores ($\%$) at a threshold of 20 pixels in terms of individual attributes on the OTB2013 
			dataset \cite{DBLP:conf/cvpr/WuLY13}. Best: bold, second best: underline.}
		\label{tb:attribute100}
		\setlength{\tabcolsep}{.2em}
		\begin{tabular}{*{16}c}
			\toprule
			& Ours-deep & \ Ours \ & MUSTer & MEEM & TGPR & \ KCF \ & DSST & \ STC \ & \ CSK \ & Struck & \ SCM \ & \ MIL \ & \ TLD \ & LSHT & \ \ CT \ \ \\
			& & & \cite{Hong_2015_CVPR} &
			\cite{DBLP:conf/eccv/ZhangMS14} &
			\cite{DBLP:conf/eccv/GaoLHX14} &
			\cite{DBLP:journals/pami/HenriquesC0B15} & 
			\cite{DBLP:conf/bmvc/DanelljanKFW14} &
			\cite{DBLP:conf/eccv/ZhangZLZY14} & 
			\cite{DBLP:conf/eccv/HenriquesCMB12} & 
			\cite{DBLP:conf/iccv/HareST11} & \cite{DBLP:journals/tip/ZhongLY14} & 
			\cite{DBLP:journals/pami/BabenkoYB11} &
			\cite{DBLP:journals/pami/KalalMM12} & \cite{DBLP:conf/cvpr/HeYLWY13} & 
			\cite{DBLP:conf/eccv/Zhang0Y12} \\\midrule
			Illumination variation (23) & \textbf{84.4} & 77.4 & {\ul 77.7} & 75.0 & 64.4 & 70.6 & 71.8 & 55.9 & 47.6 & 55.2 & 55.9 & 31.7 & 49.8 & 49.3 & 32.5 \\
			Out-of-plane rotation (37) & \textbf{87.9} & 84.3 & {\ul 84.4} & 83.3 & 68.2 & 75.9 & 76.1 & 56.0 & 55.8 & 61.6 & 61.8 & 48.0 & 57.9 & 56.6 & 40.4 \\
			Scale variation (28) & \textbf{87.8} & 75.8 & {\ul 81.7} & 78.5 & 62.0 & 68.0 & 74.0 & 54.5 & 50.3 & 63.9 & 67.2 & 47.1 & 60.6 & 49.8 & 44.8 \\
			Occlusion (27) & \textbf{85.2} & 83.6 & {\ul 84.5} & 78.6 & 68.0 & 79.0 & 76.3 & 51.8 & 52.1 & 58.8 & 64.2 & 44.3 & 53.7 & 51.4 & 42.8 \\
			Deformation (17) & \textbf{86.9} & {\ul 86.1} & 84.5 & 83.2 & 70.0 & 80.4 & 71.0 & 51.1 & 50.8 & 55.3 & 58.3 & 48.5 & 46.5 & 56.2 & 46.2 \\
			Motion blur (12) & \textbf{85.1} & 66.4 & 69.5 & {\ul 71.5} & 53.7 & 65.0 & 60.3 & 33.0 & 34.2 & 55.1 & 33.9 & 35.7 & 51.8 & 33.1 & 30.6 \\
			Fast motion (17) & \textbf{81.7} & 66.5 & 69.5 & {\ul 74.2} & 49.3 & 60.2 & 56.2 & 29.3 & 38.1 & 60.4 & 33.3 & 39.6 & 55.1 & 33.4 & 32.3 \\
			In-plane rotation (31) & \textbf{85.9} & {\ul 80.2} & 79.9 & 80.0 & 67.5 & 72.5 & 78.0 & 51.8 & 54.7 & 61.7 & 59.7 & 45.3 & 58.4 & 54.0 & 35.6 \\
			Out of view (6) & 70.6 & \textbf{72.8} & 70.9 & {\ul 72.7} & 50.5 & 64.9 & 53.3 & 39.5 & 37.9 & 53.9 & 42.9 & 39.3 & 57.6 & 38.4 & 33.6 \\
			Background clutter (21) & \textbf{87.2} & 79.6 & {\ul 83.1} & 79.7 & 71.7 & 75.2 & 69.1 & 52.9 & 58.5 & 58.5 & 57.8 & 45.6 & 42.8 & 52.9 & 33.9 \\
			Low resolution (4) & \textbf{95.1} & 71.7 & 75.0 & {\ul 88.8} & 43.8 & 63.0 & 73.8 & 49.1 & 46.4 & 55.0 & 66.1 & 33.1 & 56.6 & 58.5 & 34.0 \\ \midrule
			Weighted Average & \textbf{87.8} & 84.8 & {\ul 86.5} & 83.0 & 74.1 & 73.9 & 64.9 & 54.7 & 65.6 & 70.5 & 47.5 & 60.8 & 56.1 & 54.5 & 40.6 \\\bottomrule
		\end{tabular} 
	\end{table*}
	
	\begin{table*}
		\centering
		\caption{Overlap success scores ($\%$) at a threshold of 0.5 IoU in terms of individual attributes on the OTB2013 dataset \cite{DBLP:conf/cvpr/WuLY13}. Best: bold, second best: underline.}
		\label{tb:attribute50}
		\setlength{\tabcolsep}{.2em}
		\begin{tabular}{*{16}c}
			\toprule
			& Ours-deep & \ Ours \ & MUSTer & MEEM & TGPR & \ KCF \ & DSST & \ STC \ & \ CSK \ & Struck & \ SCM \ & \ MIL \ & \ TLD \ & LSHT & \ \ CT \ \ \\
			& & & \cite{Hong_2015_CVPR} &
			\cite{DBLP:conf/eccv/ZhangMS14} &
			\cite{DBLP:conf/eccv/GaoLHX14} &
			\cite{DBLP:journals/pami/HenriquesC0B15} & 
			\cite{DBLP:conf/bmvc/DanelljanKFW14} &
			\cite{DBLP:conf/eccv/ZhangZLZY14} & 
			\cite{DBLP:conf/eccv/HenriquesCMB12} & 
			\cite{DBLP:conf/iccv/HareST11} & \cite{DBLP:journals/tip/ZhongLY14} & 
			\cite{DBLP:journals/pami/BabenkoYB11} &
			\cite{DBLP:journals/pami/KalalMM12} & \cite{DBLP:conf/cvpr/HeYLWY13} & 
			\cite{DBLP:conf/eccv/Zhang0Y12} \\\midrule
			Illumination variation (23) & \textbf{73.6} & \textbf{73.6} & 71.6 & 63.6 & 58.5 & 58.2 & 58.4 & 36.1 & 40.2 & 49.4 & 53.4 & 28.8 & 43.1 & 43.8 & 29.0 \\
			Out-of-plane rotation (37) & {\ul 78.3} & \textbf{79.4} & 73.0 & 66.7 & 60.6 & 62.9 & 60.1 & 38.1 & 45.2 & 52.1 & 57.4 & 36.5 & 47.6 & 45.5 & 32.6 \\
			Scale variation (28) & \textbf{74.2} & 69.3 & {\ul 70.4} & 57.0 & 48.8 & 47.7 & 50.5 & 33.8 & 35.2 & 47.1 & 63.5 & 33.5 & 49.4 & 33.6 & 34.2 \\
			Occlusion (27) & \textbf{80.6} & {\ul 79.3} & 74.8 & 66.1 & 61.7 & 64.9 & 57.7 & 32.0 & 41.8 & 51.2 & 59.9 & 37.4 & 43.6 & 40.8 & 36.1 \\
			Deformation (17) & {\ul 81.3} & \textbf{87.1} & 80.3 & 65.4 & 67.0 & 72.5 & 60.8 & 33.2 & 38.9 & 50.0 & 56.1 & 43.3 & 40.4 & 45.6 & 42.1 \\
			Motion blur (12) & \textbf{75.3} & 66.5 & {\ul 66.8} & 66.0 & 53.9 & 59.6 & 53.2 & 22.9 & 33.6 & 51.8 & 33.9 & 24.7 & 48.2 & 27.3 & 26.1 \\
			Fast motion (17) & \textbf{74.1} & {\ul 67.6} & 65.1 & 68.1 & 49.0 & 55.7 & 51.6 & 22.6 & 38.0 & 56.7 & 33.5 & 33.8 & 47.3 & 30.9 & 32.3 \\
			In-plane rotation (31) & {\ul 74.2} & \textbf{77.0} & 69.1 & 65.0 & 59.7 & 61.4 & 64.6 & 38.4 & 45.7 & 52.8 & 56.0 & 33.1 & 47.6 & 43.5 & 28.9 \\
			Out of view (6) & 68.8 & \textbf{69.9} & {\ul 69.2} & 74.8 & 54.0 & 64.9 & 54.7 & 29.6 & 41.0 & 55.0 & 44.9 & 41.6 & 51.6 & 41.9 & 40.5 \\
			Background clutter (21) & \textbf{78.8} & {\ul 76.7} & 75.0 & 73.7 & 67.5 & 67.3 & 59.1 & 39.5 & 49.1 & 54.5 & 55.0 & 41.4 & 38.8 & 48.5 & 32.3 \\
			Low resolution (4) & \textbf{57.5} & {\ul 45.2} & 43.9 & 36.9 & 30.6 & 25.8 & 33.1 & 30.2 & 27.3 & 24.0 & 55.5 & 16.1 & 32.7 & 20.0 & 14.6 \\\midrule
			Weighted Average & {\ul 79.9} & \textbf{81.3} & 78.4 & 69.6 & 62.2 & 59.3 & 61.6 & 36.5 & 55.9 & 62.8 & 37.3 & 52.1 & 45.7 & 44.3 & 34.1 \\\bottomrule
		\end{tabular}
	\end{table*}
	
	\subsection{Overall Performance}
	\label{sec:overallperformance}
	The object tracking benchmark dataset \cite{DBLP:journals/pami/WuLY15} contains two versions: 
	(1) OTB2013 \cite{DBLP:conf/cvpr/WuLY13} with 50 sequences
	and 
	(2) OTB-2015 \cite{DBLP:journals/pami/WuLY15} with 100 sequences. 
	We show the quantitative results using the one-pass evaluation (OPE), temporal robustness evaluation (TRE), and spatial robustness evaluation (SRE) criteria on both datasets in Figure \ref{fig:benchmark50} and Figure \ref{fig:benchmark100}.
	Following the protocol, we report distance precision rate at a threshold of 20 pixels, the overlap success rate at a threshold of 0.5 Intersection over Union (IoU), the average center location error, and and the average tracking speed in frames per second in Table \ref{tb:comparison}.
	We show in Table \ref{tb:comparison} that the proposed algorithm performs favorably against the representative baseline methods in both overlap success and distance precision metrics.
	
	In addition, we compare our method with the MEEM and MUSTer trackers in more detail as both two approaches explicitly incorporate re-detection modules. 
	The MEEM tracker employs multiple SVM classifiers with different learning rates and uses an entropy measure to fuse all the outputs from multiple classifiers. 
	While the MEEM tracker can recover from tracking failures, it does not handle scale changes well.
	Our method explicitly predicts scale variation and thus achieves higher overlap success rate over MEEM (81.3\% versus 69.6\%).
	The MUSTer tracker is a concurrent work with our preliminary work \cite{Ma_2015_CVPR}. 
	Both the MUSTer tracker and our approach can cope with scale changes.
	Unlike the MUSTer tracker, we update the translation filter $\mathcal{A}_\mathrm{T}$ \emph{without} considering scale changes. 
	We observe that slight inaccuracy in scale estimation would cause rapid performance degradation of the translation filter.
	Our method achieves higher overlap success rates than the MUSTer tracker: 81.3\% versus 78.4\% on the OTB2013 dataset, and 70.1\% versus 68.3\% on the OTB2015 dataset, respectively. 
	
	Regarding tracking speed, the STC, CSK and KCF trackers using only one correlation filter are faster than our approach.
	The accuracy of these trackers, however, is inferior to our approach due to their inability to recover from failures and to handle scale variation.
	Our tracker runs 20 frames per second (close to real-time) as we only activate the detector when the confidence score is below the re-detection threshold $T_r$ and avoid the computationally expensive search in sliding window.
	In terms of the TRE and SRE criteria, the proposed method does not perform as well as in the OPE evaluation. 
	This can be explained by the fact that the TRE and SRE evaluation schemes
	are designed to evaluate tracking methods without re-detection modules. 
	In the TRE evaluation criterion, a video sequence 
	is divided into several fragments, and thus the importance of the re-detection module in long-term tracking is not taken into account.
	In the SRE evaluation criterion, the trackers are initialized with the slightly inaccurate target position and scale. 
	As our tracker relies on learning correlation filters to discriminate the target from its background, inaccurate initialization in the spatial domain adversely affects the performance of the learned filter in locating targets. 
	
	We discuss several observations from the experimental results. 
	First, correlation trackers (e.g., KCF and DSST) consistently outperform the methods that use one single discriminative classifier (e.g., LSHT, CSK, and CT).
	This can be attributed to that correlation filters regress all the circularly shifted samples of target appearance into soft regression targets rather than hard-thresholded binary labels.
	Correlation filter based trackers effectively alleviate the sampling ambiguity problem.
	Second, trackers with re-detection modules (e.g., MUSTer, MEEM and the proposed tracker) outperform those without re-detection modules.
	Third, while TLD uses a re-detection module, we find that the TLD tracker does not perform well in sequences with drastic appearance changes.
	It is because the tracking module in TLD builds upon the Lucas-Kanade \cite{DBLP:conf/ijcai/LucasK81} method with an aggressive update rate. 
	Such highly adaptive model often results in drifting.
	
	\subsection{Attribute-Based Evaluation}
	The video sequences in the benchmark dataset \cite{DBLP:conf/cvpr/WuLY13} are annotated with 11 attributes to describe the various challenges in object tracking, e.g., occlusion or target disappearance of the camera view (out-of-view). 
	We use these attributes to analyze the performance of trackers in various aspects.
	In Table \ref{tb:attribute50} and Table \ref{tb:attribute100}, we show the attribute-based evaluation results in terms of overlap success and distance precision on the OTB2013 dataset. 
	In terms of overlap success rate, the proposed algorithm performs well against the baseline methods in most of the attributes. 
	Compared to the concurrent MUSTer method, our tracker achieves large performance gains in seven attributes: illumination variation (2.0\%), out-of-plan rotation (6.4\%), occlusion (4.5\%), deformation (6.8\%), in-plan rotation (7.9\%), background clutter (1.7\%) and low resolution (1.3\%).
	We attribute the improvements mainly to three reasons. 
	First, we separate the model update for learning the translation filter 
	$\mathcal{A}_\mathrm{T}$ and the scale filter $\mathcal{A}_\mathrm{S}$.
	While this approach appears to be sub-optimal for inferring target states when compared to the MUSTer tracker, we find that it effectively avoids the degradation of the translation filter caused by inaccuracy of scale estimation.
	Second, the HOI features are based on the local histogram of intensities, which strengthen the distinction between target objects and background in the presence of rotation. This helps the translation filter locate target objects precisely. 
	Third, we maintain the long-term memory of target appearance as a holistic template using correlation filters. The MUSTer tracker instead uses a pool of local key-point descriptors (i.e., SIFT \cite{DBLP:journals/ijcv/Lowe04}) for capturing the long-term memory of target appearance. 
	In the presence of significant deformation and rotation, there are significantly fewer key points to discriminate target objects.
	As a result, our tracker is more robust to these challenges than the MUSTer tracker.
	Table \ref{tb:attribute100} shows that our method achieves the best results in deformation (86.1\%), in-plane rotation (80.2\%) and out-of-view movement (72.8\%) based on the distance precision rate.
	These results demonstrate the effectiveness of our method in handling large appearance changes and recovering targets from failure cases. 
	With the use of a similar re-detection module, the MEEM tracker performs favorably in dealing with motion blur, fast motion, and low resolution.

	\subsection{Ablation Studies}

	\begin{figure}
		\centering
		\setlength{\tabcolsep}{.5mm}
		\begin{tabular}{cc}
			\includegraphics[width=0.235\textwidth]{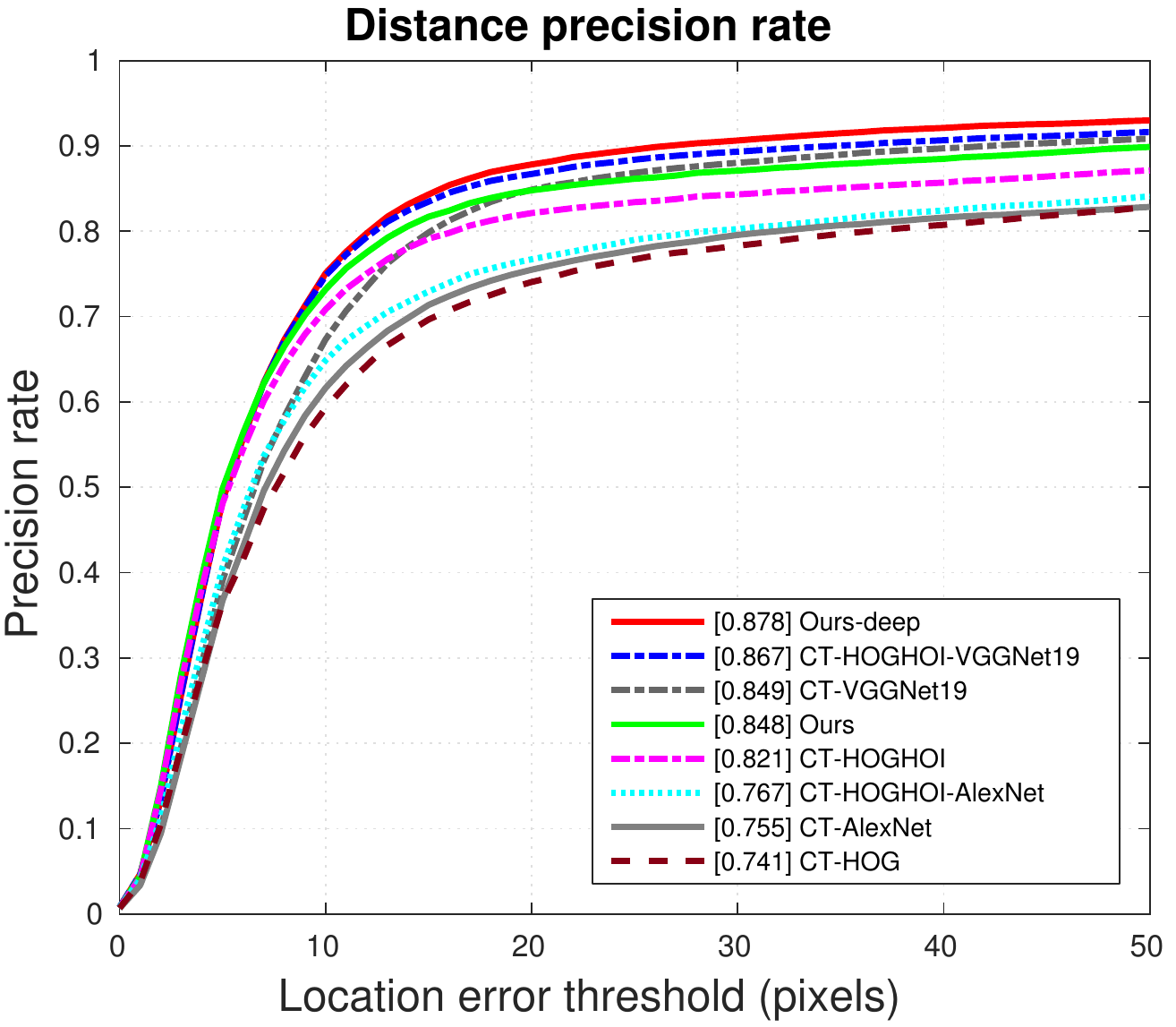} & 
			\includegraphics[width=0.235\textwidth]{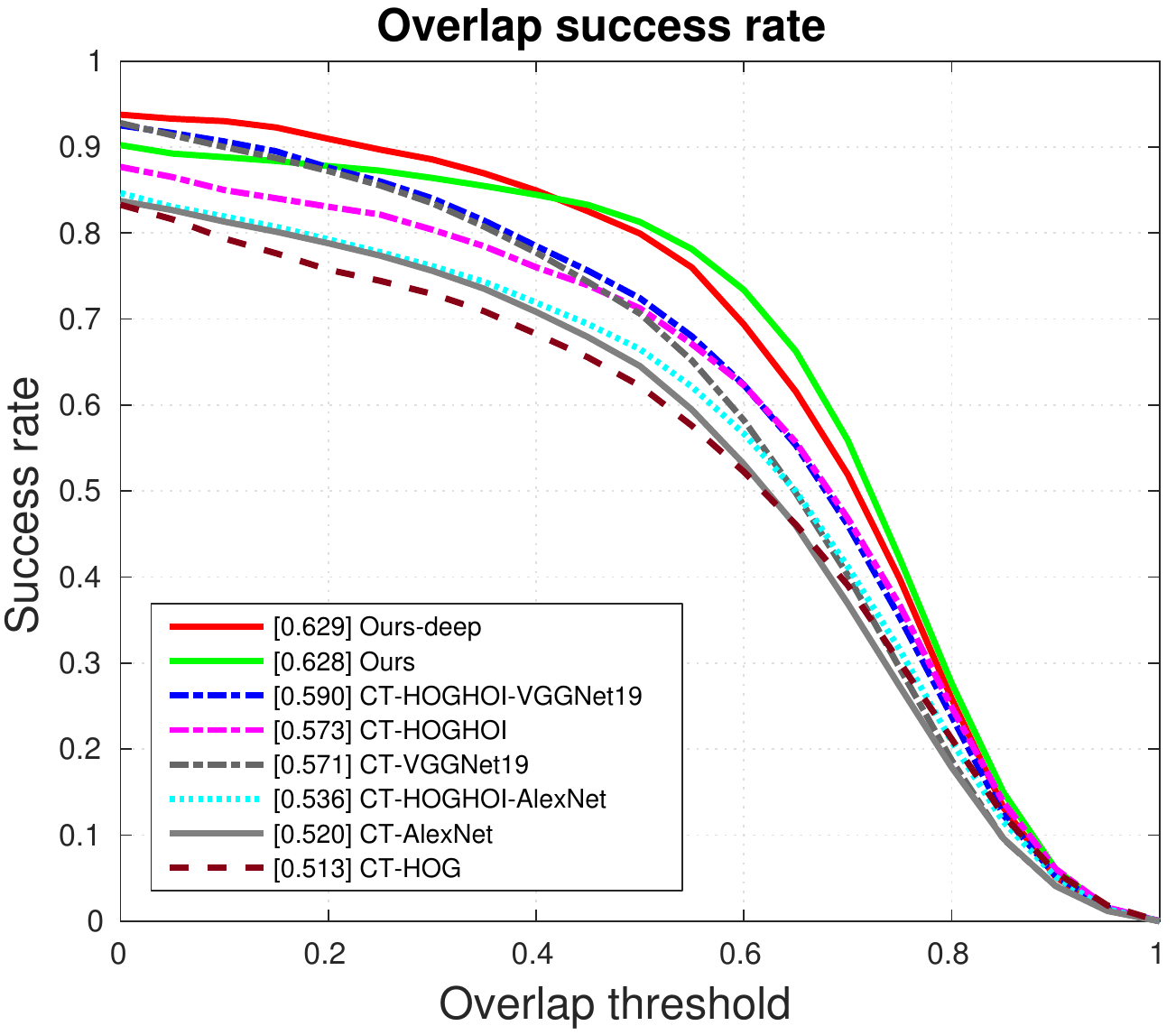} \\
		\end{tabular}
		\caption{
			\textbf{Feature analysis for learning translation filters on the OTB2013 dataset}. 
			The baseline trackers (CT-*) do not incorporate re-detection modules. 
			Using both deep and handcrafted features, the CT-HOGHOI-VGGNet19 method outperforms other alternatives. 
		}
		\label{fig:component-translation}
	\end{figure}

	\begin{figure}
		\centering
		\setlength{\tabcolsep}{.5mm}
		\begin{tabular}{cc}
			\includegraphics[width=0.235\textwidth]{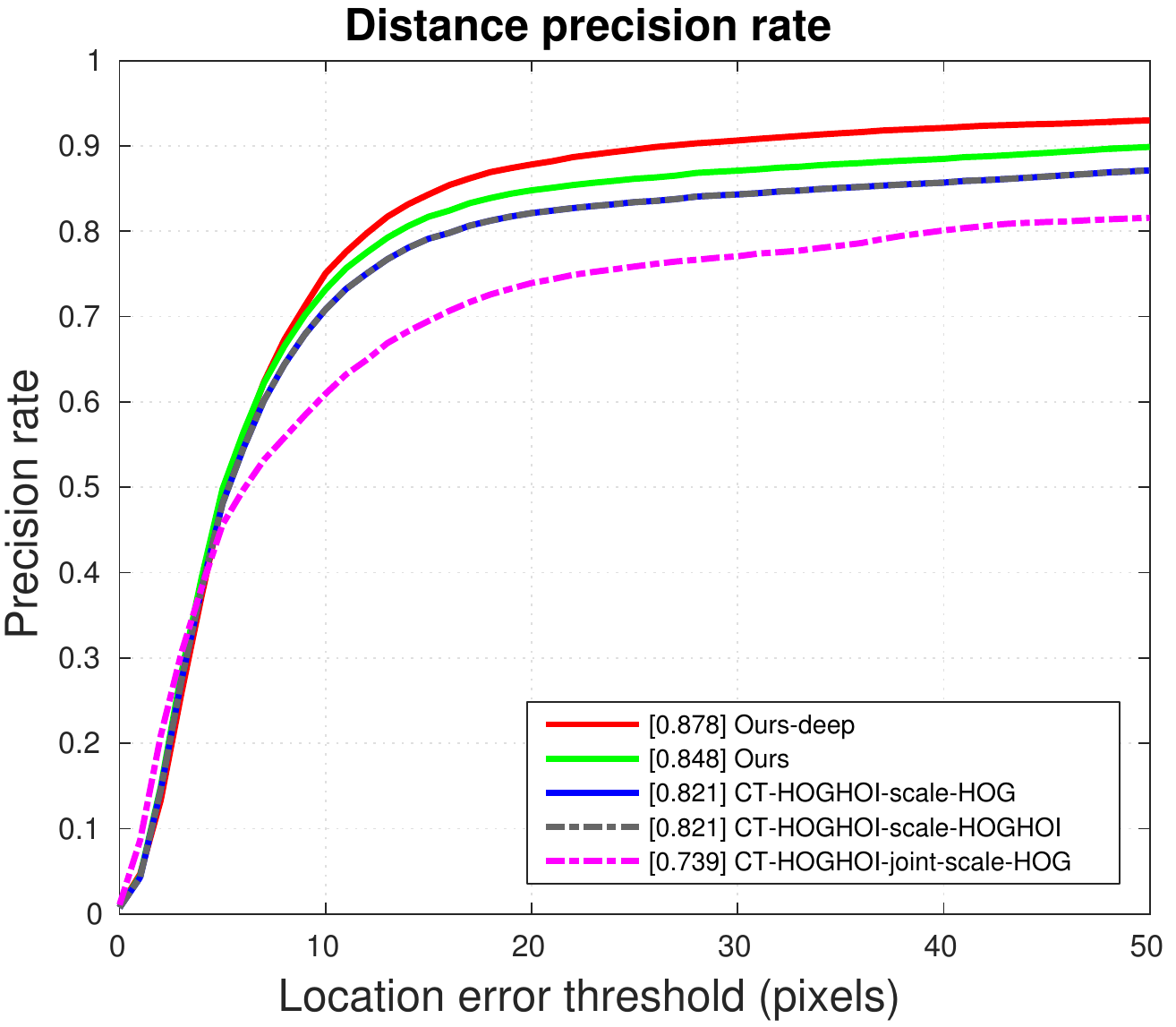} & 
			\includegraphics[width=0.235\textwidth]{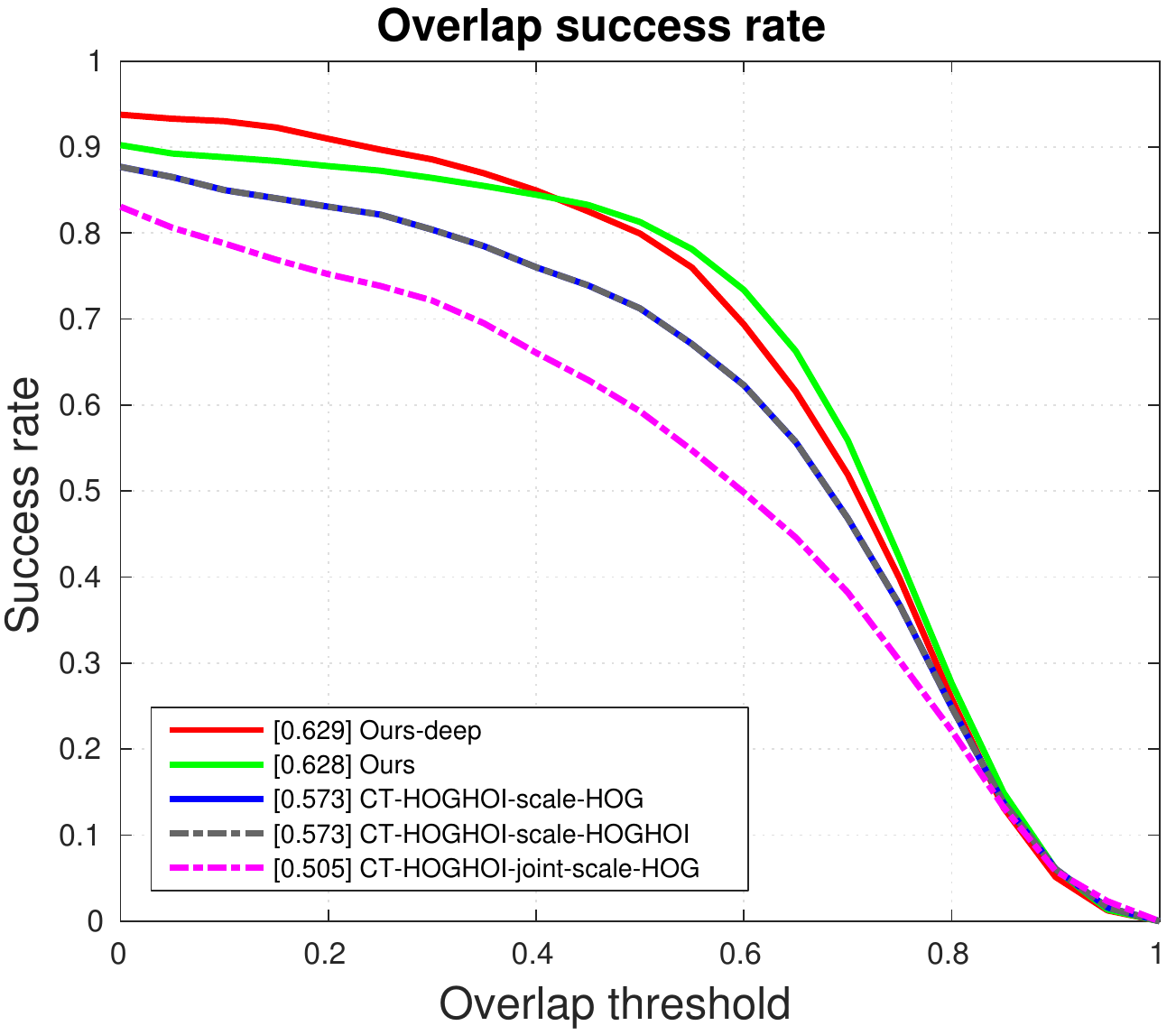} \\
		\end{tabular}
		\caption{
			\textbf{Analysis of scale estimation schemes on the OTB2013 dataset}. 
			We first analyze features for learning scale filters. Adding the HOI features for scale estimation (CT-HOGHOI-scale-HOGHOI) does not improve tracking accuracy when compared to the CT-HOGHOI-scale-HOG method. 
			The CT-HOGHOI-joint-scale-HOG method updates the translation filter $\mathcal{A}_\mathrm{T}$ using the \emph{estimated} scale change in each frame. 
			In contrast, we use the ground-truth scale in the first frame to update the translation filter $\mathcal{A}_\mathrm{T}$).
		}
		\label{fig:component-scale}
	\end{figure}

\begin{figure}
		\centering
		\setlength{\tabcolsep}{.5mm}
		\begin{tabular}{cc}
			\includegraphics[width=0.235\textwidth]{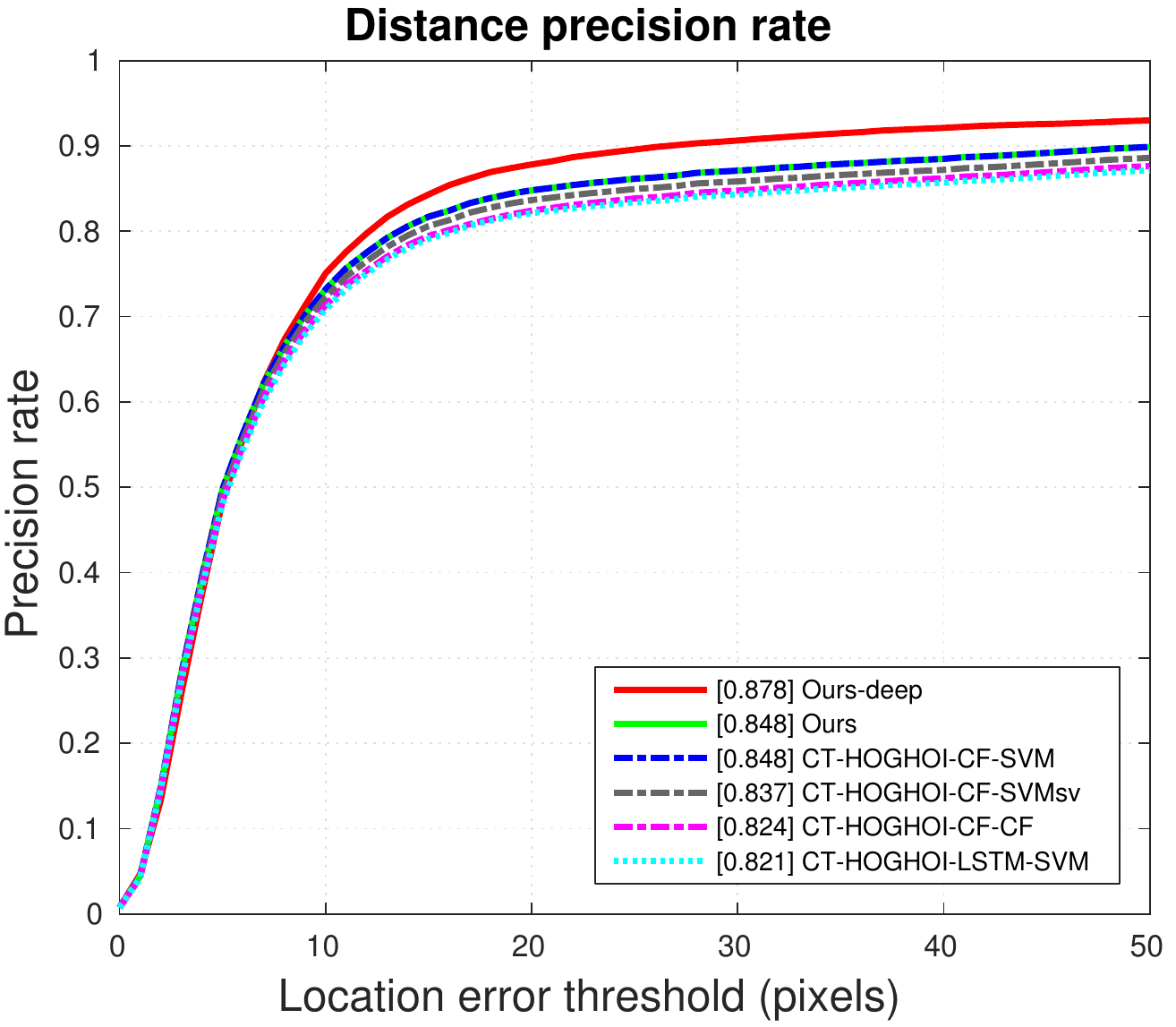} & 
			\includegraphics[width=0.235\textwidth]{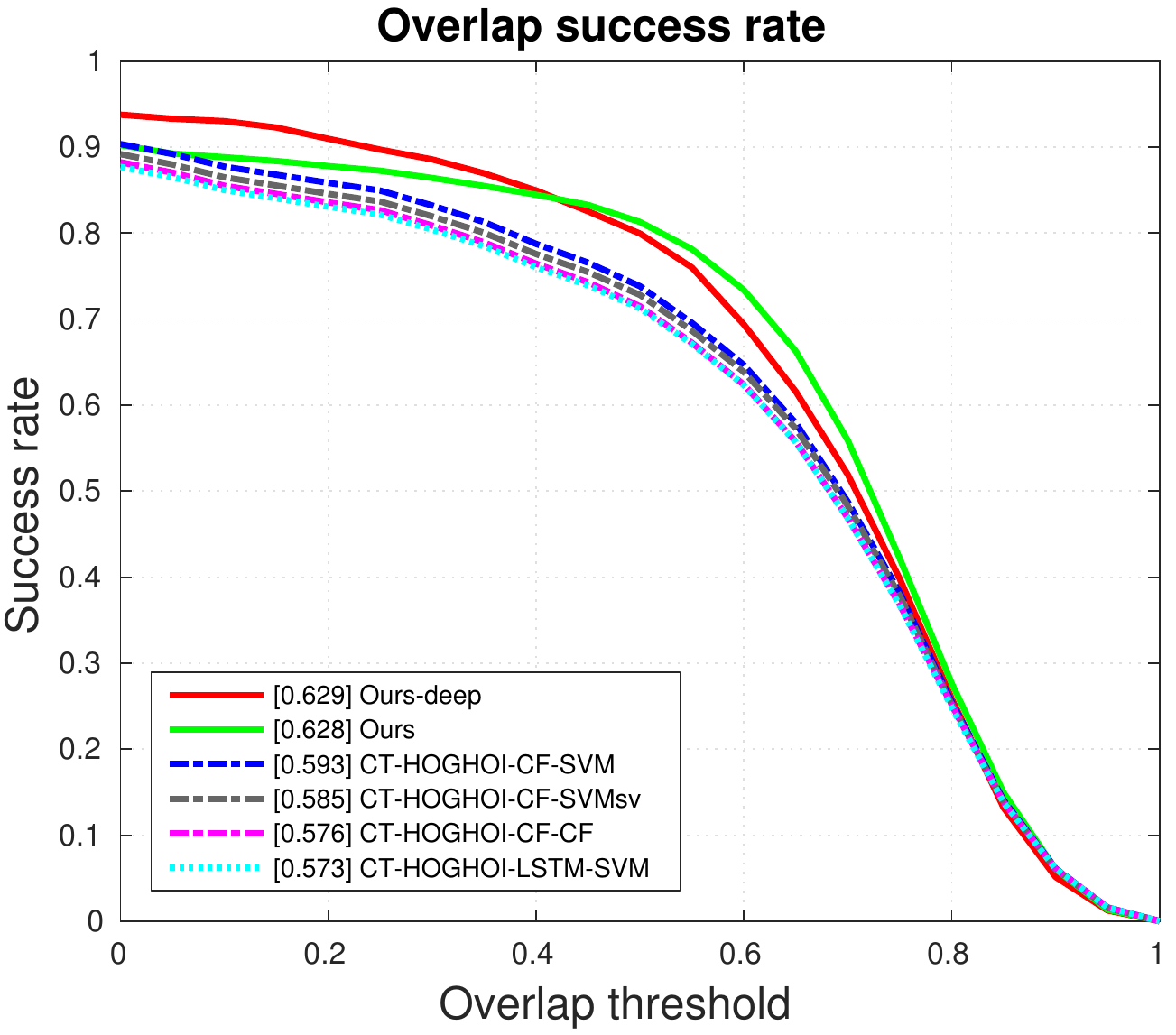} \\
		\end{tabular}
		\caption{
			\textbf{Analysis of re-detection modules on the OTB2013 dataset.}
			%
			Using the CT-HOGHOI method, we evaluate the results of using four re-detection schemes. 
			With the correlation filter based long-term filter, we use two different schemes to update SVM using (1) the proposed passive-aggressive scheme (CT-HOGHOI-CF-SVM) 
			and (2) the support vector update scheme (CT-HOGHOI-CF-SVMsv) described in \cite{DBLP:conf/eccv/ZhangMS14}. 
			(3) The CT-HOGHOI-CF-CF method uses the long-term filter itself as a detector. 
			(4) The CT-HOGHOI-LSTM-SVM method using the hidden states of an LSTM network as the long-term filter. 
			The proposed CT-HOGHOI-CF-SVM scheme performs well against other alternative approaches.
		}
		\label{fig:component-detector}
	\end{figure}

	\label{sec:ablation}
	To better understand the contributions of each component of the proposed tracker, we carry out three group of ablation studies by comparing with other alternative design options.
	Figure \ref{fig:component-translation}-\ref{fig:component-detector} show the overall tracking performance and comparisons with alternative approaches for developing the translation filter, scale filter, and the re-detection module on the OTB2013 dataset \cite{DBLP:conf/cvpr/WuLY13} using the one pass evaluation (OPE) protocol. 
	The legend of precision plots shows scores at a threshold of 20 pixels. 
	The legend of success plots contains the values of the area under the curve (AUC). 
	For clarity, we add the proposed methods that incorporate all components using deep features (ours-deep) or handcrafted features (ours) in 
	Figure \ref{fig:component-translation}-\ref{fig:component-detector}.
	
	{\flushleft \bf Feature Analysis on Translation Filter.} 
	We first demonstrate the effectiveness of using different types of features for learning the translation filter.
	Note that all the baseline methods (except ours and ours-deep) in Figure \ref{fig:component-translation} \emph{do not} incorporate the scale filter and the re-detection module. 
	From Figure \ref{fig:component-translation}, we have the following observations: 
	(1) Deeper CNN features facilitate correlation filter based trackers in locating target object (the CT-VGGNet19 method outperforms both the CT-AlexNet and CT-HOGHOI methods) as the encoded semantic information within deep features are robust to significant appearance changes. 
	(2) Handcrafted features (HOG-HOI) with fine-grained spatial details are helpful for estimating scale changes. 
	The CT-HOGHOI method outperforms the CT-VGGNet19 method in terms of overlap success. 
	The CT-HOGHOI method performs better than the CT-AlexNet method using the AlexNet \cite{DBLP:conf/nips/KrizhevskySH12}. 
	Note that the CT-HOGHOI method significantly outperforms the CT-HOG features. 
	The results show the effectiveness of the proposed HOI features. 
	(3) The CT-HOGHOI-VGGNet19 method exploits both merits 
	of deep and handcrafted features and outperforms other alternative approaches in both distance precision and overlap success.

	{\flushleft \bf Feature Analysis on Scale Filter.} 
	We evaluate different types of features for learning the scale filter based on the CT-HOGHOI implementation.
	Figure \ref{fig:component-scale} shows that the scale filter using HOG features do not outperform that using both HOG and HOI features. 
	Consequently, we learn the scale filter using only HOG features for efficiency. 
	In addition, we implement the CT-HOGHOI-joint-scale-HOG approach, which updates the translation filter $\mathcal{A}_\mathrm{T}$ using the estimated scale in each frame as in the DSST \cite{DBLP:conf/bmvc/DanelljanKFW14} and MUSTer \cite{Hong_2015_CVPR} 
	methods. 
	We observe the CT-HOGHOI-joint-scale-HOG approach performs worst 
	among the compared methods as slight inaccuracy in scale estimation always causes a rapid degradation of the translation filter. 
	As a result, we use the ground-truth scale in the first frame to update the translation filter. 
	
	{\flushleft \bf Re-Detection Module.} 
	We evaluate four re-detection schemes with the baseline CT-HOGHOI method. 
	Using the long-term filter based on correlation filter (CT), we compare two different schemes to update the SVM detector. 
	We implement the CT-HOGHOI-CF-SVM method 
	using the proposed passive-aggressive update scheme (see Section \ref{sec:svm}), 
	while the CT-HOGHOI-CF-SVMsv using the support vector update scheme \cite{DBLP:conf/eccv/ZhangMS14}. 
	Figure \ref{fig:component-detector} shows that the proposed passive-aggressive scheme performs slightly better. 
	The reason is that, by directly updating the hyperplane $\bh$ in \eqref{equ:svmupdate}, the passive-aggressive scheme makes use of \emph{all} the training data. In contrast, the support vector scheme uses a small \emph{subset} of training data (support vectors) to update model.
	As the long-term filter can be used as a detector as well, we implement the CT-HOGHOI-CF-CF method by replacing the SVM detector with the long-term filter. 
	However, the CT-HOGHOI-CF-CF methd does not perform as well as the CT-HOGHOI-CF-SVM. 
	For the CT-HOGHOI-LSTM-SVM method, we use the hidden states of an LSTM network as the long-term filter. 
	Due to limited training data, the CT-HOGHOI-LSTM-SVM method does not perform as well as the CT-HOGHOI-CF-SVM, which uses the correlation filter based long-term filter. 

		\begin{figure}
		\centering
		\setlength{\tabcolsep}{0.5mm}
		\begin{tabular}{cc}
			\includegraphics[width=0.235\textwidth]{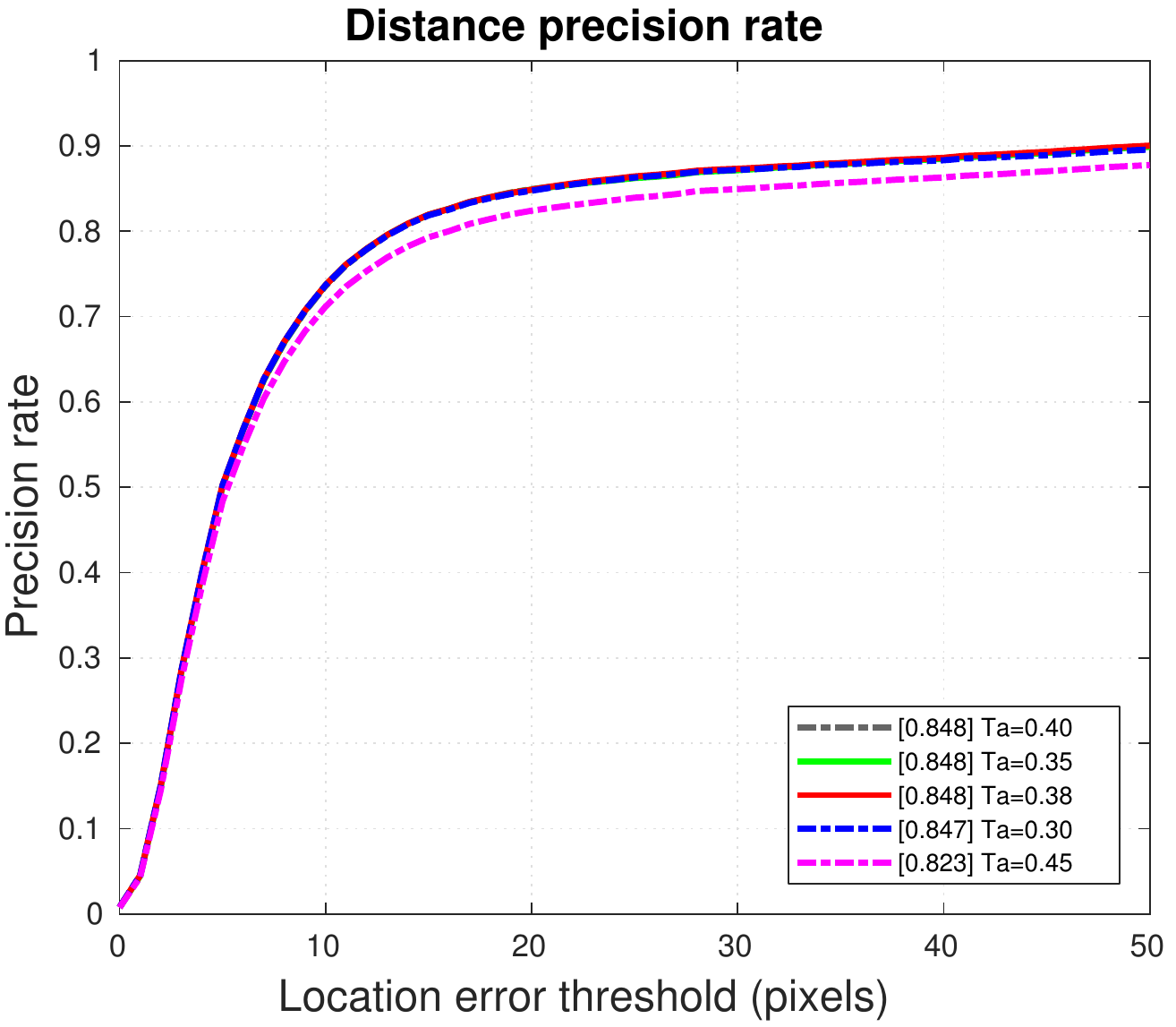} & 
			\includegraphics[width=0.235\textwidth]{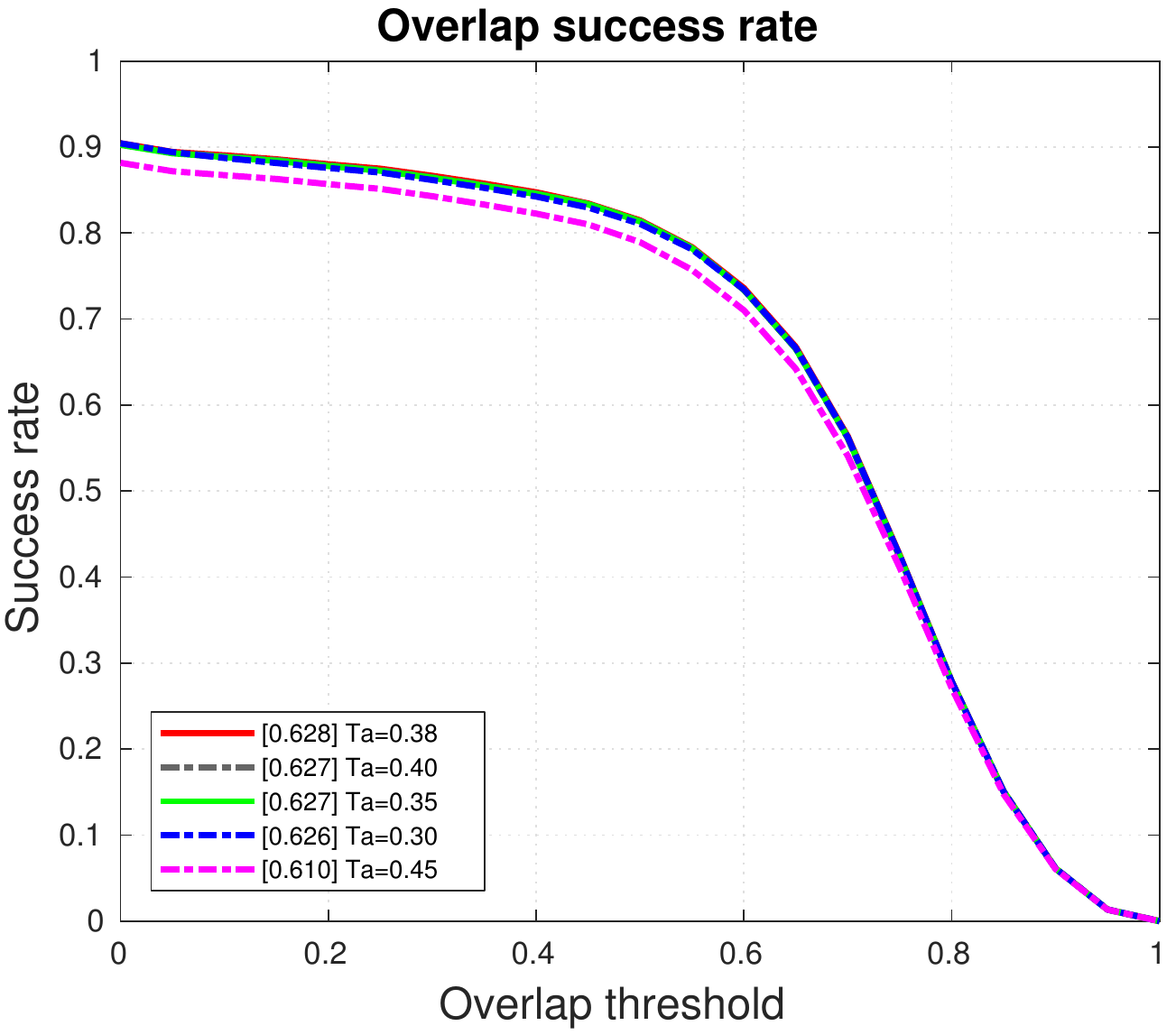} \\
		\end{tabular}
		\caption{\textbf{Sensitivity analysis on the OTB2013 dataset \cite{DBLP:conf/cvpr/WuLY13} under one pass evaluation (OPE)}. The legend of precision plots shows the distance precision scores at 20 pixels. The legend of success plots contains the overlap success scores with the area under the curve (AUC).
		}
		\label{ablation-ta}
	\end{figure}
	
	\subsection{Sensitivity Analysis}
\begin{figure}
		\centering
		\includegraphics[width=.42\textwidth]{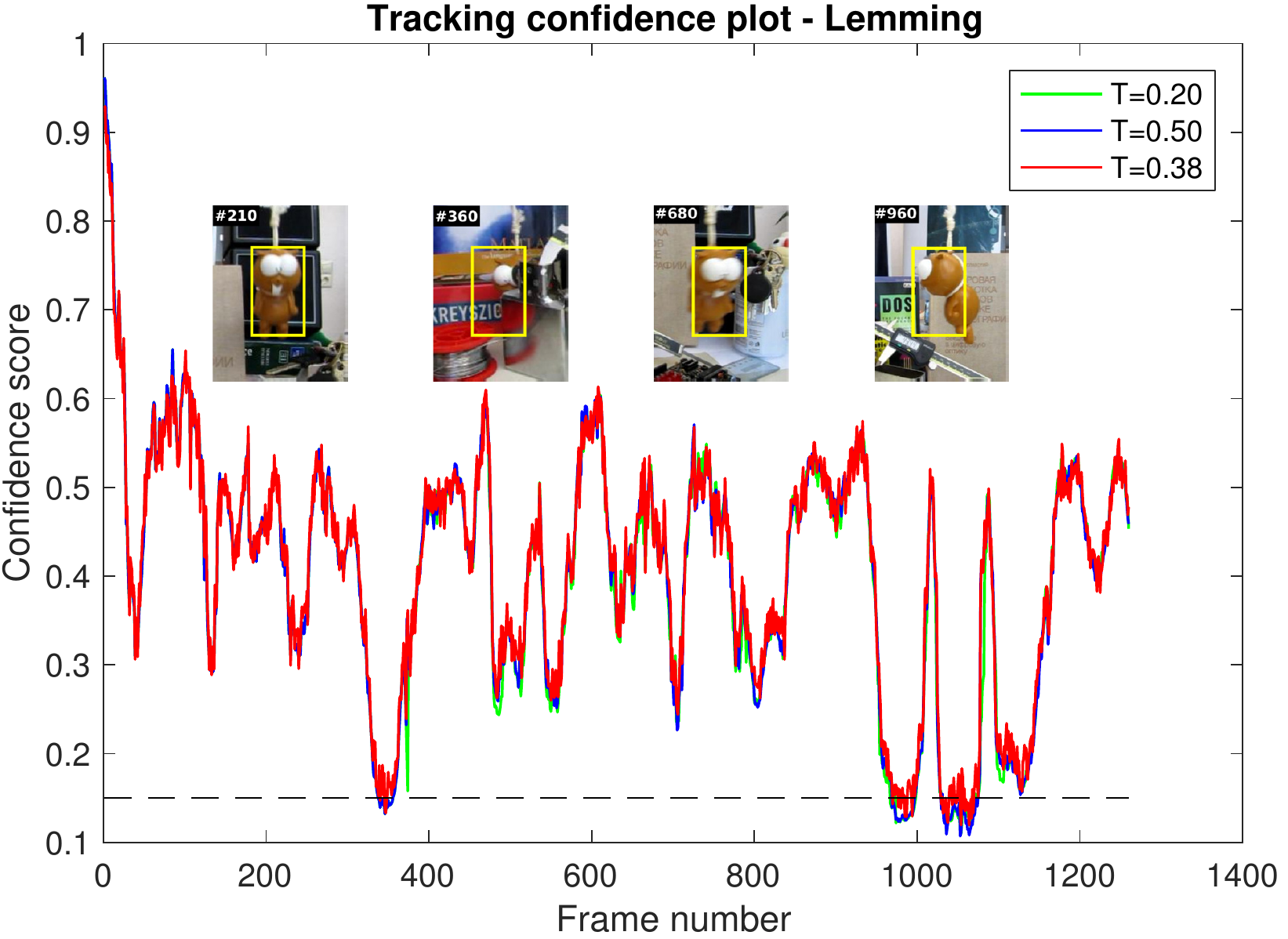} \\
		\caption{
			\textbf{Sensitivity of the threshold selection.}
			The tracking confidence scores of the proposed method on the \textit{lemming} sequence \cite{DBLP:conf/cvpr/WuLY13} are computed with different stability threshold values to update the long-term filter. 
			The confidence scores are not sensitive with the threshold values between 0.2 to 0.5.}
		\label{fig:confidence-update}
	\end{figure}

	We discuss how we set three important thresholds: 
	(1) re-detection threshold $T_r$ for activating the detection module; 
	(2) acceptance threshold $T_a$ for adopting detection results; and 
	(3) stability threshold $T_s$ for conservatively updating the long-term filter. 
	We use the tracking results on the \textit{lemming} sequence for illustration. 
	As shown in Figure~\ref{fig:confidence-nodetection}, 
	we implement a baseline CT-HOGHOI method, which does not incorporate a
	re-detection module and fails to track the target after the 360-th frame. 
	The tracked results cover a variety of tracking successes and failures.
	We apply the long-term filter to compute the confidence scores of the tracked results. 
	We fine-tune the stability threshold values for conservatively updating the long-term filter and examine the correlation between the confidence scores and the overlap success rates. 
	We empirically find that when targets undergo occlusion, the confidence scores are generally smaller than 0.15. 
	As such, we set the re-detection threshold $T_r$ to 0.15. 
	For setting the acceptance threshold $T_a$, we use a larger value to accept the detection results conservatively. 
	We initialize the acceptance threshold $T_a$ two times of the re-detection threshold $T_r$.
	We use the grid search and empirically set the acceptance threshold $T_a$ to 0.38 for better
	results. Figure \ref{ablation-ta} shows that the performance is not sensitive to $T_a$ between 0.3 and 0.45.
	For setting the stability threshold $T_s$, we show in Figure~\ref{fig:confidence-update}
	the confidence scores using different threshold values to update the long-term filter. 
	Figure~\ref{fig:confidence-update} shows that the performance is not sensitive to a reasonable selection of stability threshold $T_s$ (0.2 - 0.5). 
	We thus set $T_s$ equal to $T_a$ as 0.38. \\

	\subsection{Exploiting Contextual Cues}
	\label{sec:pa}
\begin{figure}
		\centering
		\setlength{\tabcolsep}{0pt}
		\begin{tabular}{cc}
			\includegraphics[width=.25\textwidth]{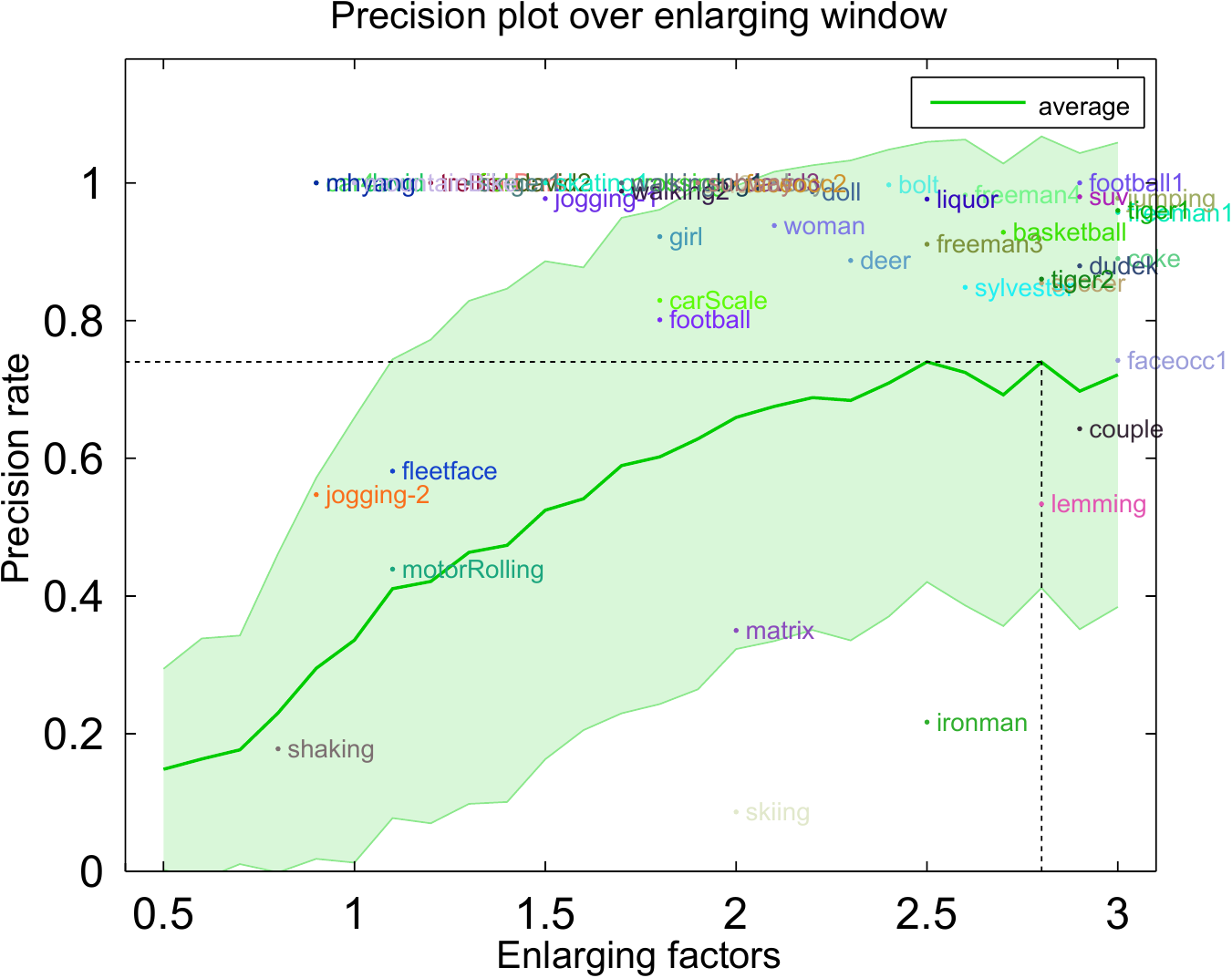} &
			\includegraphics[width=.25\textwidth]{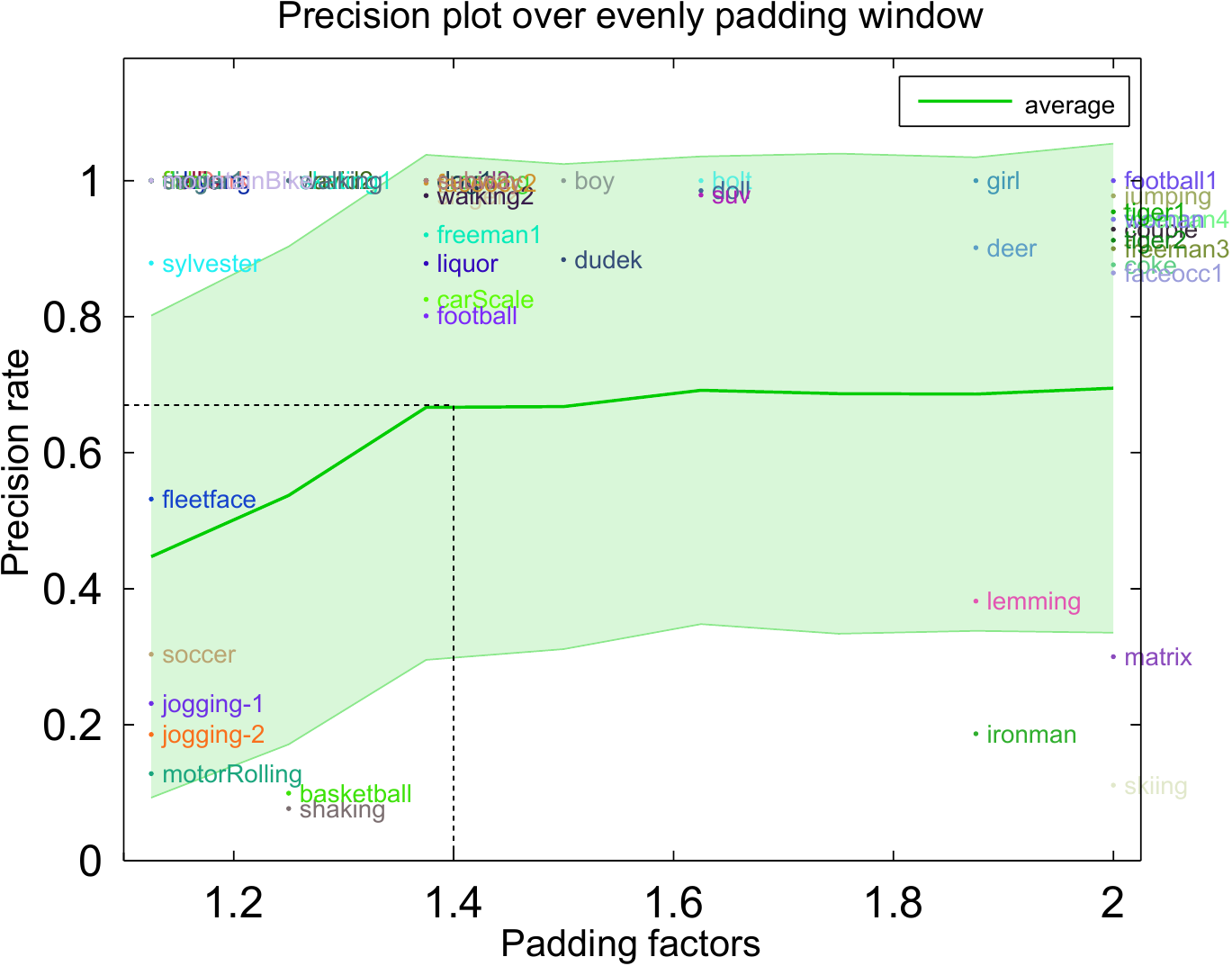} \\
			(a) & (b) \\
		\end{tabular} 
		\caption{
			\textbf{Validation of incorporating context.}
			Distance precision results on learning translation filter with different sizes of context area on the OTB2013 dataset \cite{DBLP:conf/cvpr/WuLY13}. (a) Enlarge the bounding box of target by a given factor to incorporate surrounding context. (b) Uniformly pad the width and height of target bounding box by a factor proportional to the target size. 
			The horizontal axis indicates the context area size that gives rise to the best result on the particular sequence. 
			The green line is the averaged result, and the shaded area shows the standard deviation.}
		\label{fig:context}
	\end{figure}

	We explore two approaches for incorporating surrounding context for learning the translation filter $\mathcal{A}_\mathrm{T}$: 
	(1) scaling: enlarging the target bounding box by scaling the bounding box with a given factor, and (2) padding: evenly padding the width and height of the bounding box with a certain size.
	We plot the tracking results in terms of distance precision on 50 benchmark sequences \cite{DBLP:conf/cvpr/WuLY13} in Figure \ref{fig:context}.
	The results show that the performance of the translation filter is sensitive to the padding size of surrounding context on target objects.
	For the target objects with smaller aspect ratios, e.g., \textit{jogging} and \textit{walking}, it performs better with evenly-padded context areas in precise localization. 
	This motivates us to exploit the merits of these two approaches simultaneously.
	From Figure \ref{fig:context}(a), we find that a scaling factor of 2.8 leads to good results.
	For the target object with a small aspect ratio (e.g., a pedestrian), we find that padding the bounding box with 1.4 times of height in the vertical directions yields improved results as shown in Figure \ref{fig:context}(b).
	This heuristic, despite its simplicity, provides a moderate improvement in locating target objects. For example, the overall distance precision rate on the OTB2013 dataset increases from 81.6\% to 84.8\%.

	\begin{figure*}
		\centering
		\footnotesize
		\setlength{\tabcolsep}{0pt}
		\begin{tabular}{cccccc}
			\includegraphics[trim = 0mm 20mm 30mm 0mm, clip,width=.165\textwidth]{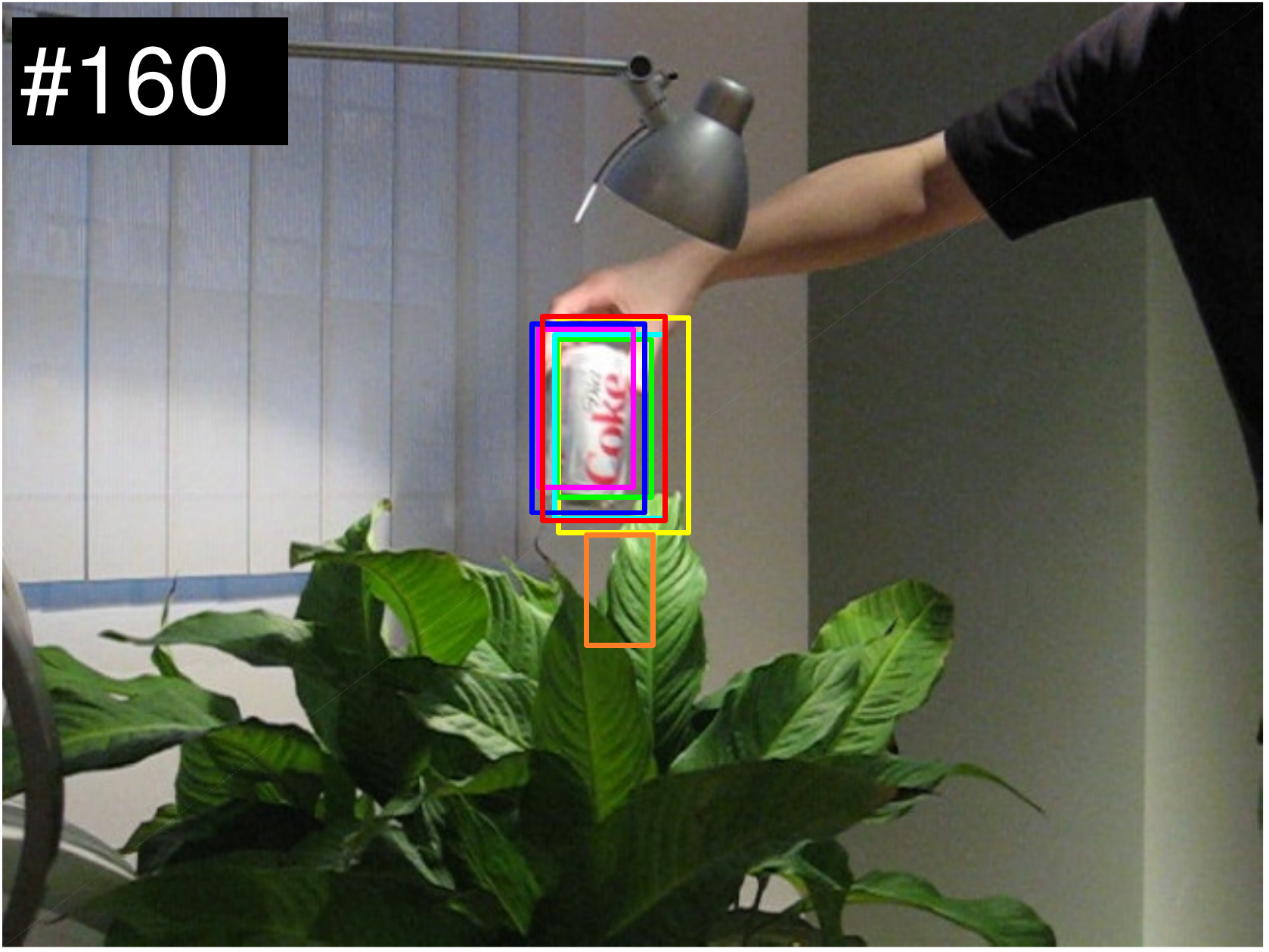} &
			\includegraphics[trim = 0mm 20mm 30mm 0mm, clip,width=.165\textwidth]{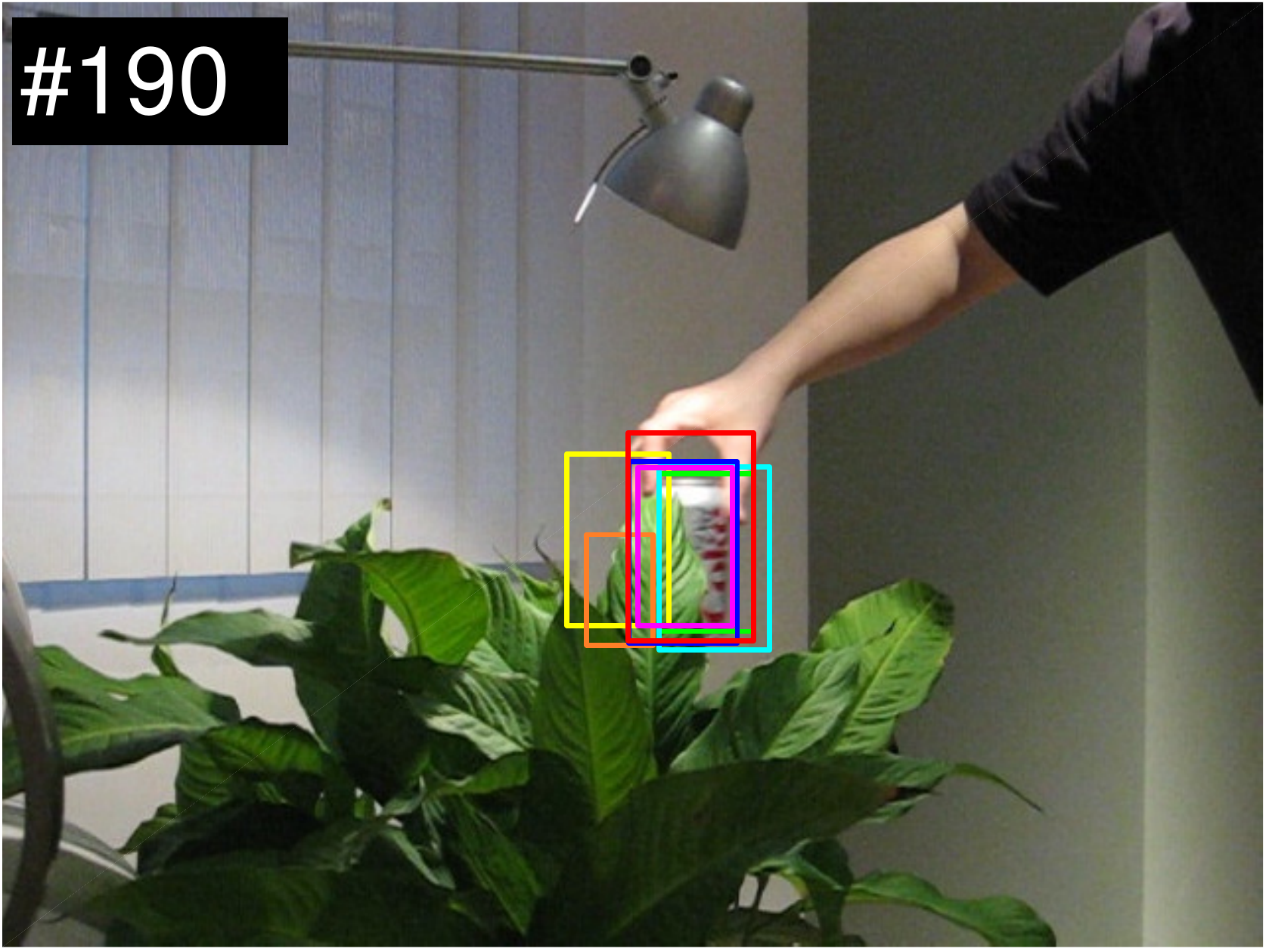} &
			\includegraphics[trim = 0mm 20mm 30mm 0mm, clip,width=.165\textwidth]{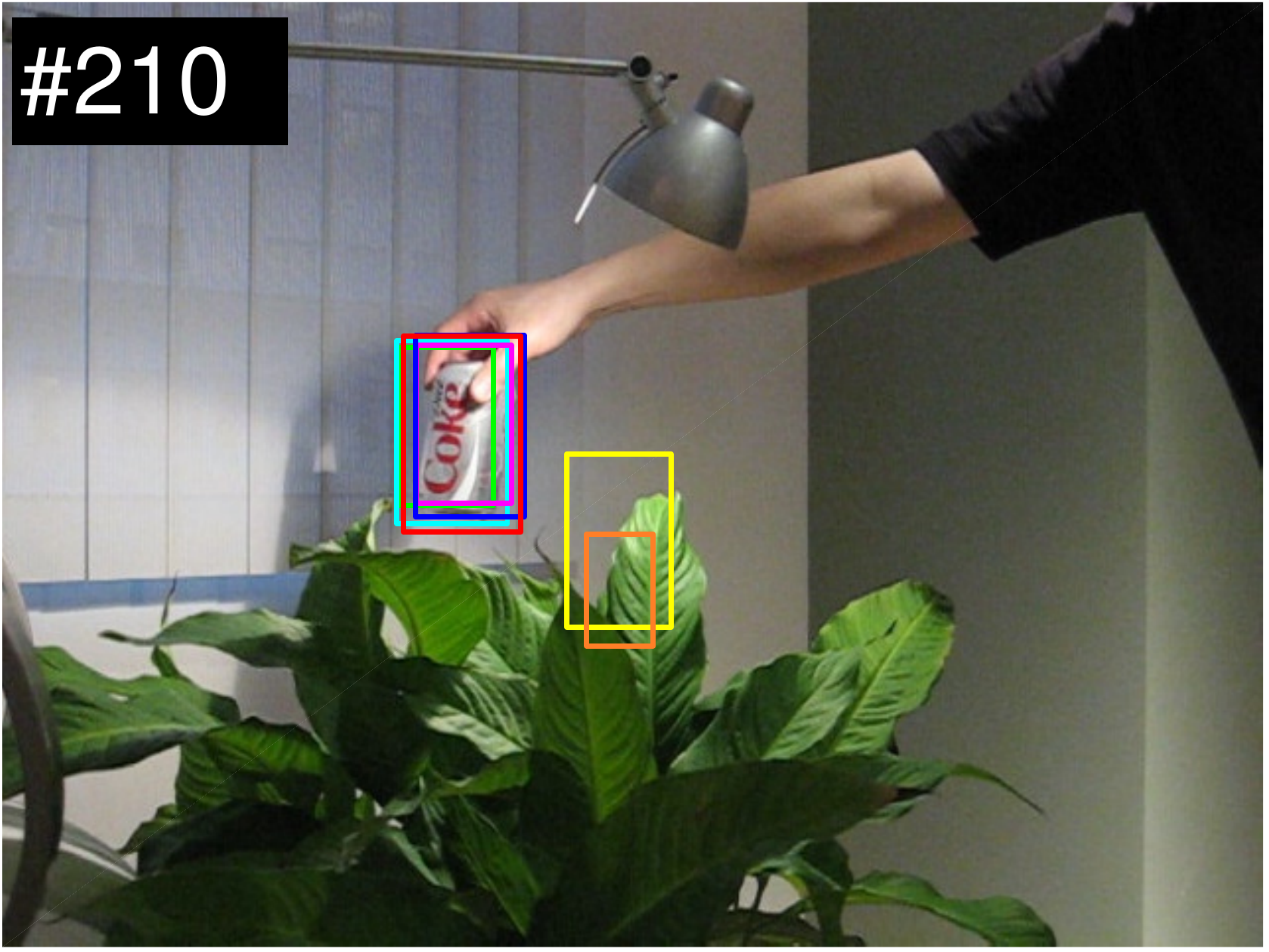} &
			\includegraphics[trim = 0mm 20mm 30mm 0mm, clip,width=.165\textwidth]{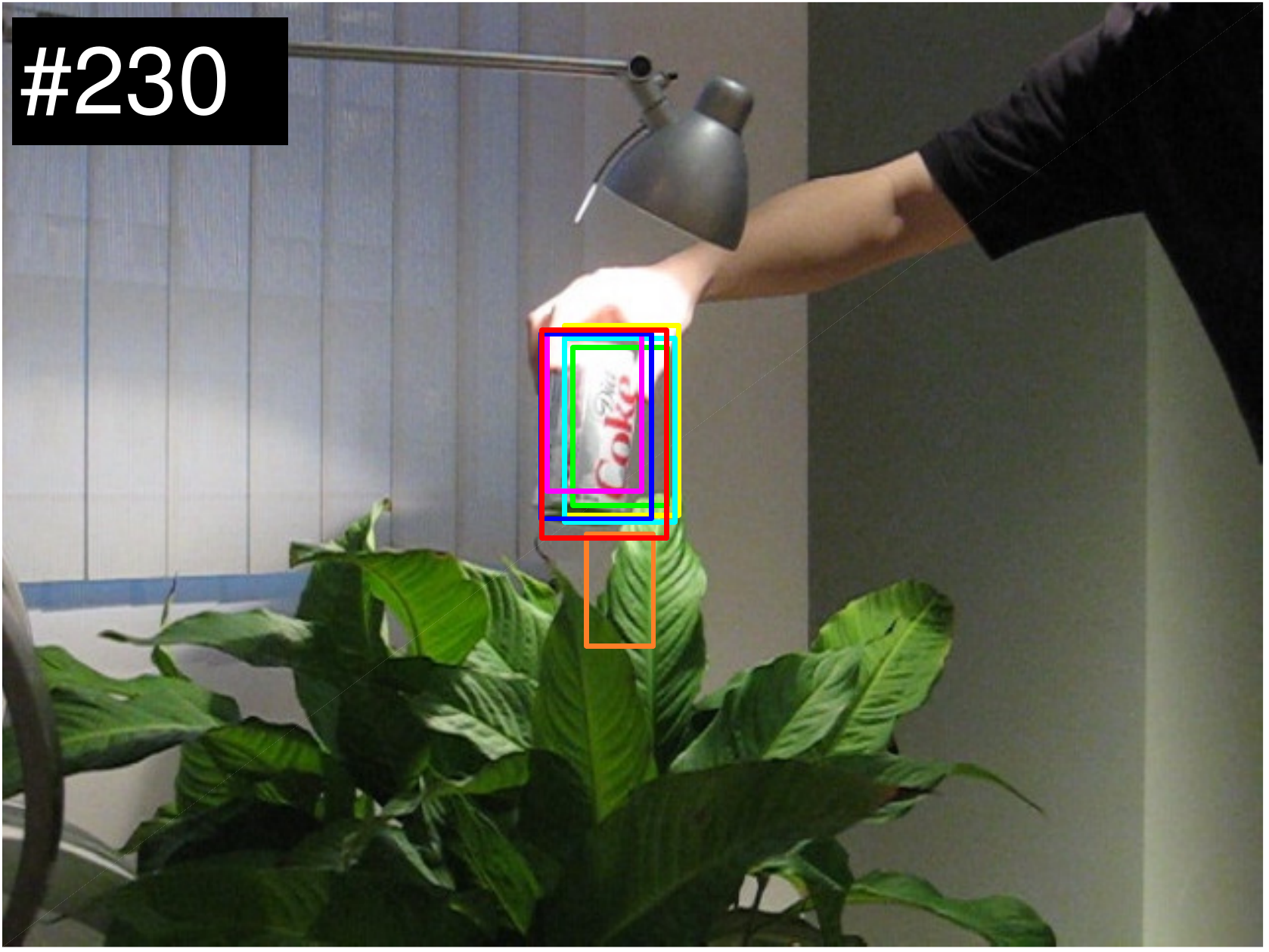} &
			\includegraphics[trim = 0mm 20mm 30mm 0mm, clip,width=.165\textwidth]{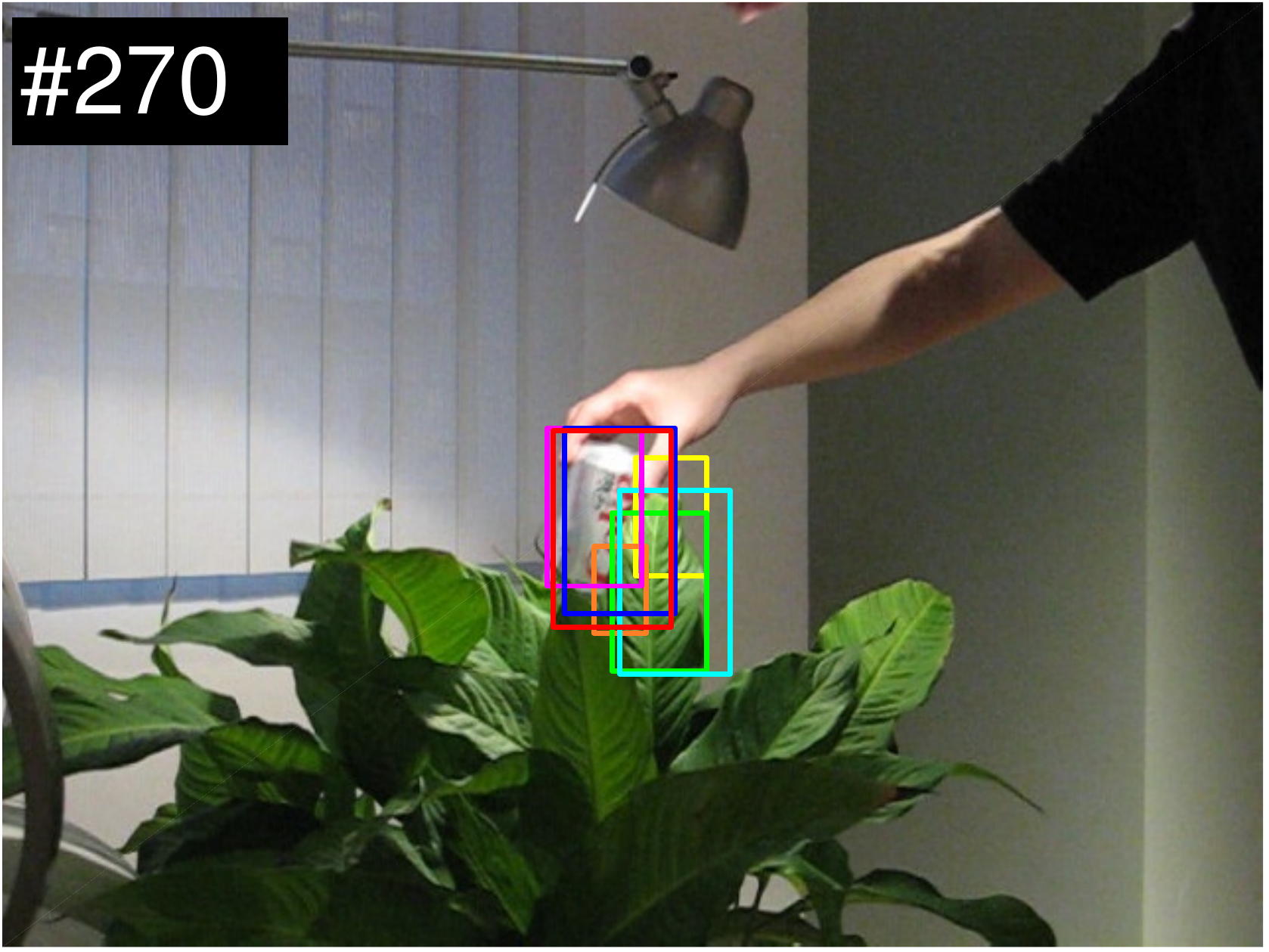} &
			\includegraphics[trim = 0mm 20mm 30mm 0mm, clip,width=.165\textwidth]{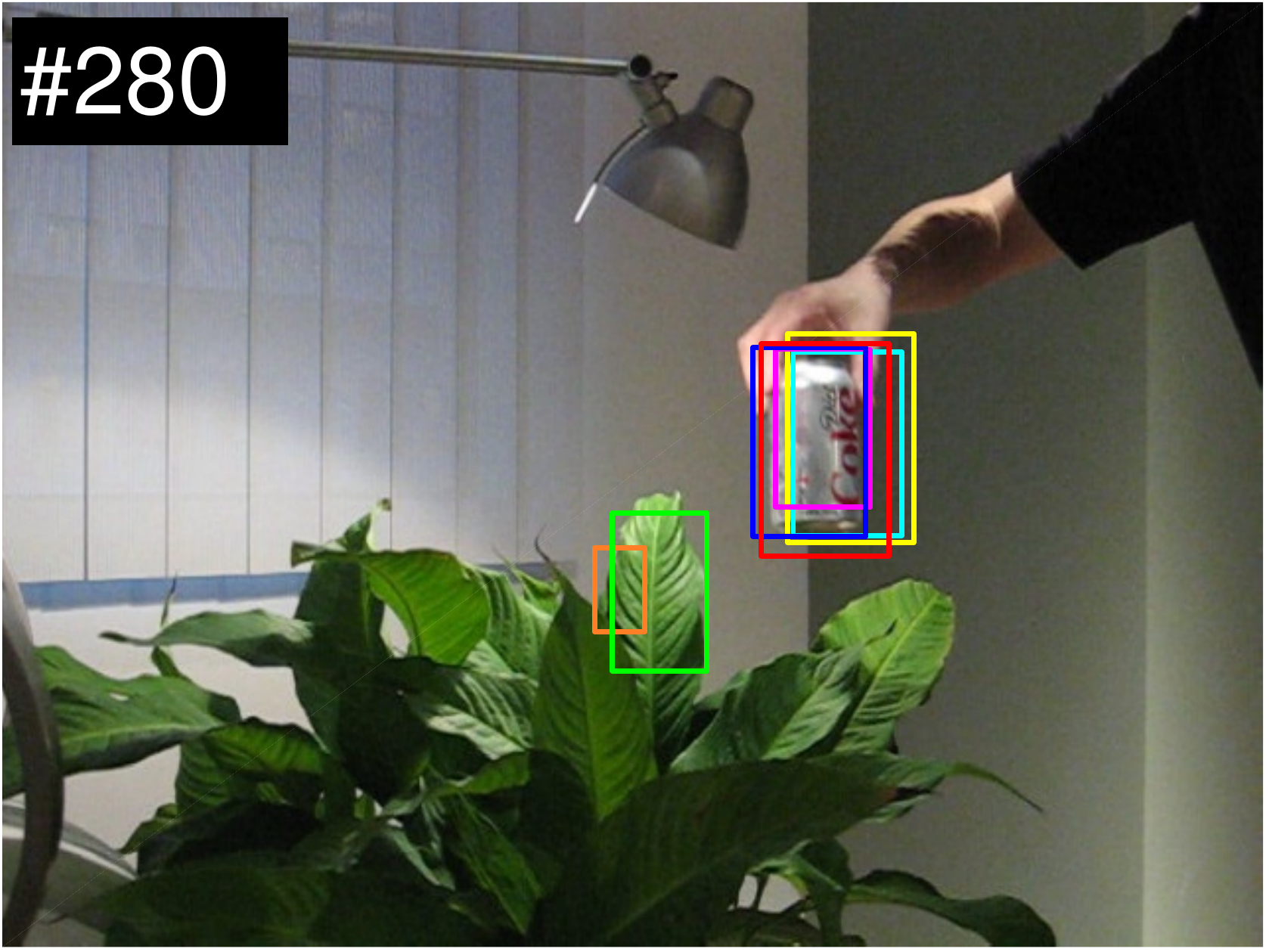} \\
			\includegraphics[trim = 0mm 10.5mm 50mm 0mm, clip,width=.165\textwidth]{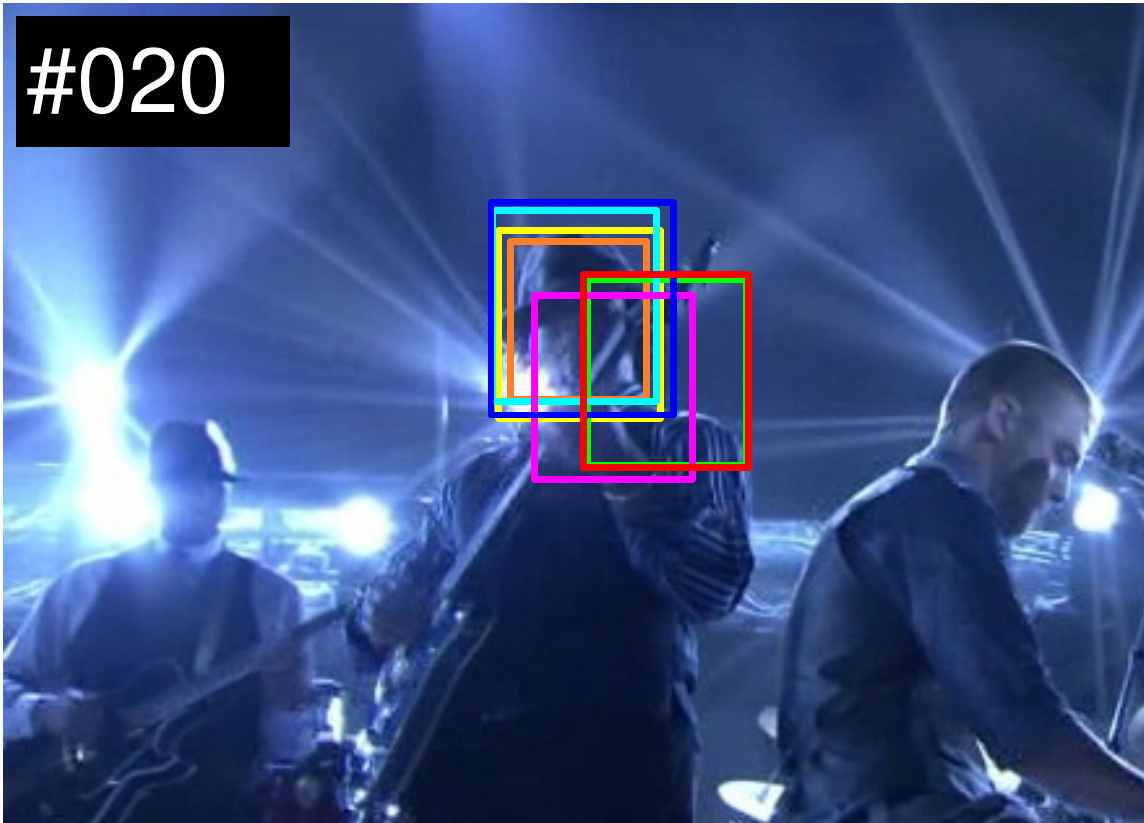} &
			\includegraphics[trim = 0mm 10.5mm 50mm 0mm, clip,width=.165\textwidth]{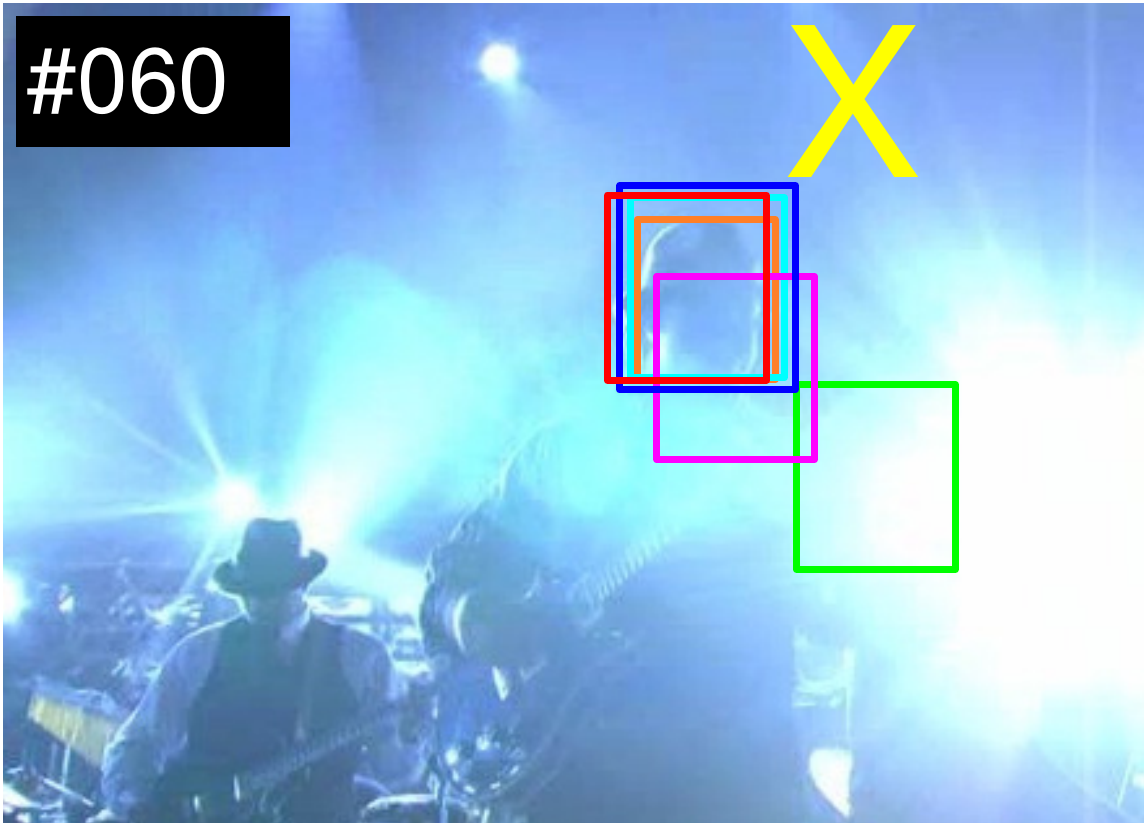} &
			\includegraphics[trim = 0mm 10.5mm 50mm 0mm, clip,width=.165\textwidth]{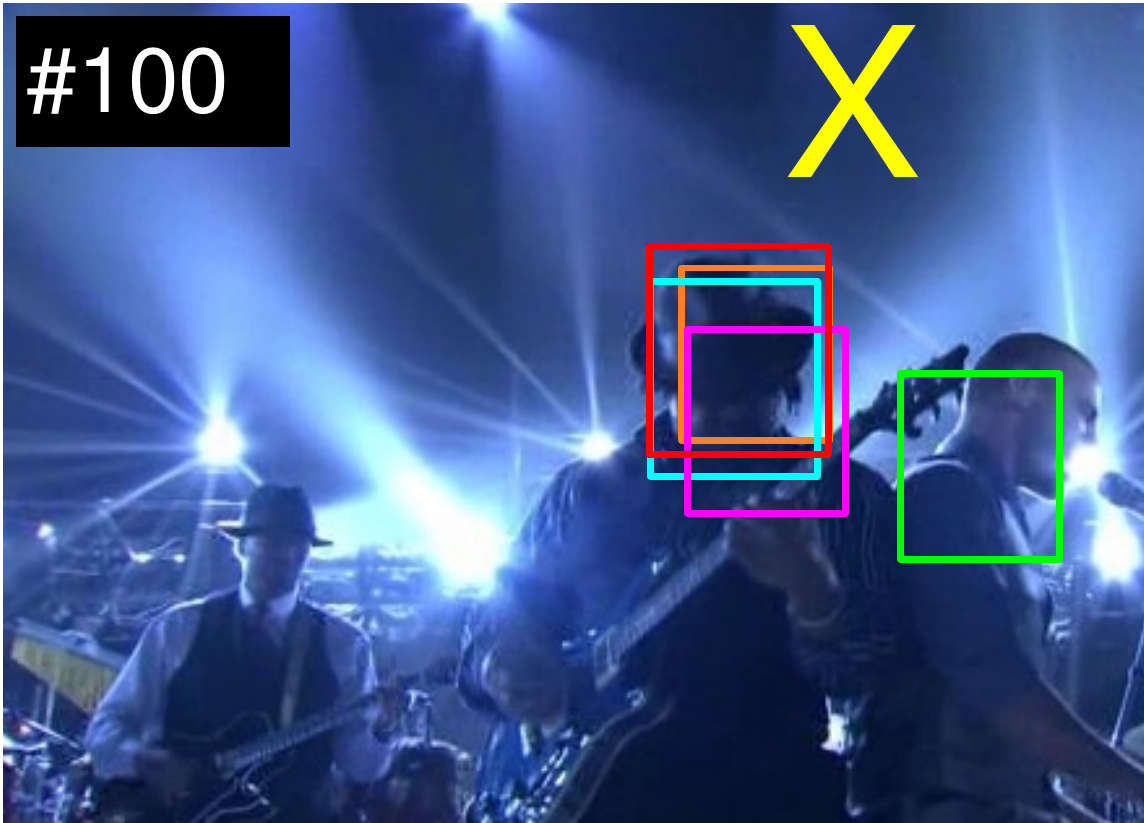} &
			\includegraphics[trim = 0mm 10.5mm 50mm 0mm, clip,width=.165\textwidth]{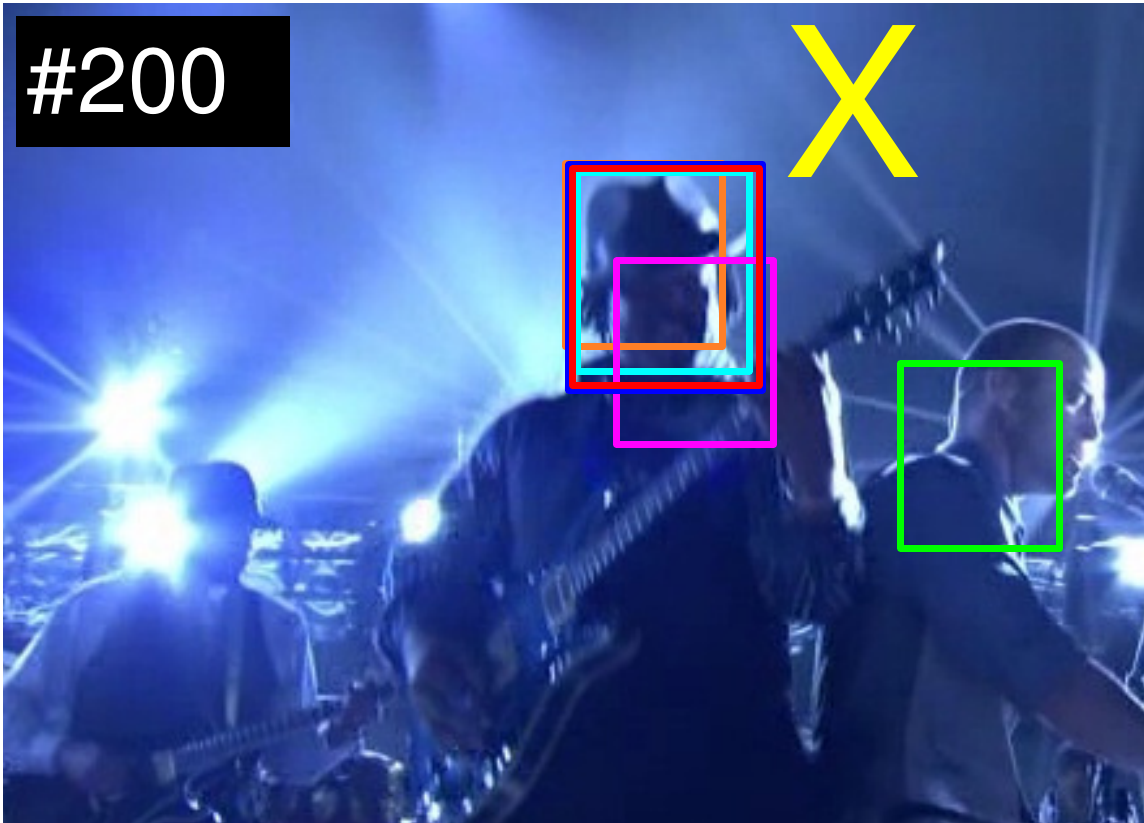} &
			\includegraphics[trim = 0mm 10.5mm 50mm 0mm, clip,width=.165\textwidth]{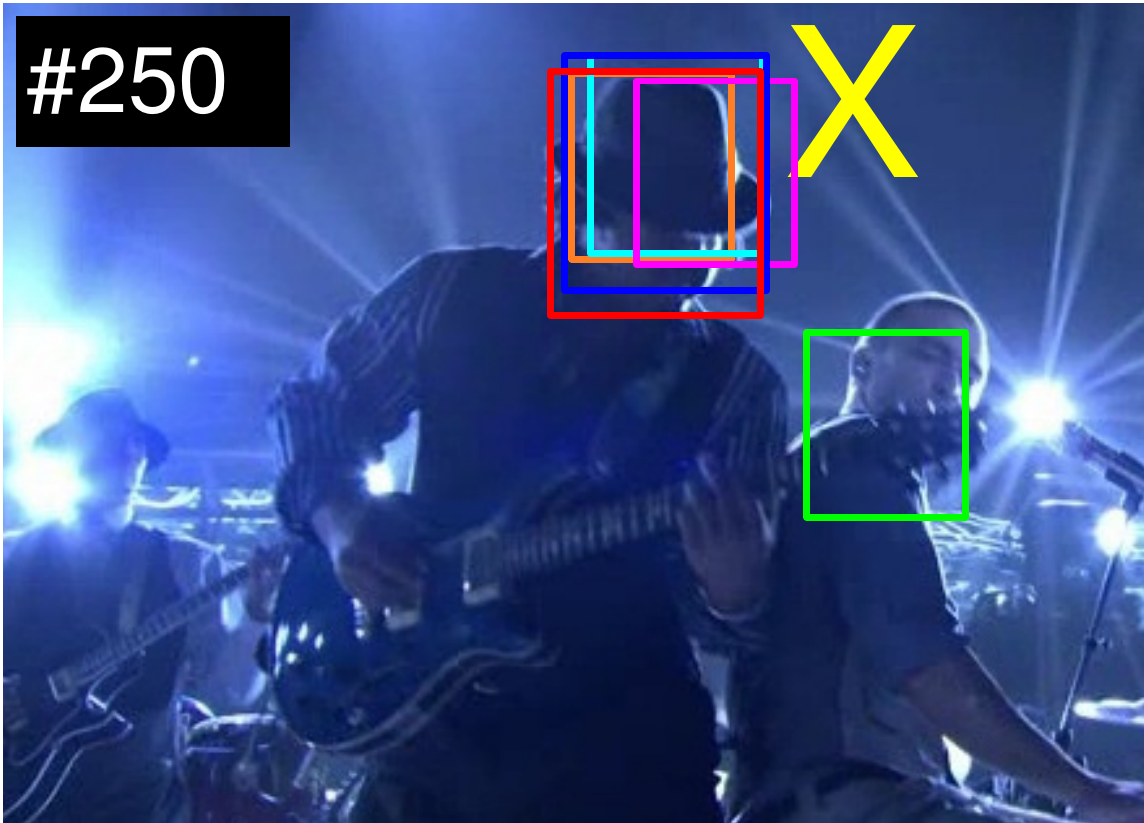} &
			\includegraphics[trim = 0mm 10.5mm 50mm 0mm, clip,width=.165\textwidth]{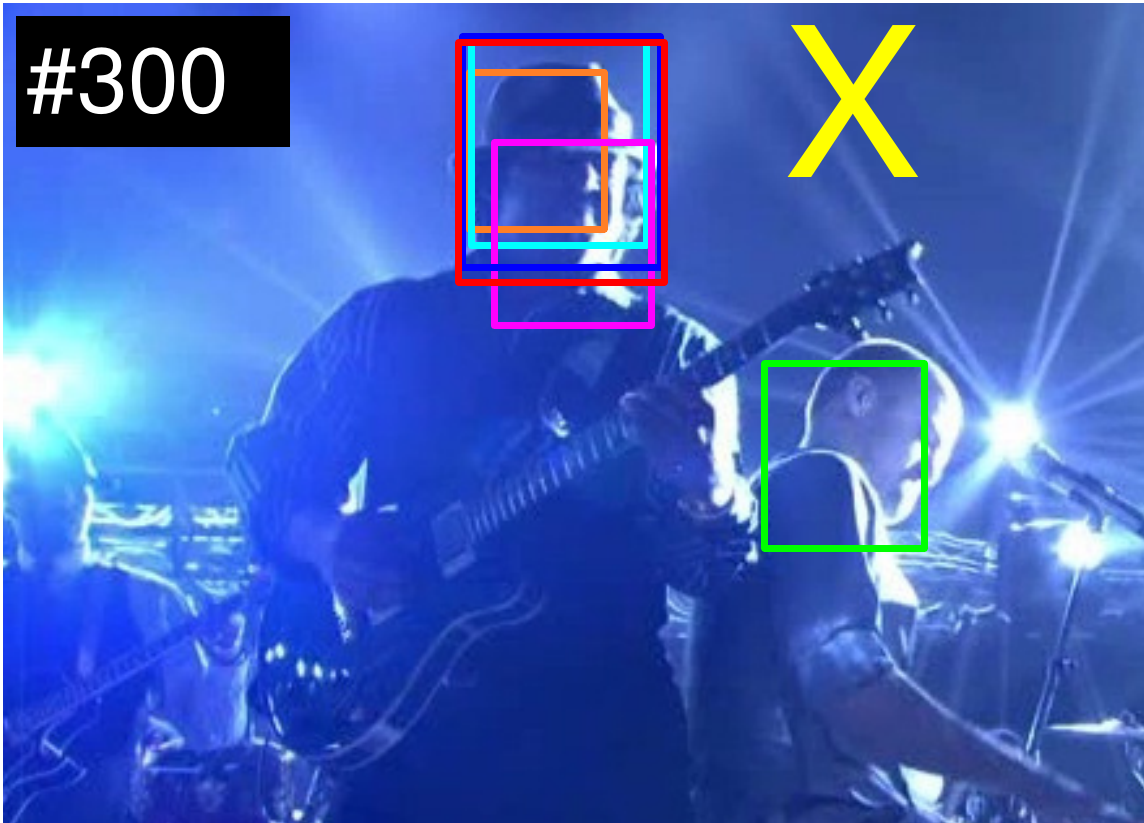} \\
			\includegraphics[trim = 0mm 10mm 50mm 0mm, clip,width=.165\textwidth]{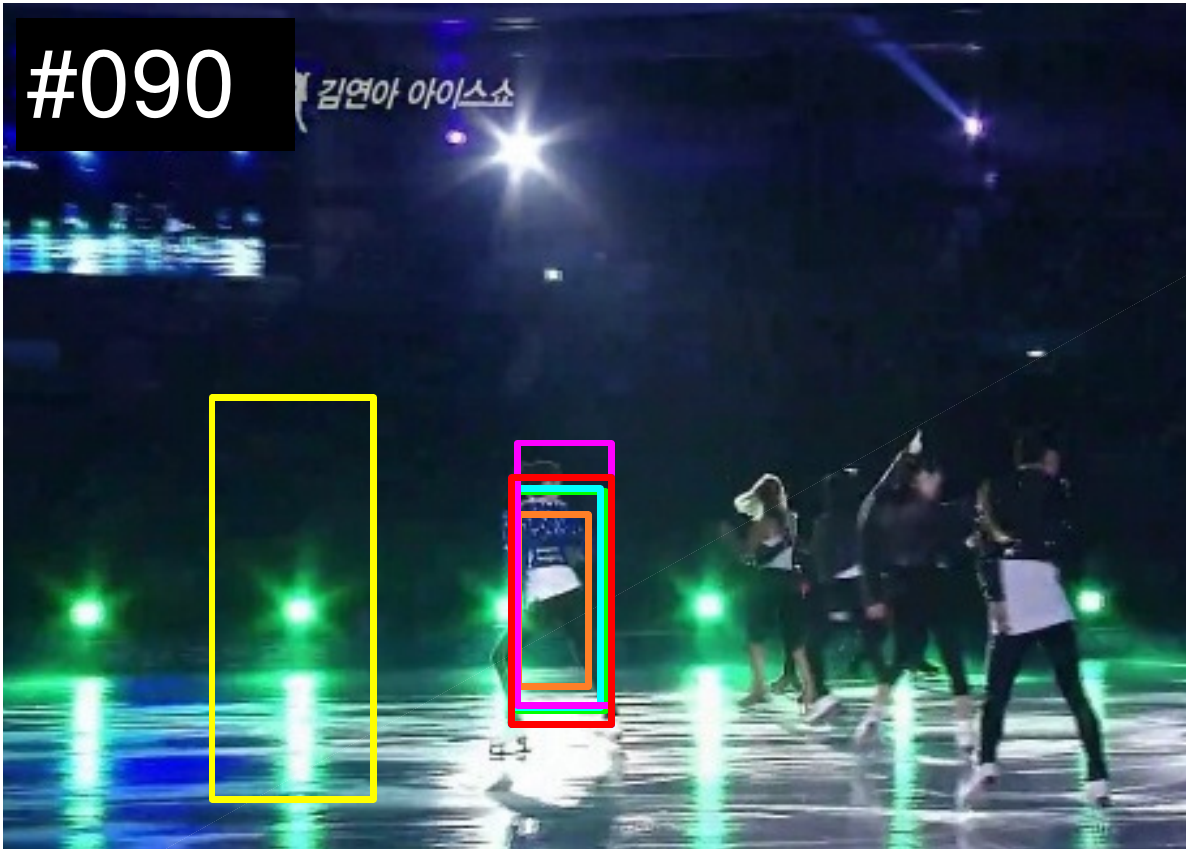} &
			\includegraphics[trim = 0mm 10mm 50mm 0mm, clip,width=.165\textwidth]{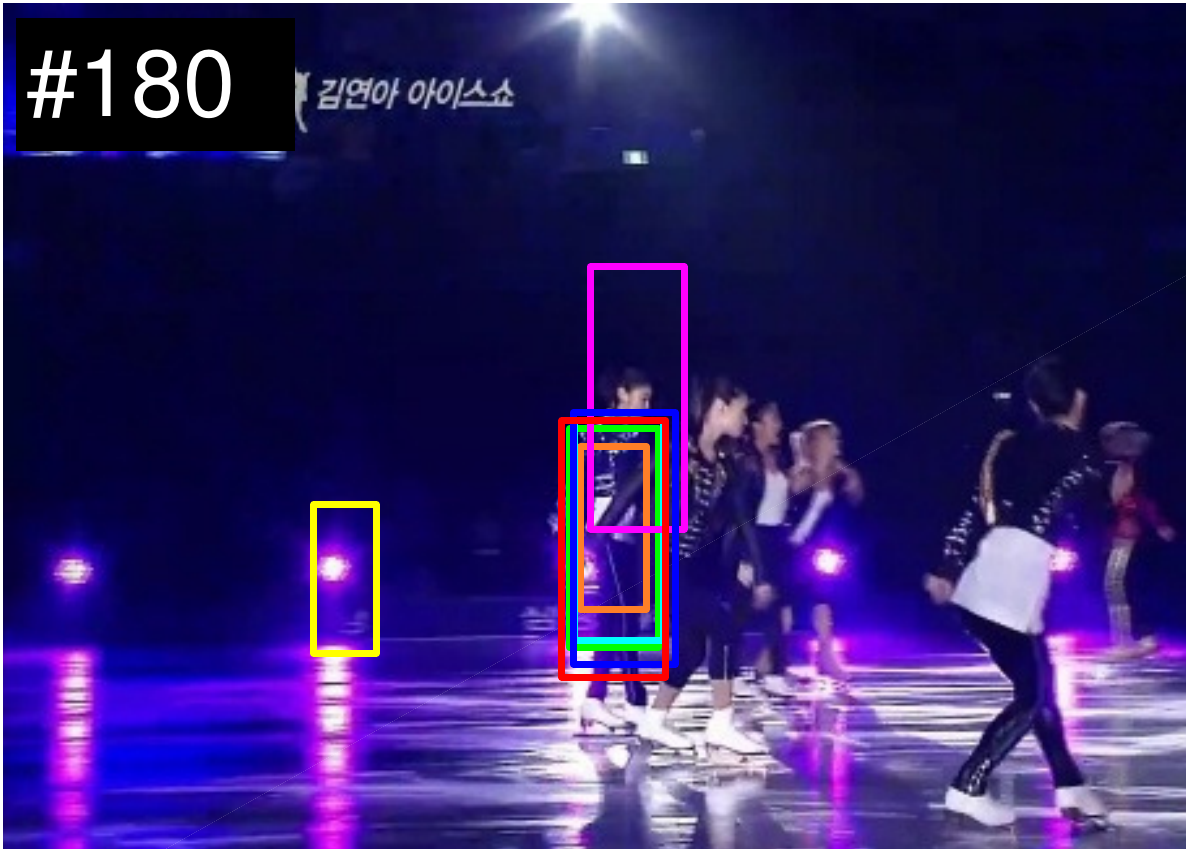} &
			\includegraphics[trim = 0mm 10mm 50mm 0mm, clip,width=.165\textwidth]{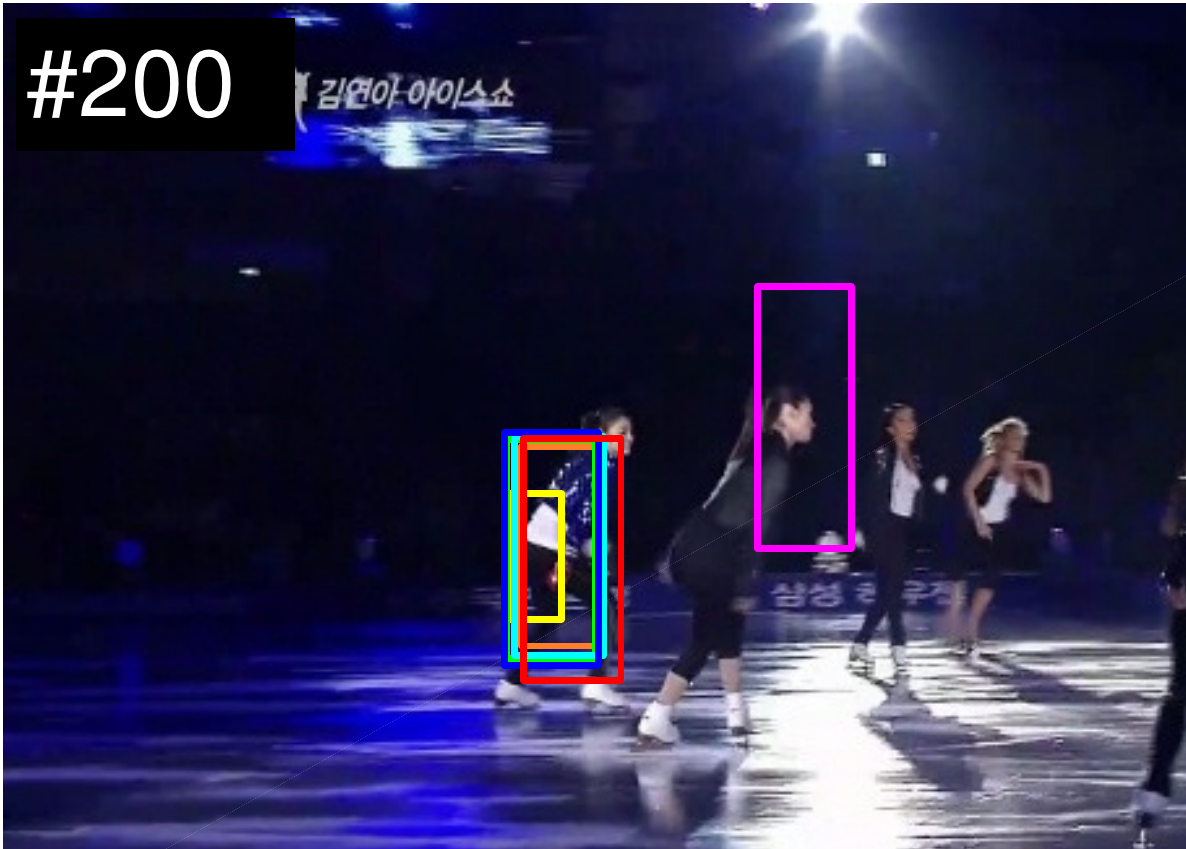} &
			\includegraphics[trim = 0mm 10mm 50mm 0mm, clip,width=.165\textwidth]{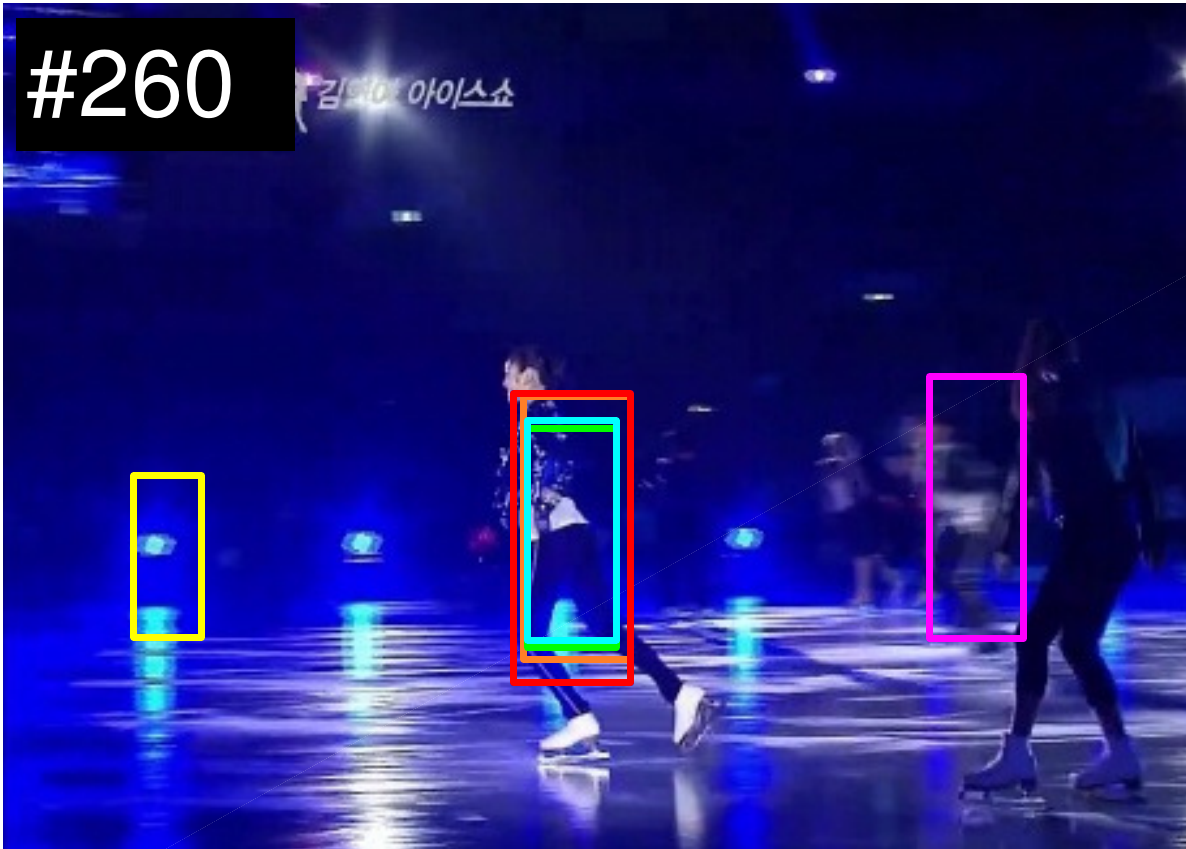} &
			\includegraphics[trim = 0mm 10mm 50mm 0mm, clip,width=.165\textwidth]{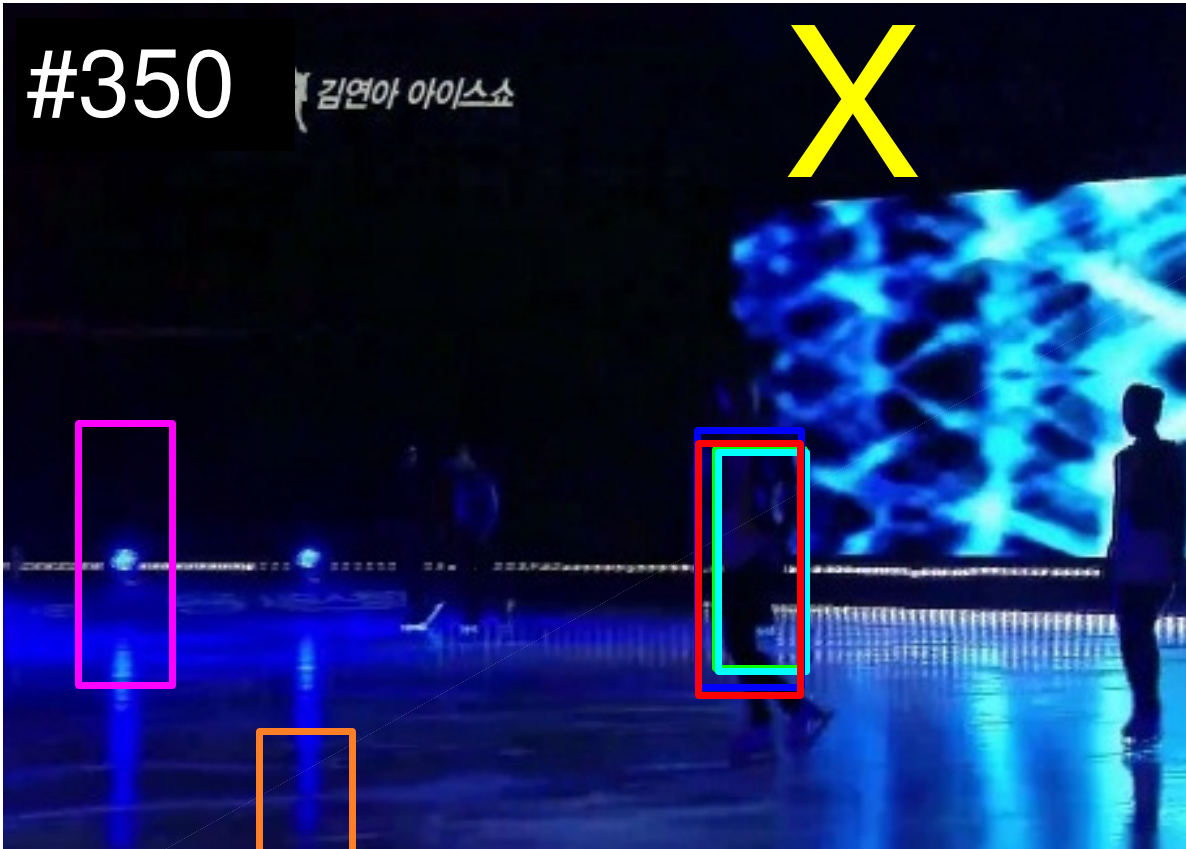} & 
			\includegraphics[trim = 0mm 10mm 50mm 0mm, clip,width=.165\textwidth]{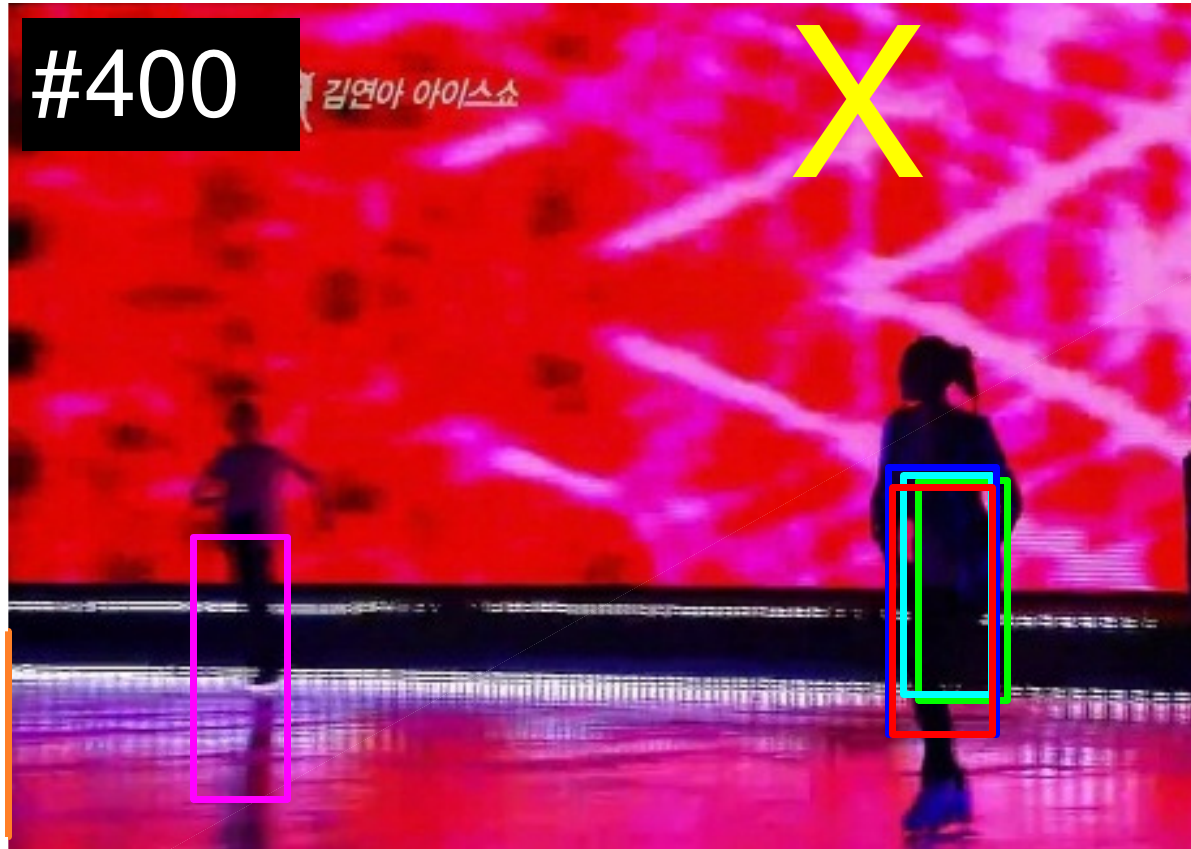} \\
			\includegraphics[trim = 0mm 2mm 2mm 0mm, clip,width=.165\textwidth]{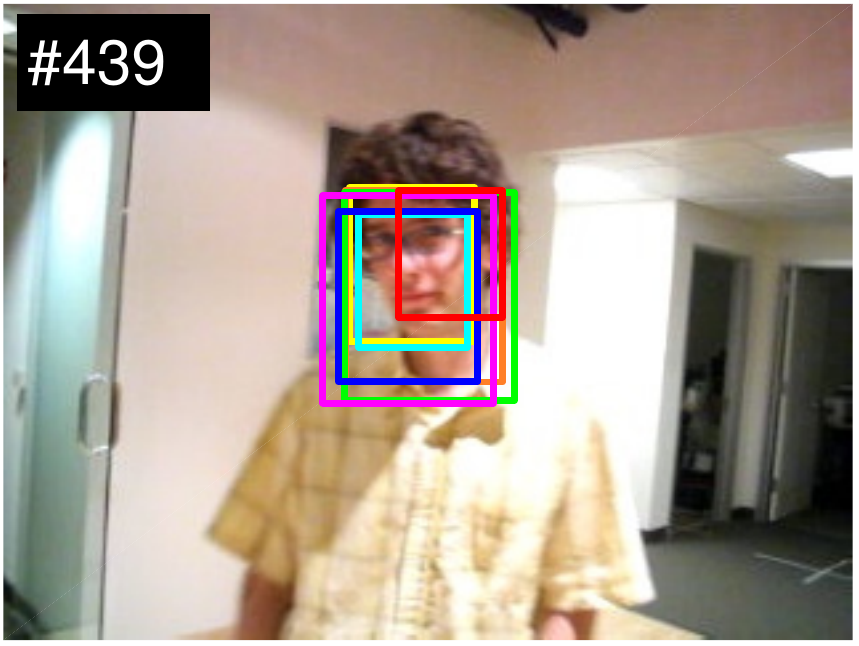} &
			\includegraphics[trim = 0mm 2mm 2mm 0mm, clip,width=.165\textwidth]{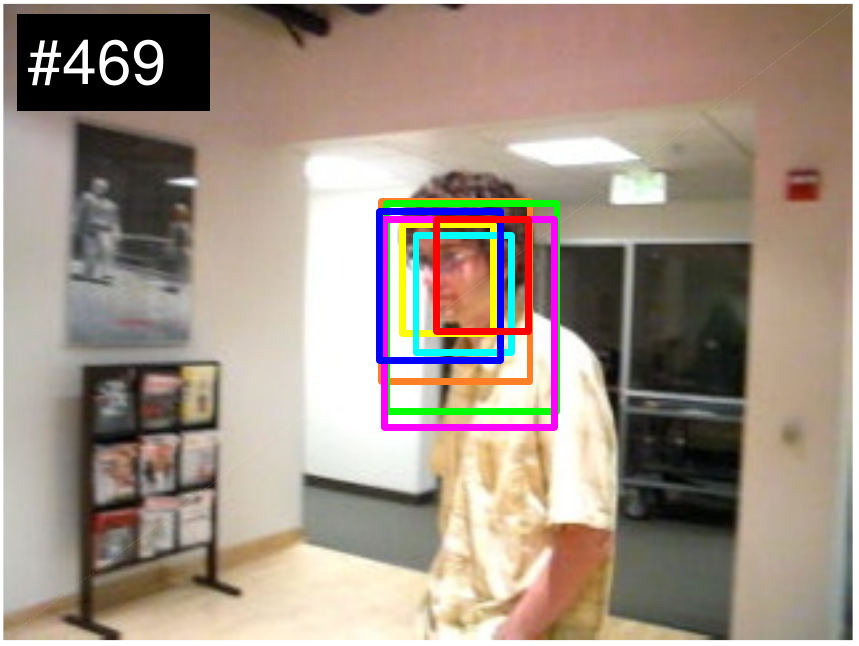} &
			\includegraphics[trim = 0mm 2mm 2mm 0mm, clip,width=.165\textwidth]{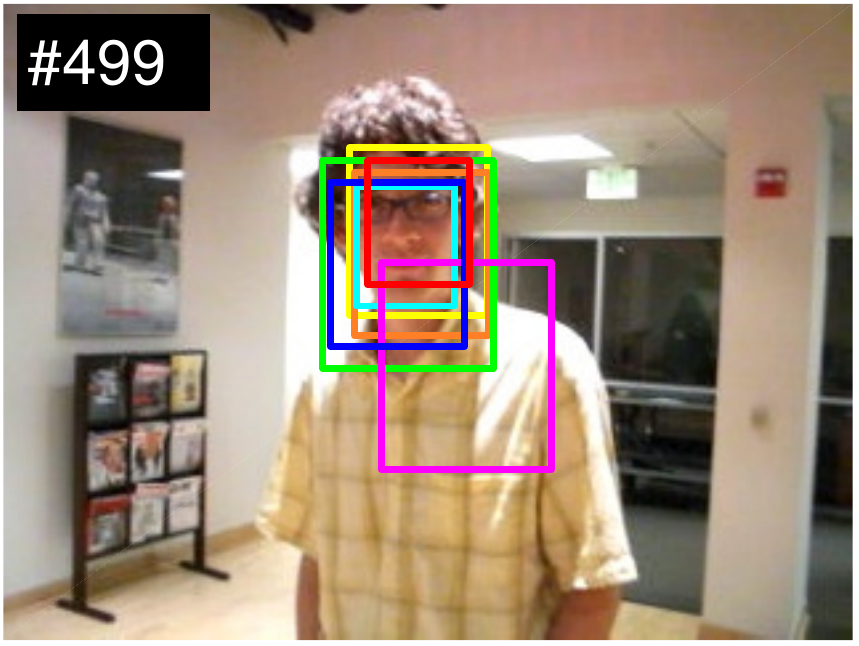} &
			\includegraphics[trim = 0mm 2mm 2mm 0mm, clip,width=.165\textwidth]{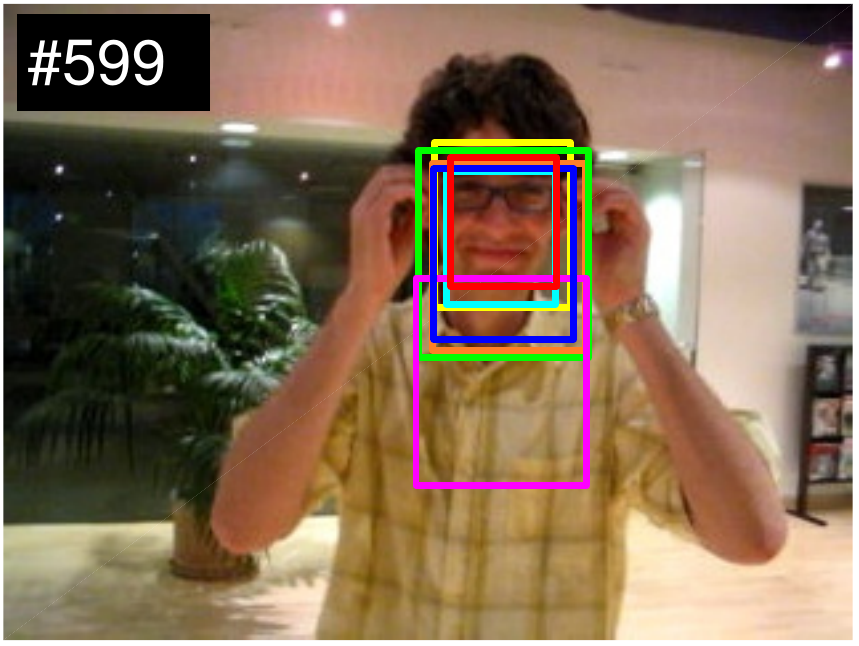} &
			\includegraphics[trim = 0mm 2mm 2mm 0mm, clip,width=.165\textwidth]{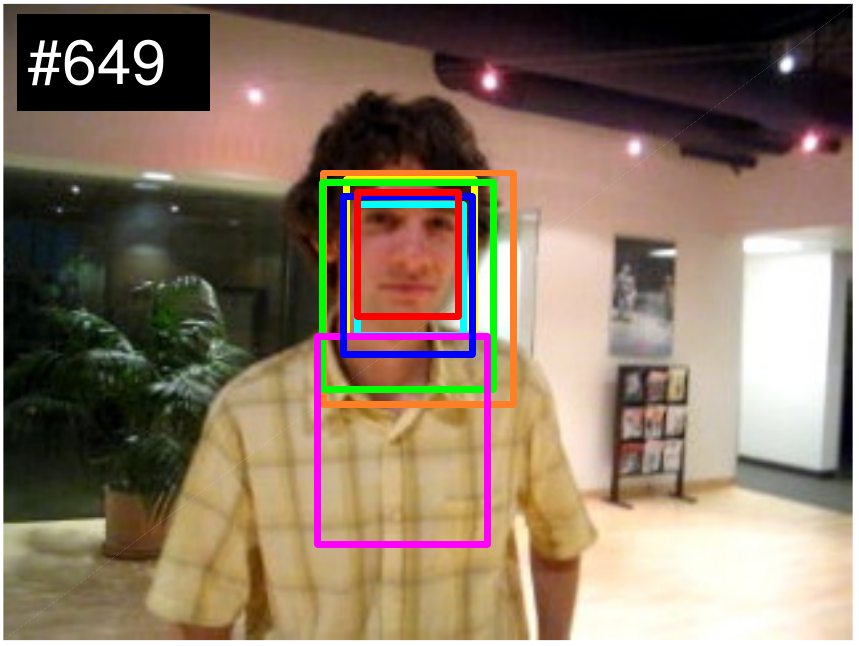} &
			\includegraphics[trim = 0mm 2mm 2mm 0mm, clip,width=.165\textwidth]{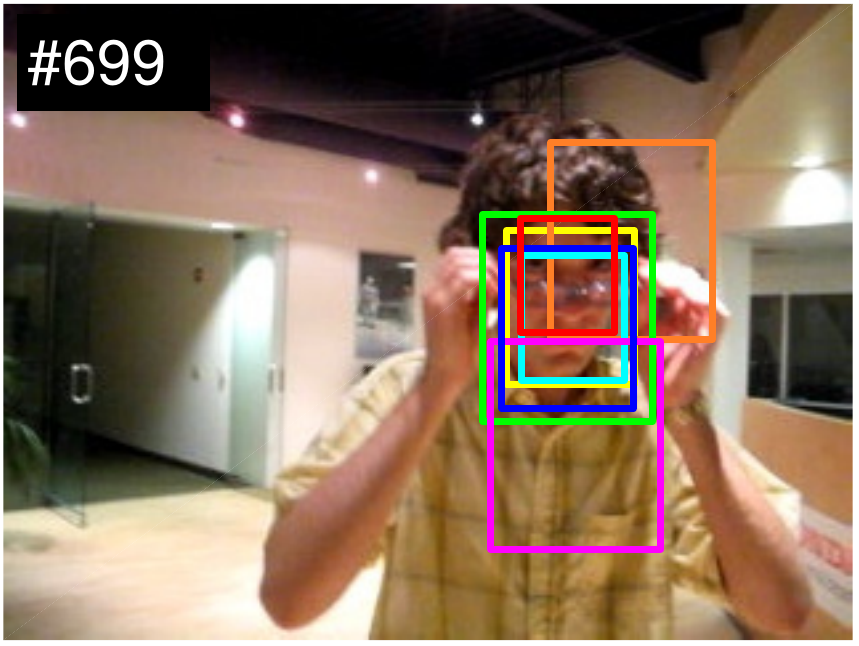} \\
			\includegraphics[trim = 0mm 2mm 2mm 0mm, clip,width=.165\textwidth]{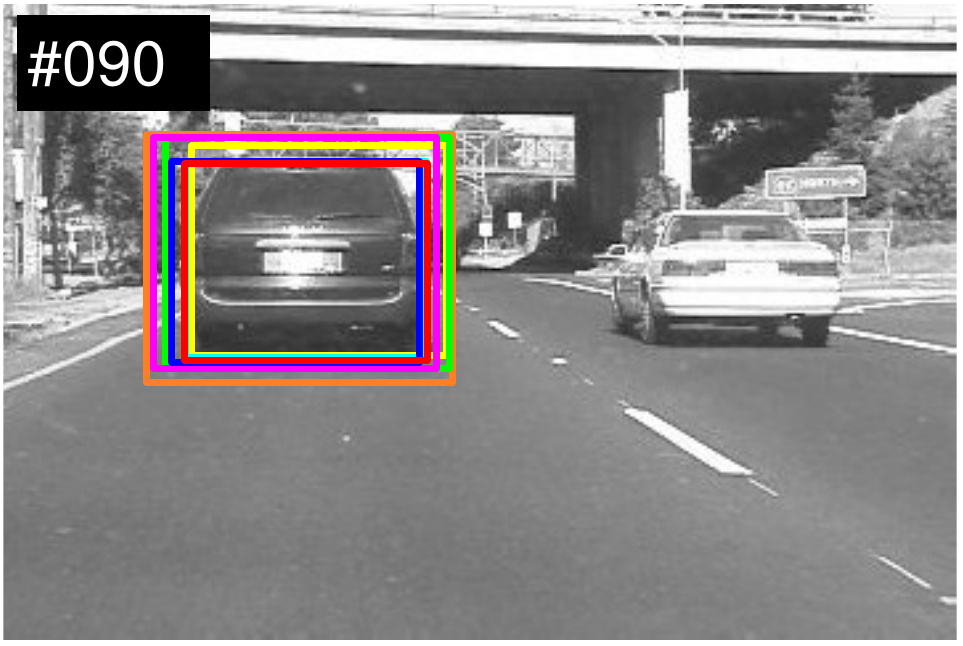} &
			\includegraphics[trim = 0mm 2mm 2mm 0mm, clip,width=.165\textwidth]{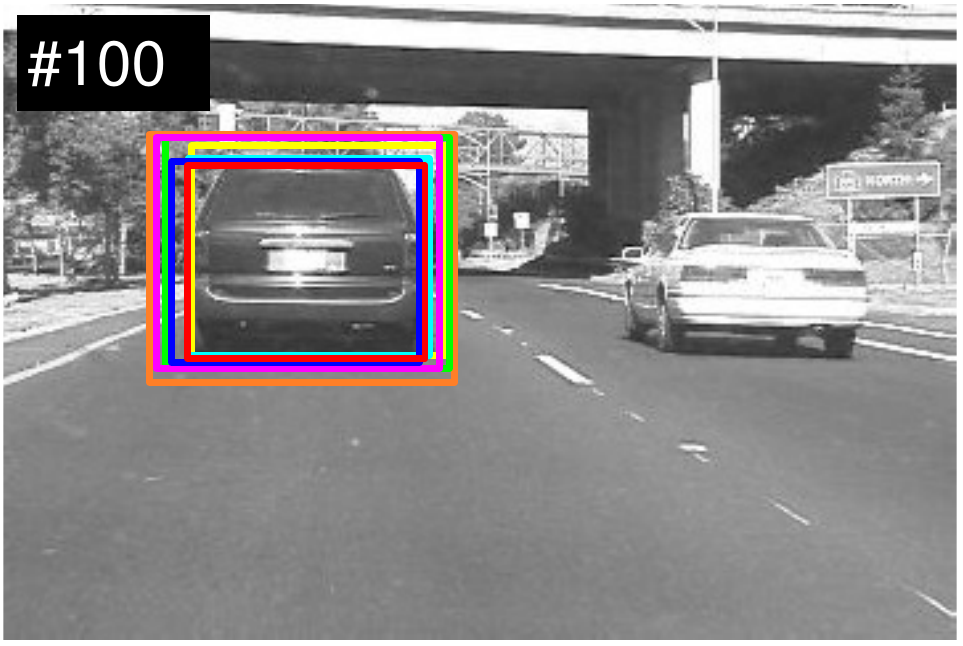} &
			\includegraphics[trim = 0mm 2mm 2mm 0mm, clip,width=.165\textwidth]{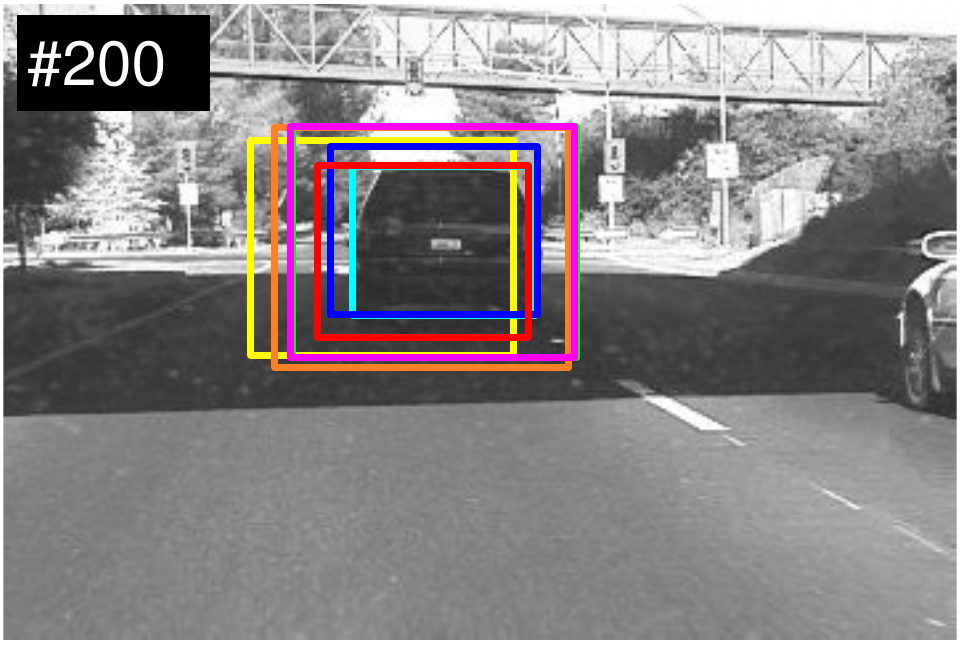} &
			\includegraphics[trim = 0mm 2mm 2mm 0mm, clip,width=.165\textwidth]{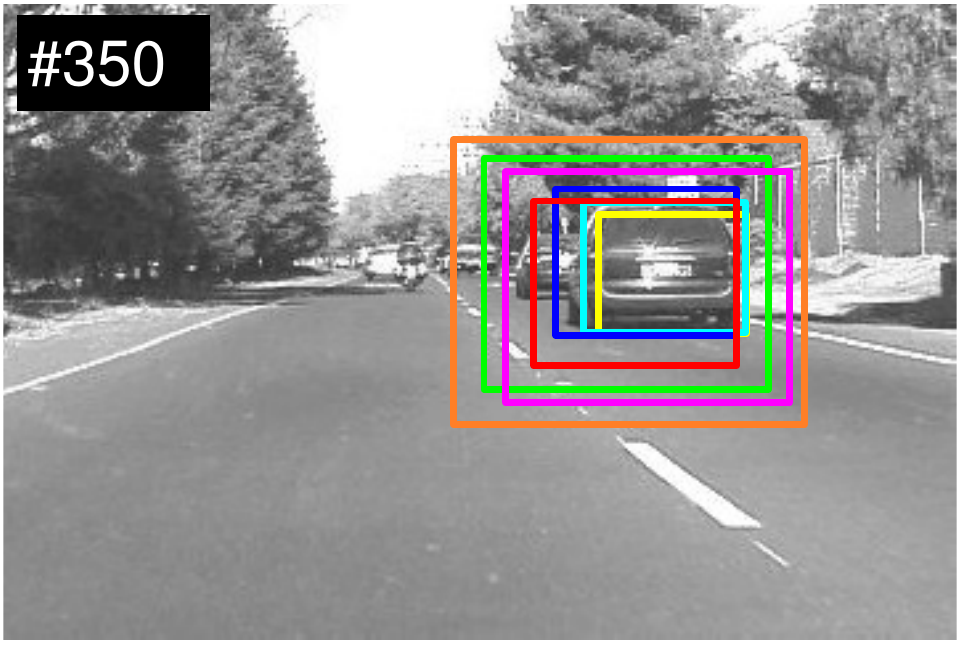} &
			\includegraphics[trim = 0mm 2mm 2mm 0mm, clip,width=.165\textwidth]{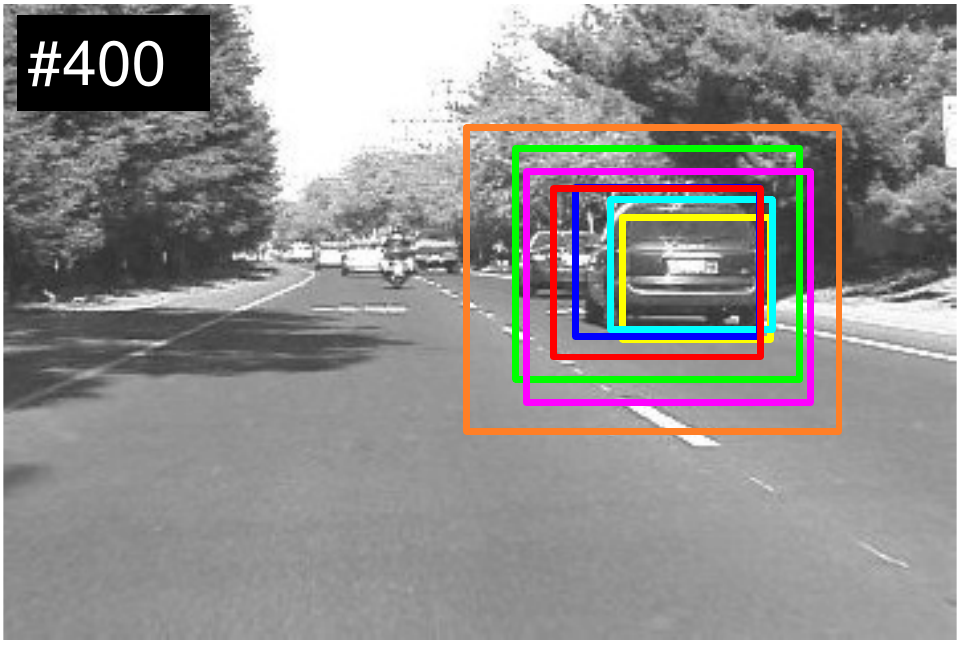} &
			\includegraphics[trim = 0mm 2mm 2mm 0mm, clip,width=.165\textwidth]{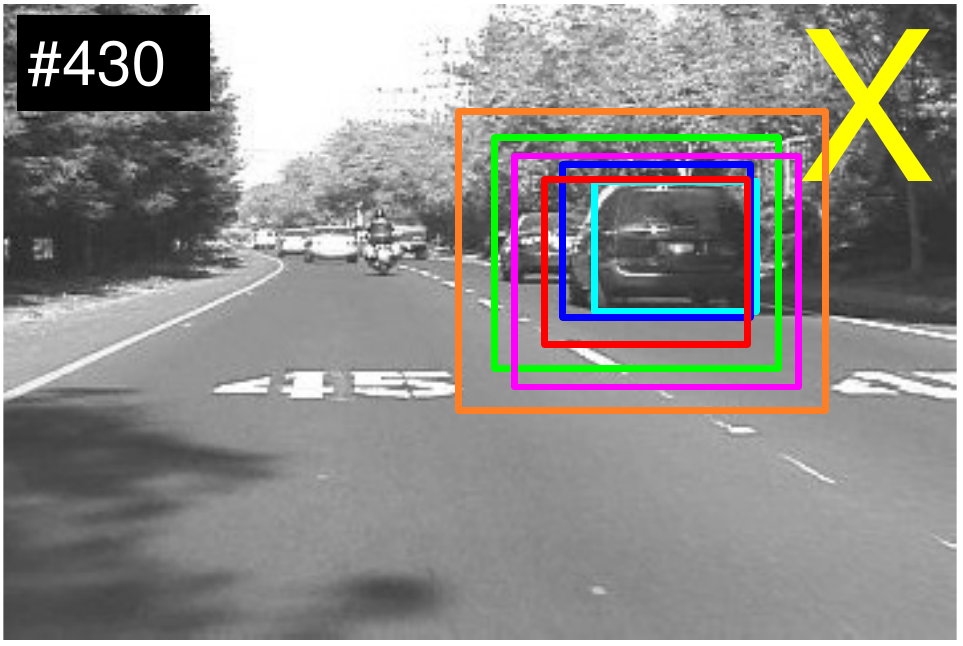} \\
			\includegraphics[trim = 0mm 2mm 2mm 0mm, clip,width=.165\textwidth]{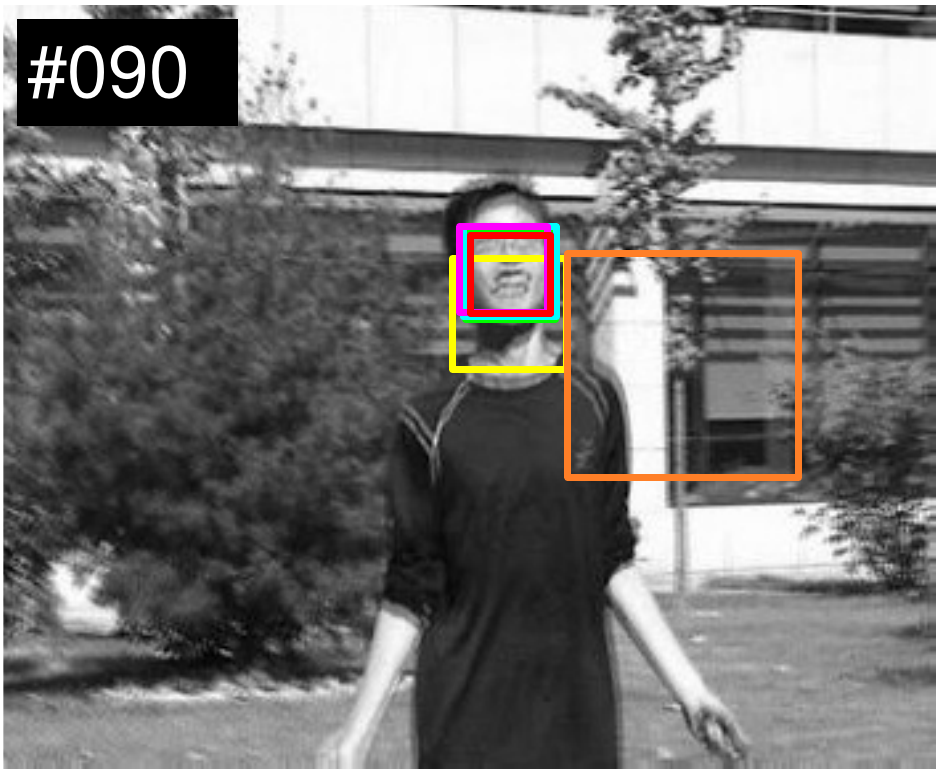}& 
			\includegraphics[trim = 0mm 2mm 2mm 0mm, clip,width=.165\textwidth]{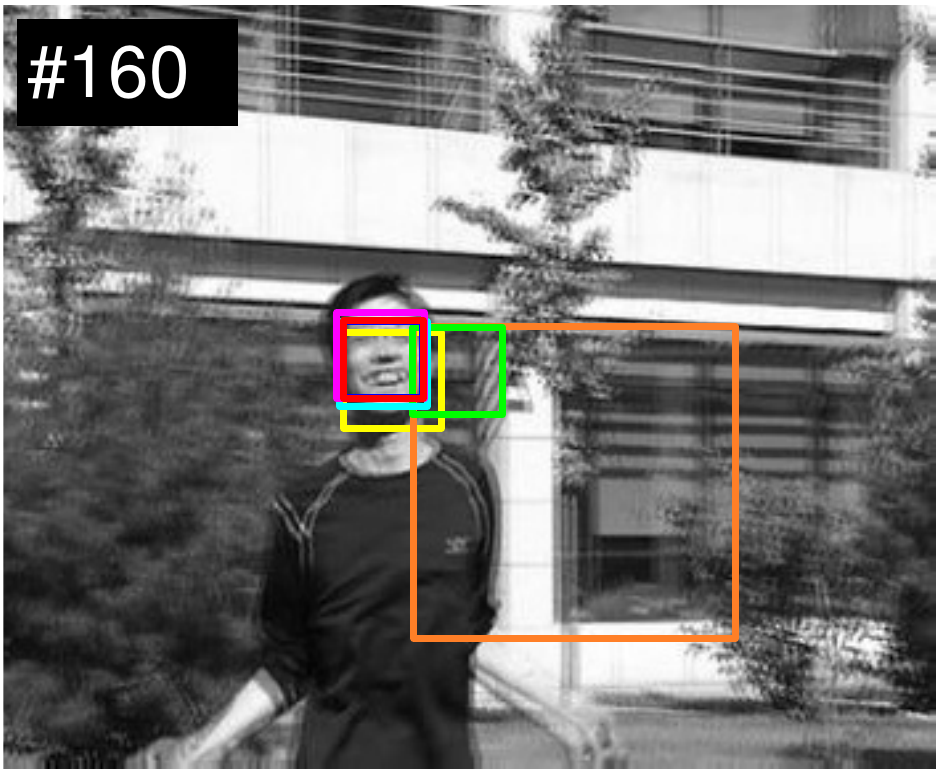}& 
			\includegraphics[trim = 0mm 2mm 2mm 0mm, clip,width=.165\textwidth]{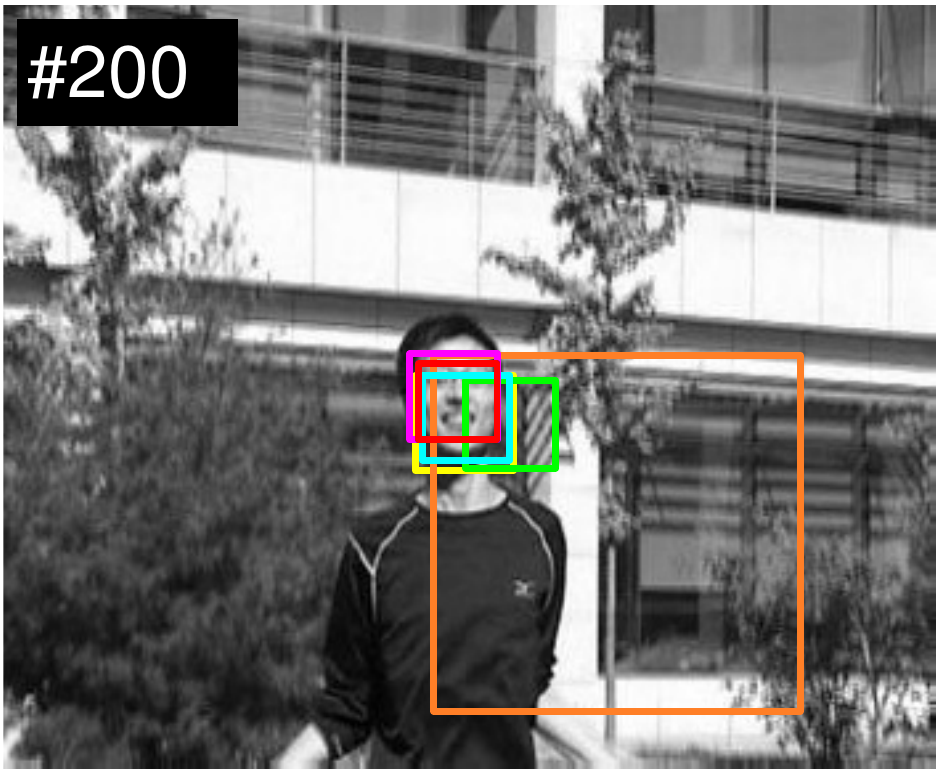}& 
			\includegraphics[trim = 0mm 2mm 2mm 0mm, clip,width=.165\textwidth]{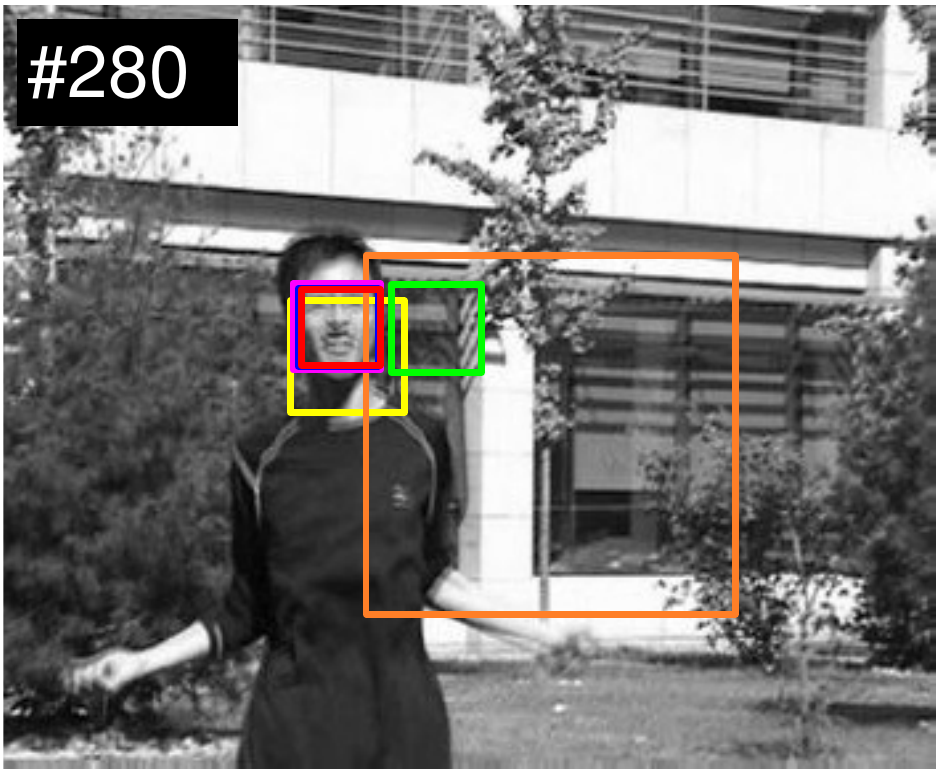} &
			\includegraphics[trim = 0mm 2mm 2mm 0mm, clip,width=.165\textwidth]{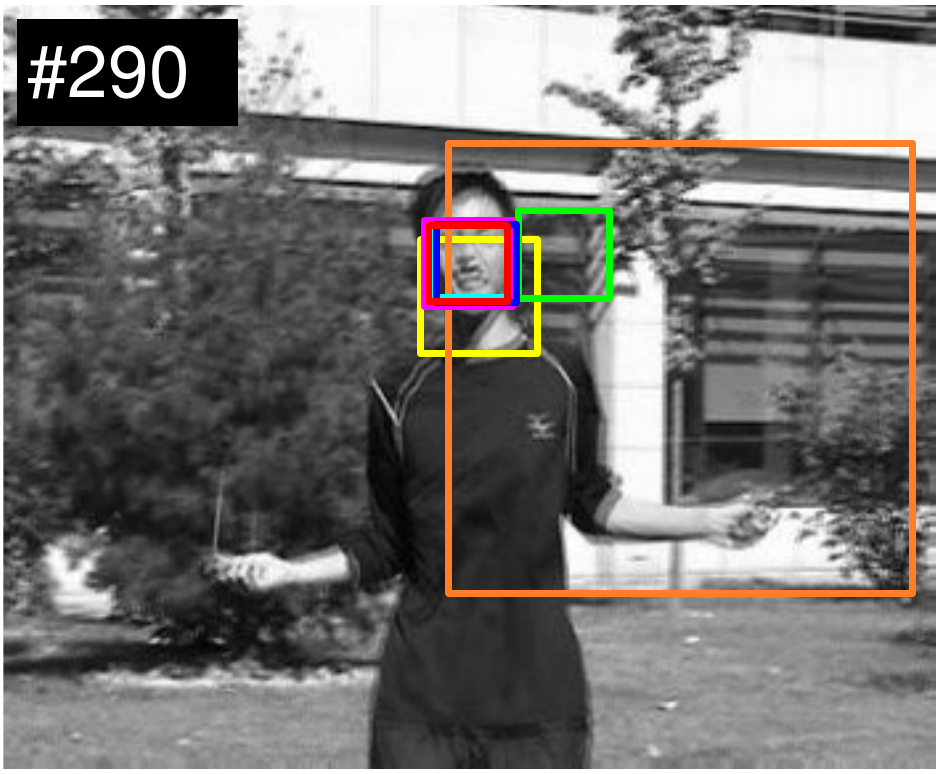} &
			\includegraphics[trim = 0mm 2mm 2mm 0mm, clip,width=.165\textwidth]{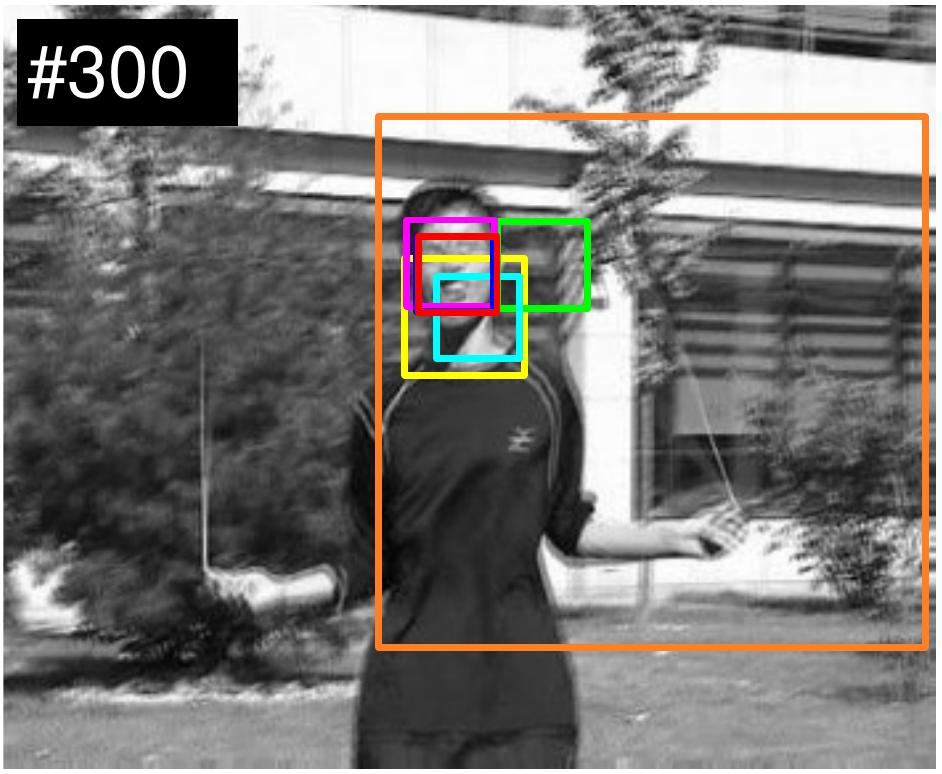} \\
			\includegraphics[trim = 0mm 2mm 2mm 0mm, clip,width=.165\textwidth]{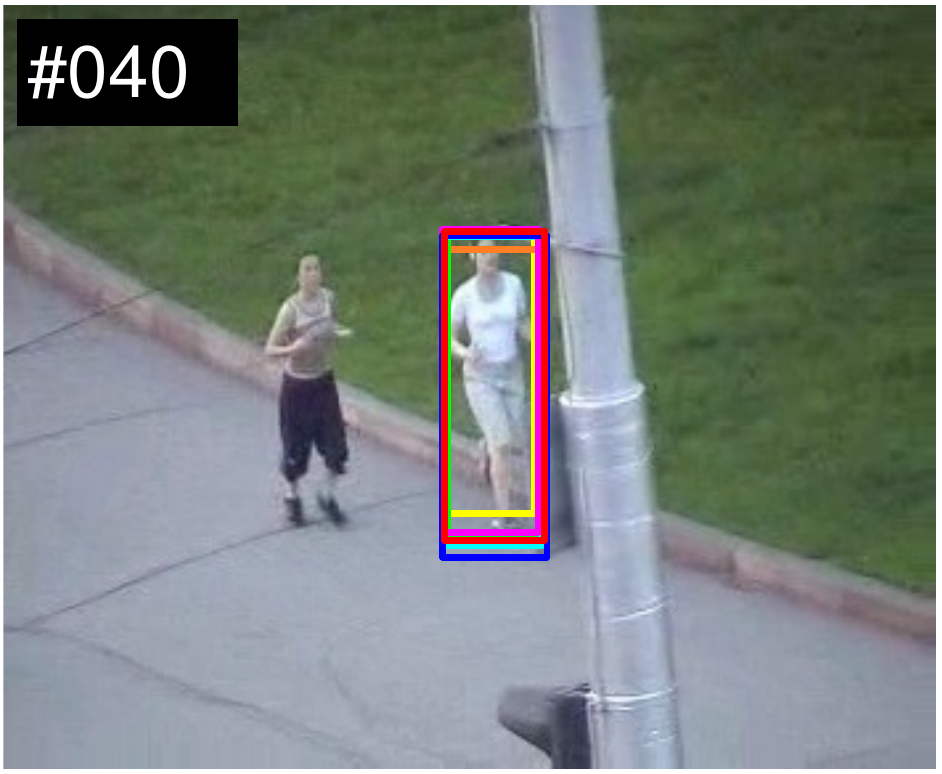}& 
			\includegraphics[trim = 0mm 2mm 2mm 0mm, clip,width=.165\textwidth]{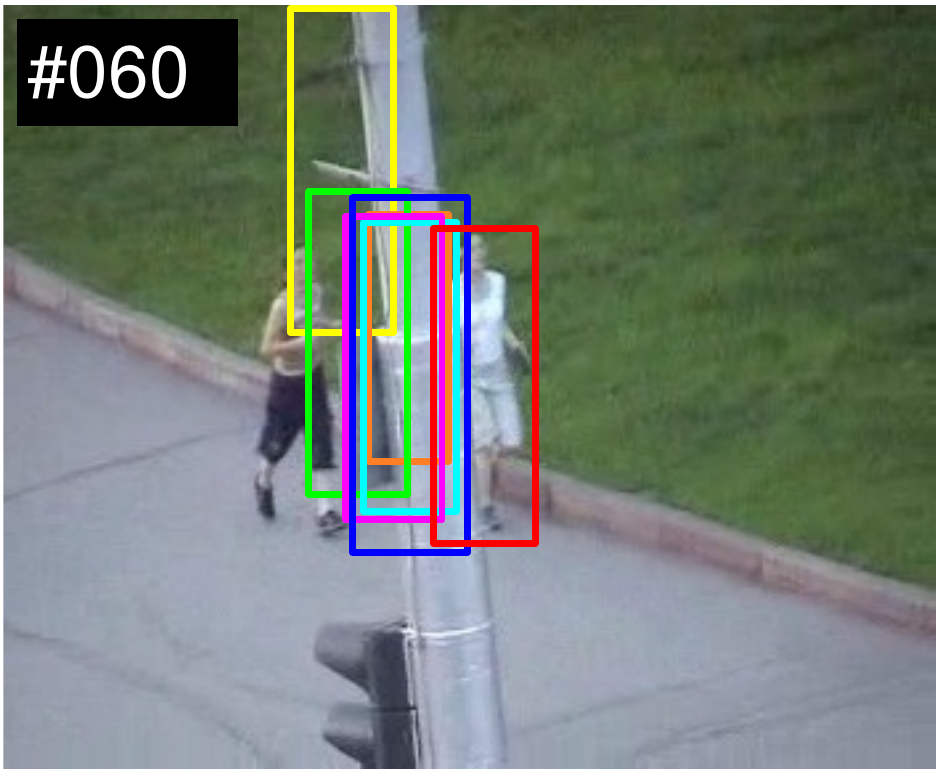} &
			\includegraphics[trim = 0mm 2mm 2mm 0mm, clip,width=.165\textwidth]{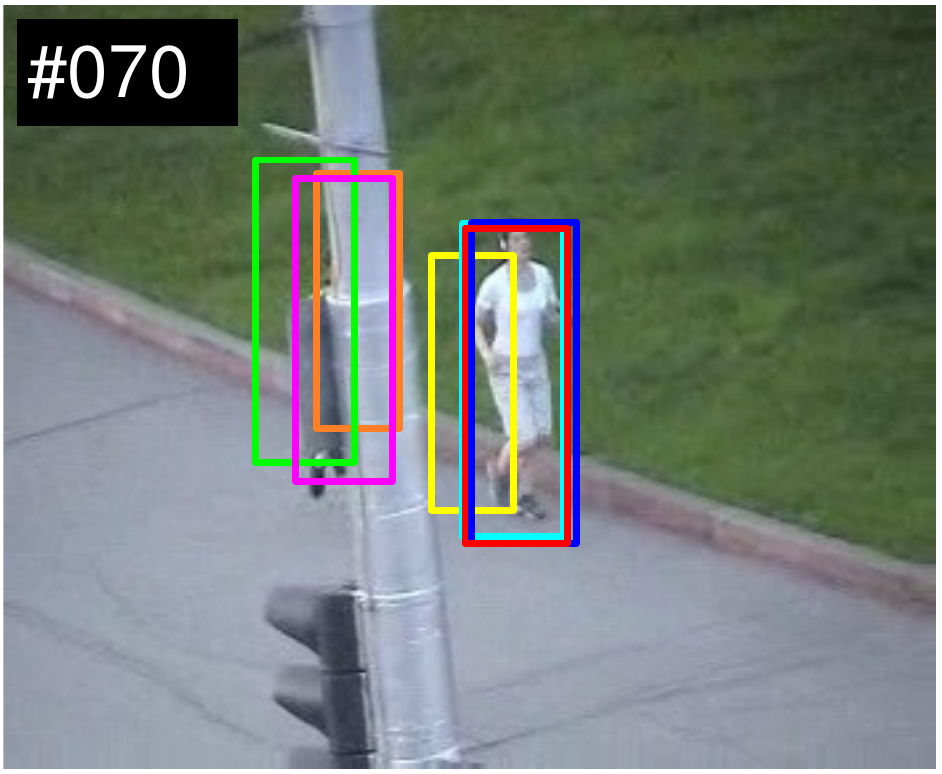} &
			\includegraphics[trim = 0mm 2mm 2mm 0mm, clip,width=.165\textwidth]{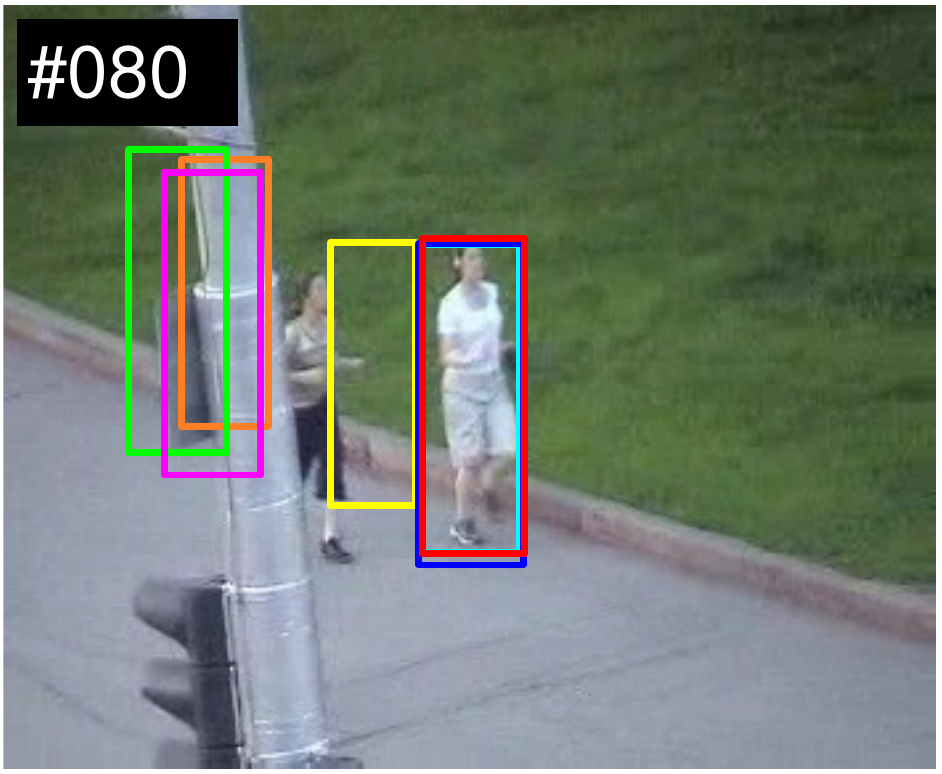} &
			\includegraphics[trim = 0mm 2mm 2mm 0mm, clip,width=.165\textwidth]{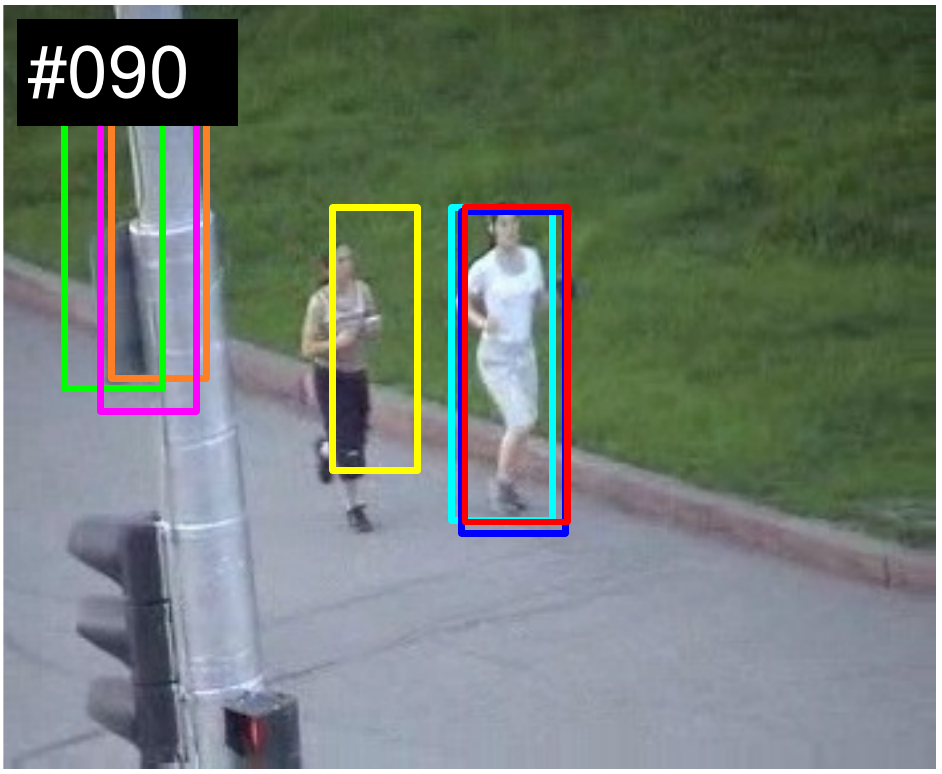} &
			\includegraphics[trim = 0mm 2mm 2mm 0mm, clip,width=.165\textwidth]{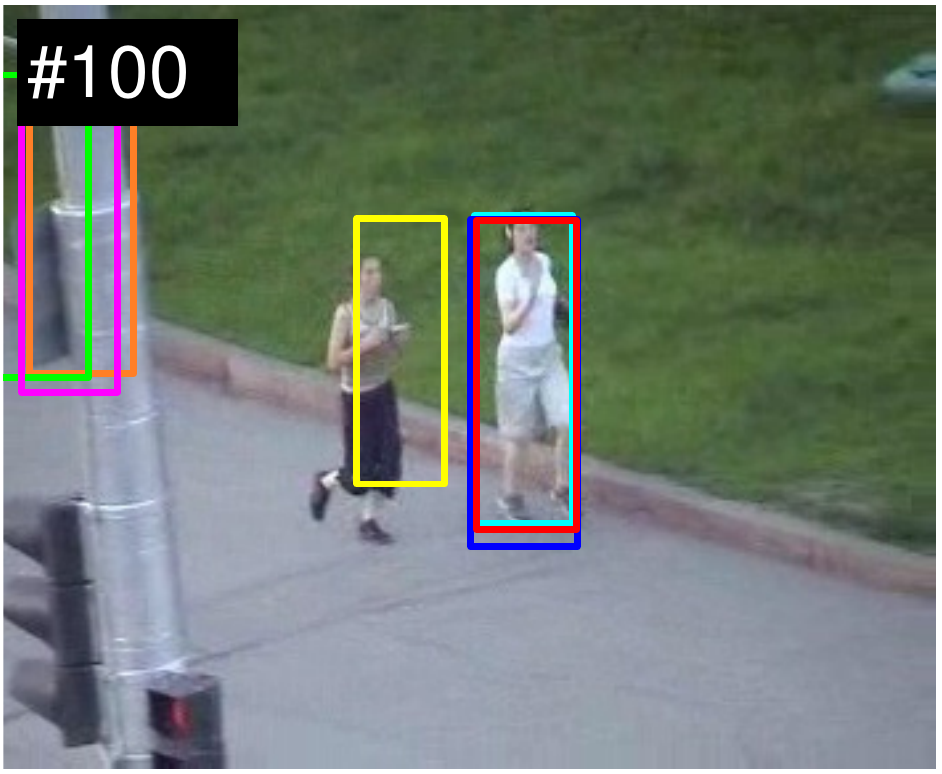} \\ 
		\end{tabular}
		\includegraphics[width=.92\textwidth]{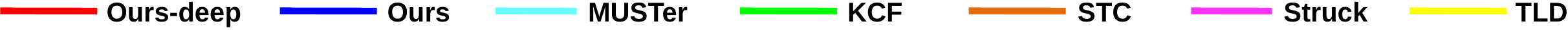} \\
		\caption{
			\textbf{Qualitative comparison.}
			Tracking results on the seven challenging sequences are from our approach, the MUSTer \cite{Hong_2015_CVPR}, KCF \cite{DBLP:journals/pami/HenriquesC0B15},
			STC \cite{DBLP:conf/eccv/ZhangZLZY14}, Struck \cite{DBLP:conf/iccv/HareST11} and TLD \cite{DBLP:journals/pami/KalalMM12} algorithms ($\times$: no tracking output). The proposed method performs favorably against the baseline trackers in terms of both translation estimation and scale estimation. From first to last row: \textit{coke, shaking, skating1, david, car4, jumping} and \textit{jogging-2}.}
		\label{fig:result}
	\end{figure*}

	\begin{figure}
		\centering
		\setlength{\tabcolsep}{.1em}
		\begin{tabular}{ccc}
			\includegraphics[height=.11\textwidth]{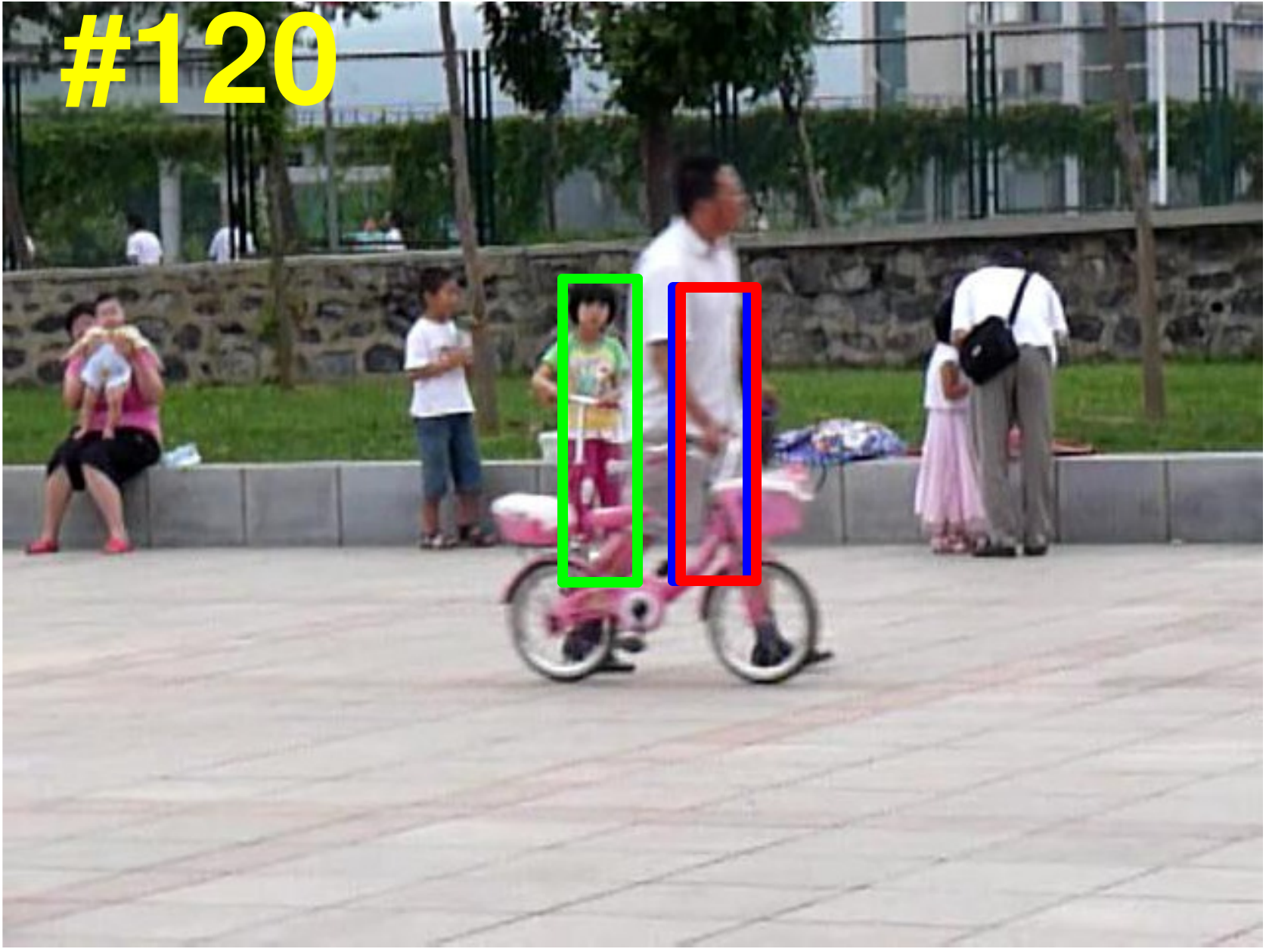} &
			\includegraphics[trim = 0mm 0mm 30mm 0mm, clip, height=.11\textwidth]{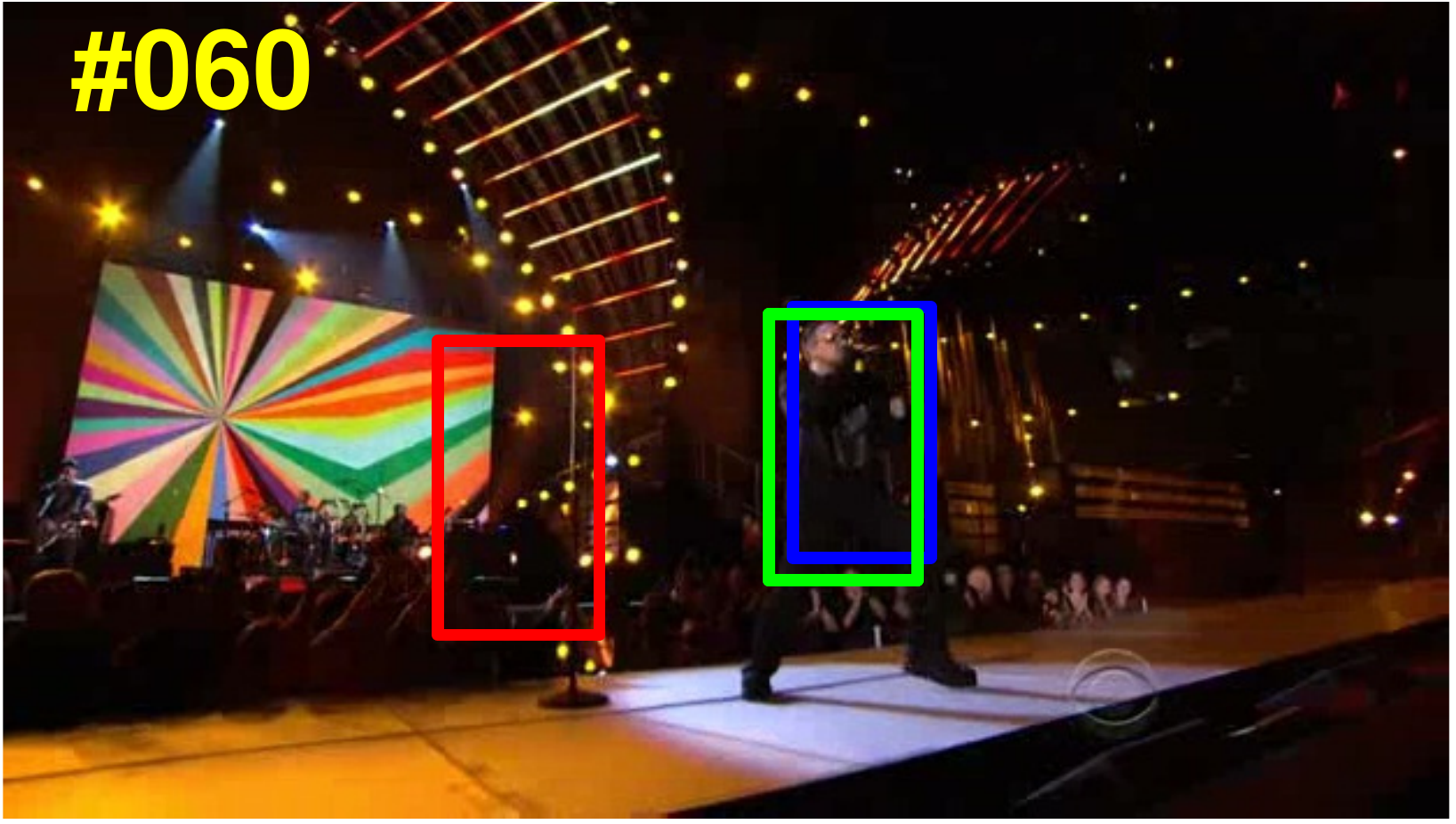} &
			\includegraphics[trim = 0mm 0mm 30mm 0mm, clip, height=.11\textwidth]{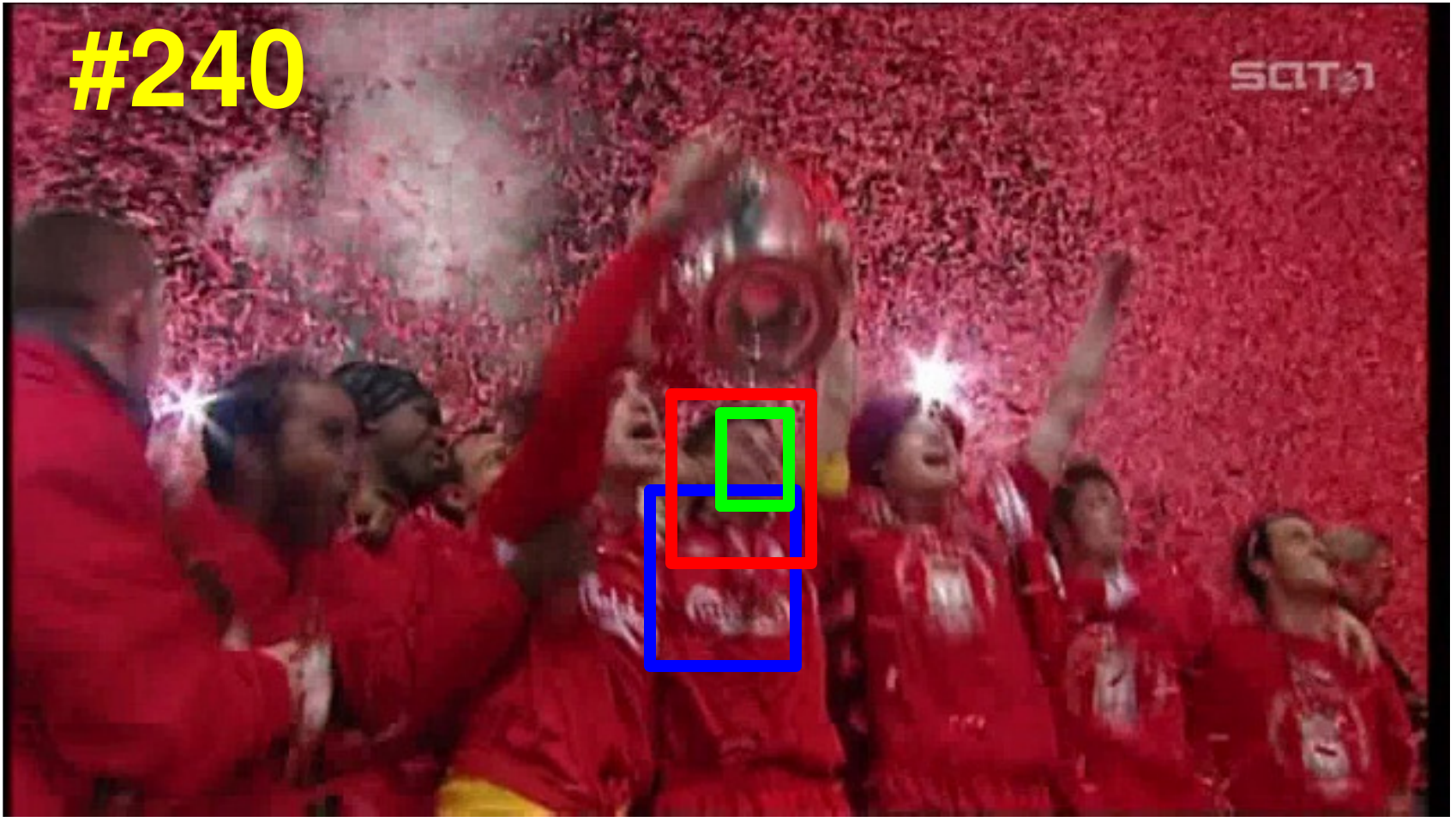} 
		\end{tabular}
		\caption{\textbf{Failure cases} on the \textit{girl2}, \textit{singer2} and \textit{soccer} sequences \cite{DBLP:conf/cvpr/WuLY13}. Red: ours-deep; blue: ours; green: ground truth.}
		\label{fig:failure}
	\end{figure}
	
	\subsection{Qualitative Evaluation}
	We evaluate the proposed algorithm with five state-of-the-art trackers (MUSTer \cite{Hong_2015_CVPR}, KCF \cite{DBLP:journals/pami/HenriquesC0B15}, 
	STC \cite{DBLP:conf/eccv/ZhangZLZY14}, Struck \cite{DBLP:conf/iccv/HareST11}, and TLD \cite{DBLP:journals/pami/KalalMM12}) on seven sequences with representative challenging attributes in Figure \ref{fig:result}. 
	The MUSTer tracker contains similar tracking components as our approach, i.e., translation and scale filters as well as a re-detection module, and performs well against the other methods.
	However, the translation filters in MUSTer are learned from color attribute features \cite{DBLP:conf/cvpr/DanelljanKFW14}, which are not robust to background clutters (\textit{coke}) and fast motion (\textit{jumping}). 
	With the use of the kernelized correlation filter learned from HOG features, the KCF tracker (similar to the baseline CT-HOG method in Figure \ref{fig:component-scale}) performs well in handling significant deformation and fast motion (\textit{david}).
	However, the KCF tracker tends to drift when the target undergoes temporary occlusion (\textit{coke}) and fails to recover from tracking failures (\textit{jogging-2}).
	Furthermore, the KCF tracker does not perform well for scenes with background clutters (\textit{shaking}) due to the presence of noisy image gradients. 
	Although the STC tracker can estimate scale changes, it does not perform well when the target objects undergo significant scale changes or abrupt motion (\textit{jumping}).
	This is because the STC tracker uses intensity as features and the scale is estimated from the response map of one single translation filter.
	The Struck tracker does not perform well when the target objects undergo out-of-plane rotation (\textit{david}), heavy occlusion, background clutter (\textit{coke}), or out-of-view movement(\textit{jogging-2}), as one single classifier is unlikely to balance model stability and adaptivity well.
	The TLD tracker can recover target objects from tracking failures by performing detection in each frame.
	However, the tracking component in TLD is updated too aggressively to locate objects 
	undergoing significant deformation and fast motion (\textit{shaking} and \textit{jumping}). 
	As the TLD tracker updates the detector in each frame, drifting (\textit{skating1}) and false positive re-detections are likely to occur as well (\textit{jogging-2}).
	
	The proposed tracker performs well in estimating both the translation and scale changes on these challenging sequences. 
	We attribute the favorable performance to three reasons.
	First, we learn the translation filter $\mathcal{A}_\mathrm{T}$ over a complementary set of features: HOG and HOI.
	Our tracker is thus less sensitive to illumination and background clutter (\textit{shaking} and \textit{singer2}), rotation (\textit{david}), and partial occlusion (\textit{coke}).
	Second, the scale filter $\mathcal{A}_\mathrm{S}$ and the translation filter $\mathcal{A}_\mathrm{T}$ are updated independently.
	This design effectively alleviates the degradation of the translation filter caused by inaccuracy in scale estimation as in the MUSTer tracker.
	It also helps alleviate the drifting problem caused by scale change, e.g., rapid performance loss in scale estimation on the \textit{jumping} sequence for the STC tracker.
	Third, the online trained detector can re-detects target objects in case of tracking failure, e.g., in the presence of the heavy occlusion 
	(\textit{coke}) or target disappearance from the camera view (\textit{jogging-2}).
	
	We show sample tracking failures by the proposed trackers in Figure~\ref{fig:failure}. 
	For the \textit{girl2} sequence, when long-term occlusions occur, 
	the proposed re-detection scheme is not activated due to the high similarity between the target and surrounding people. 
	In the \textit{singer2} sequence, our method using deep features (ours-deep) fails to track the target as deep features capture the semantics, which is not effective in differentiating the dark foreground from the bright background. 
	In contrast, our method with the handcrafted features encodes the spatial fine-grained details and performs well over the entire sequence. 
	For the \textit{soccer} sequence, the cluttered background yields a large amount of spatial details that lead our method not using deep features to drift, while the semantics within deep features are robust to such appearance variations.

	\begin{table}
		\centering
		\caption{Overlap success rates ($\%$) on the MEEM dataset. The best and second best results are highlighted by bold and underline.} 
		\label{tb:osmeem}
		\setlength{\tabcolsep}{.4em}
		\begin{tabular}{*{7}c}
			\hline
			& ~Ours~ & MUSTer & ~KCF~ & DSST & ~STC~ & ~TLD~ \\
			& & \cite{Hong_2015_CVPR} & 
			\cite{DBLP:journals/pami/HenriquesC0B15} &
			\cite{DBLP:conf/eccv/ZhangZLZY14} & 
			\cite{DBLP:conf/bmvc/DanelljanKFW14} & 
			\cite{DBLP:journals/pami/KalalMM12} \\\hline
			\textit{ball} & 65.3 & \underline{96.7} & \textbf{97.2} & 57.9 & 44.3 & 6.72 \\
			\textit{billieJean} & \underline{53.4} & 52.4 & \textbf{53.6} & 12.4 & 11.6 & 10.3 \\
			\textit{boxing1} & \textbf{46.5} & 14.4 & 29.3 & \underline{40.5} & 22.7 & 7.88 \\
			\textit{boxing2} & \textbf{97.0} & 27.5 & \underline{52.9} & 35.2 & 12.8 & 40.9 \\
			\textit{carRace} & \underline{97.3} & \textbf{98.1} & 33.6 & 33.6 & 6.53 & 0.33 \\
			\textit{dance} & \textbf{36.1} & \underline{26.5} & 26.0 & 25.4 & 22.7 & 13.9 \\
			\textit{latin} & 41.7 & \textbf{44.4} & \underline{44.4} & 36.8 & 12.0 & 14.6 \\
			\textit{ped1} & \textbf{100} & 8.1 & 50.9 & 55.6 & 48.3 & \underline{56.8} \\
			\textit{ped2} & \textbf{15.2} & 12.9 & 12.0 & \underline{14.3} & 14.2 & 3.95 \\
			\textit{rocky} & \textbf{100} & 38.5 & \textbf{100} & \textbf{100} & 74.0 & 10.7 \\\hline
			Average & \textbf{65.3} & 41.9 & \underline{50.0} & 41.2 & 26.9 & 16.6 \\\hline 
		\end{tabular}
	\end{table}
	
	\begin{table}
		\centering
		\caption{Distance precision rates ($\%$) on the MEEM dataset. The best and second best results are highlighted by bold and underline.}
		\label{tb:dpmeem}
		\setlength{\tabcolsep}{.4em}
		\begin{tabular}{*{7}c}
			\hline
			& ~Ours~ & MUSTer & ~KCF~ & DSST & ~STC~ & ~TLD~ \\
			& & \cite{Hong_2015_CVPR} & 
			\cite{DBLP:journals/pami/HenriquesC0B15} &
			\cite{DBLP:conf/eccv/ZhangZLZY14} & 
			\cite{DBLP:conf/bmvc/DanelljanKFW14} & 
			\cite{DBLP:journals/pami/KalalMM12} \\\hline 
			\textit{ball} & 66.4 & \underline{96.5} & \textbf{96.7} & 56.6 & 56.8 & 6.72 \\
			\textit{billieJean} & 60.2 & \underline{64.9} & \textbf{71.9} & 14.7 & 13.1 & 10.3 \\
			\textit{boxing1} & \textbf{49.2} & 20.3 & 26.5 & \underline{46.4} & 29.0 & 3.03 \\
			\textit{boxing2} & \textbf{93.6} & 27.3 & \underline{47.9} & 34.3 & 24.2 & 34.4 \\
			\textit{carRace} & \underline{95.8} & \textbf{97.0} & 33.5 & 33.6 & 21.2 & 0.33 \\
			\textit{dance} & \textbf{25.1} & 22.5 & \underline{22.8} & 21.0 & 19.3 & 10.2 \\
			\textit{latin} & 30.5 & \textbf{44.9} & \underline{44.6} & 37.3 & 29.8 & 3.17 \\
			\textit{ped1} & \textbf{100} & 53.4 & 55.1 & 59.0 & 53.0 & \underline{69.2} \\
			\textit{ped2} & \textbf{15.6} & 13.9 & 12.3 & \underline{15.3} & 15.2 & 3.95 \\
			\textit{rocky} & \textbf{100} & 38.5 & \textbf{100} & \textbf{100} & \textbf{100} & 16.3 \\\hline
			Average & \textbf{63.6} & 47.9 & \underline{51.1} & 41.8 & 36.2 & 15.8 \\\hline
		\end{tabular}
	\end{table}

	\begin{figure*}
		\centering
		\footnotesize
		\setlength{\tabcolsep}{0pt}
		\begin{tabular}{cccc}
			\includegraphics[width=.24\textwidth]{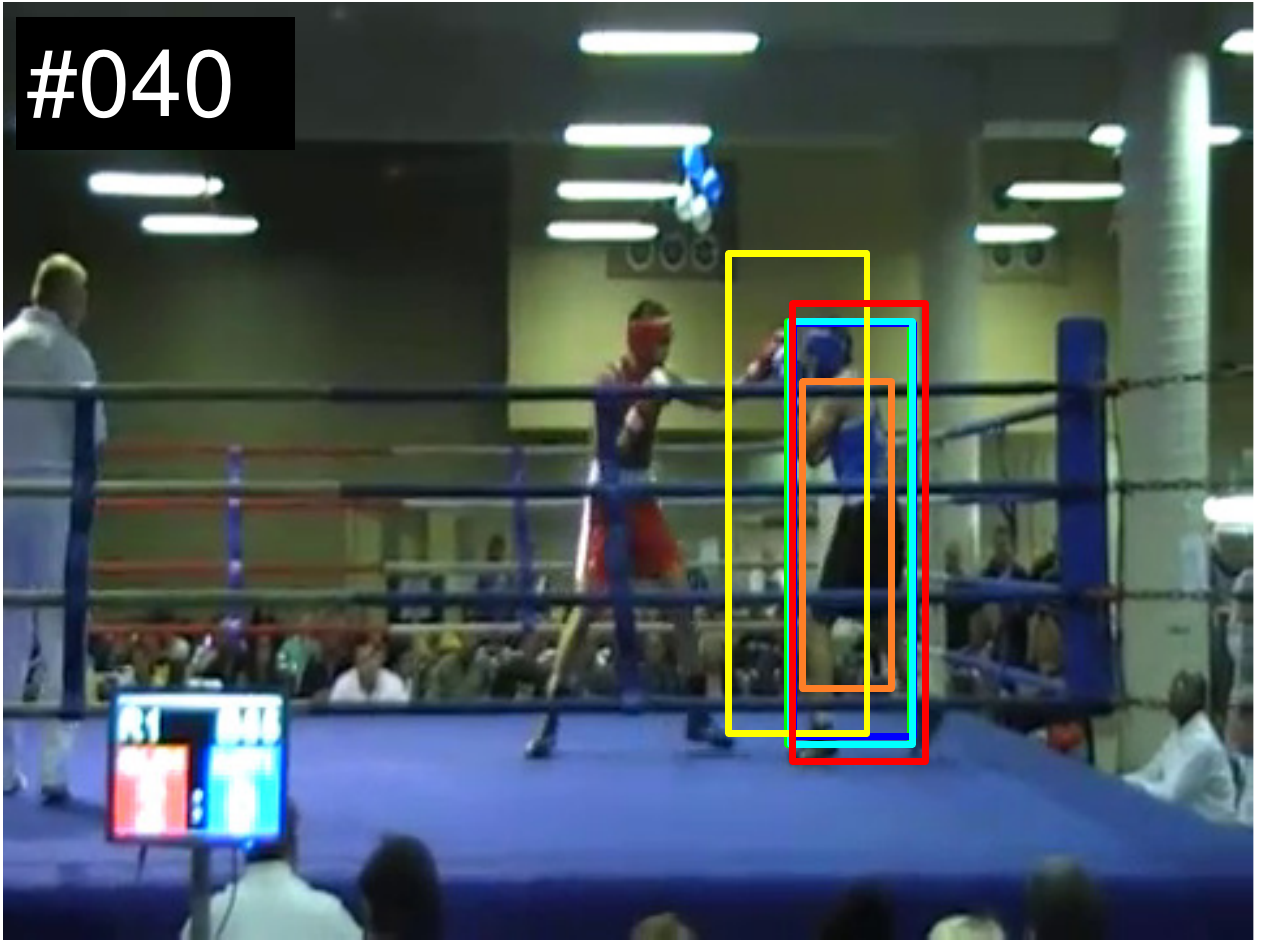} &
			\includegraphics[width=.24\textwidth]{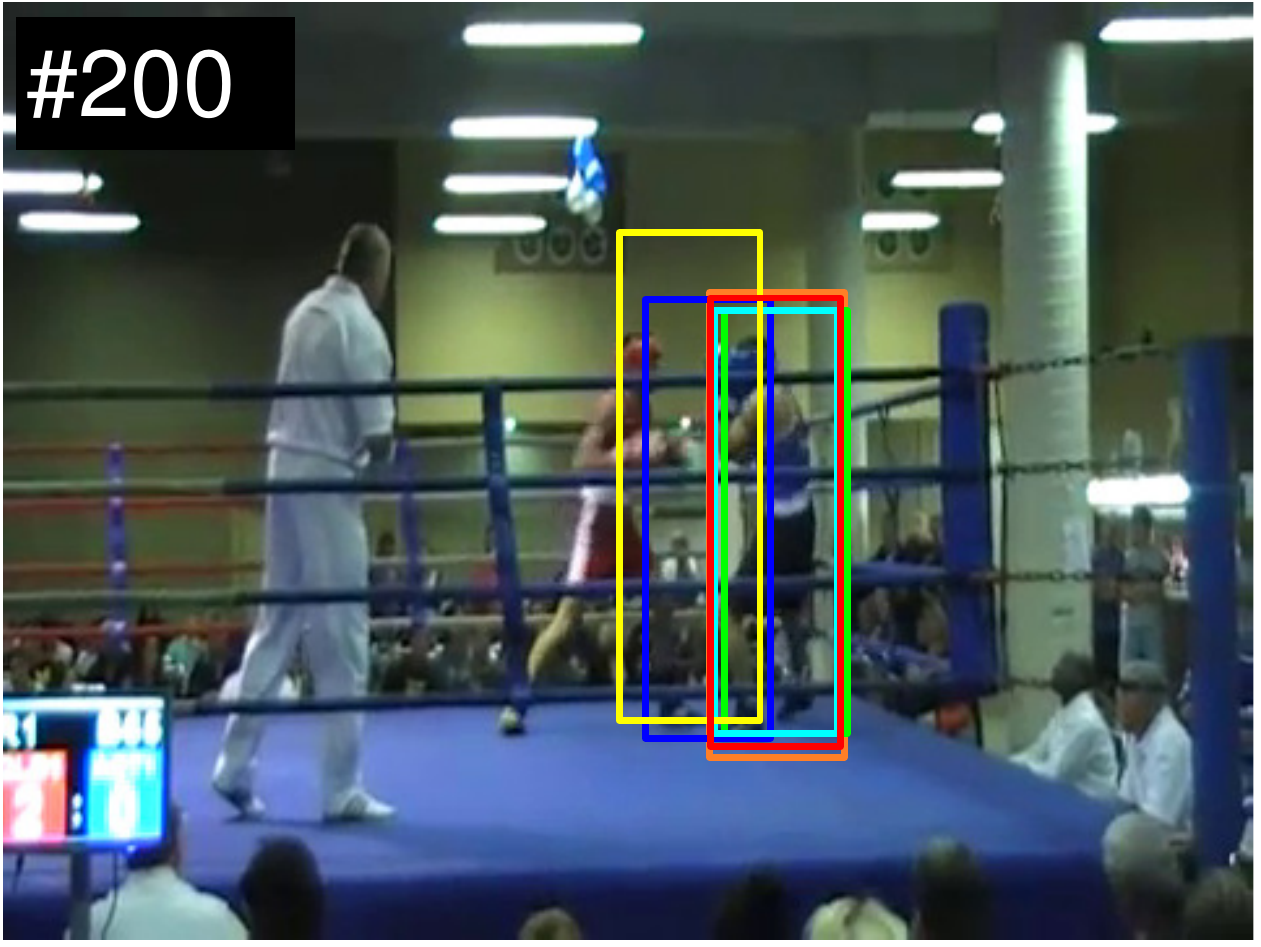} &
			\includegraphics[width=.24\textwidth]{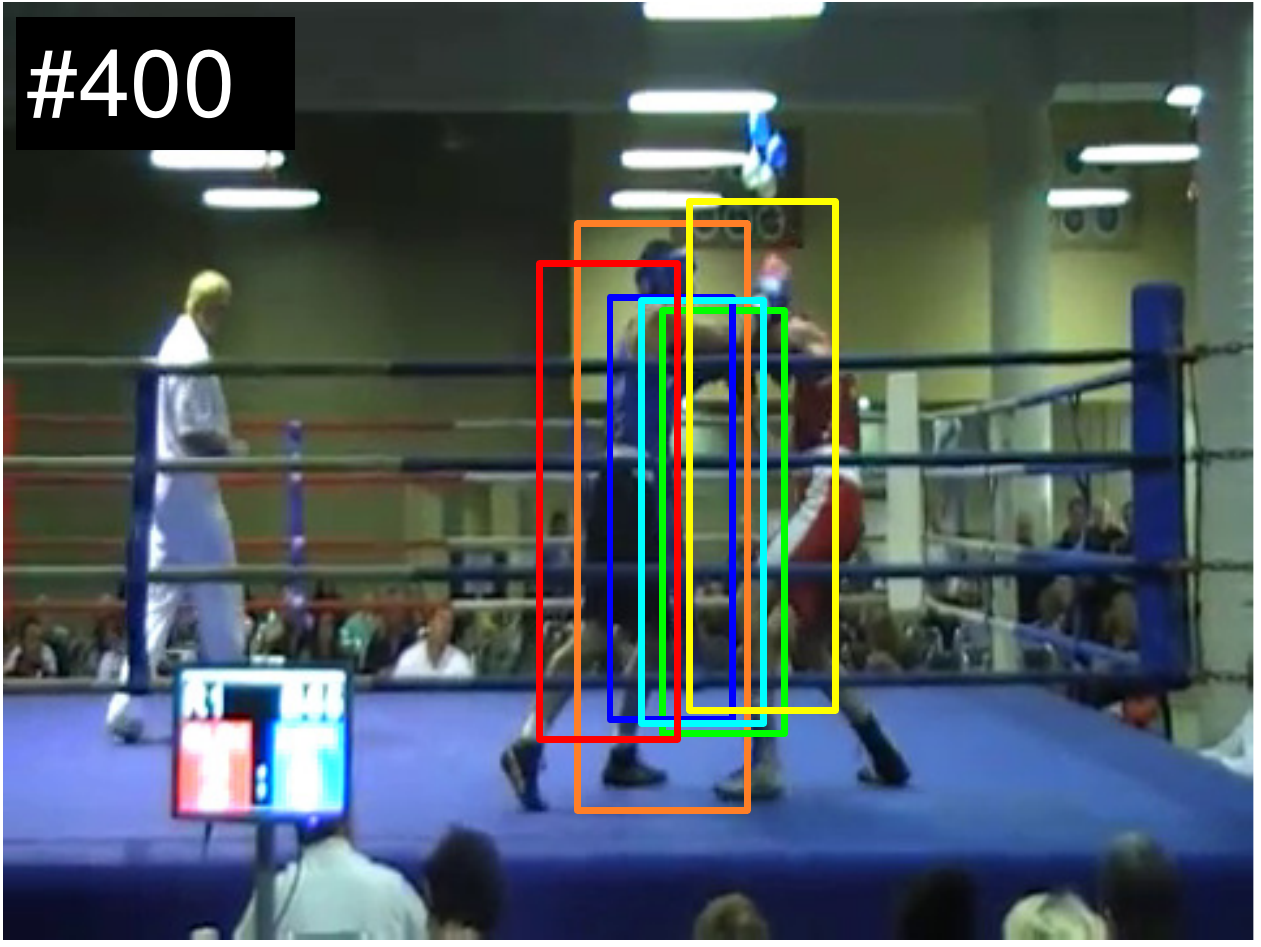} &
			\includegraphics[width=.24\textwidth]{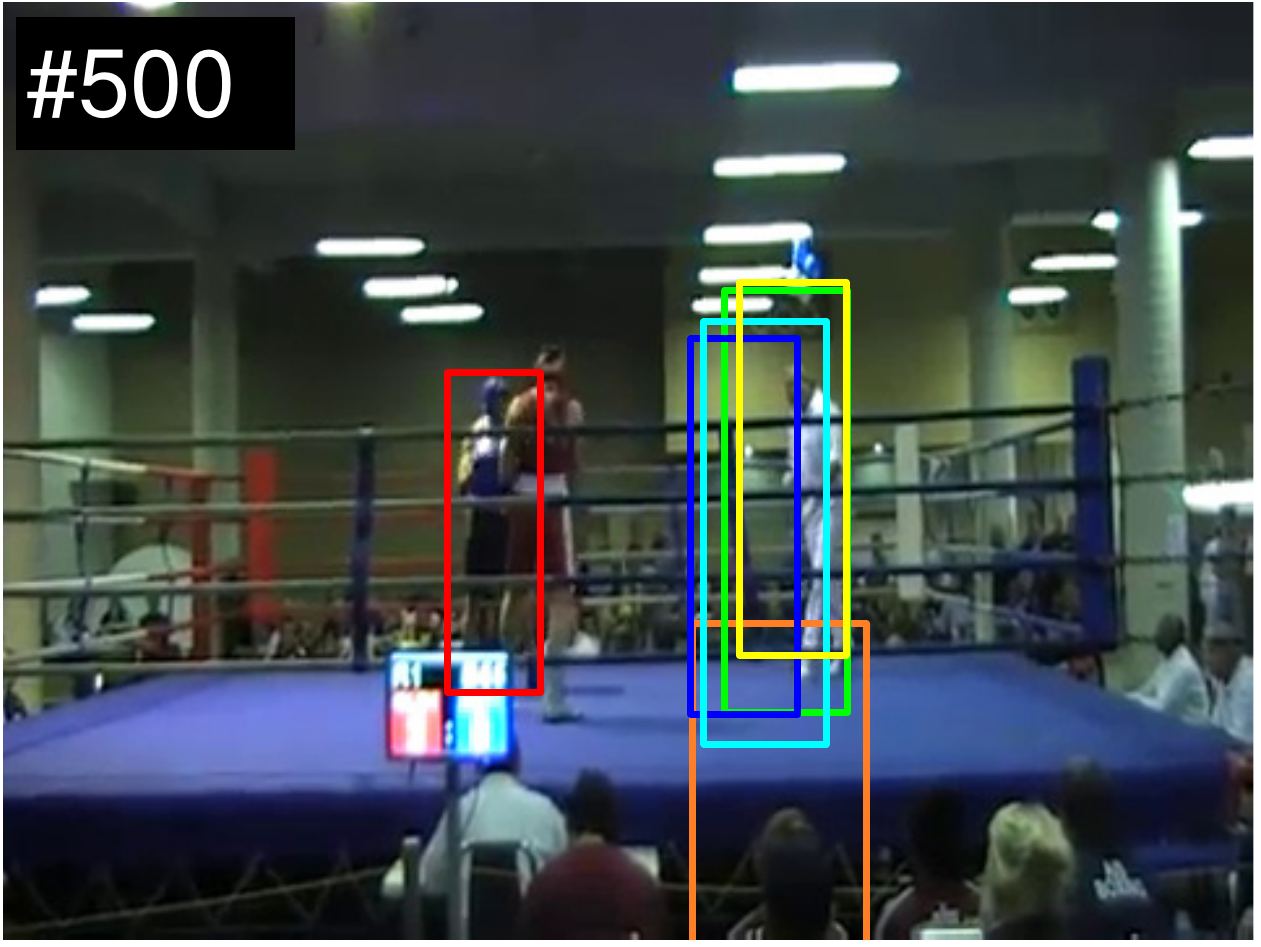} \\
			\includegraphics[width=.24\textwidth]{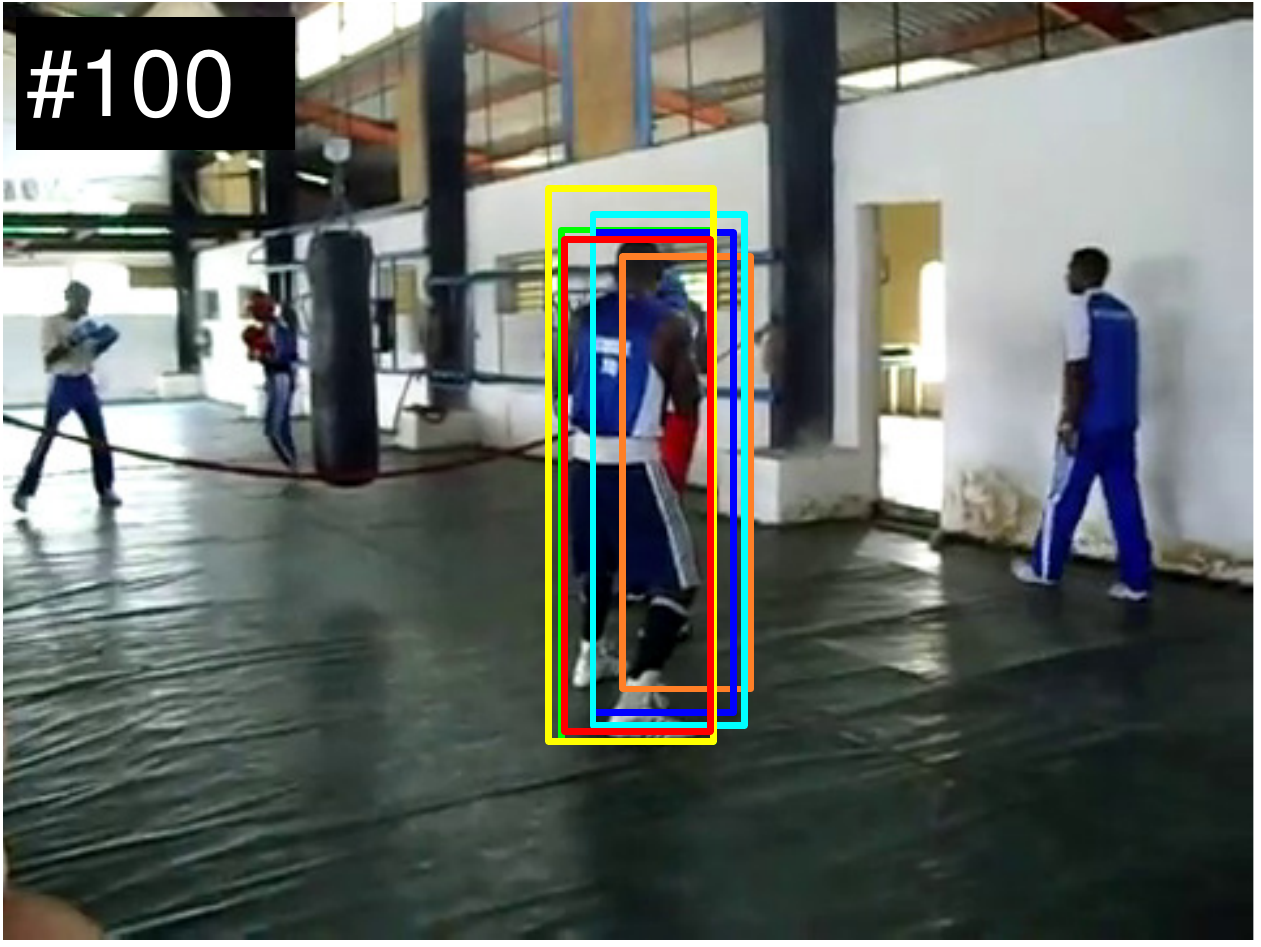} &
			\includegraphics[width=.24\textwidth]{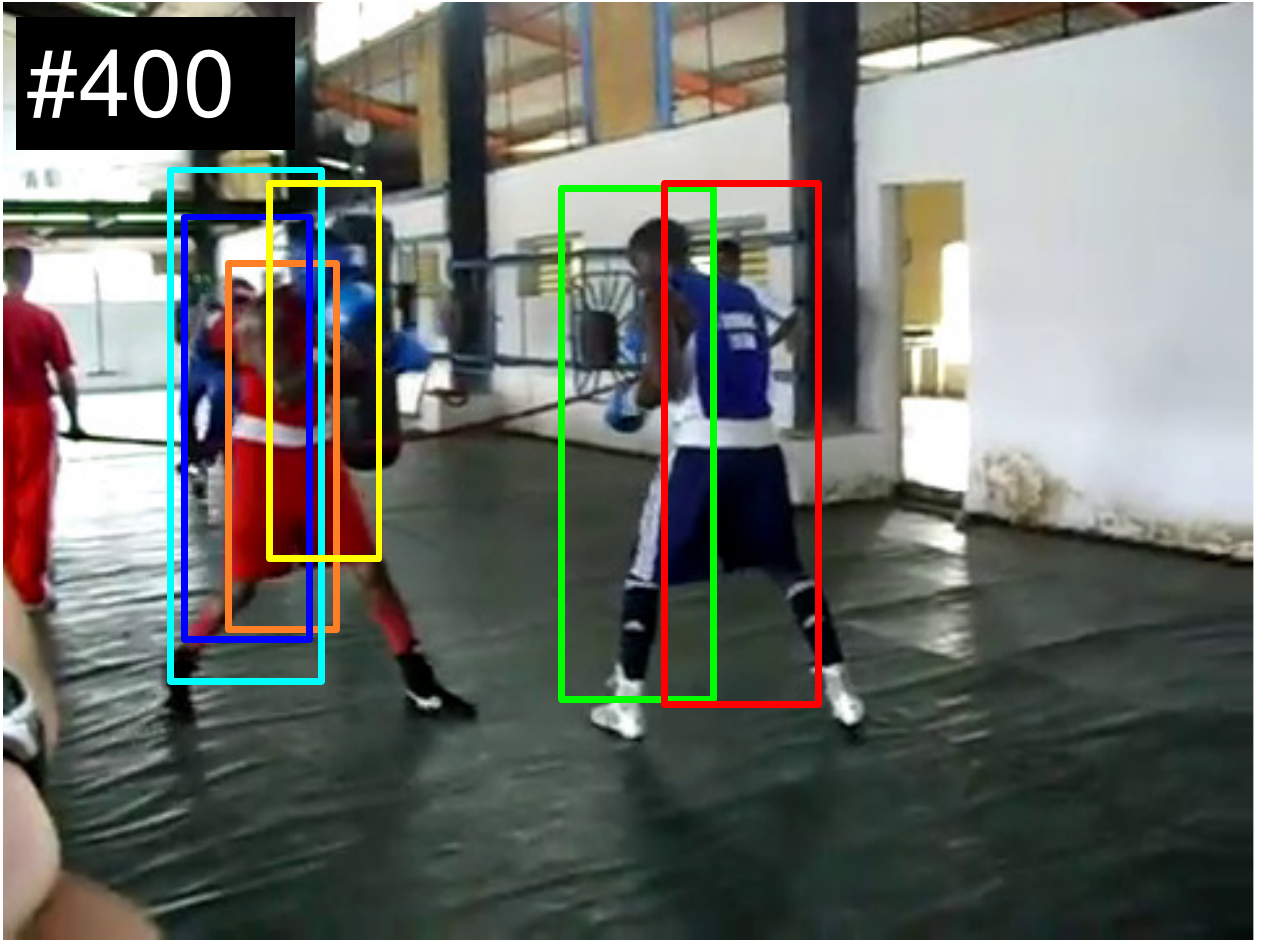} &
			\includegraphics[width=.24\textwidth]{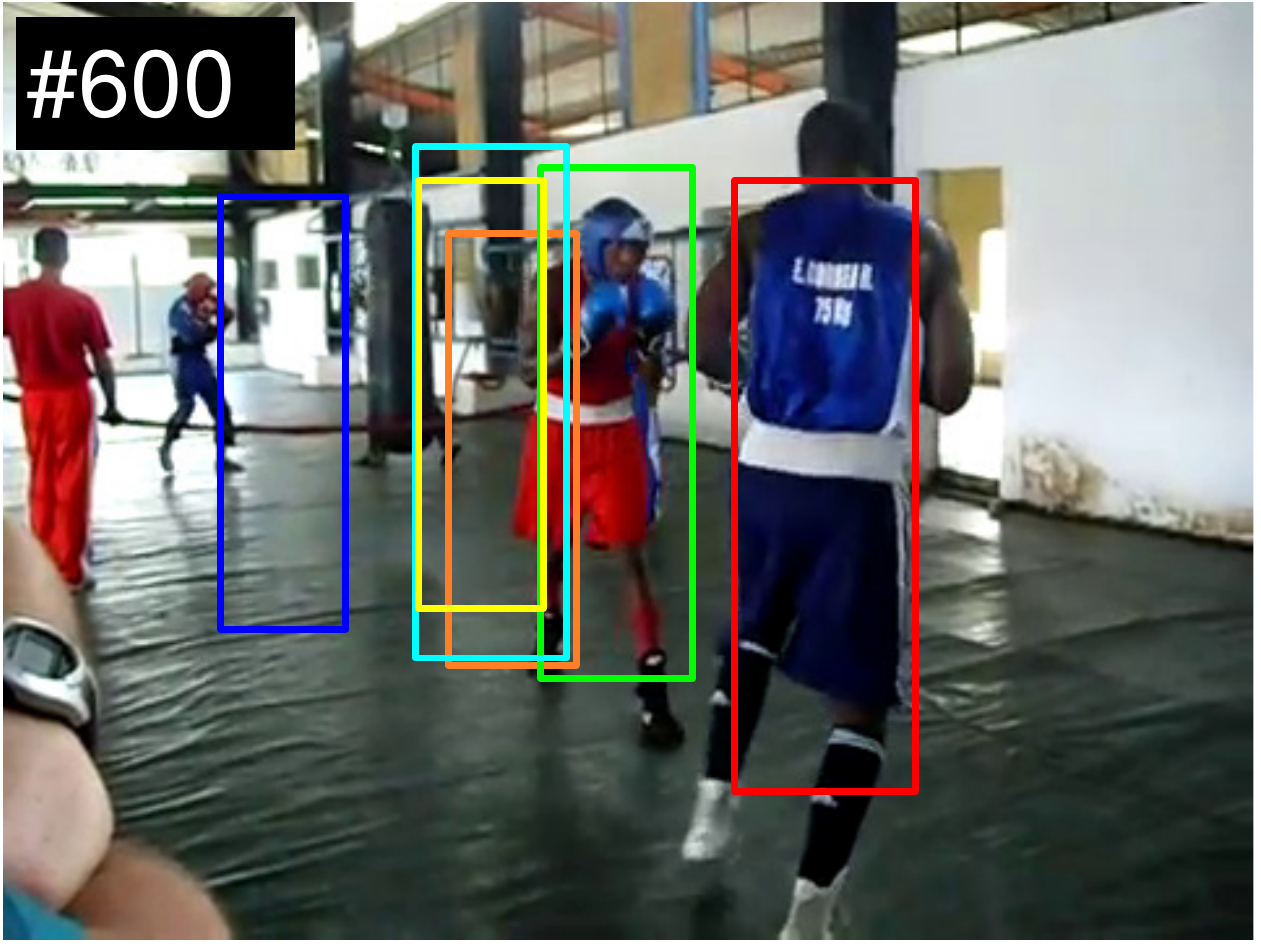} &
			\includegraphics[width=.24\textwidth]{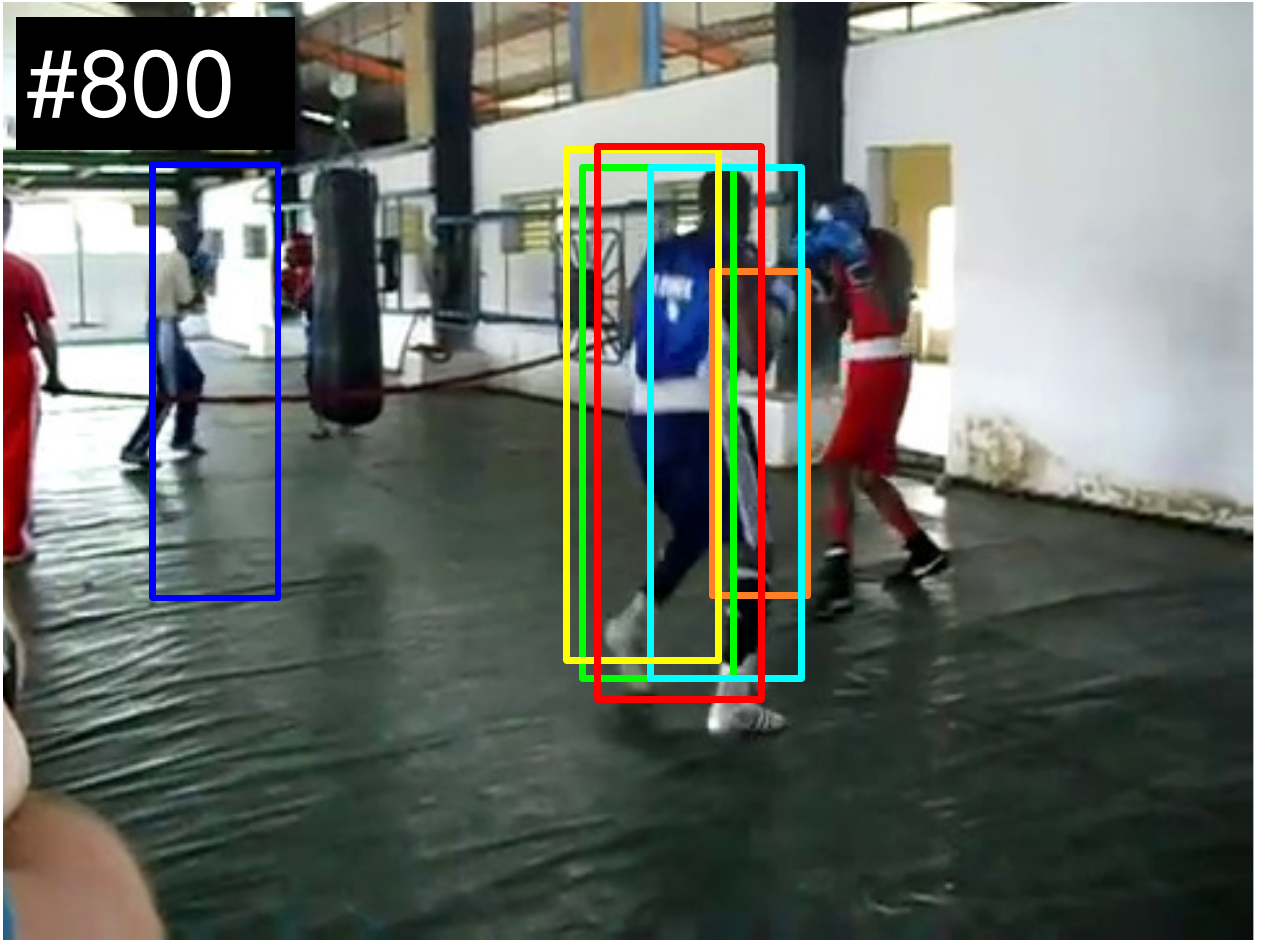} \\
			\includegraphics[width=.24\textwidth]{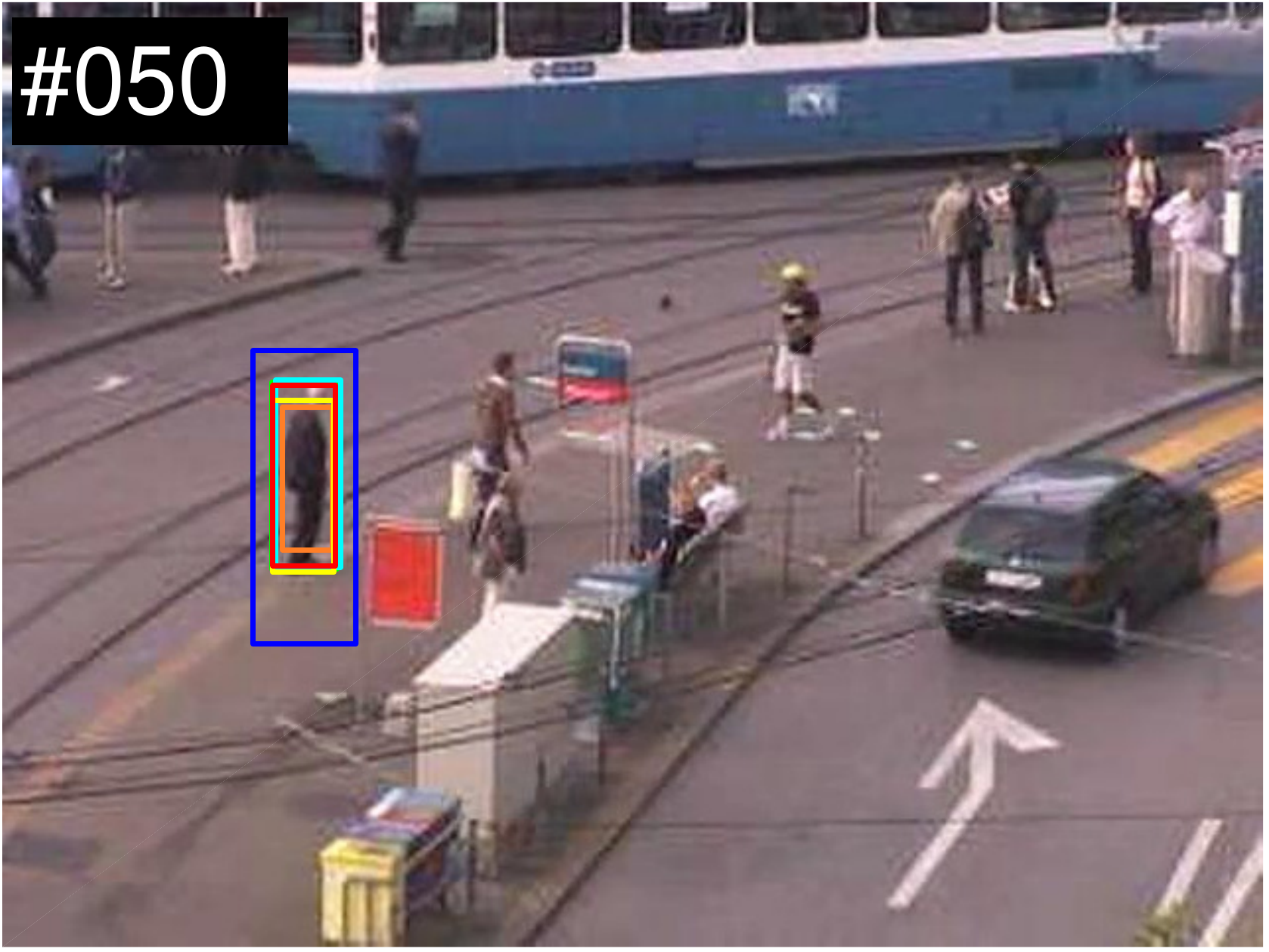} &
			\includegraphics[width=.24\textwidth]{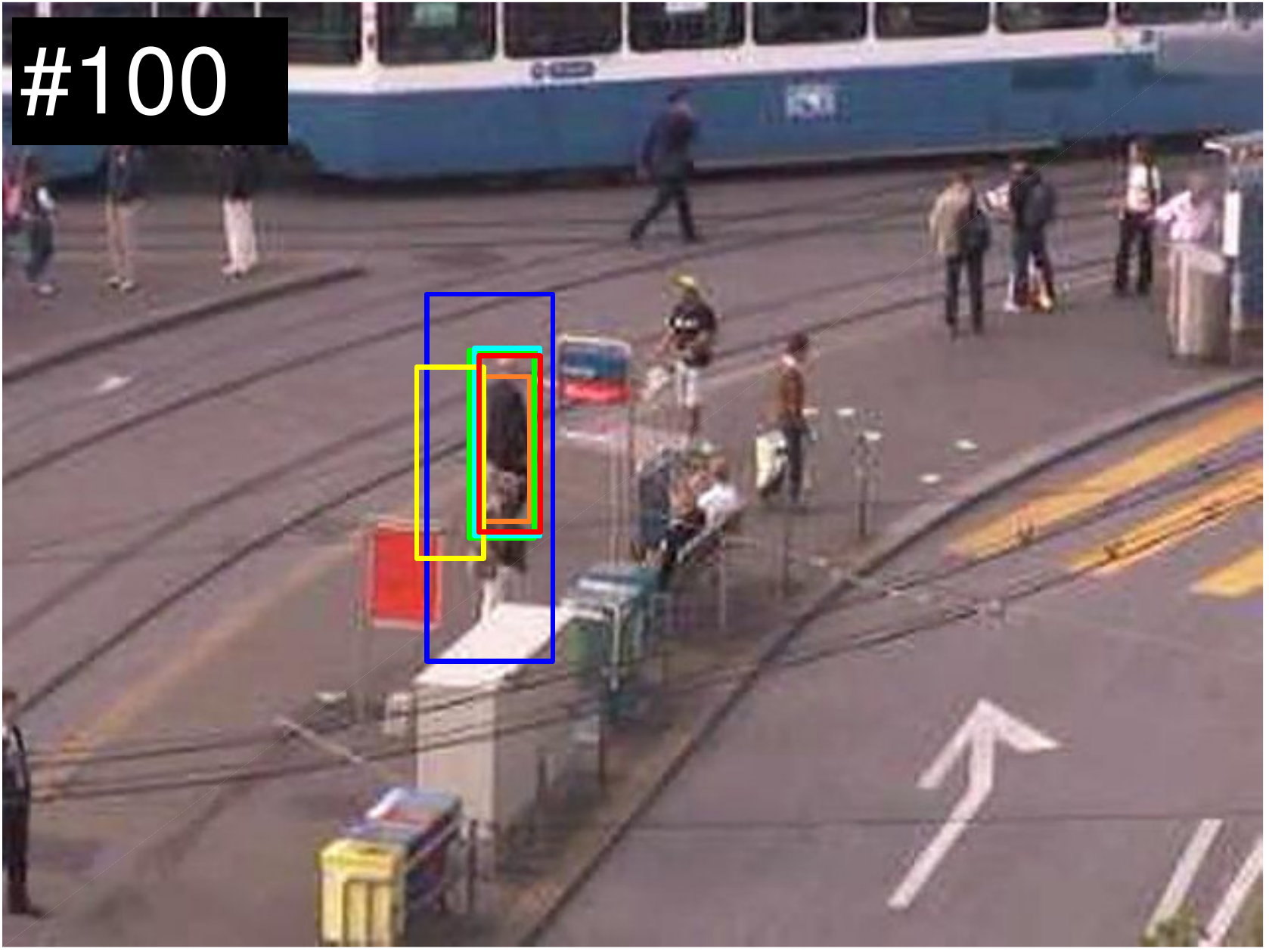} &
			\includegraphics[width=.24\textwidth]{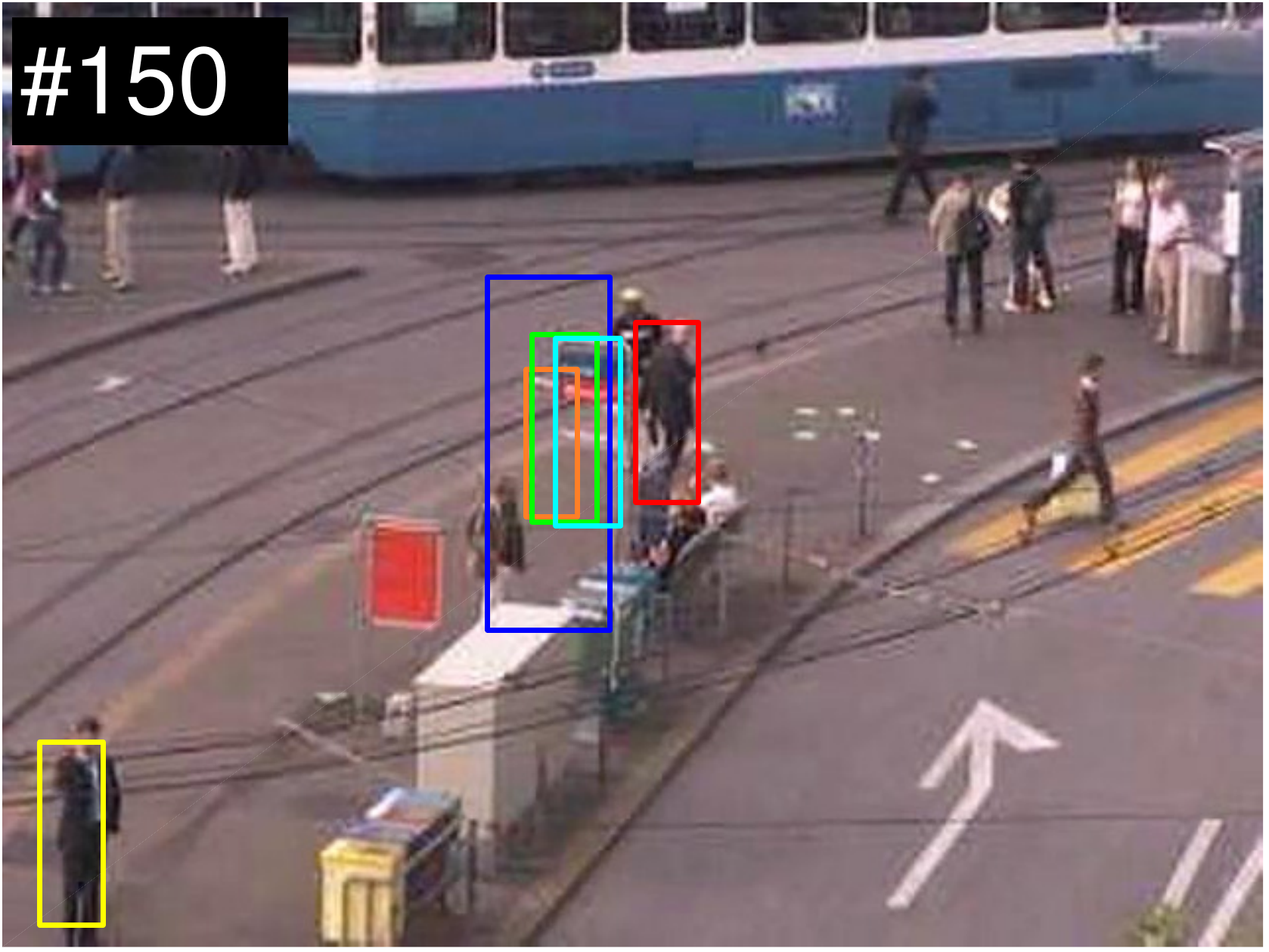} &
			\includegraphics[width=.24\textwidth]{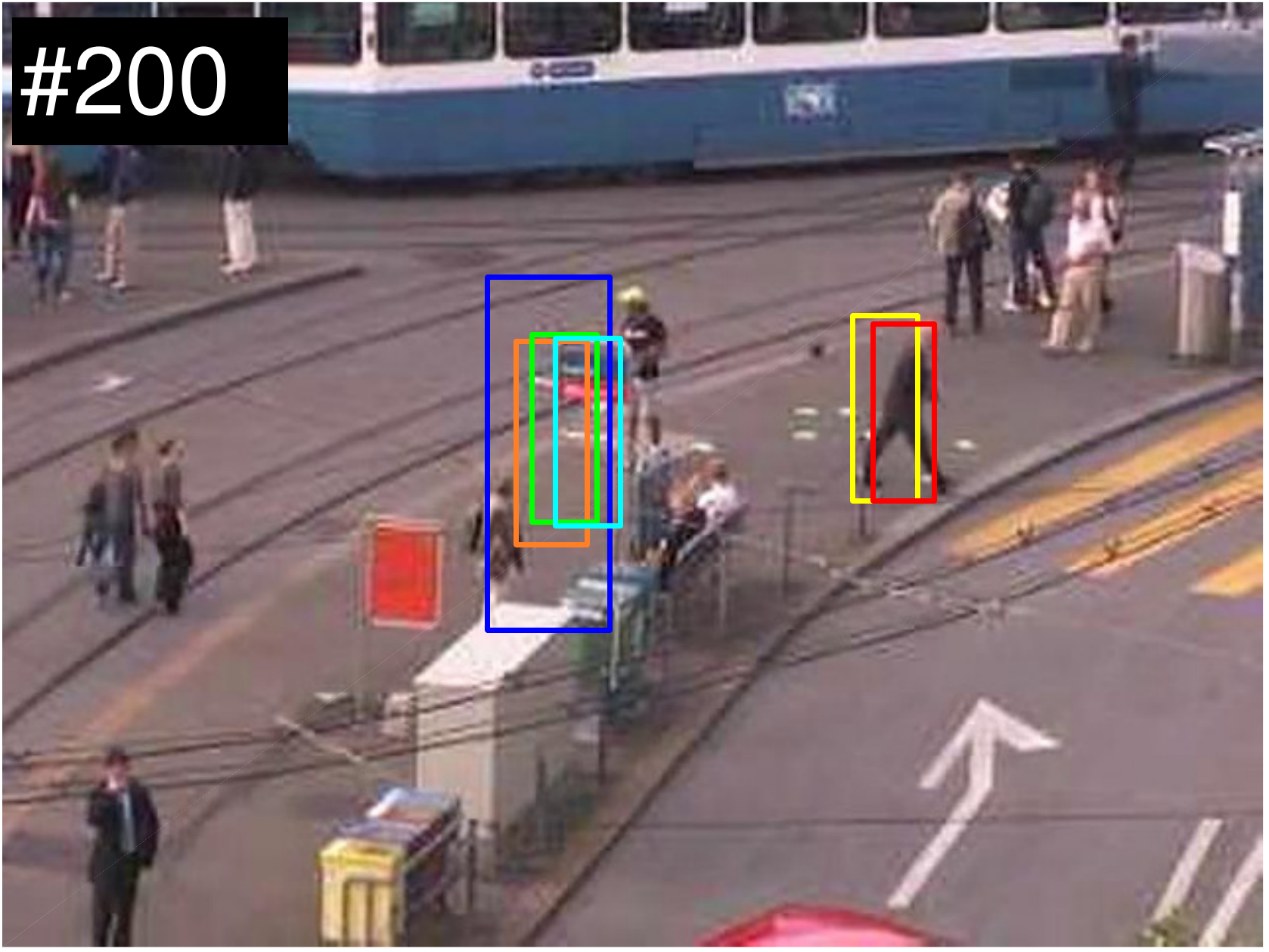} \\
			\includegraphics[width=.24\textwidth]{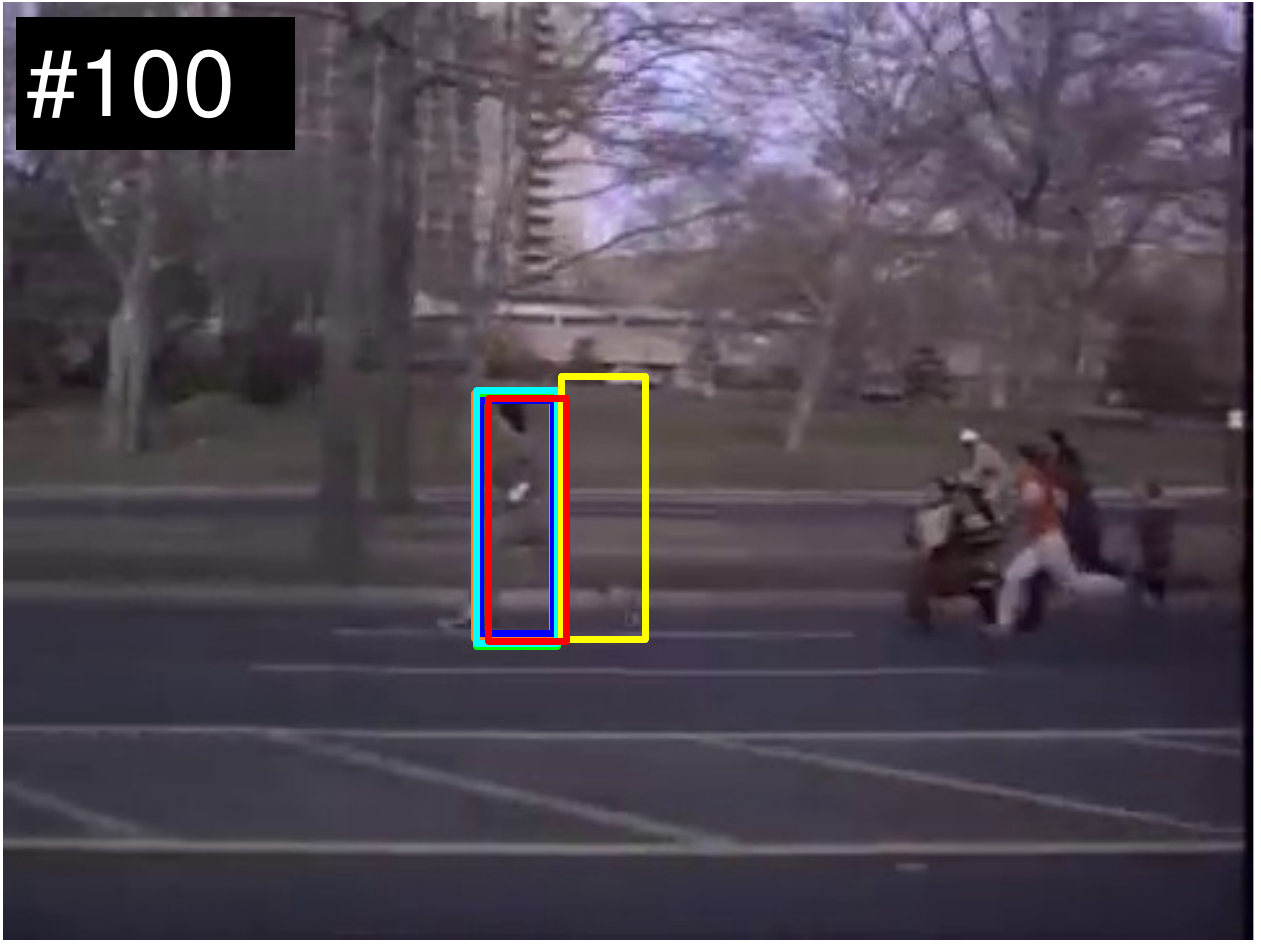} &
			\includegraphics[width=.24\textwidth]{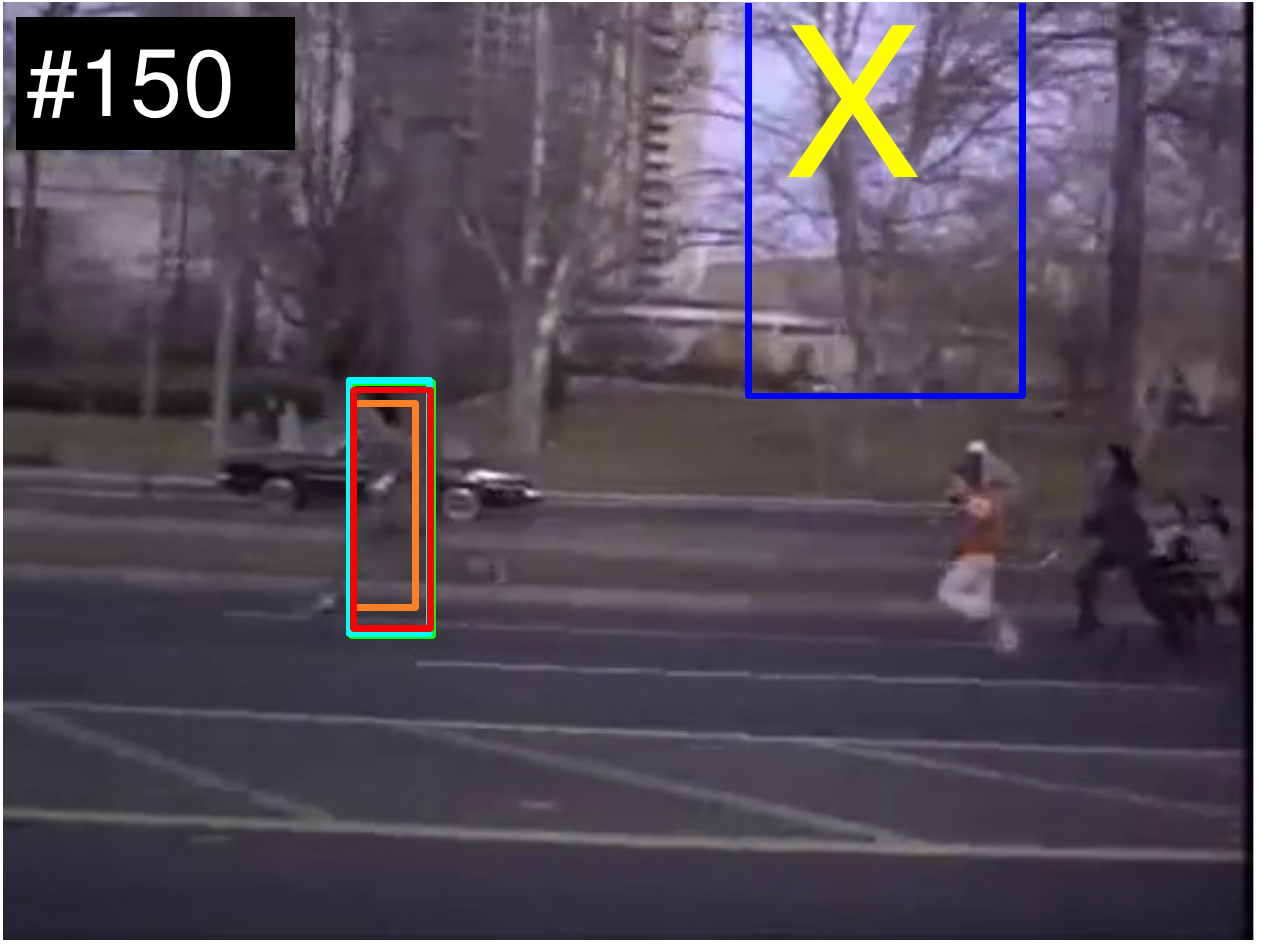} &
			\includegraphics[width=.24\textwidth]{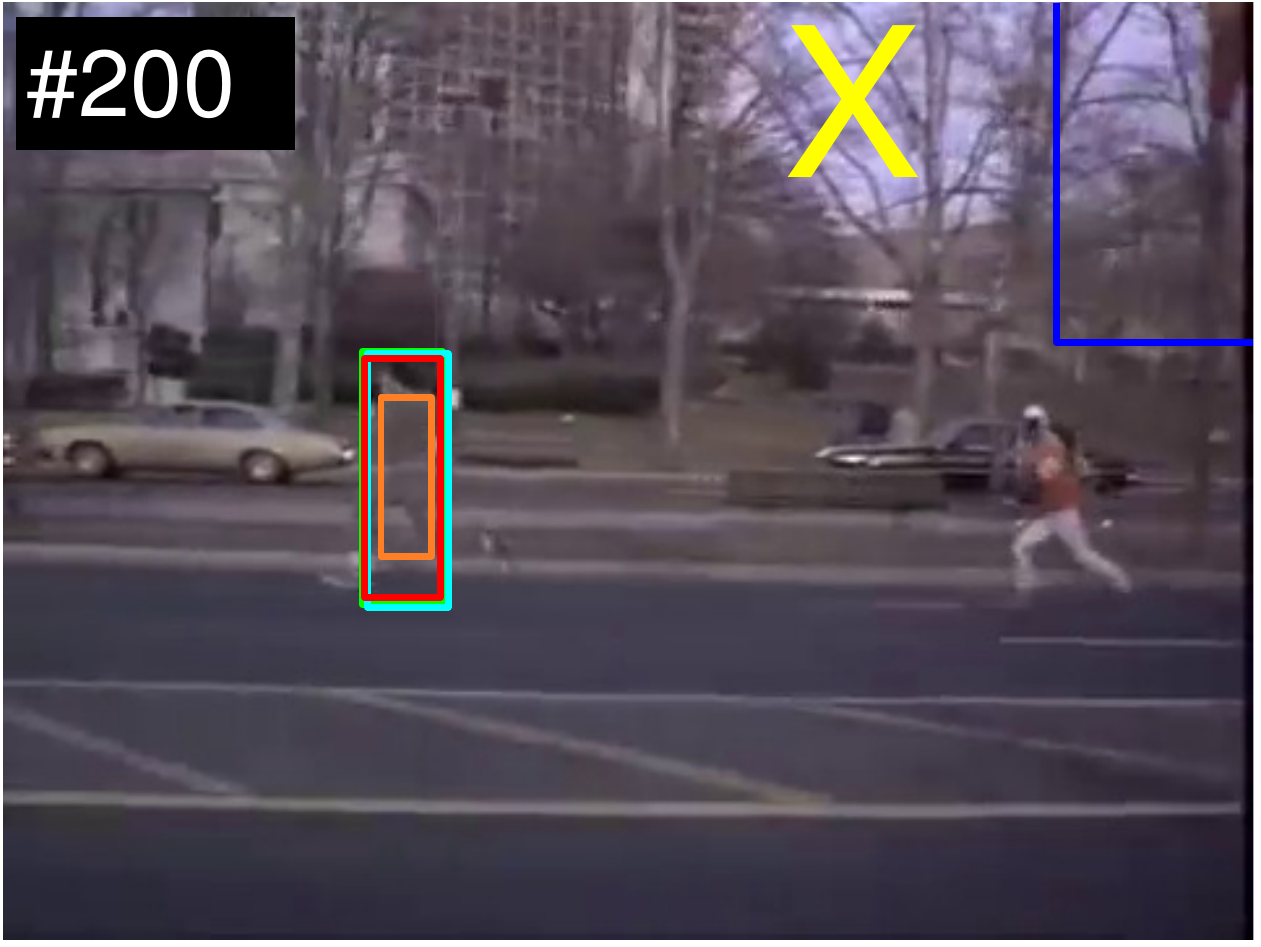} &
			\includegraphics[width=.24\textwidth]{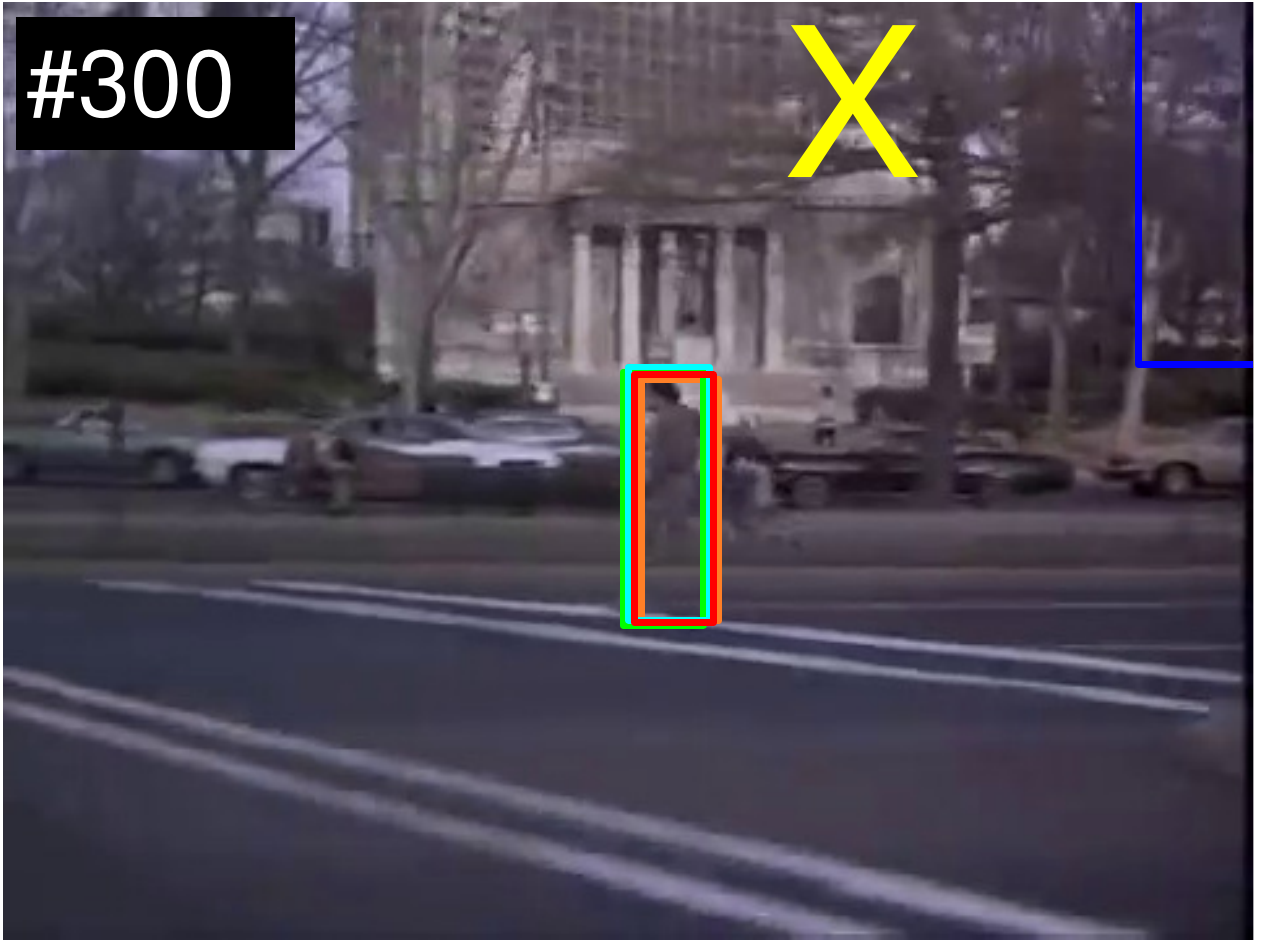} \\
		\end{tabular}
		\includegraphics[width=.86\textwidth]{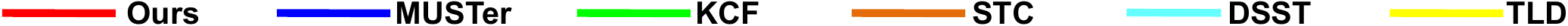} \\
		\caption{
			\textbf{Qualitative comparison in long-term tracking.}
			Results on four challenging sequences: \textit{boxing1, boxing2, ped1}, and \textit{rocky} are from our approach, the MUSTer \cite{Hong_2015_CVPR}, KCF \cite{DBLP:journals/pami/HenriquesC0B15},
			STC \cite{DBLP:conf/eccv/ZhangZLZY14}, DSST \cite{DBLP:conf/bmvc/DanelljanKFW14} and TLD \cite{DBLP:journals/pami/KalalMM12} algorithms
			($\times$: no tracking output for TLD). Our approach performs well against the baseline methods.}
		\label{fig:resultmeem}
	\end{figure*}
	\begin{figure}
		\centering
		\setlength{\tabcolsep}{0em}
		\begin{tabular}{cc}
			\includegraphics[width=.24\textwidth]{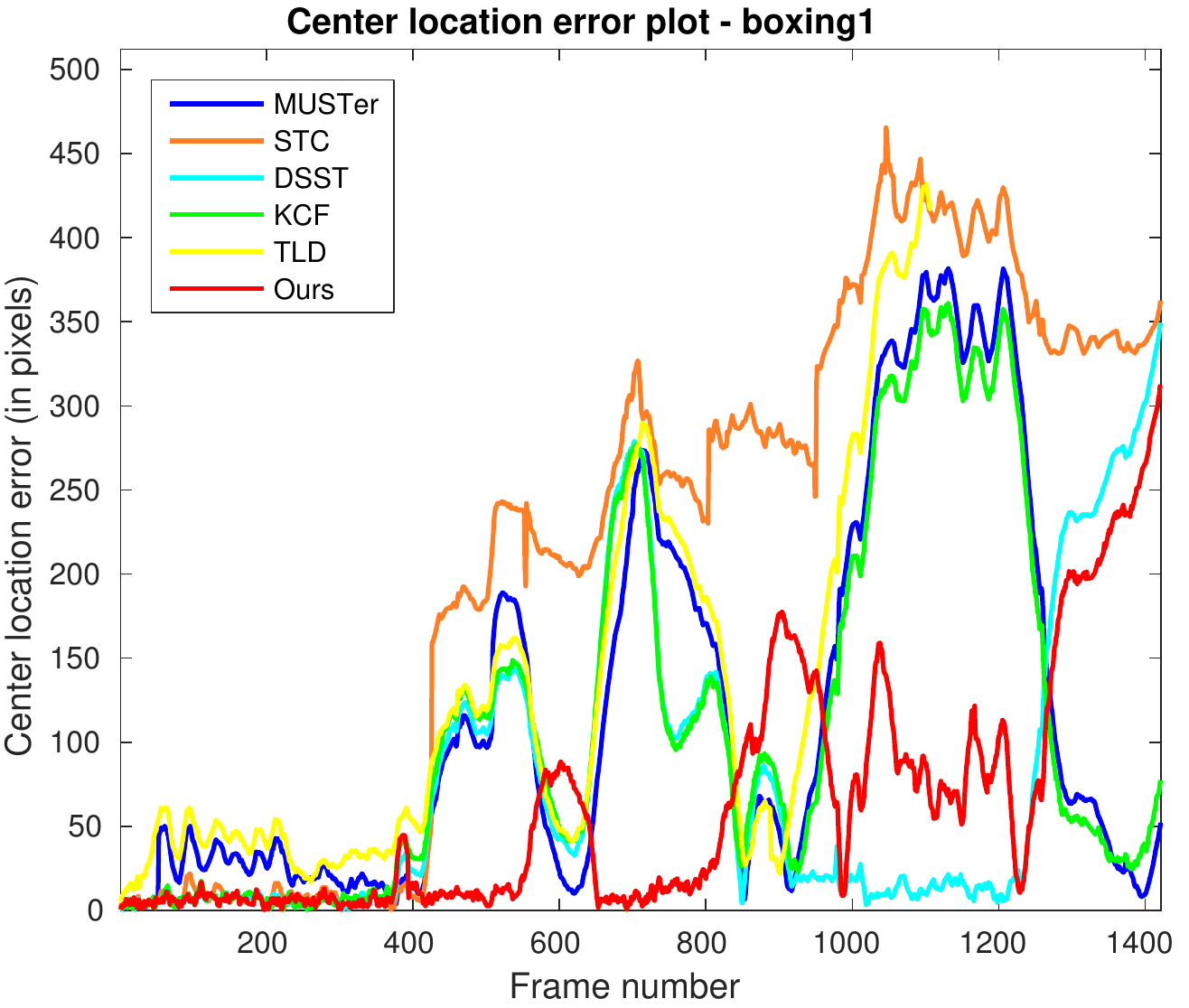} &
			\includegraphics[width=.24\textwidth]{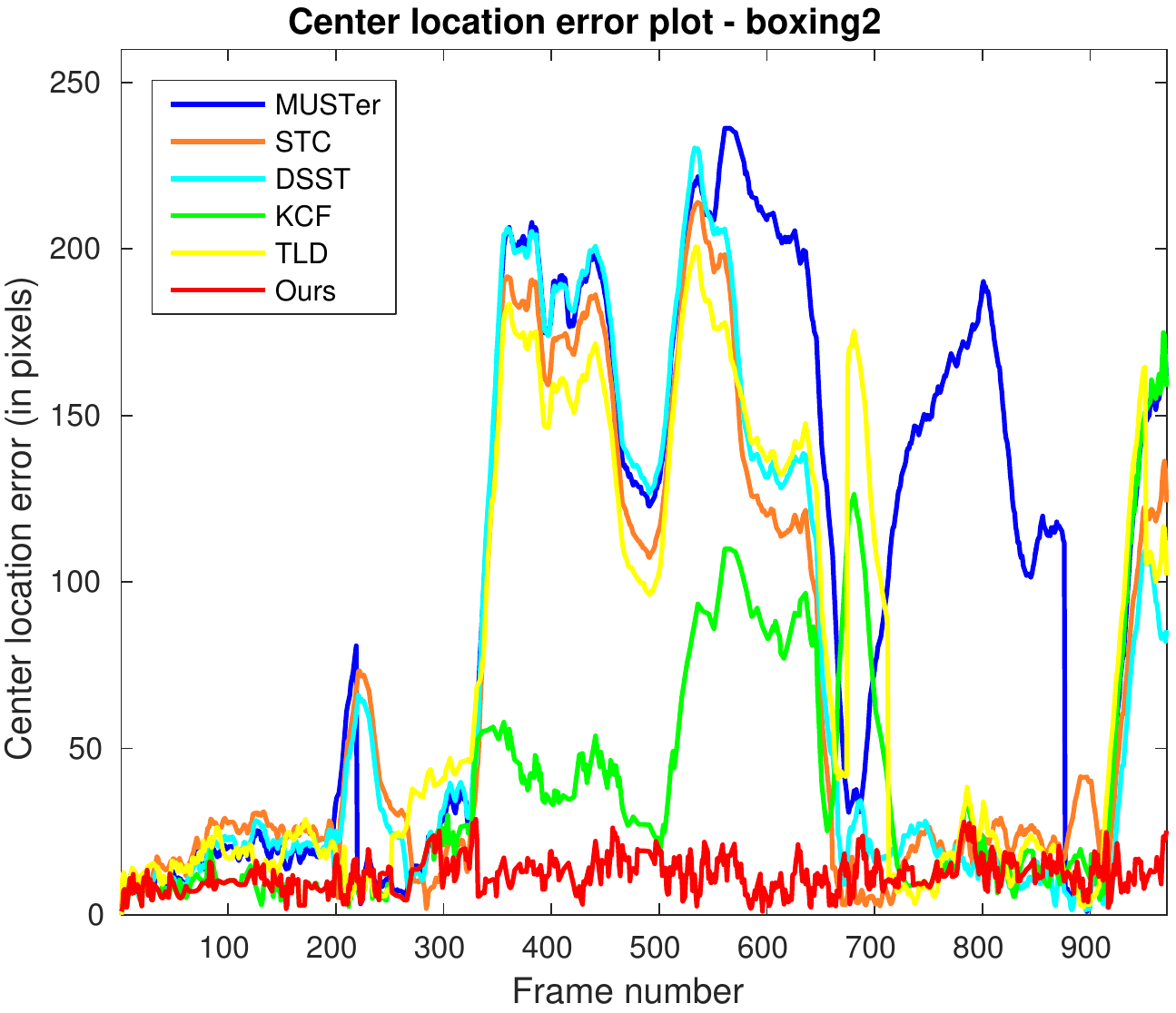} \\
			\includegraphics[width=.24\textwidth]{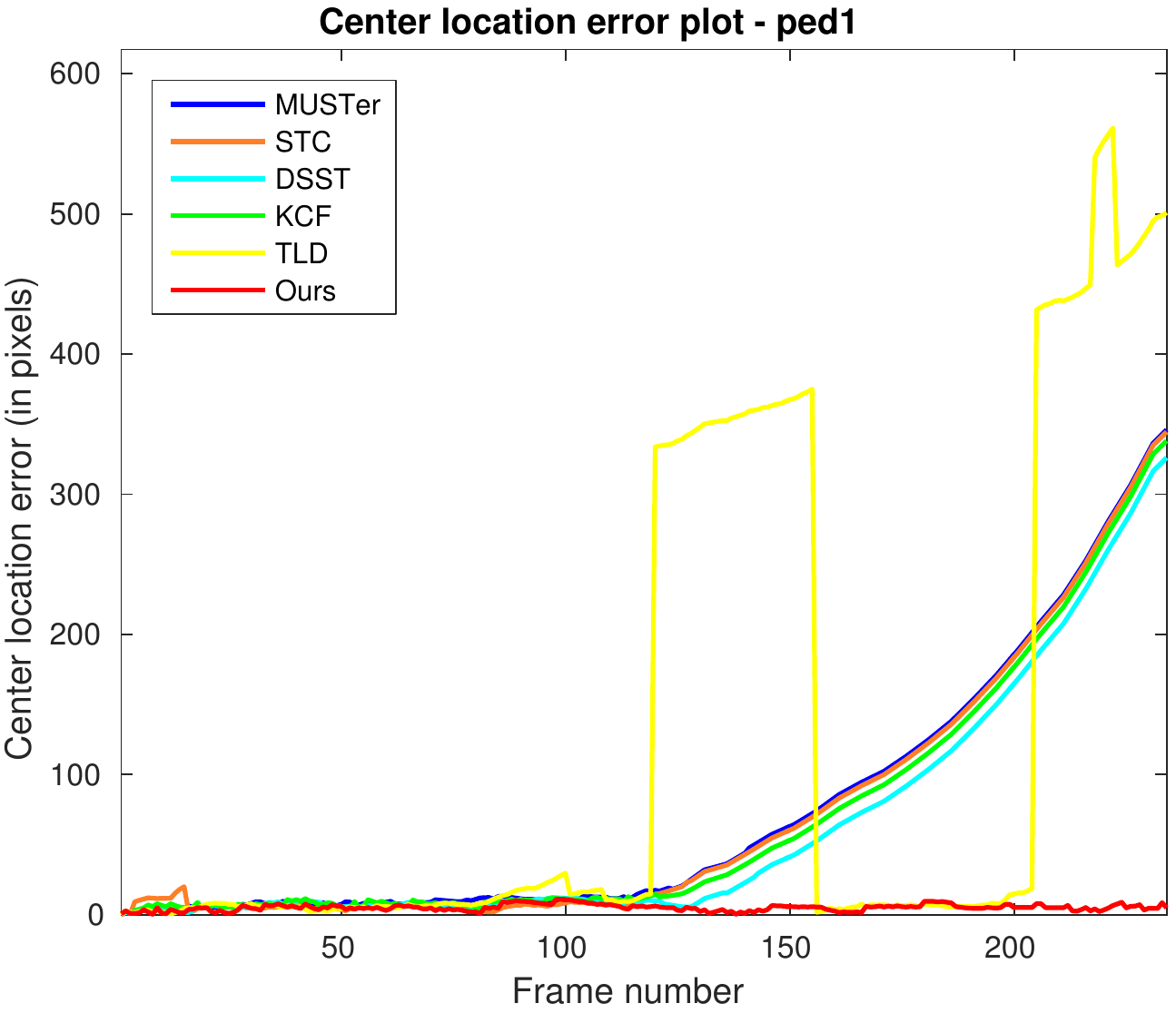} &
			\includegraphics[width=.24\textwidth]{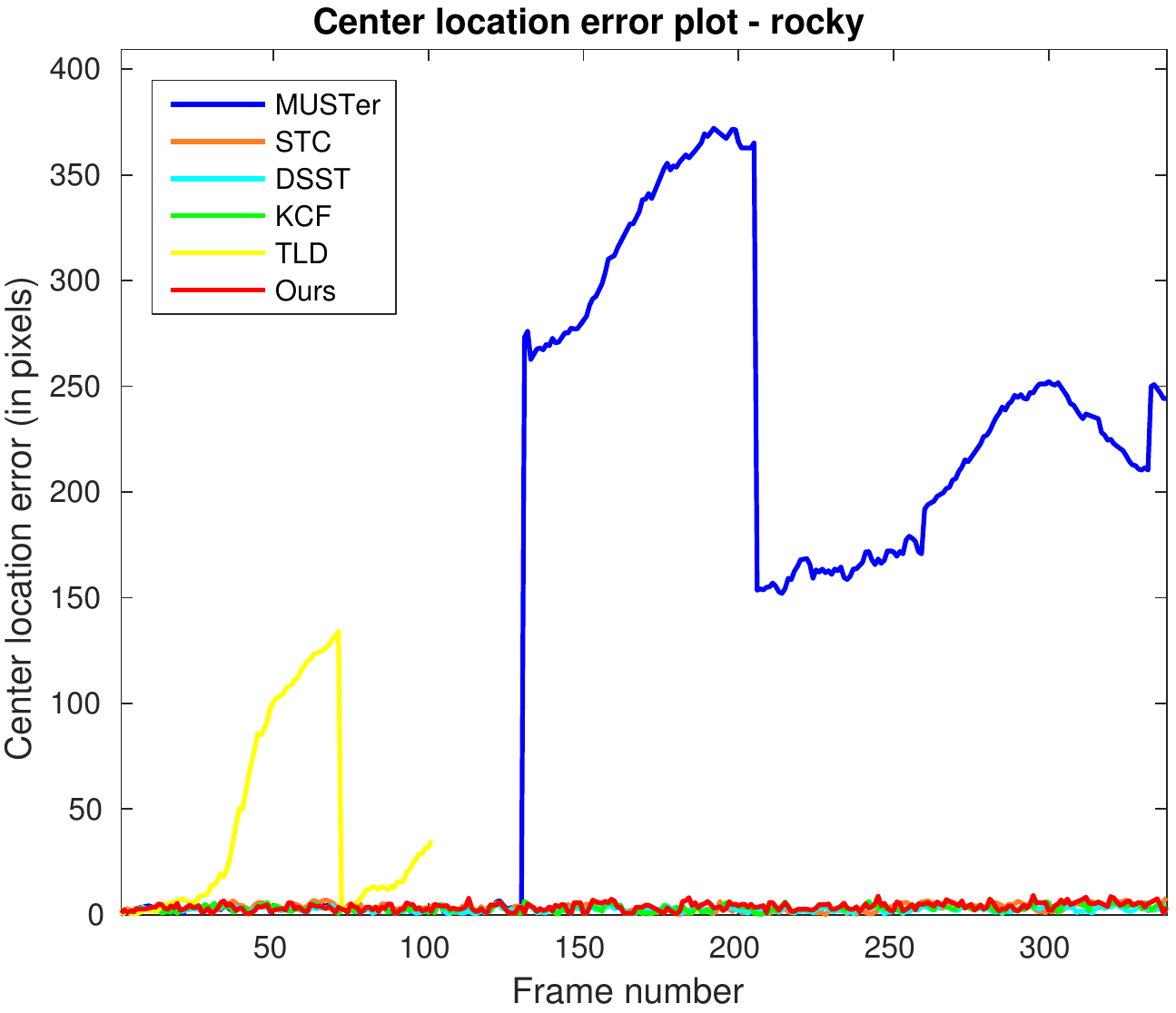} \\
		\end{tabular}
		\caption{\textbf{Quantitative results in center location error on the four challenging sequences in Figure \ref{fig:resultmeem}.} 
			Our method performs favorably against the compared trackers.
		}
		\label{fig:clemeem}
	\end{figure}

	\subsection{MEEM Dataset}
	In addition to the OTB2013 and OTB2015 datasets, we compare with correlation filter based trackers (MUSTer, KCF, STC, and DSST) on the 
	dataset used in the MEEM method ({\small \url{http://cs-people.bu.edu/jmzhang/MEEM/MEEM.html}}).
	The MEEM dataset contains 10 sequences with featured challenging attributes, 
	such as heavy occlusion (\textit{dance}, \textit{boxing1}, \textit{boxing2}, \textit{ped1}, 
	and \textit{ped2}), abrupt illumination changes (\textit{carRace}, and \textit{billieJean}), low contrast
	(\textit{ball}, \textit{ped2}, \textit{rocky}, and \textit{billieJean}), and significant non-rigid deformation
	(\textit{latin}, \textit{ball}, \textit{carRace}, \textit{dance}, and \textit{billieJean}). 
	This dataset contains approximately 7500 frames in total. 
	We report the results of our method without using deep features for fair comparison.
	We also include the TLD tracker in our comparison to analyze the effectiveness of the re-detection module. 
	For fair comparisons, we fix all the parameters as used in the OTB2013 and OTB2015 benchmark studies. 
	
	Table \ref{tb:osmeem} and Table \ref{tb:dpmeem} show that the proposed algorithm performs favorably against the state-of-the-art trackers with more than 10\% gains in both overlap success and distance precision rate.
	Among the other correlation trackers, the KCF method performs well due to the robustness of kernelized correlation filters. 
	The DSST and MUSTer trackers update the translation filters by taking the scale changes into consideration.
	We observe that such an update scheme does not perform well on these challenging sequences as slight inaccuracy in scale estimation causes significant performance loss of the translation filter and the detection module.
	The MUSTer tracker is sensitive to false positive detections and thus fails to track target objects.
	
	We show the qualitative tracking results on four challenging sequences in Figure \ref{fig:resultmeem} and compare their center location error in Figure \ref{fig:clemeem}.
	Figure \ref{fig:resultmeem} shows that correlation trackers without re-detection modules (e.g., KCF, STC and DSST) are unable to recover target objects from heavy occlusion (\textit{boxing1}, \textit{boxing2}, and \textit{ped1}).
	We compare our approach with the MUSTer and TLD trackers in greater details.
	In the \textit{boxing1} sequence, the target boxer in blue is occluded by the other boxer and ropes.
	The MUSTer tracker uses a pool of local key-point features as the long-term memory of target appearance
	and fails to handle heavy occlusion as few reliable key points are detected in such case. 
	For the TLD tracker, the detector is learned on the thresholded intensity features, which are less discriminative in representing the target undergoing fast motion and frequent occlusion. 
	In the \textit{boxing2} sequence, the MUSTer tracker aggressively updates the detector online and yields a false positive detection in the 400th frame. This false positive detection inaccurately reinitializes the tracking component and causes rapid performance loss of both the tracker and detector in subsequent frames.
	Instead, our tracker alleviates the noisy update problem through a conservative update scheme and thus increases tracking precision. 
	In the \textit{ped1} sequence, the MUSTer tracker does not estimate scale correctly at the beginning of the sequence. 
	The errors in scale estimation get accumulated in subsequent frames and adversely affects the translation estimation and the long-term memory module.
	Our method updates the translation and scale filters independently and does not from such error accumulation.
	For the \textit{rocky} sequence, parts of the target object are similar to the tree branches in the background due to low image resolution.
	As such, false positive detections cause both the MUSTer and TLD trackers to lose the target object. 
	
	Overall, the proposed tracker effectively exploits multiple correlation filters for robust object tracking. 
	Both the qualitative (Figure \ref{fig:resultmeem}) and quantitative (Figure \ref{fig:clemeem}) results demonstrate that the proposed tracking algorithm performs favorably against the state-of-the-art trackers.

	\section{Conclusion}
	
	In this paper, we propose an effective algorithm for robust object tracking. 
	Building on the recent success of correlation filter based tracking algorithm, we extended it in several aspects. 
	First, we address the stability-adaptivity dilemma by exploiting three correlation filters: (1) translation filter, (2) scale filter, and (3) long-term filter. 
	These three filters work collaboratively to capture both the short-term and long-term memory of target object appearance. 
	Second, we propose to learn correlation filter using HOI features in addition to the commonly used HOG features for improving localization accuracy. We further investigate the appropriate size of surrounding context and learning rates to improve the tracking performance. 
	Third, we explicitly handle tracking failures by incrementally learning an online detector to recover the targets. 
	We provide a comprehensive ablation study to justify our design choices and understand the trade-off.
	Extensive experimental results show that the proposed algorithm performs favorably against the state-of-the-art methods in terms of efficiency, accuracy, and robustness.

	\section*{Acknowledgments}
	
This work is supported in part by the National Key Research and Development Program of China (2016YFB1001003), NSFC (61527804, 61521062) and the 111 Program (B07022).
	
	\bibliographystyle{spmpsci} 
	\bibliography{bib_long_title,bib_pami15_tracking} 
	
	\newpage
	
	\section*{SUPPLEMENTARY DOCUMENT}
	
	In this supplementary document, we present three additional ablation studies on the OTB2013 dataset. 
	First, we show the results of updating the correlation filters by directly minimizing the errors over all the tracked results. 
	Second, we analyze the robustness of the proposed method by spatially shifting the ground truth bounding boxes. 
	Third, we investigate the effect of training data for the LSTM tracker.

\begin{figure}
		\centering
		\setlength{\tabcolsep}{1mm}
		\begin{tabular}{cc}
			\includegraphics[width=0.22\textwidth]{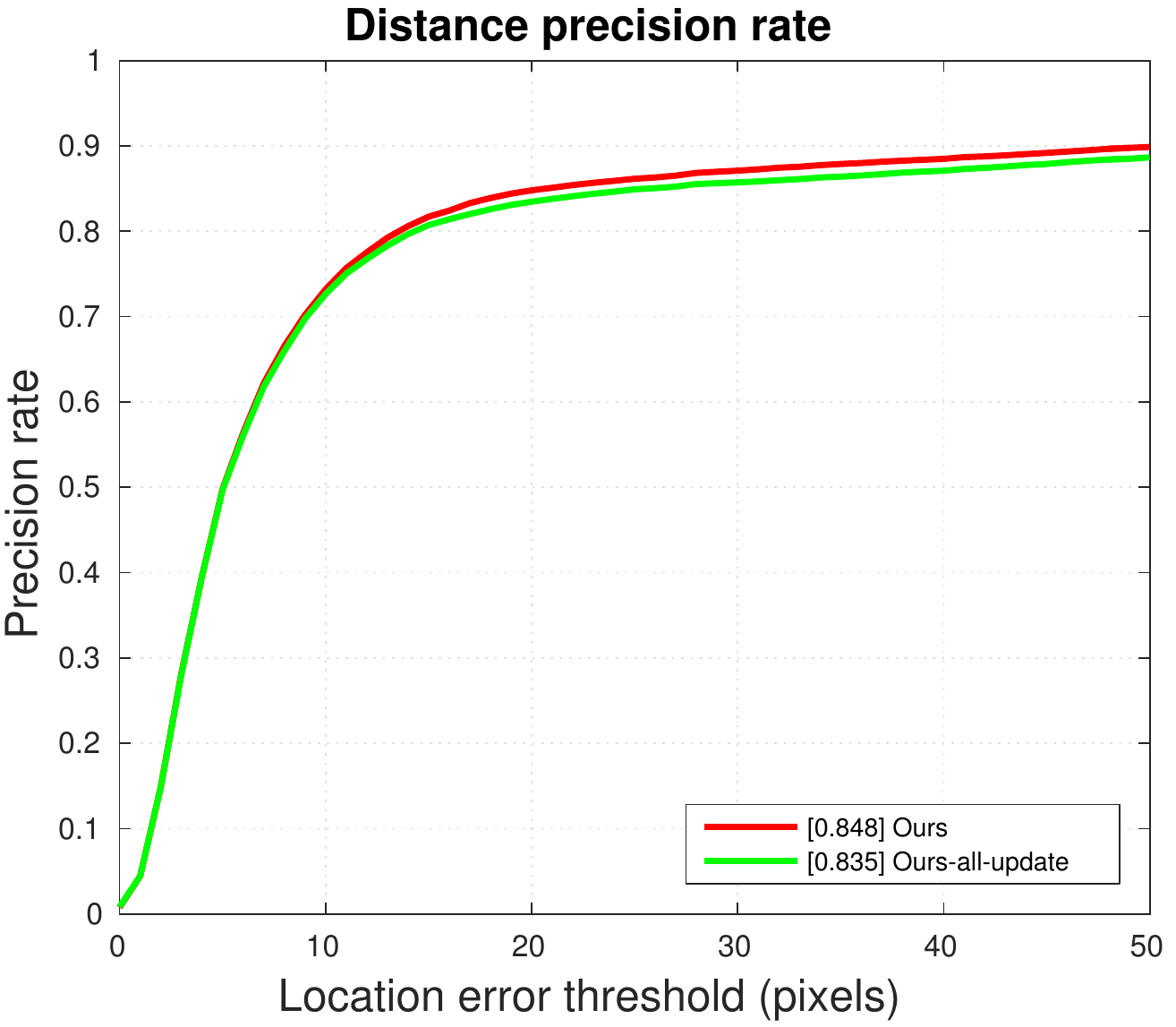} & 
			\includegraphics[width=0.22\textwidth]{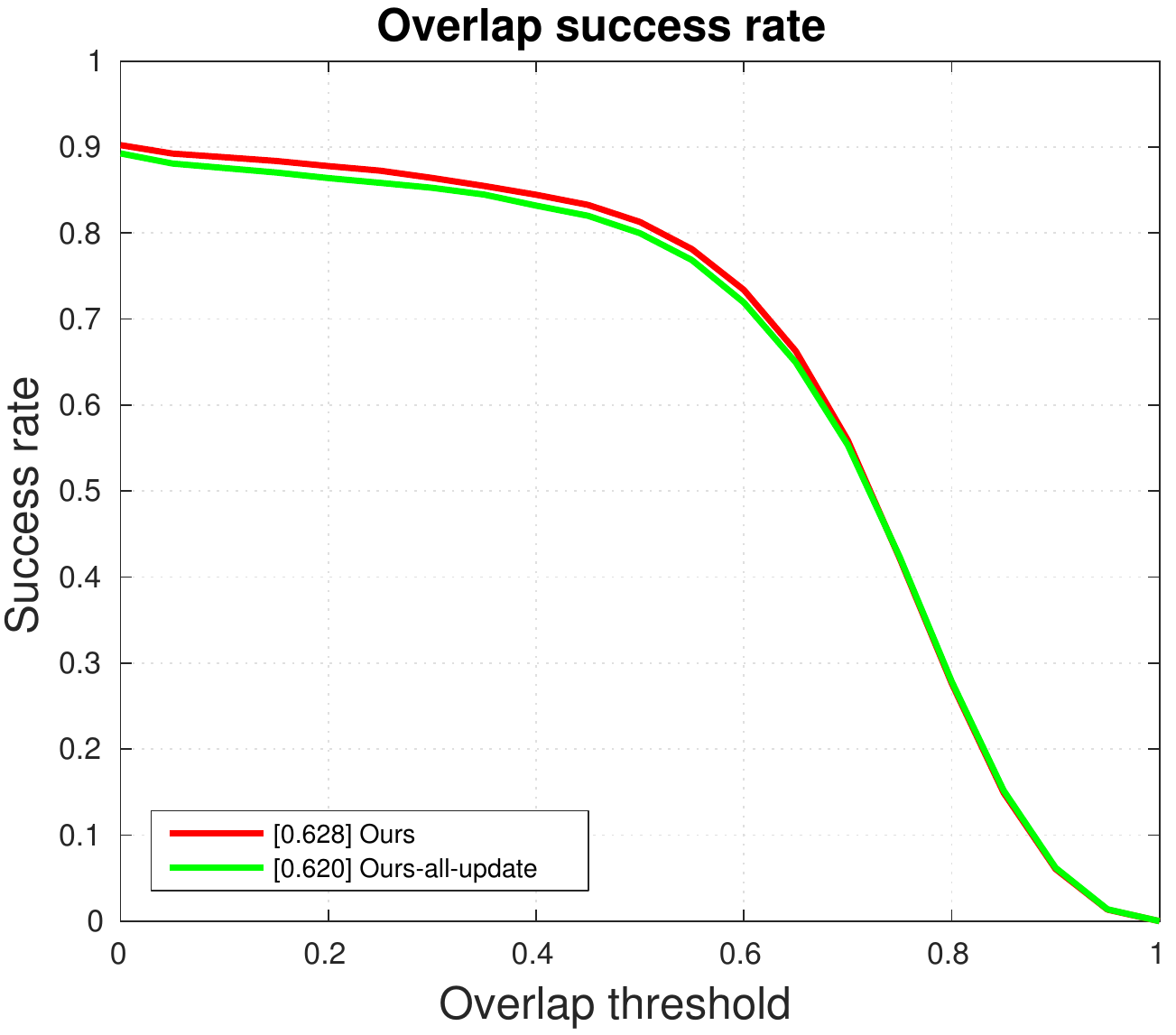} \\
		\end{tabular}
		\caption{Performance of different update schemes on the OTB2013 dataset \cite{DBLP:conf/cvpr/WuLY13} under one pass evaluation (OPE). Considering all the tracked results (ours-all-update) to update the translation filter does not improve tracking performance. The legend of precision plots shows the distance precision scores at 20 pixels. The legend of success plots contains the overlap success scores with the area under the curve (AUC).
		}
		\label{ablation-update}
	\end{figure}

	By directly minimizing the errors over all the tracked results, we consider \emph{all} the  extracted appearances $\{x_j,j=1,2,\ldots,p\}$ of the target object from the first frame up to the current frame $p$. 
	The cost function is the weighted average quadratic error over these $p$ frames.
	We assign each frame $j$ with a weight $\beta_j \ge 0$ and learn correlation filter $w$ by minimizing the following objective function:
	\begin{equation}
	\label{sequ:filter}
	\min_w\sum_{j=1}^p\beta_j\left( \sum_{m,n}\left| \left\langle \phi\left(x_{m,n}^j\right), w^j \right\rangle -y^j(m,n) \right|^2 +\lambda\left\langle w^j,w^j\right\rangle \right),
	\end{equation}
	where $w^j=\sum_{k,l}a(k,l)\phi(x_{k,l}^j)$. 
	We have the solution to \eqref{sequ:filter} in the Fourier domain as:
	\begin{equation}
	\mathcal{A}^p=\frac{\sum_{j=1}^p\beta_j\bK_x^j\odot\overline{\bY}}{\sum_{j=1}^p\beta_j\bK_x^j\odot\left(\bK_x^j+\lambda\right)},
	\label{eqA}
	\end{equation} 
	where $\bK_x^j=\mathscr{F}\left\{k_x^j\right\}$ and $k_x^j(m,n)=k(x_{m,n}^j,x^j)$. 
	We perform a grid search and set the weight $\beta_j=0.01$ and the update rate $\lambda=10^{-4}$ for the best accuracy. 
	We restore the parameter $\{\bK_x^j\}$, $j=1,2,\ldots,p-1$, to update the correlation filter in frame $j$. 
	Note that such an update scheme is not applicable in practice as it requires a linearly increasing computation and memory storage over the increase of frame number $p$. 
	The average tracking speed is 2.5 frames per second (fps) vs. 20.8 fps (ours) on the OTB2013 dataset. 
	However, Figure \ref{ablation-update} shows that this update scheme does not improve performance. 
	The average distance precision is 83.5\% vs. 84.8\% (ours), and the average overlap success is 62.0\% vs. 62.8\% (ours). 
	%

\begin{figure}
		\centering
		~ \includegraphics[width=.46\textwidth]{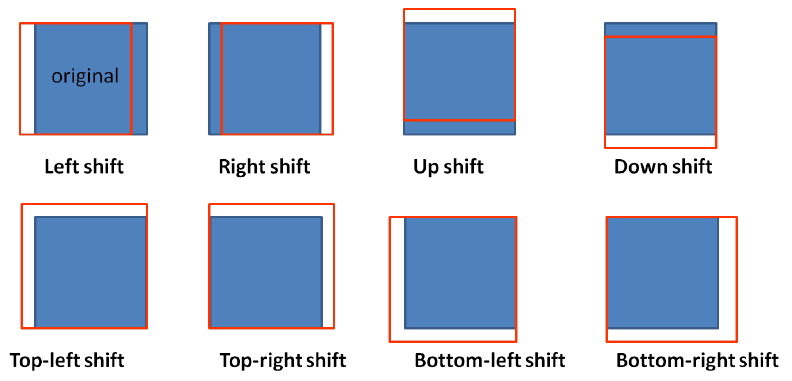} \\
		\caption{Spatial shifts. The amount of shift is 10\% of width or
			height of the ground-truth bounding box.}
		\label{fig:shift}
	\end{figure}

\begin{figure}
		\centering
		\includegraphics[width=.22\textwidth]{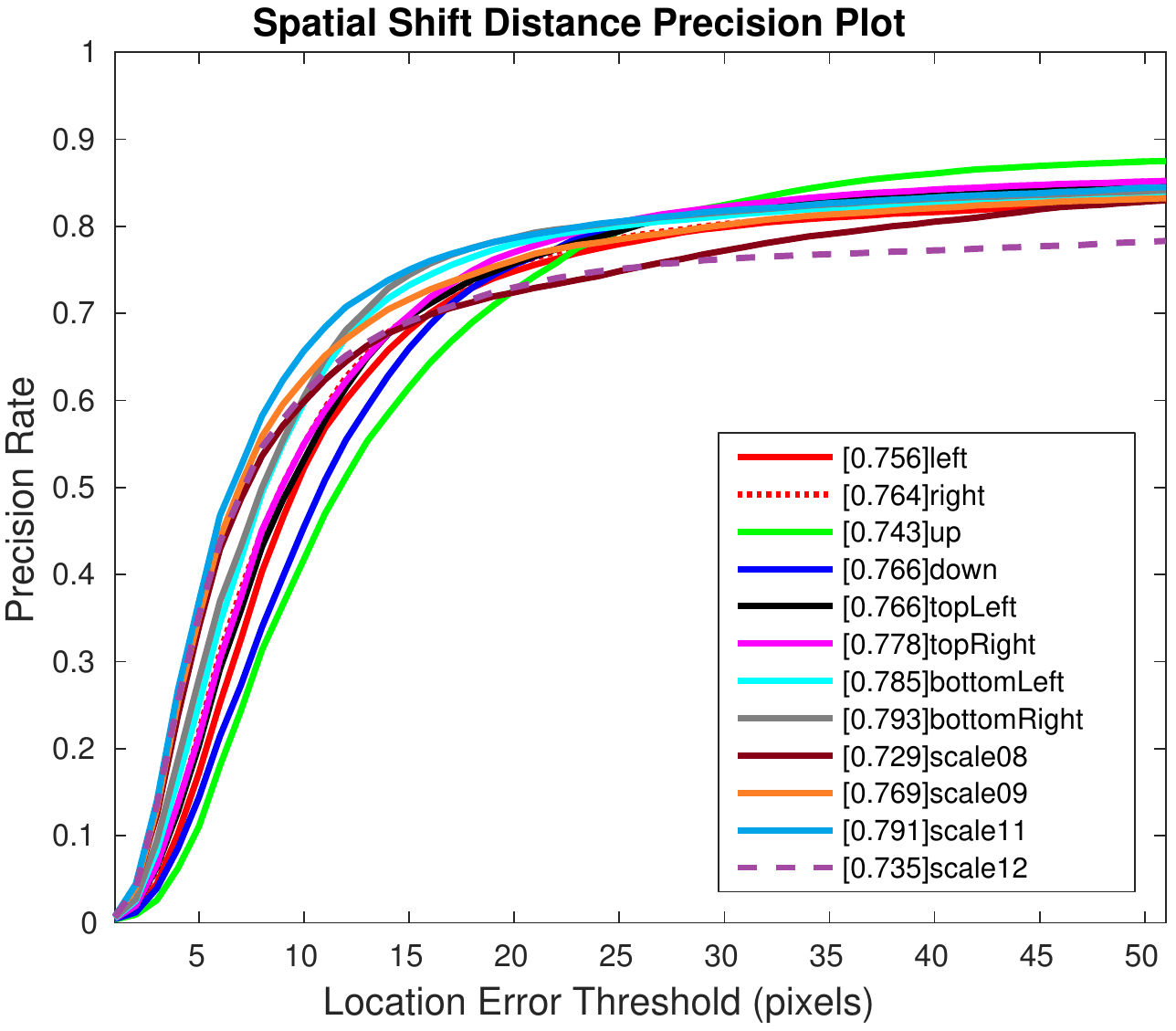} 
		\includegraphics[width=.22\textwidth]{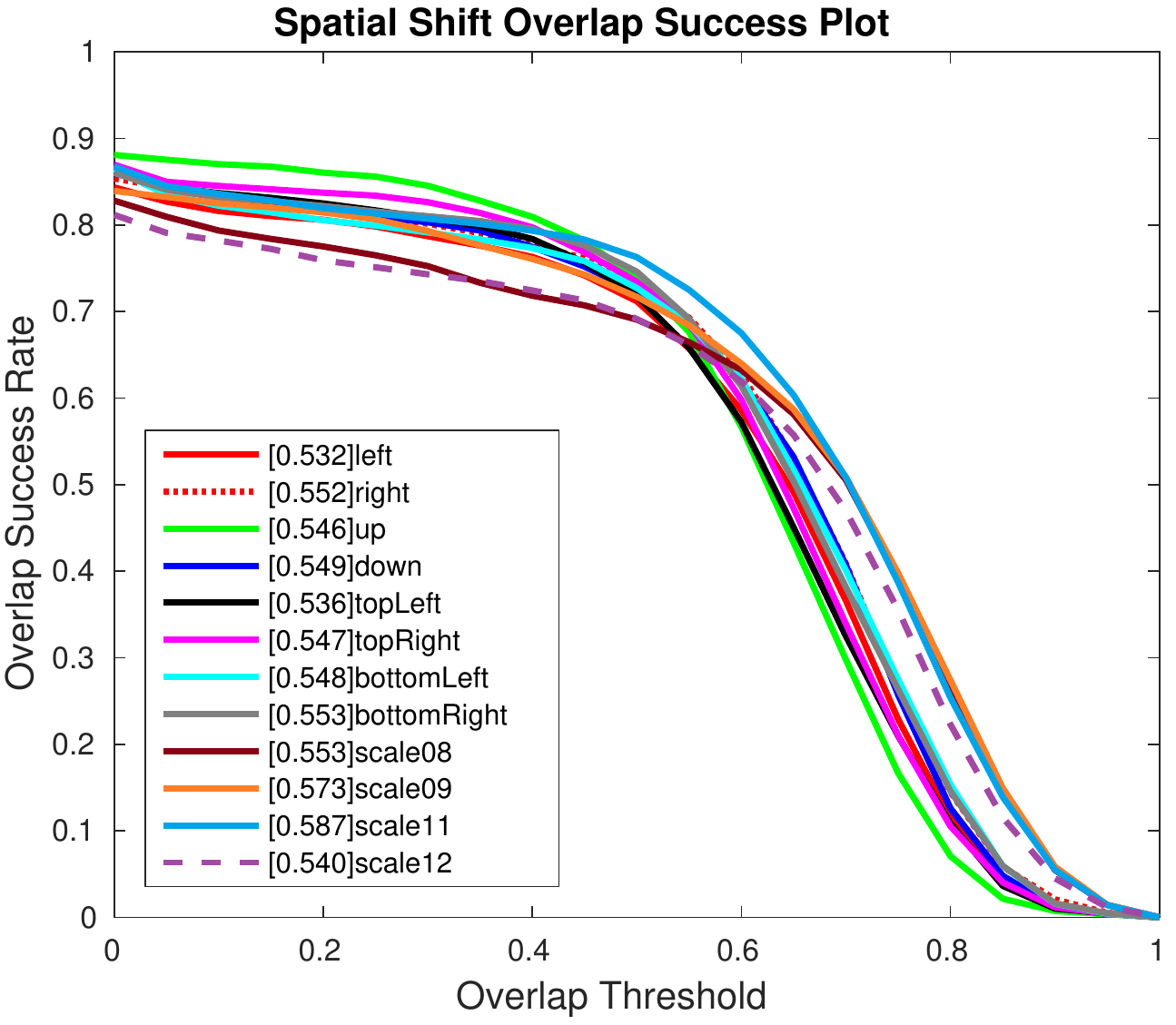}
		\caption{Tracking performance with spatially shifted ground truth bounding boxes on the OTB2013 dataset \cite{DBLP:conf/cvpr/WuLY13} under one pass evaluation (OPE).}
		\label{fig:shiftresults}
	\end{figure}

	
	We spatially shift the ground truth bounding boxes with eight directions (Figure \ref{fig:shift}) and rescale the ground truth bounding boxes with scaling factors 0.8, 0.9, 1.1 and 1.2. 
	Figure \ref{fig:shiftresults} shows that slightly enlarge the ground truth bounding boxes (with scaling factor 1.1) does not significantly affect the tracking performance. 
	%
	
\begin{figure}
		\centering
		\setlength{\tabcolsep}{.5mm}
		\begin{tabular}{cc}
			\includegraphics[width=0.22\textwidth]{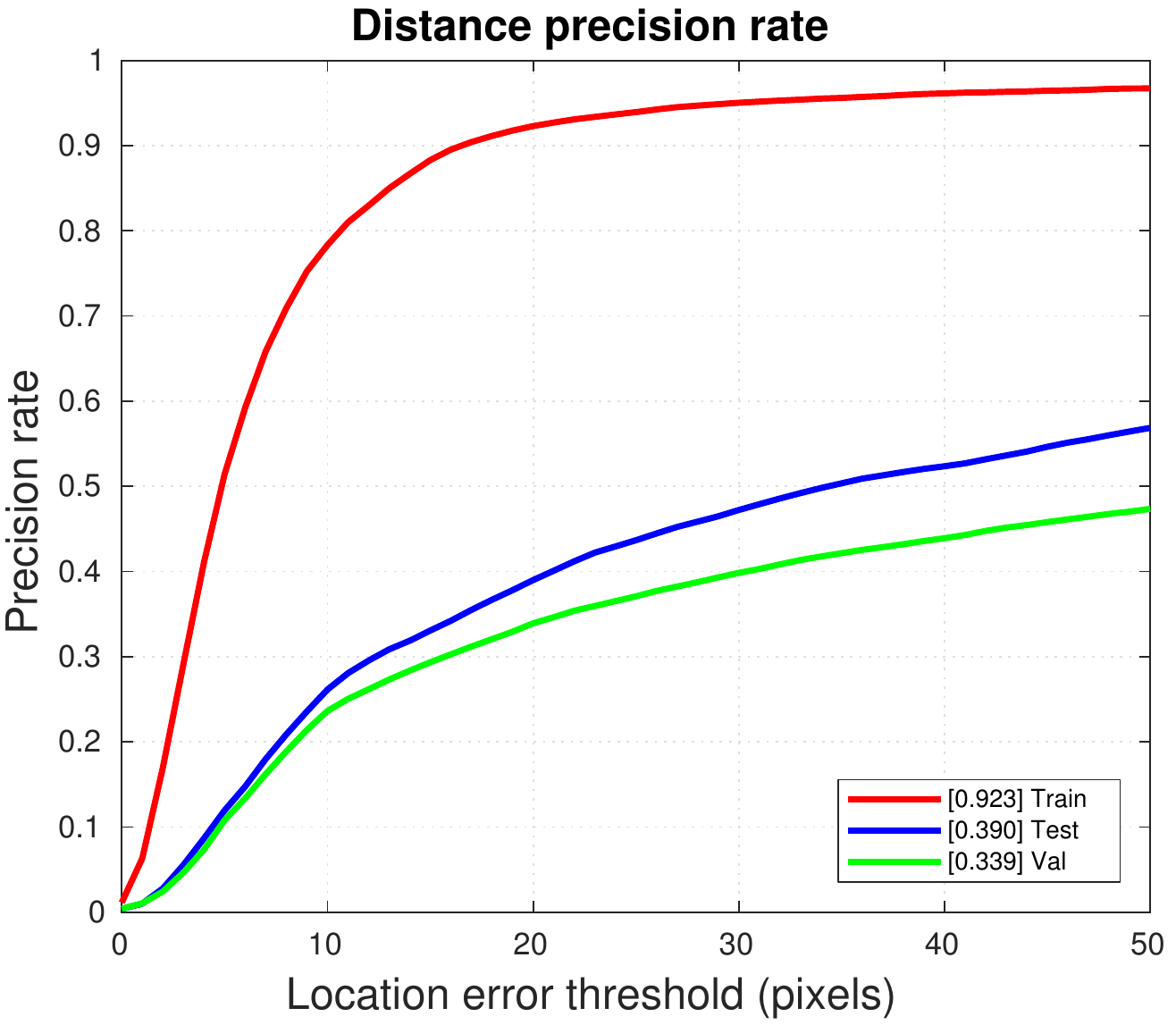} & 
			\includegraphics[width=0.22\textwidth]{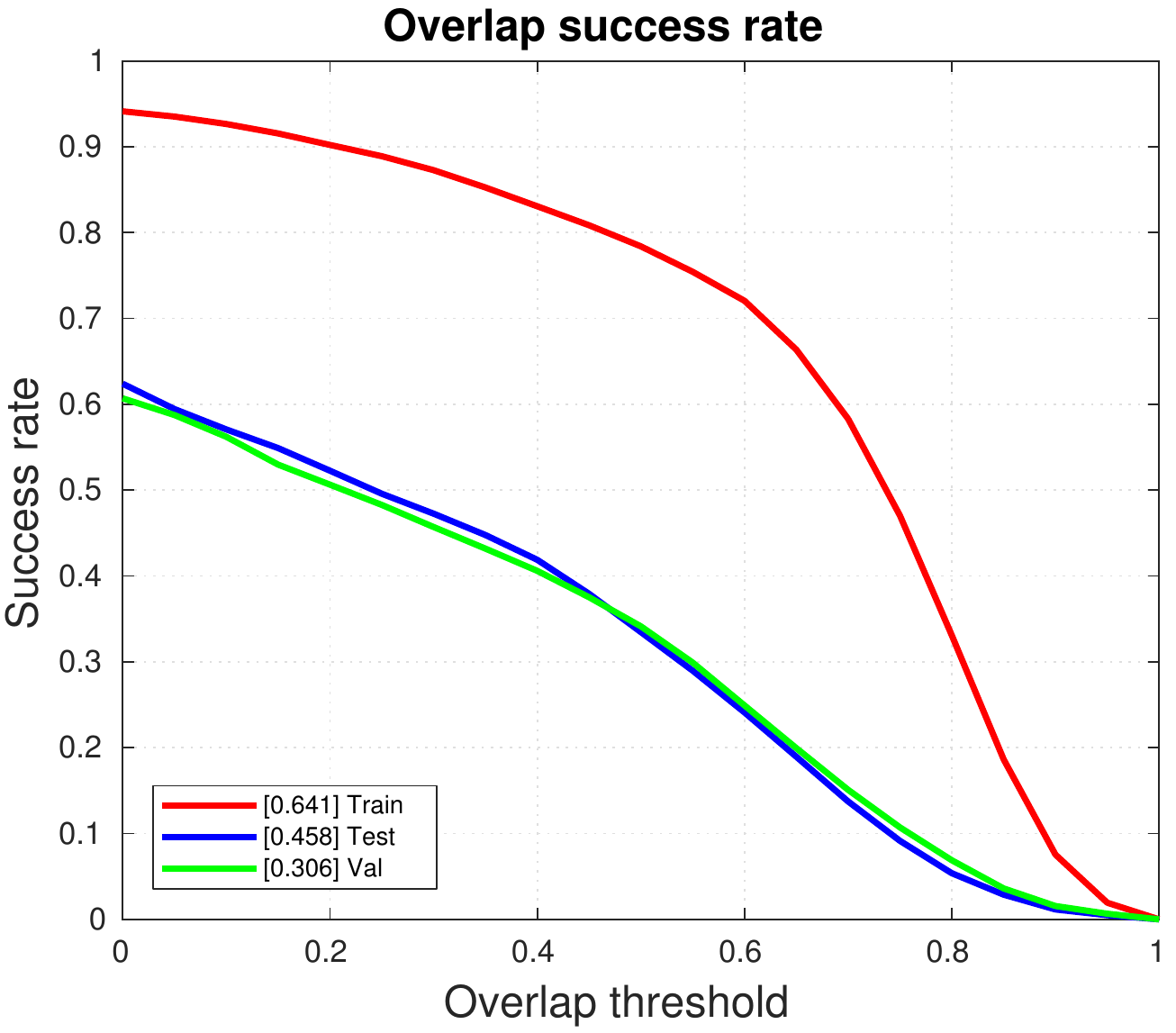} \\
		\end{tabular}
		\caption{\textbf{Performance comparison of the baseline LSTM tracker \cite{rolo}} on training, validation, and test sets under one pass evaluation (OPE). The legend of precision plots shows the distance precision scores at 20 pixels. The legend of success plots contains the overlap success scores with the area under the curve (AUC).
		}
		\label{ablation-lstm}
	\end{figure}

	We follow the project \cite{rolo} (\url{https://github.com/Guanghan/ROLO}) to implement a baseline tracker, which uses the LSTM cell to produce tracking results.  We use the same 50 sequences from the OTB2013 dataset as the test set and the remaining sequences of OTB2015 as the validation set. Figure \ref{ablation-lstm} shows the tracking performance on training, validation, and test sets. The large performance gap on training and validation/test sets is due to limited training data.
	
\end{document}